\documentclass[runningheads]{llncs}

\usepackage{eccv}

\usepackage{eccvabbrv}

\usepackage{graphicx}
\usepackage{booktabs}
\usepackage{bm}
\usepackage{tikz}
\usepackage{colortbl}
\usepackage{multirow}
\usepackage[normalem]{ulem}
\useunder{\uline}{\ul}{}

\usepackage[accsupp]{axessibility}  

\usepackage{hyperref}

\usepackage{orcidlink}

\begin{document}

\title{DLGStream: Dynamic Language-embedded Guassian Splatting for Open-vocabulary Enabled Free-viewpoint Video Streaming} 

\titlerunning{DLGStream}

\author{Zhihui Ke\orcidlink{0009-0002-4042-7044} \and
Yuyang Liu\orcidlink{0009-0008-2706-3745} \and
Xiaobo Zhou\orcidlink{0000-0002-7772-458X}\thanks{Corresponding author.} \and
Tie Qiu\orcidlink{0000-0003-2324-2523}}

\authorrunning{Zhihui~Ke et al.}

\institute{School of Computer Science and Technology, Tianjin University, China
\email{\{kezhihui,yvyang,xiaobo.zhou,qiutie\}@tju.edu.cn}}

\maketitle

\begin{abstract}

3D Gaussian Splatting~(3DGS) has emerged as a promising paradigm for reconstructing streamable free-viewpoint video~(FVV) from multi-view videos. However, 3DGS-based FVVs typically lack user interaction and editing capabilities, which diminishes the immersive experience. Recent research has integrated language features from CLIP into 3DGS via distillation, enabling open-vocabulary queries and supporting many downstream applications. Nevertheless, the stringent requirements of FVV, low frame size and high FPS, make current language Gaussian representations unsuitable for language-embedded FVV. In this paper, we propose DLGStream, a novel language-embedded FVV representation that streams time-varying language features alongside Gaussian attributes to support 4D environment interaction, scene editing, and spatial intelligence. Specifically, we propose a dual-opacity dynamic language Gaussian representation, which maintains two opacity attributes for color and language features to deal with performance degradation that occurs when colors and features are jointly optimized. Furthermore, we introduce an interpolation-based deformation field to reduce temporal redundancy. This deformation field can also be used for 4D frame interpolation, boosting FVV sequences from low to high FPS. Experimental results demonstrate that DLGStream achieves superior performance in both on open-vocabulary segmentation and reconstruction quality with an average frame size of merely 43 KB. The code is available on \href{https://github.com/kkkzh/DLGStream}{https://github.com/kkkzh/DLGStream}.
  \keywords{Language Guassian Splatting \and Dynamic Scene Reconstruction \and Free-viewpoint Video}
\end{abstract}

\section{Introduction}
\label{sec:intro}
The groundbreaking development of 3D Gaussian Splatting~(3DGS)~\cite{kerbl20233d} for static scene reconstruction has prompted the creation of 3DGS-based Free-Viewpoint Video~(FVV) methods. These methods extend 3DGS to dynamic scenes and support frame streaming just like traditional video. Due to their superior reconstruction quality and FPS, 3DGS-based FVV methods have rapidly supplanted those based on Neural Radiance Field~(NeRF). However, these FVV representations are fundamentally composed of neural network parameters, which precludes users from performing tasks such as 3D scene editing, environment interaction, and so on.

Recently, LangSplat~\cite{qin2024langsplat} uses Segment Anything Model~(SAM)~\cite{kirillov2023segment} to segment large, medium, and small objects in multi-view images and extracts their language features from CLIP~\cite{radford2021learning}. Then, these multi-level language features are treated as pseudo ground-truth and distilled into the original 3DGS representation by incorporating language feature attributes. Through language-embedded Gaussian splatting, open-vocabulary queries can be achieved and naturally support various downstream tasks, including 4D environment interaction, scene editing, and embodied and spatial intelligence. 

Based on LangSplat, subsequent research has further enhanced query accuracy and expanded its application scope. For example, 4DLangSplat~\cite{li20254d} introduces language feature distillation into a dynamic 3D Gaussian representation to enable time-sensitive queries. 4DLangSplat adheres to existing training procedures for language Gaussian representation, resulting in three separate language-embedded dynamic 3D Gaussian models, one for each level of language features. The training process first fixes the parameters of the dynamic 3D Gaussians, and then distills language features. However, this scheme is ill-suited for language-embedded FVV. The aims of FVV reconstruction are high reconstruction quality and low frame size, yet 4DLangSplat have to transmit three distinct models triples the frame size. Moreover, executing open-vocabulary queries necessitates multiple rendering passes to obtain the multi-level language features, substantially decreasing FPS. The separate training of language features also prolongs the total training time. Crucially, when we try to jointly train language feature and color, the reconstruction performance degrades~\cite{he2025joint}, especially for color, which is unacceptable for FVV.

To address the aforementioned challenges of language-embedded FVV, we propose DLGStream, a novel language-embedded FVV representation that enables many downstream tasks through streaming time-related language features while preserving the high reconstruction quality and low frame size essential for FVV. To solve the performance degradation when jointly training language features and color, we propose a dual-opacity dynamic language Gaussian representation. This is based on a key insight that color and language feature have different physical properties. While attributes such as position, rotation, and scaling solely define the geometric distribution of the scene. Opacity, however, is integrally involved in the rendering of both color and language feature. Consequently, the discrepancy between color and language features drives opacity optimization in diverging directions. Thus, we use two separate opacity attributes to mitigate the interference between language features and color. 

Furthermore, existing FVV representations learn independent temporal features or Gaussian residuals per frame, which results in significant temporal redundancy. This redundancy is further exacerbated by the incorporation of language features. In this paper, we propose a novel deformation field based on time feature interpolation, inspired by video interpolation. The interpolation-based deformation field maintains key time features at fixed intervals and derives non-key time features via interpolation. This strategy substantially reduces temporal redundancy and frame size. The interpolated time features are then fed into attribute-specific MLPs to predict Gaussian attribute offsets. By modeling continuous temporal changes through interpolation, our deformation field enables 4D frame interpolation for FVV, effectively enhancing a model trained at a low FPS to a higher one. This low FPS training also reduces the cost of extracting language feature labels. Moreover, our proposed interpolation-based deformation field is decoupled from other Gaussian attributes, making it applicable to a wide range of Gaussian-based representations. Finally, we design a GOP-by-GOP training strategy to address background flickering across different GOPs.
The contributions are summarized as follows:
\begin{itemize}
   \item We propose a novel language-embedded FVV representation, DLGStream, for real-time FVV streaming that supports various downstream tasks via 4D open vocabulary queries.
   \item We introduce a dual-opacity dynamic language Gaussian representation to solve the performance degradation during the joint optimization of language features and color.
   \item We develop an interpolation-based deformation field to reduce temporal redundancy and frame size, which also supports 4D frame interpolation.
   \item Our method is applicable to various Gaussian representations. Experiments conducted on 3DGS and Scaffold-GS demonstrate that our method achieves a 9\% mIOU improvement on 4D open-vocabulary query tasks while maintaining state-of-the-art~(SoTA) reconstruction quality, with an average frame size of only 43KB.
\end{itemize}

\section{Related Work}

\subsection{3DGS-based Dynamic Scene Reconstruction}
NeRF-based dynamic scene reconstruction methods~\cite{xian2021space,gao2021dynamic,li2021neural,tretschk2021non, pumarola2021d, park2021nerfies, fang2022fast, park2021hypernerf, du2021neural,li2022nv3d,mihajlovic2023resfields,li2023dynibar,peng2023representing,yu2023dylin,yan2023nerf,wang2023flow,liu2023robust,tian2023mononerf,zhan2024kfd,li2022streaming,wang2022mixed,guo2023forward,fridovich2023k,cao2023hexplane,shao2023tensor4d,yang2025dmit,lou2024darenerf,wang2024masked,wang2023f2,park2023temporal,song2023nerfplayer,attal2023hyperreel} struggle to achieve real-time performance even on high-performance GPUs. Benefiting from fast rendering performance of 3DGS~\cite{kerbl20233d}, dynamic 3DGS has rapidly emerged and been applied to dynamic scene reconstruction. Four primary categories exist: (1) Deformable 3DGS~\cite{luiten2024dynamic, bae2025per, lu20243d, shaw2024swings, yang2024deformable, huang2024sc, zhao2024gaussianprediction, wansuperpoint, wu20244d, lu2024dn, duisterhof2023md, li2024spacetime, katsumata2025compact, lin2024gaussian, xu2024grid4d,yan20244d, zhu2024motiongs,kim20244d,kong2025efficient,lu2025bard,liang2025himor,wu2025swift4d,wang2025degauss,yoonsplinegs,wu2025localdygs,chen2025dash}, which constructs canonical 3D Gaussians and employs a deformation field~(e.g., MLPs, HexPlanes, or Hash Grids) to predict attribute residuals for each canonical Gaussian based on its position and timestamps. (2) 4D Gaussian representation~\cite{yang2023gs4d,duan20244d,yuan20251000+}, which incorporates a temporal dimension into 3D Gaussian attributes to create 4D Gaussian primitives. (3) Other foundation representations~\cite{lee2024fully,yoonsplinegs,wang2025freetimegs,li2025trace}. (4) Sparse dynamic scene reconstruction~\cite{wang2025monofusion}. However, these methods utilize a single model to represent an entire dynamic scene, making them unsuitable for streaming applications.

\subsection{Free-viewpoint Video Reconstruction}
Implicit neural representations become a new paradigm for reconstructing FVV from multi-view videos. StreamRF~\cite{li2022streaming} proposes a frame-by-frame training strategy, where an initial NeRF model is trained using the first frame, following by incremental training of sequential model residuals based on previous NeRF model. This method enables the streaming of model residuals rather than the entire NeRF model. Subsequent works, such as ReRF~\cite{wang2023neural}, VideoRF~\cite{wang2024videorf}, HPC~\cite{zheng2024hpc}, \cite{zhang2024rate}, FSVFG~\cite{yin2024fsvfg}, and VRVVC~\cite{hu2025vrvvc} further developed this training strategy, primarily to reduce the storage requirements.
With the advent of 3DGS, methods like 3DGStream~\cite{sun20243dgstream}, HiCoM~\cite{gao2024hicom}, QUEEN~\cite{girish2024queen}, OR2~\cite{yun2025compensating}, iFVC~\cite{tang2025compressing}, and 4DGC~\cite{hu20254dgc} extended the frame-by-frame training strategy to 3DGS for FVV reconstruction. 3DGStream~\cite{sun20243dgstream}, HiCoM~\cite{gao2024hicom}, OR2~\cite{yun2025compensating} and 4DGC~\cite{hu20254dgc} employ deformation fields to predict position and rotation residuals and deform previous Gaussians to the current timestamp, subsequently streaming deformation fields. IGS~\cite{yan2025instant} utilize a feed-forward model to predict position and rotation residuals. However, since only position and rotation attributes are varied per frame, they struggle to handle non-rigid deformation and the appearance of new objects in the following frames. In contrast, QUEEN~\cite{girish2024queen}, V$^3$~\cite{wang2024v}, H3D-DGS~\cite{he2024h3d}, and iFVC~\cite{tang2025compressing} learn all Gaussian attribute residuals but suffer from high frame size. Moreover, frame-by-frame training strategy inevitably leads to a continuous decline in performance due to accumulation errors. To mitigate this problem, TeTriRF~\cite{wu2024tetrirf}, V$^3$~\cite{wang2024v}, GIFStream~\cite{li2025gifstream}, 4DGCPro~\cite{zheng20254dgcpro}, and StreamSTGS~\cite{ke2025streamstgs} divide a video into multiple GOPs for training within each GOP to limit error accumulation. However, these methods require learning the corresponding temporal features per frame, resulting in significant temporal redundancy. Moreover, the above FVV representations do not support environment interaction, which diminishes the user's immersive experience. DLGStream distills language features into FVV to support many download tasks.

\subsection{3D Language Gaussian Field}
Current 3D language Gaussian representations~\cite{qin2024langsplat,zhou2024feature,jun2025dr,tian2025ccl,jiao2025clip} distill features from 2D foundation model into 3D Gaussians to enable 3D open-vocabulary queries. These methods typically use SAM to obtain objects, parts, and subparts within multi-view images, and then extract multi-level language features from CLIP~\cite{radford2021learning}. Subsequently, they train separate language Gaussian fields for each feature level. 4DLangSplat~\cite{li20254d} adopts this training method and further extends language Gaussian field into dynamic scene, thereby supporting time-related 4D open-vocabulary queries. However, 4DLangSplat is not directly suitable for language-embedded FVV, due to it greatly reduces FPS, increases frame size, and adversely affects reconstruction quality and training time. In contrast, DLGStream successfully overcomes these challenges through dual-opacity language Gaussian representation and interpolation-based deformation field.

\subsection{3D Gaussian Compression}
The 3DGS Compression methods comprises two main categories: (1) Constructing sparse 3DGS representation~\cite{lee2024compact,fan2024lightgaussian,niedermayr2024compressed,mallick2024taming,sun2024f,zhanglp} through pruning less important Gaussians, aided by attribute quantization and codebook compression. (2) Employing rate-distortion optimized entropy encoding models to encode the probability distributions of the features and attributes end-to-end during training ~\cite{girish2024eagles,chen2024hac,liu2024compgs,wang2024end,wang2024contextgs,zhan2025cat}. These studies are design for compressing static 3D Gaussians. DLGStream's compression of canonical Gaussian attribute is independent of its time feature compression, allowing canonical Gaussians to be compressed using different methods. This decoupled design means that advancements in static Gaussian compression can be integrated into DLGStream framework.

\section{Method}

\begin{figure}[t!]
   \centering
   \includegraphics[width=1\textwidth]{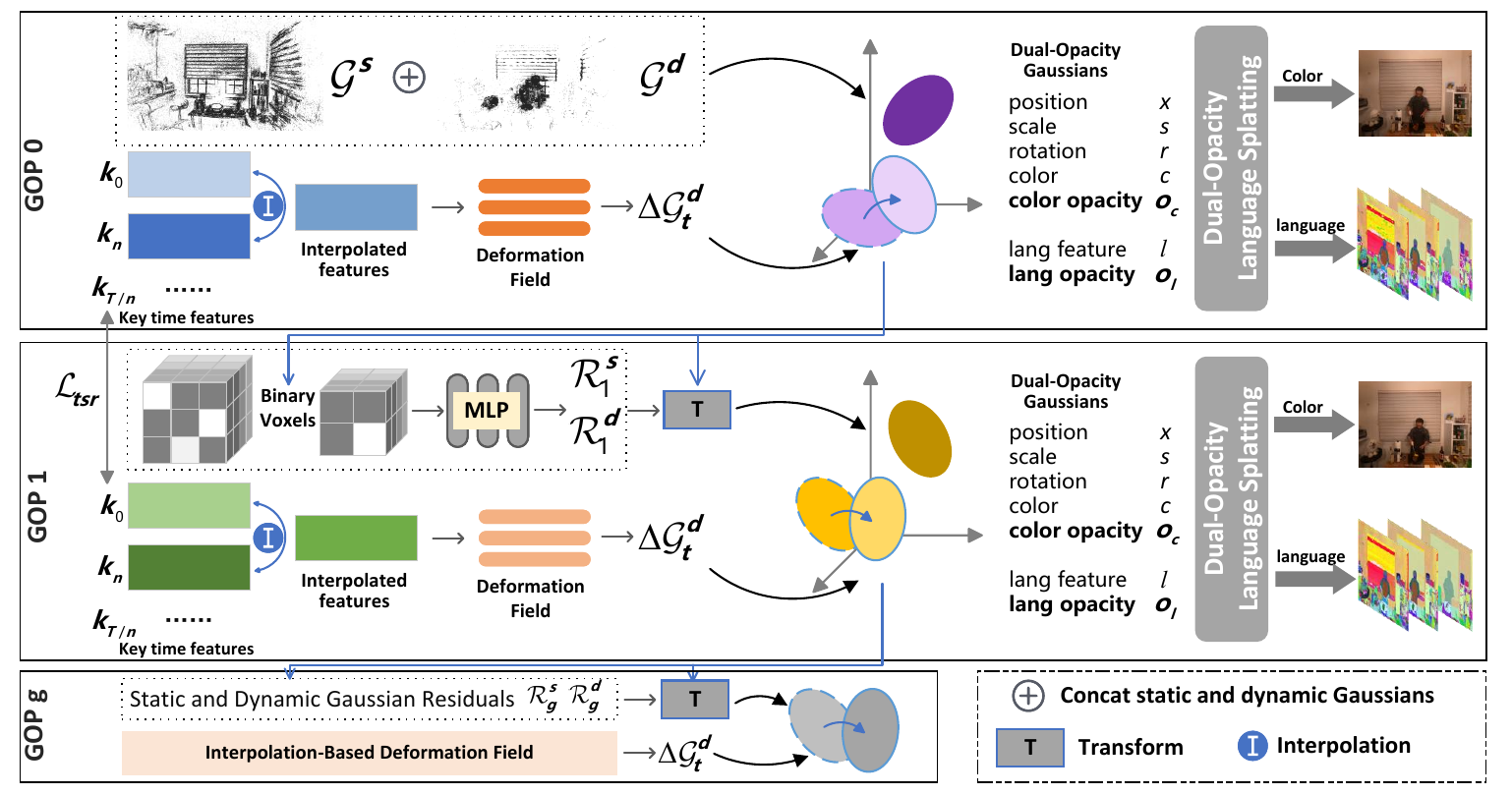}
   \caption{Overview of the DLGStream framework. We adopt a GOP-by-GOP training strategy. In the first GOP, static and dynamic Gaussians are decoupled, while in subsequent GOPs, the residuals of these Gaussians are learned from the previous GOP. Besides, we interpolate key time features and use a deformation field to predict deformed dynamic Gaussian attributes. Finally, dual-opacity language Gaussian splatting is applied to obtain color and language features in one rendering.}
   \label{fig:framework}
   \vspace{-3mm}
\end{figure}

Given multi-view videos, we split them into multiple GOPs to address the performance degradation from accumulated errors inherent in frame-by-frame training. Typically, real-world dynamic scenes captured by multiple cameras contain many static elements. Deforming these static elements is computationally inefficient and resource-intensive. Thus, we perform a static-dynamic 3D Gaussian decomposition, following Swift4D~\cite{wu2025swift4d}~(More details are provided in Appendix). This decomposition yields a static 3D Gaussian representation $\mathcal{G}^s$ for static elements, ensuring that only the dynamic 3D Gaussians $\mathcal{G}^d$ learn deformation. Then, we propose a dual-opacity language Gaussian representation in Section~\ref{sec:dual-opacity} for both static and dynamic Gaussians to facilitate the learning of language features for downstream tasks as shown in Fig.~\ref{fig:framework}. To reduce the temporal redundancy of existing FVV representations, we introduce an interpolation-based deformation field in Section~\ref{sec:deform}, which maintains only key time features for dynamic Gaussians while non-key time features are obtained via interpolation. Finally, we design a GOP-by-GOP training and compression strategy in Section~\ref{sec:gop} to further reduces frame size and eliminate inter-GOP flickering.

\subsection{Dual-Opacity Language Gaussian Splatting}
\label{sec:dual-opacity}

4DLangSplat~\cite{li20254d} constructs separate models for multi-level language features, which significantly impacts training time, frame size, and FPS, thereby degrading FVV performance. To address these limitations, we constructed a unified language field. Specifically, we employ 9-dimensional Gaussian language features $\mathcal{F}_l^s \in \mathbb{R}^9$ and $\mathcal{F}_l^d \in \mathbb{R}^9$ to learn all three language feature levels. The feature vector is partitioned such that the first, middle, and final three dimensions correspond to the three distinct language levels, respectively. This design needs only a single language Gaussian splatting operation, bypassing the need for three separate rendering processes and consequently achieving a higher FPS.

To reduce training time, we initially attempted to optimize language features jointly with color reconstruction. However, the physical attribute differences between colors and language features cause shared attributes to be optimized in different directions, leading to performance degradation both in open-vocabulary query accuracy and color rendering quality. Considering that position, rotation, and scaling only define the geometric distribution of the scene. However, opacity plays a decisive role in the rendering of color and linguistic features. Therefore, we learn two independent opacity attributes, $o^c \in  \mathcal{O}^c$ for color and $o^l \in \mathcal{O}^l$ for language features, which decouples the color and language feature computations during the alpha-blending stage of splatting, as defined below:
\begin{equation}
\label{alpha-bleding}
    R_c(x) = \sum_{m \in N} c_m o_{m}^c \prod_{j=1}^{m-1} (1 - o^c_j),
\end{equation}
\begin{equation}
    R_l(x) = \sum_{m \in N} f_m o_{m}^l \prod_{j=1}^{m-1} (1 - o^l_j),
\end{equation}
where $R_c$ and $R_l$ is rendered color and language feature images. $c_m$ and $f_m$ represent the color and language feature of Gaussian $m$, respectively. To prevent surface localization mismatches between color and language features, we use the color transparency accumulation truncation for the language feature computation. Moreover, the color and language features are optimized with separate gradients during backpropagation.

The language feature loss can be expressed as:
\begin{equation}
    \mathcal{L}_l = L1(R_l, concate(I_l^{small}, I_l^{mid}, I_l^{large}))
\end{equation}
where $concate$ is a tensor concatenation operation where we concatenate language feature labels of three levels together.

After training, we follow 4D open-vocabulary query method of 4DLangSplat, which first renders multi-level language feature images for each frames and then calculates the relevance score~\cite{kerr2023lerf} between these features images and a given text query to generate a segmentation mask for each frame.

\subsection{Interpolation-based Deformation Field}
\label{sec:deform}

Existing FVV reconstruction methods require learning corresponding time features for every frame. The incorporation of additional language feature attributes further escalates the frame size required per frame. Considering that objects in real-world videos exhibit continuous motion, video interpolation~\cite{jain2024video} utilizes this characteristic to reduce temporal redundancy. Inspired by video interpolation, we interpolate time features, thereby eliminating the need to learn time features for every individual frame and substantially reducing frame size.

Our interpolation-based deformation field maintains several key time features $\mathcal{K}=\{k_0, k_1, \cdots, k_{T/n}\}$, where $T$ represents the total number of frames in a GOP and $n$ denotes the interval. Given timestamp $t$, adjacent key time features $k_i$ and $k_{i+1}$ are obtained using the index $i=\lfloor \frac{t}{n} \rfloor$. Then, we utilize linear interpolator to obtain non-key time feature $k_t = Lerp(k_i, k_{i+1},t)$. For each dynamic Gaussian attribute $a^d \in \mathcal{G}^d$\footnote{For 3DGS, $\mathcal{G}$=\{Position, Scale, Rotation, Opacity, and Color\}, while for Scaffold-GS, $\mathcal{G}$=\{Position, Scale, Feature, Offset\}.}, we utilize Gaussian attribute predictors $\mathcal{D}_{a}$ to predict corresponding attribute offset $\Delta a^d_t \in \Delta \mathcal{G}^d_t$:
\begin{equation}
\Delta a^d_T = \mathcal{D}_{a}(k_t), \quad a_t^d = a^d + \Delta a^d_t.
\end{equation}
As a result, our interpolation-based deformation field significantly reduces storage requirements by storing $T/n$ key temporal features instead of the full $T$.

To enable real-time streaming, an effective time feature compression method is essential. Although several compression frameworks have been developed for encoding static 3D Gaussians, real-time streaming requires the consideration of both frame size and decoding latency. Recent context-aware entropy encoding methods~(e.g., HAC~\cite{chen2024hac}) offer remarkable compression efficiency, but their long decoding latency makes them unsuitable for this application. Thus, we propose to utilize video codec to encode time features within GOPs.

We organize the time features $\mathcal{K}$ into a sequence of 2D images and encode them as a feature video using video codecs~(e.g., H.264 and HEVC). These video codecs are highly optimized, facilitating real-time encoding and decoding on a wide range of devices. 
The frame size of the feature video is highly determined by the similarity between adjacent time features. Because of the temporal continuity of object motion in real-world videos, time features should also have similar continuity. To enforce this, we introduce temporal regularization $\mathcal{L}_{tsr}$ as follows:
\begin{equation}
\mathcal{L}_{tsr} = L1(k_i, k_{i+1}), \ i=\left\lfloor t/n \right\rfloor.
\end{equation}

To balance storage and performance, the GOP size is set to $60$, and the key time feature interval is set to $n=10$. Moreover, time features are encoded as a video using libx265 with a quality parameter $crf=6$.

\subsection{GOP-by-GOP training and Compression}
\label{sec:gop}

We observe that previous GOP-based FVV training methods~\cite{wu2024tetrirf,wang2024v,li2025gifstream,ke2025streamstgs} exhibit noticeable flickering across different GOPs. This artifact arises because the Gaussian color parameters learned by these methods represent the average color within a single GOP. Since these average values differ across GOPs and no temporal correlation constraints are enforced between them, inconsistent brightness becomes apparent. To solve this problem, we leverage the Gaussian attributes of the previous GOP $g-1$ as initialization rather than training from scratch. We optimize Gaussian attribute residuals of the static and dynamic Gaussians from the previous GOP $g-1$:
\begin{equation}
\mathcal{G}^s_{g} = \mathcal{G}^s_{g-1} + \mathcal{R}^s_{g}, \quad \mathcal{G}^d_{g} = \mathcal{G}^d_{g-1} + \mathcal{R}^d_{g},
\end{equation}
where $\mathcal{R}^s_{g}$ and $\mathcal{R}^s_{g}$ are predicted residuals for each attribute. The interpolation-based deformation field requires training from initialization. This GOP-by-GOP training strategy stores not the complete set of Gaussian attributes, but only their residuals, which further reduces the frame size. The core challenge, therefore, becomes the effective compression of these residual values. 

The residuals of Gaussian attributes are sensitive to numerical precision and quantization operation that can significantly degrade reconstruction quality. Motivated by BiRF~\cite{shin2023binary} that utilizes binary voxels with a tiny MLP to model static scene. The parameters of these binary voxels are constrained to values of either $+1$ or $-1$, rather than floating-point numbers, which achieves a substantial reduction in storage. Hence, we propose employing binary voxels to model the Gaussian attribute residuals for both static and dynamic canonical 3D Gaussians. Specifically, for a Gaussian at position $\mathcal{X}$, we obtain spatial feature $f_g$ from multi-level binary voxels $\mathcal{V}$:
\begin{equation}
    f_g = Sigmoid(interp(\mathcal{X}, \mathcal{V})),
\end{equation}
Then, a tiny MLP is utilized to predict the Gaussian attribute residuals $\mathcal{R}_g^s$ and $\mathcal{R}_g^d$ as follows:
\begin{equation}
    \mathcal{R}_g^s, \mathcal{R}_g^d = MLP(f_g).
\end{equation}

To reduce the size of binary voxels, we follow previous work~\cite{chen2024hac} to estimate bit consumption and minimize it:
\begin{equation}
    \mathcal{L}_e = M_{+}(-log(v_f)) + M_{-}(-log(1-v_f)),
\end{equation}
where, $v_f$ is the occurrence frequency of the parameter $+1$ in binary voxels. $M_+$ and $M_{-}$ is the total number of $+1$ and $-1$, respectively.

Since the uniform number of dynamic Gaussians across GOPs, we introduce an inter-GOP temporal regularization, which enables us to encode temporal features from different GOPs into a unified feature video, thereby further reducing data size.
\begin{equation}
\mathcal{L}_{tsr} = L1(k^{g}_{T/n}, k^{g-1}_0).
\end{equation}

\paragraph{Parallel GOP Training.} To accelerate training speed, the GOP-by-GOP training strategy can be adapted to facilitate parallel training. Specifically, after the first GOP is trained, subsequent GOPs can be trained based on the first GOP without waiting for the previous GOP to finish training, i.e., $\mathcal{G}^s_{g} = \mathcal{G}^s_{0} + \mathcal{R}^s_{g},\  \mathcal{G}^d_{g} = \mathcal{G}^d_{0} + \mathcal{R}^d_{g}$.

\subsection{Implementation}

The dual-opacity language Gaussian representation, interpolation-based deformation field, and GOP-by-GOP training strategy are agnostic to the fundamental 3D Gaussians representation and can therefore be integrated into various frameworks. To demonstrate this broad applicability, we implement our method using two prevalent representations: 3DGS~\cite{kerbl20233d} and Scaffold-GS~\cite{lu2024scaffold}. We empoly SOG~\cite{jain2024video} to compress 3DGS-based static and dynamic Gaussians, and HAC~\cite{chen2024hac} to compress Scaffold-GS based static and dynamic anchors.

Since 3DGS representation differs from Scaffold-GS representation, our dual-opacity language Gaussian implementation is tailored to each. For 3DGS, we augment both static and dynamic Gaussians by adding learnable language feature attributes $\mathcal{F}_l^s \in \mathbb{R}^9$ and $\mathcal{F}_l^d \in \mathbb{R}^9$ alongside their corresponding opacity attributes $\mathcal{O}_l^s$ and $\mathcal{O}_l^d$. Furthermore, the deformation field is extended with a tiny MLP to predict residuals for these language features. As for Scaffold-GS, we simply incorporate an additional attribute decoding MLP that takes anchor features as input and predicts Gaussian language features of dimension $[N \times k, 9]$. Simultaneously, the opacity decoding MLP is modified to output $[N \times k, 2]$ opacity attributes.

The implementation of the interpolation-based deformation field is the same for both 3DGS and Scaffold-GS. It is worth noting that deformation is not applied to the dynamic Gaussian attributes decoded from Scaffold-GS, but rather directly to the positions, features, scales, and offsets of the dynamic anchors . We have provided more details in the Appendix.

\paragraph{Loss.} When training the first GOP, the total loss is $\mathcal{L}=\alpha \mathcal{L}_{rgb} + (1 - \alpha) \mathcal{L}_{ssim} + \beta \mathcal{L}_{tsr} + \psi \mathcal{L}_l$. For subsequent GOPs, the binary regularization term $\mathcal{L}_e$ needs to be incorporated:
\begin{equation}
\mathcal{L} = \alpha \mathcal{L}_{rgb} + (1 - \alpha) \mathcal{L}_{ssim} + \beta \mathcal{L}_{tsr} + \psi \mathcal{L}_l + \gamma \mathcal{L}_e,
\end{equation}
where $\alpha$, $\beta$, $\psi$, and $\gamma$ are hyper-parameters. We have not shown the regularization terms introduced by SOG or HAC. They are only used to train the first GOP and are not needed for training subsequent GOPs.

\section{Experiment}
\subsection{Experimental Setup}
\textbf{Datasets}. We evaluate our method on three multi-view dynamic scene datasets: (1) \textbf{N3DV}~\cite{li2022nv3d} dataset have six scenes featuring large motions, dynamic illuminations, and specular highlights. These multi-view videos were captured by 18 to 21 cameras at a resolution of $2704 \times 2028$ with 30 FPS. Following prior work, we downsample the resolution to $1352 \times 1014$. (2) \textbf{MeetRoom}~\cite{li2022streaming} dataset, which includes three scenes captured by 12 cameras at a resolution of $1280 \times 720$ and a frame rate of 30. (3) \textbf{WideRange4D}~\cite{yang2025widerange4d} dataset is different from the two previous forward-facing datasets, as it provides a 360-degree capture of dynamic scenes with less overlap between camera views, thereby establishing a new evaluation benchmark. We selected two outdoor scenes, each comprising over 240 frames. For each scene, 60 camera views were captured, of which 59 were used for training and one for testing.


\begin{table}[t!]
\caption{mIoU(\%)/mAcc(\%) comparison of 4D open-vocabulary querying on the N3DV dataset.}
\label{tab:query-miou}
\centering
\resizebox{\textwidth}{!}{%
\begin{tabular}{c|ccccccc}
\toprule
Method           & Coffee Martini & Cook Spinach   & Cut Beef       & Flame Salmon     & Flame Steak       & Sear Steak       & Average \\ \midrule
LangSplat~\cite{qin2024langsplat}  & 68.0/\textbf{98.5}  & 78.3/\textbf{98.6}  & 36.5/97.0   & 66.0/82.2   & 64.1/97.8    & 78.3/98.6    & 61.49/91.9            \\
4DLangSplat~\cite{li20254d}   & 76.0/80.0    & 77.4/62.7   & 67.4/100   & 84.5/100   & 74.9/100    & 72.9/100       & 75.5/90.4            \\
Ours(3DGS)       & \textbf{87.8/80.0}    & \textbf{88.5/97.3}   & {\ul 85.0/100}  & \textbf{86.8/100}  & \textbf{87.0/100}  & \textbf{81.7/100}  & \textbf{85.8/96.2}      \\
Ours(ScaffoldGS) & {\ul 87.4/80.0} & {\ul 88.3/96.0} & \textbf{86.3/100} & {\ul 84.6/100}   & {\ul 84.3/100}     & {\ul 77.3/100}  & {\ul 84.7/ 96.0}   \\ \bottomrule
\end{tabular}%
}
\end{table}

\begin{table}[t]
\caption{Quantitative comparison of joint optimizing color and language features on N3DV dataset.}
\label{tab:query-rec}
\centering
\begin{tabular}{c|ccccc}
\toprule
Method           & PSNR $\uparrow$ & SSIM $\uparrow$  & LPIPS $\downarrow$ & \begin{tabular}[c]{@{}c@{}}Storage \\ (KB) $\downarrow$\end{tabular}  & FPS $\uparrow$ \\ \midrule
4DLangSplat~\cite{li20254d}      & 30.99 & 0.9351 & 0.1571 & 431   & 15                      \\
Ours(3DGS)       & \textbf{32.15} & {\ul 0.9442} & \textbf{0.1413} & {\ul 97.11} & \textbf{83}                       \\
Ours(ScaffoldGS) & {\ul 32.04} & \textbf{0.9446} & {\ul 0.1419} & \textbf{46.33} & {\ul 48}                       \\ \bottomrule
\end{tabular}%
\end{table}

\begin{figure*}[t!]
   \centering
   \includegraphics[width=0.95\textwidth]{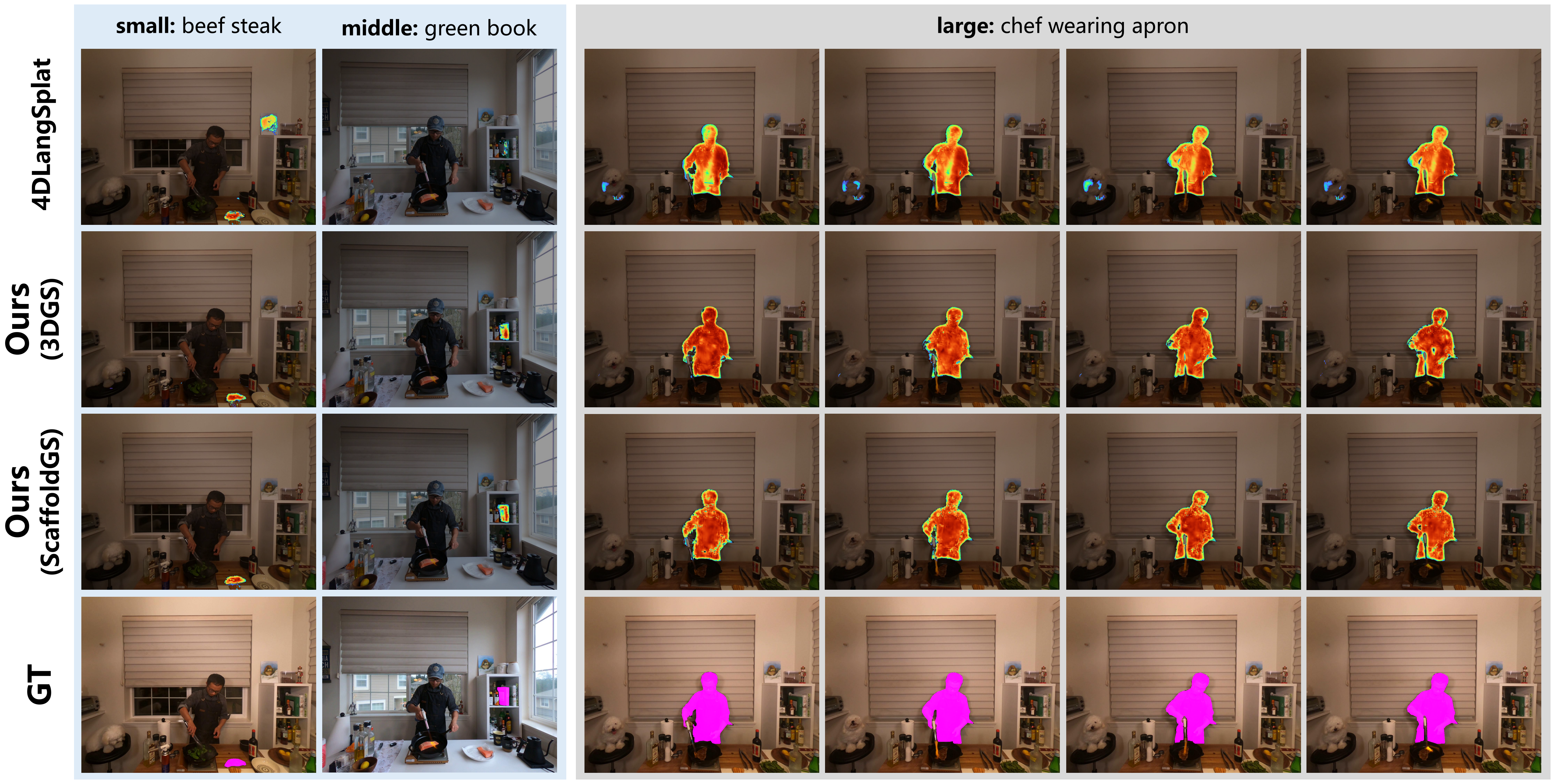}
   \caption{Qualitative comparisons of 4D open-vocabulary querying.}
   \label{fig:query}
   \vspace{-1mm}
\end{figure*}

\noindent \textbf{Baseline.} For 4D open-vocabulary querying task, we select 4DLangSplat~\cite{li20254d} and LangSplat~\cite{qin2024langsplat} as baselines for evaluation. As for FVV reconstruction task, we use two different 3D Gaussian representations~(i.e., 3DGS and Scaffold-GS), so we also divided the current benchmark into two categories. 3DGS-based FVV representations include TeTriRF~\cite{wu2024tetrirf}, 3DGStream~\cite{sun20243dgstream}, HiCoM~\cite{gao2024hicom}, QUEEN~\cite{girish2024queen}, OR2~\cite{yun2025compensating}, 4DGC~\cite{hu20254dgc}, and StreamSTGS~\cite{ke2025streamstgs}; Scaffold-GS-based FVV representations include GIFStream~\cite{li2025gifstream} and iFVC~\cite{tang2025compressing}. Considering that some methods utilize de-distorted images for training, while others do not. We follow the dataset preprocessing pipeline established by 4DGaussian~\cite{wu20244d} and re-trained several benchmarks using this pipeline to ensure a fair comparison.

\subsection{4D Open-vocabulary Querying Results}
\paragraph{Quantitative Comparisons}.
We perform 4D open-vocabulary querying on N3DV dataset, with mIoU(\%) and mAcc(\%) results detailed in Table~\ref{tab:query-miou}. Compared to 4DLangSplat, our 3DGS-based DLGStream improves mIoU by 10.3\%, and Scaffold-GS-based DLGStream achieves a 9.2\% improvement. As for mAcc metric, our DLGStream achieves a 6\% improvement. As shown in Table~\ref{tab:query-rec}, our method reduces the frame size by 10X and increases the FPS by 5X compared to 4DLangSplat. These enhancements primarily stem from our proposed dual-opacity language Gaussian representation, which effectively eliminates the interference between color and language features during joint training. Moreover, this representation enables the simultaneous learning of multi-level language features, thereby acquiring all feature maps in a single rendering pass.

\paragraph{Qualitative Comparisons}.
Fig.~\ref{fig:query} shows the segmentation results for the 4D open-vocabulary querying task. We queried objects of small, medium and large sizes across different dynamic scenarios. According to the qualitative results in Fig.~\ref{fig:query}, our method accurately segments the target objects, whereas 4DLangSplat either identifies non-target objects or fails to segment the target completely. More quantitative and qualitative comparison results are provided in the \textbf{Appendix}.

\begin{figure}[!t]
   \centering
   \subfloat[N3DV]{
      \includegraphics[width=0.48\textwidth]{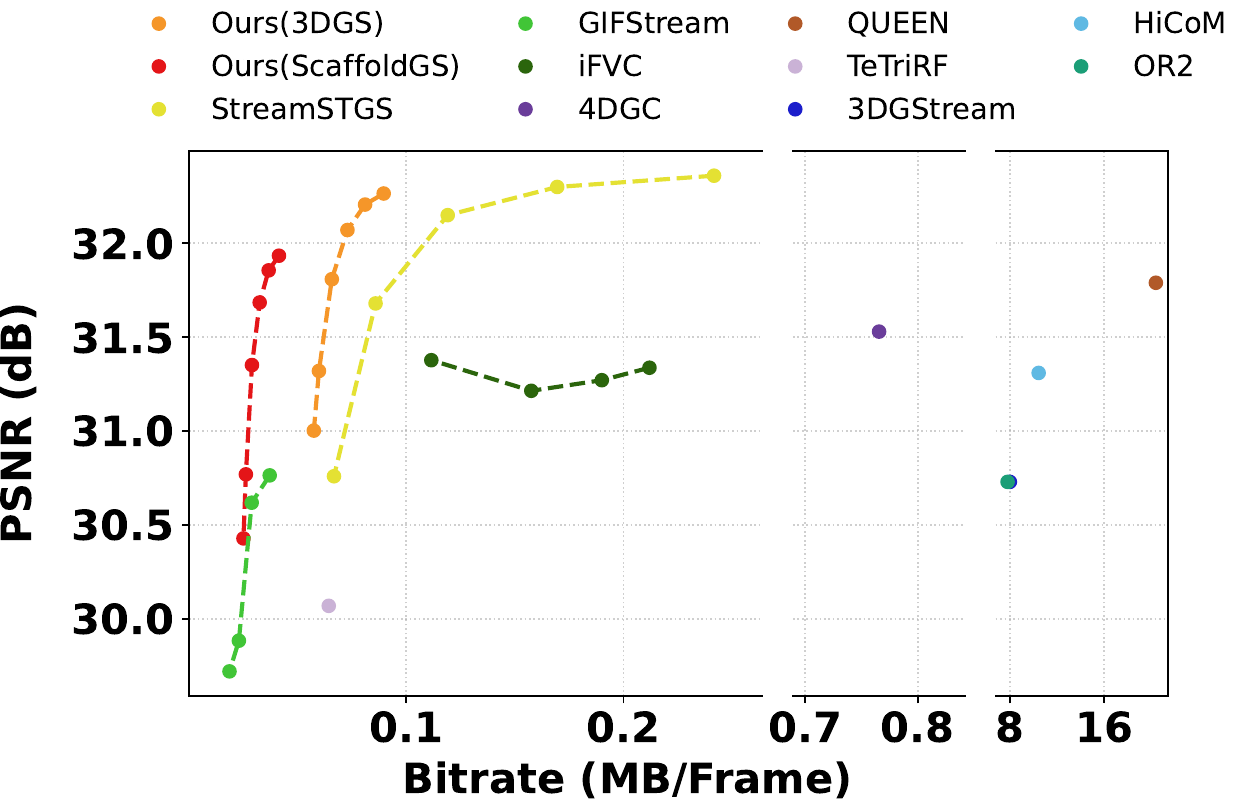}
   }
   \subfloat[MeetRoom]{
      \includegraphics[width=0.48\textwidth]{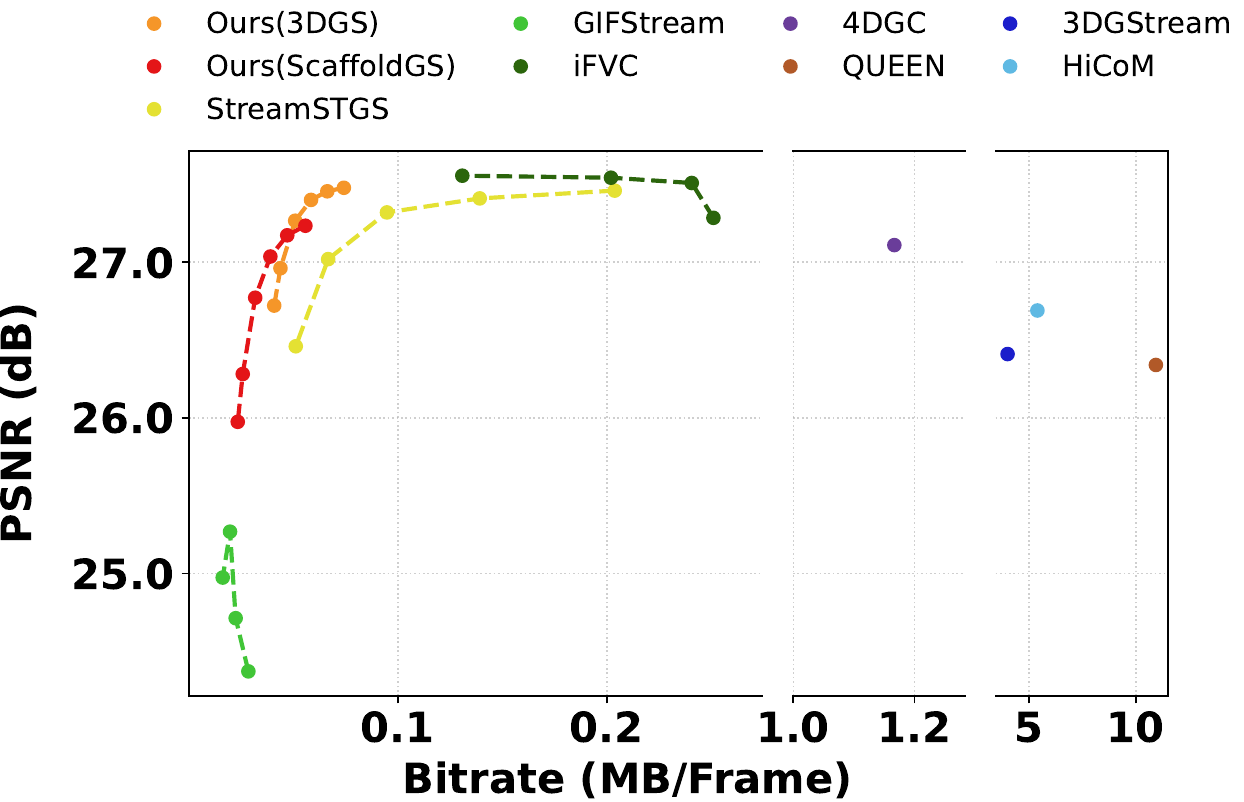}
   }
   \caption{RD-Cave results of N3DV and MeetRoom datasets.}
   \label{fig:rd}
   \vspace{-1mm}
\end{figure}

\begin{table*}[t!]
  \caption{Quantitative comparison of FVV reconstruction on N3DV dataset.}
  \label{tab:n3dv}
  \resizebox{\textwidth}{!}{%
  \begin{tabular}{ccccccccccc}
    \toprule
    \multirow{2}{*}{Method} & \multicolumn{5}{c}{\textbf{N3DV}} & \multicolumn{5}{c}{\textbf{MeetRoom}} \\ \cmidrule(lr){2-6} \cmidrule(lr){7-11}
           & PSNR $\uparrow$ & SSIM $\uparrow$ & LPIPS $\downarrow$
           & Storage $\downarrow$ & FPS $\uparrow$ 
           & PSNR $\uparrow$ & SSIM $\uparrow$ & LPIPS $\downarrow$
           & Storage $\downarrow$ & FPS $\uparrow$ \\ \midrule
    
    TeTriRF~\cite{wu2024tetrirf}
      & 30.07 & 0.9003 & 0.2986 & 65.89 &1.53
      & - & - & - & - & - \\ \midrule
        & \multicolumn{10}{c}{\textbf{3DGS-based representations}} \\
    3DGStream~\cite{sun20243dgstream}
      & 30.73 & 0.9348 & 0.1470 & 8205 & 72 
      & 26.41 & 0.8990 & 0.2370 & 4108 & 121.44 \\
    HiCoM~\cite{gao2024hicom}
      & 31.31 & 0.9390 & 0.1475 & 10704 & \cellcolor[HTML]{FFCE93}163 
      & 26.69 & 0.9035 & 0.2315 & 5535 & \cellcolor[HTML]{FFCE93}275 \\
    QUEEN~\cite{girish2024queen}
      & 31.79 & 0.9449 & \cellcolor[HTML]{FFFC9E}0.1415 & 700 & \cellcolor[HTML]{FF9E98}251 
      & 26.34 & \cellcolor[HTML]{FF9E98}0.9185 & \cellcolor[HTML]{FFFC9E}0.2164 & 11201 & \cellcolor[HTML]{FF9E98}497 \\
    4DGC~\cite{hu20254dgc}
      & 31.53 & 0.9412 & 0.1431 & 784 & 79 
      & \cellcolor[HTML]{FFFC9E}27.11 & 0.9093 & 0.2306 & \cellcolor[HTML]{FFFC9E}1195 & 110 \\
    OR2~\cite{yun2025compensating}
      & 30.73 & 0.9336 & 0.1577 & 8005 & \cellcolor[HTML]{FFFC9E}134
      & - & - & - & - & - \\
    4DGCPro~\cite{zheng20254dgcpro}
      & 31.64 & 0.9440 & -      & 655.4 & -
      & - & - & - & - & - \\
    Motion Matters~\cite{chen2026motion}
      & \cellcolor[HTML]{FFFC9E}32.12 & \cellcolor[HTML]{FFFC9E}0.9450 & \cellcolor[HTML]{FF9E98}0.1290 & \cellcolor[HTML]{FFCE93}109.6 & -
      & - & - & - & - & - \\
    ReCon-GS~\cite{fu2026recon}
      & 31.89 & \cellcolor[HTML]{FFCE93}0.9450 & 0.1410 & 450.6  & -
      & - & - & - & - & - \\
    StreamSTGS~\cite{ke2025streamstgs}
      & \cellcolor[HTML]{FF9E98}32.30 & 0.9436 & 0.1474 & \cellcolor[HTML]{FFFC9E}174 & 100 
      & \cellcolor[HTML]{FFCE93}27.41 & \cellcolor[HTML]{FFCE93}0.9181 & \cellcolor[HTML]{FFCE93}0.2157 & \cellcolor[HTML]{FFCE93}143 & 100 \\
    Ours(3DGS)
      & \cellcolor[HTML]{FFCE93}32.26 &
        \cellcolor[HTML]{FF9E98}0.9452 &
        \cellcolor[HTML]{FFCE93}0.1388 &
        \cellcolor[HTML]{FF9E98}92.04 & 
        107 
     & \cellcolor[HTML]{FF9E98}27.48 &
        \cellcolor[HTML]{FFFC9E}0.9179 &
        \cellcolor[HTML]{FF9E98}0.2080 &
        \cellcolor[HTML]{FF9E98}75.92 & 
        \cellcolor[HTML]{FFFC9E}166 \\ \midrule
    & \multicolumn{10}{c}{\textbf{Scaffold-GS-based representations}} \\

    GIFStream~\cite{li2025gifstream}
      & 30.76 & 0.9368 & 0.1470 & \cellcolor[HTML]{FF9E98}38.04 & \cellcolor[HTML]{FF9E98}97 
      & 24.37 & 0.8807 & 0.2387 & \cellcolor[HTML]{FF9E98}29.11 & \cellcolor[HTML]{FF9E98}126 \\
    iFVC~\cite{tang2025compressing}
      & \cellcolor[HTML]{FFCE93}31.38 & \cellcolor[HTML]{FFCE93}0.9419 & \cellcolor[HTML]{FF9E98}0.1357 & 114 & 14.65
      & \cellcolor[HTML]{FF9E98}27.29 & \cellcolor[HTML]{FFCE93}0.9120 & \cellcolor[HTML]{FF9E98}0.2036 & 170 & 17.90 \\
    Ours(ScaffoldGS)
      & \cellcolor[HTML]{FF9E98}31.93 &
        \cellcolor[HTML]{FF9E98}0.9448 &
        \cellcolor[HTML]{FFCE93}0.1408 &
        \cellcolor[HTML]{FFCE93}42.42 &
        \cellcolor[HTML]{FFCE93}88.27
      & \cellcolor[HTML]{FFCE93}27.28 &
        \cellcolor[HTML]{FF9E98}0.9172 &
        \cellcolor[HTML]{FFCE93}0.2059 &
        \cellcolor[HTML]{FFCE93}49.31 & 
        \cellcolor[HTML]{FFCE93}123\\

    \bottomrule
  \end{tabular}}
\end{table*}

\subsection{FVV Reconstruction Results}

As presented results in Table~\ref{tab:n3dv}, our method achieves superior FVV reconstruction performance, requiring only $92$KB and $42$KB for 3DGS-based  and Scaffold-GS-based representations, respectively. This performance is primarily attributed to our interpolation-based deformation field, which effectively reduces temporal redundancy, and GOP-by-GOP training strategy further reduces the redundancy of Gaussian spatial attributes. Although GIFStream has a smaller frame size than our method, the rate-distortion~(RD) curves in Fig.~\ref{fig:rd} demonstrate that our method achieves superior quality at similar bitrates. Moreover, GIFStream and iFVC use entropy encoding loss to control frame size, but it is not stable. Our method controls frame size through video codecs, resulting in better RD performance. We also evaluated our method on an outdoor dataset WideRange4D with 360-degree captures. As shown in Table~\ref{tab:widerange4d}, our method maintains SoTA FVV reconstruction performance. More results are provided in the \textbf{Appendix}.

\begin{table}[t!]
  \centering
  \caption{Quantitative comparison on WideRange4D dataset.}
  \label{tab:widerange4d}
  \begin{tabular}{cccccc}
    \toprule
    Method & PSNR $\uparrow$ & SSIM $\uparrow$ & LPIPS $\downarrow$
           & \begin{tabular}[c]{@{}c@{}}Storage\\(KB) $\downarrow$\end{tabular} & FPS $\uparrow$ 
           
           \\
    \midrule

    iFVC~\cite{tang2025compressing}& 27.06& 0.7960& 0.2647 &263.80 &37.10 \\
    
    Ours
     & 27.58 &0.7975 &0.2814 &226.66 &36.68
    \\
    \bottomrule
  \end{tabular}
\end{table}

\begin{table}[t!]
\caption{Quantitative results of frame interpolation on N3DV dataset. We skip frames during training but evaluate all frames to simulate frame interpolation.}
\label{tab:flerp}
\centering
\begin{tabular}{c|c|ccccc}
\toprule
Method                      & Skipped frames & mIoU$\uparrow$ &mAcc$\uparrow$  & PSNR$\uparrow$  & SSIM$\uparrow$   & Storage$\downarrow$  \\ \midrule
\multirow{5}{*}{Ours(3DGS)} & 1 (15 FPS)     & 83.51 & 94.63 & 32.04 & 0.943 & 79.19 \\
                            & 2 (10 FPS)     & 84.72 & 94.54 & 32.24 & 0.944 & 100.8  \\
                            & 4 (6 FPS)      & 83.79 & 95.11 & 31.62 & 0.942  & 92.87 \\
                            & 9 (3 FPS)      & 82.64 & 95.58 & 30.94 & 0.936  & 90.01 \\
                            & 0 (30 FPS)     & 85.78 & 96.22 & 32.15 & 0.944 & 97.11  \\ \midrule
\multirow{5}{*}{Ours(ScaffoldGS)}    & 1 (15 FPS)     & 85.49 & 96.67 & 31.79 & 0.945 & 59.81 \\
                                     & 2 (10 FPS)     & 85.29 & 96.00 & 31.90 & 0.946 & 65.6  \\
                                     & 4 (6 FPS)      & 85.18 & 93.33 & 31.80 & 0.945  & 64.08 \\
                                     & 9 (3 FPS)      & 82.44 & 94.74 & 31.26 & 0.941  & 59.32 \\
                                     & 0 (30 FPS)     & 84.69 & 96.00 & 32.04 & 0.945 & 46.33  \\ 
\bottomrule
\end{tabular}%

\end{table}

\subsection{4D Frame Interpolation}
Based on our proposed interpolation-based deformation field, we investigate the 4D frame interpolation task. This task is simulated by excluding specific frames during the training. For example, when one frame is skipped, only odd-numbered frames are used for training. As presented in Table~\ref{tab:flerp}, our method achieves promising results when one or two frames are interpolated. Even with 9 frames skipped, namely conducting training at a rate of 3 FPS, competitive results were still attained. This outcome indicates that our interpolation-based deformation field effectively captures the temporal correlations. As a result, our method can be trained on sequences with a lower frame rate (e.g., 10 FPS), while higher frame rates (e.g., 30 or 60 FPS) can be achieved via interpolation during rendering. This approach also significantly reduces the costs associated with preparing pseudo language labels.

\subsection{Ablation Study}

\begin{figure}[!t]
\centering
    \subfloat[GT]{
       \includegraphics[width=0.31\columnwidth]{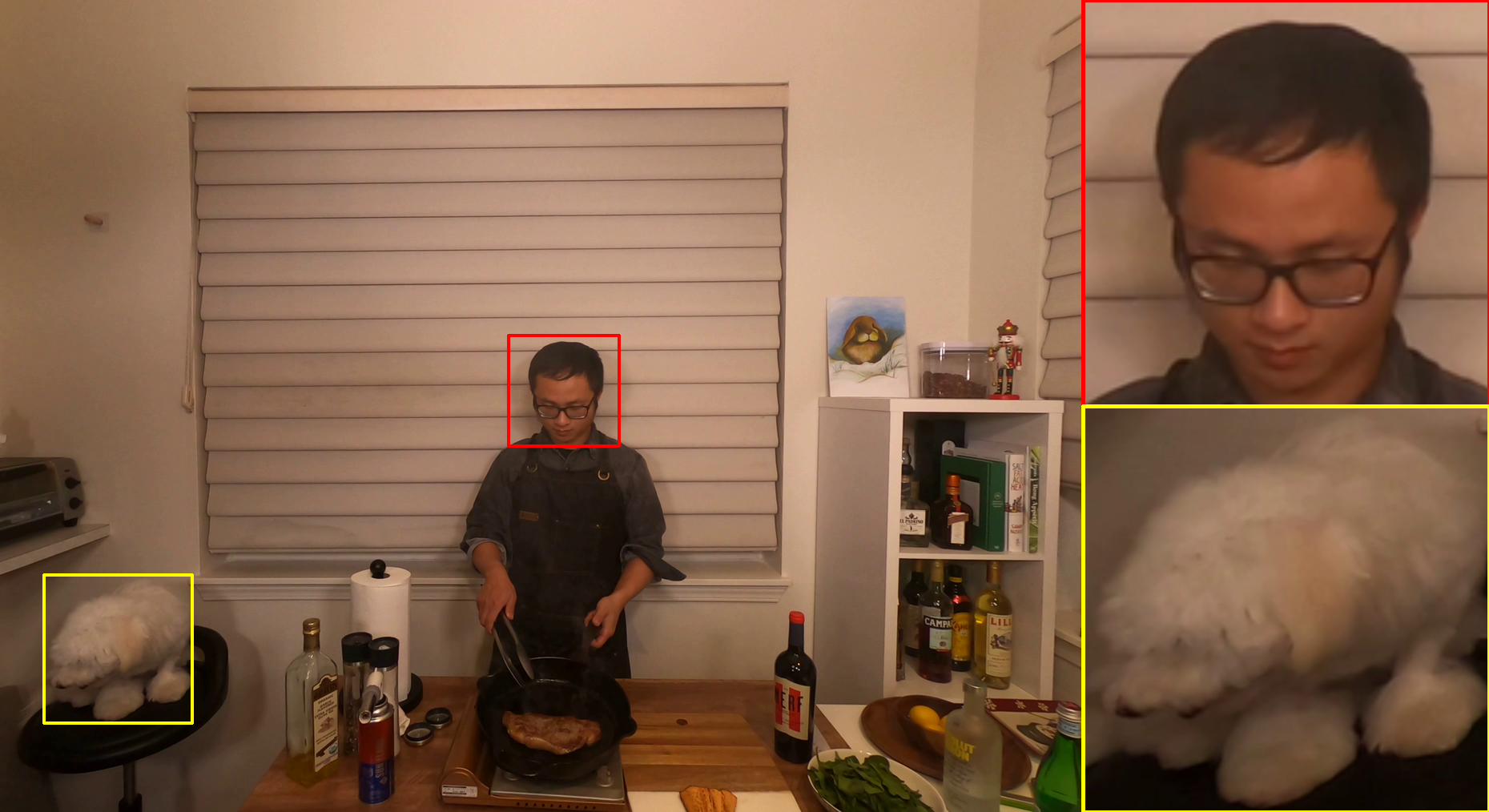}
    }
    \subfloat[Single Opacity]{
       \includegraphics[width=0.31\columnwidth]{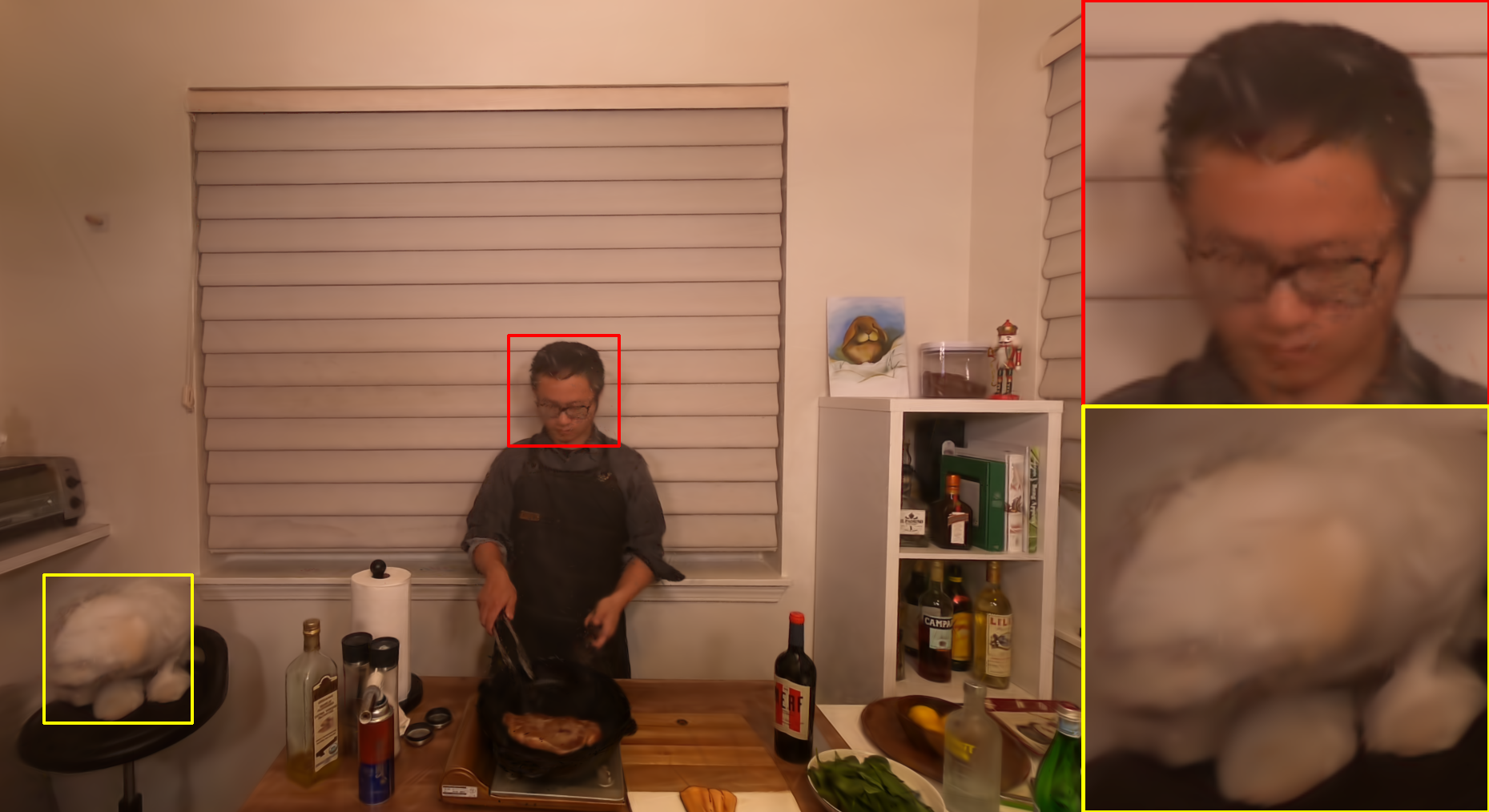}
    }
    \subfloat[Dual Opacity]{
       \includegraphics[width=0.31\columnwidth]{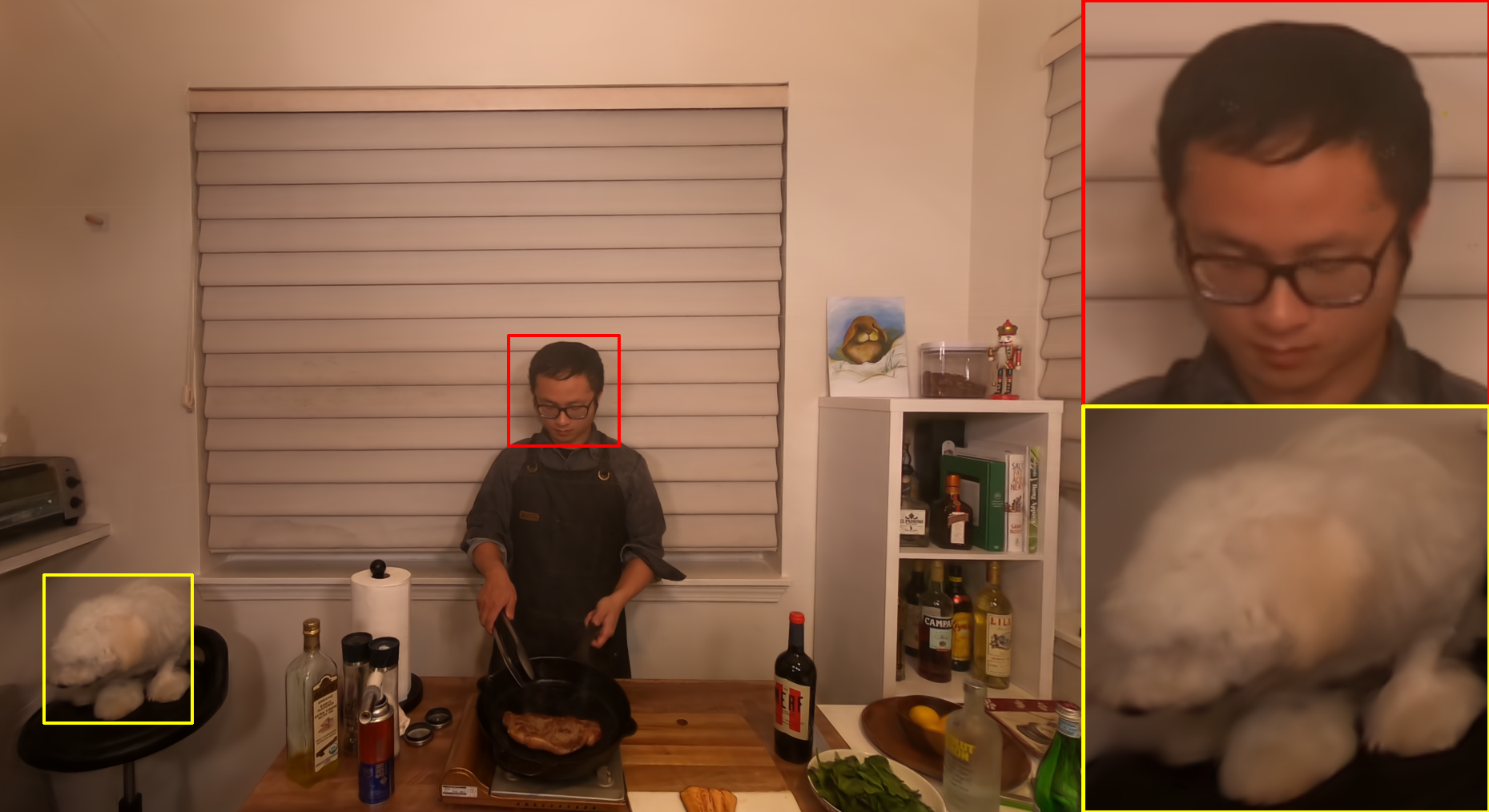}
    }
    \vspace{-2mm}
    \caption{Qualitative comparison of opacity ablation.}
   \label{fig:ablation-so}
\end{figure}

\textbf{Dual-opacity Gaussian representation.} We conducted an ablation study by removing the dual-opacity design, thereby sharing a single opacity between color and language features. The results, presented in Table~\ref{tab:ablation-so}, show a significant decrease in mIoU, mAcc, PSNR, and SSIM. Fig.~\ref{fig:ablation-so} shows noticeable artifacts in the dynamic regions when color and feature share the same opacity. This finding confirms that using separate opacity attributes for color and language features is essential for effective joint optimization.

\begin{table}[t]
\caption{Ablation results of dual-opacity Gaussian representation on the N3DV dataset. 'SO' indicates single opacity.}
\label{tab:ablation-so}
\centering
\begin{tabular}{c|ccccc}
\toprule
Method              & mIoU$\uparrow$ & mAcc$\uparrow$ & PSNR$\uparrow$ & SSIM$\uparrow$  & Storage$\downarrow$           \\ \midrule
Ours(3DGS)-SO       & 78.99          & 92.54          & 31.73          & 0.9423          & \textbf{96.25} \\
Ours(3DGS)          & \textbf{85.78} & \textbf{96.22} & \textbf{32.15} & \textbf{0.9442} & 97.11          \\ \midrule
Ours(ScaffoldGS)-SO & 79.62          & 94.67          & 31.88          & 0.9440          & 46.80          \\
Ours(ScaffoldGS)    & \textbf{84.69} & \textbf{96.00} & \textbf{32.04} & \textbf{0.9446} & \textbf{46.33} \\ \bottomrule
\end{tabular}%
\vspace{-2mm}
\end{table}

\begin{figure}[!t]
\centering
    \subfloat[Opacity of Cook Spinach]{
       \includegraphics[width=0.31\columnwidth]{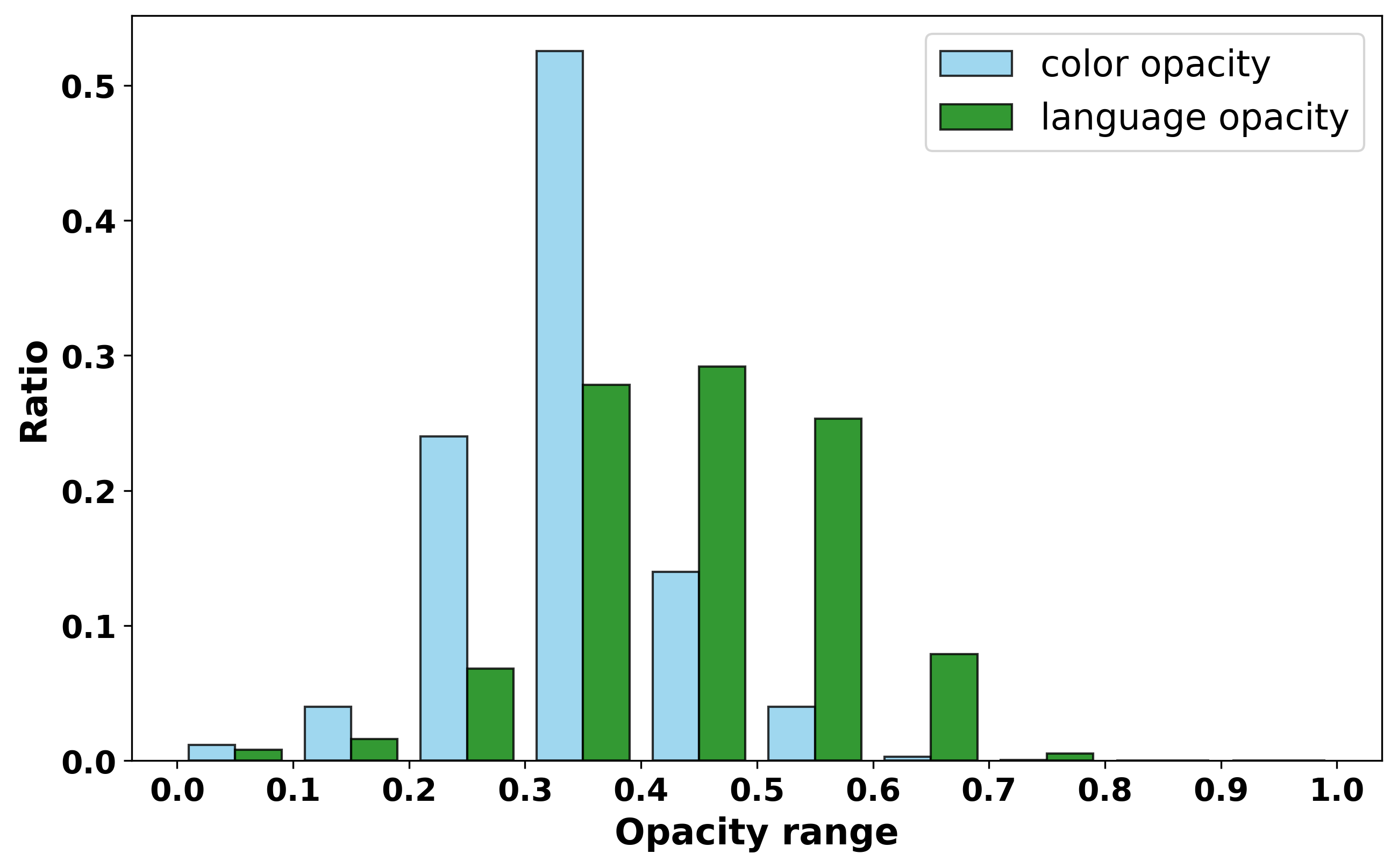}
    }
    \subfloat[Opacity of Flame Salmon]{
       \includegraphics[width=0.31\columnwidth]{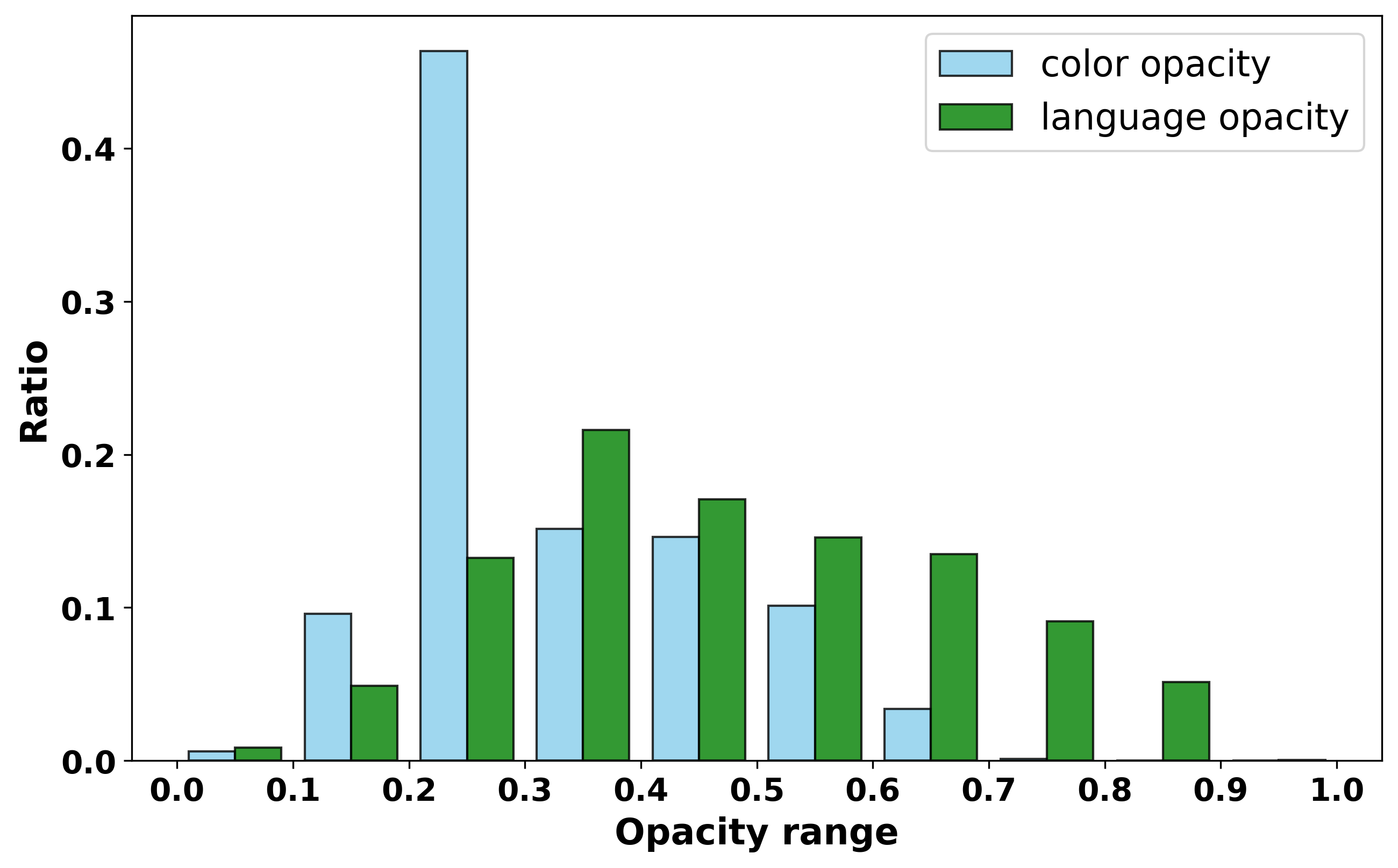}
    }
    \subfloat[Opacity of Sear Steak]{
       \includegraphics[width=0.31\columnwidth]{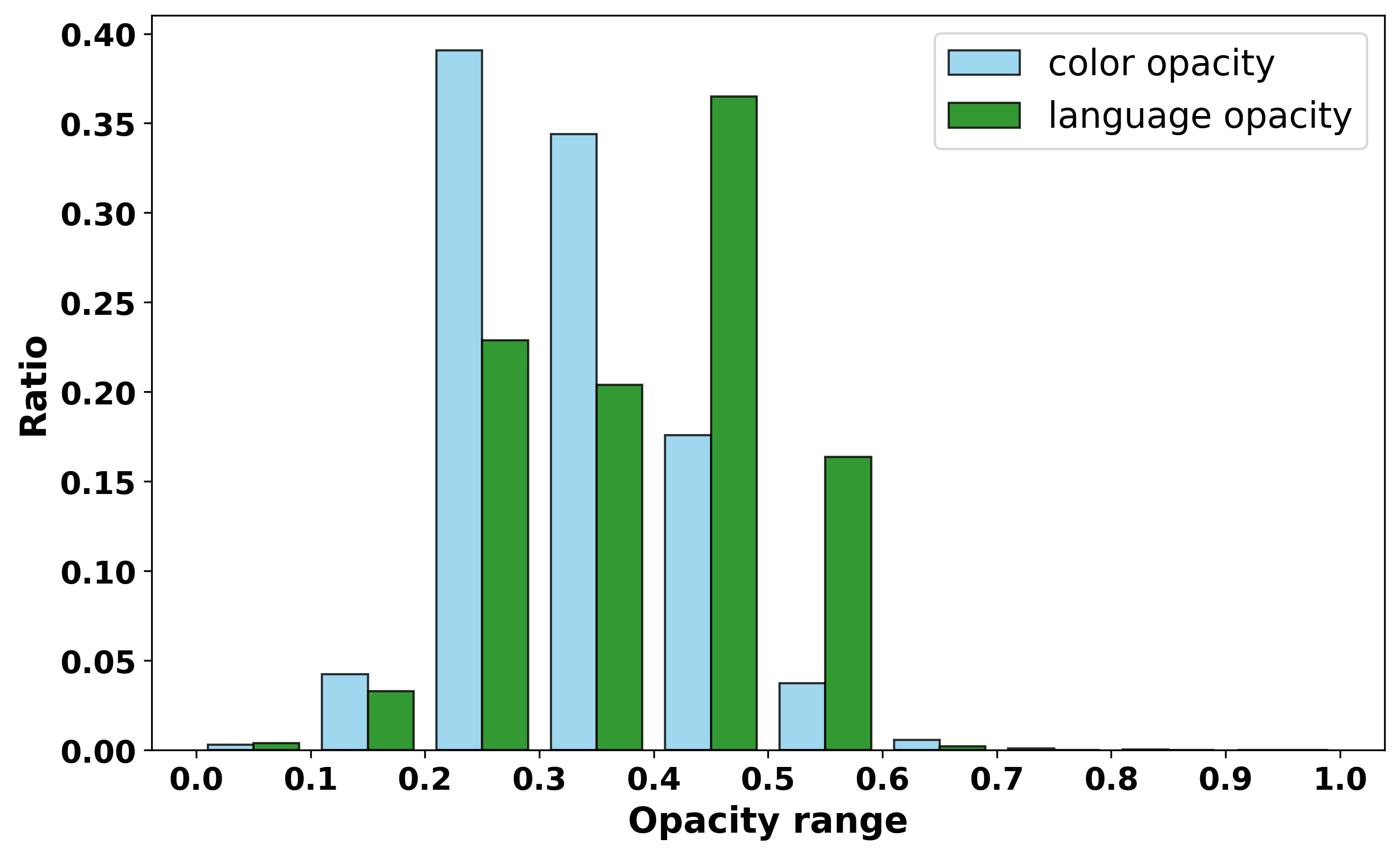}
    }
    \ 
    \subfloat[Opacity time difference of Cook Spinach]{
       \includegraphics[width=0.48\columnwidth]{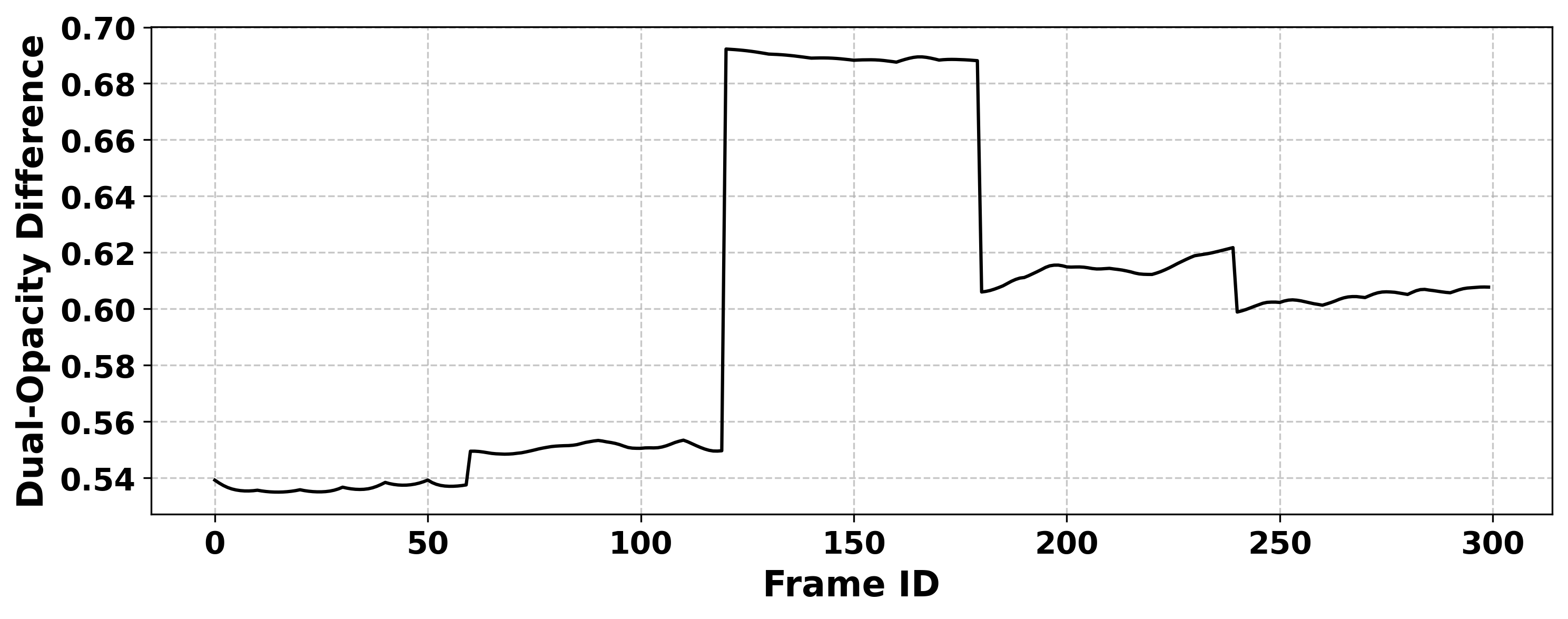}
    }
    \subfloat[Opacity time difference of Flame Salmon]{
       \includegraphics[width=0.48\columnwidth]{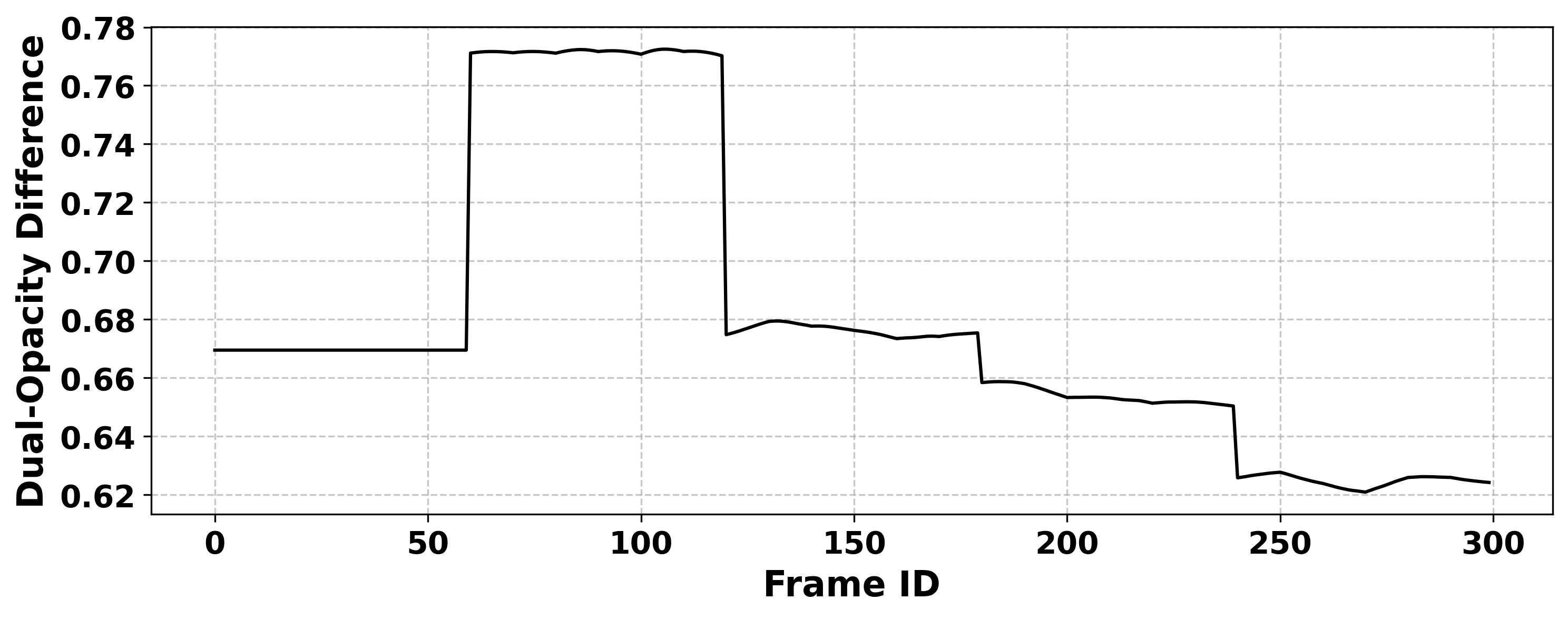}
    }
    \caption{Opacity statistical distributions of color and language feature.}
   \label{fig:dual-opacity}
   \vspace{-4mm}
\end{figure}

\noindent \textbf{Quantitative analysis of dual-opacity.} Fig.~\ref{fig:dual-opacity}(a)-(c) illustrate a distinct distributional discrepancy: feature opacity follows a Gaussian distribution centered at 0.5, whereas color opacity is centered at 0.3. This is because color requires lower opacity to represent the rich texture details of objects, while semantics of an object are consistent and do not exhibit translucency. The dynamic object in Fig.~\ref{fig:ablation-so}(b) appears overly smooth and lacks detailed textures, which confirms that sharing opacity can interfere with the learning of color and semantics. Fig.~\ref{fig:dual-opacity}(d) and (e) further depict the temporal evolution of per-Gaussian opacity differences, revealing substantial variations over time. These observations prove a significant physical distinction between color opacity and language feature opacity.

\noindent \textbf{Details of storage.} We present detailed information on the storage in Table~\ref{tab:dsize}. From the table, it can be seen that the temporal feature is only 20 to 30 KB, thanks to our proposed interpolation-based deformation field, which greatly reduces time redundancy. In addition, the Gaussian residual $R_g^s$ and $R_g^d$ between GOPs is also very small, only about 2KB. 
Note that Gaussian attributes still occupy a high proportion, and the advancement of static Gaussian compression methods can further reduce the storage.

\begin{table}[t!]
\caption{Details of average frame size (KB) in N3DV dataset.}
\label{tab:dsize}
\centering
\begin{tabular}{c|cccc}
\toprule
Method          & Attributes  & Temporal features  & MLPs  & $R_g^s$ \& $R_g^d$  \\ \midrule
Ours(3DGS)      & 64.03 & 29.33 & 0.3854  & 3.3595 \\ 
Ours(ScaffoldGS)& 22.53 & 21.18 & 0.7077  & 2.5405 \\ 
\bottomrule
\end{tabular}%
\vspace{-2mm}
\end{table}

\noindent \textbf{Ablation of interpolation method.} Table~\ref{tab:interpolate} shows bilinear interpolation is sufficient for real-world dynamic scenes. The complex bicubic yield a minimal gains, whereas nearest-neighbor performs poorly. 


\begin{table}[t!]
\caption{Ablation of interpolation method in N3DV dataset.}
\label{tab:interpolate}
\centering
\begin{tabular}{c|ccccc}
\toprule
Interpolation method          & PSNR$\uparrow$  & SSIM$\uparrow$  & LPIPS$\downarrow$  & Storage$\downarrow$   & Size of Temp. features$\downarrow$   \\ \midrule
Nearest                       & 31.35 & 0.943 & 0.143  & 66.48  & 37.59  \\
Bicubic                       & 32.20 & 0.946 & 0.136  & 80.24  & 53.12 \\
Bilinear(Ours)                & 31.93 & 0.945 & 0.141  & 42.42  & 19.52 \\ 
\bottomrule
\end{tabular}%
\vspace{-2mm}
\end{table}

\begin{table}[t!]
\caption{Ablation of intra-GOP Guassian residual learning.}
\centering
\begin{tabular}{c|ccccc}
\toprule
Configure              & PSNR$\uparrow$  & SSIM$\uparrow$  & LPIPS$\downarrow$  & Storage$\downarrow$    & Size of $R_g^s$\&$R_g^d$$\downarrow$  \\ \midrule
MLP$_{32\times2}$             & 31.92 & 0.945 & 0.141  & 42.15   & 2.18  \\
MLP$_{128\times2}$            & 31.93 & 0.945 & 0.141  & 42.74   & 2.85  \\
Voxel$_{256 to 2048}$  & 31.93 & 0.945 & 0.141  & 42.12   & 2.12      \\
Voxel$_{1024 to 8192}$ & 31.94 & 0.945 & 0.141  & 42.62   & 2.85      \\
Default/Ours(ScaffoldGS)   & 31.93 & 0.945 & 0.141  & 42.42   & 2.54  \\ 
\bottomrule
\end{tabular}%
\label{tab:gop_res}
\vspace{-6mm}
\end{table}

\noindent \textbf{Ablation of Gaussian residual learning.} The default voxels setting are [512, 1024, 2048, 4096] alongside 2-layers MLP with 64 neurons per layer. Table~\ref{tab:gop_res} demonstrates that scaling the voxel grids or MLP architecture yields only marginal differences in performance.

\section{Conclusion}
We propose DLGStream, a novel language-embedded streamable FVV representation that supports 4D open-vocabulary querying, 4D video interpolation, and so on. Our dual-opacity dynamic language Gaussian splatting efficiently solve the performance degradation during the joint optimization of color and language features. Moreover, we introduce an interpolation-based deformation field that stores only key time features and interpolates non-key time features, thereby significantly reducing the frame size. Finally, we develop GOP-by-GOP training and compression to further reduce frame size while simultaneously eliminate flickering artifacts across GOPs.


\section*{Acknowledgements}
This work is supported by the Beijing-Tianjin-Hebei Basic Research Cooperation Project (No. 24JCZXJC00050).

%
%
\bibliographystyle{splncs04}
\bibliography{main}

\section*{Appendix}
\subsection{Overview}
We provide more implementation details of static-dynamic decomposition and interpolation-based deformation field in Section~\ref{implementation}. Then, in Section~\ref{sec:long}, we demonstrate the performance of our method in reconstructing long video sequences. In Section~\ref{sec:parallel}, we provide experimental results for parallel training. In Section~\ref{sec:deforming}, we explained why we chose to deform the anchor features of Scaffold-GS instead of the decoded Gaussian attributes, because the performance was poor. In Section~\ref{appendix-experiment}, we show more ablation studies and baseline comparisons. In Section~\ref{appendix-qualitative}, we demonstrate the qualitative comparison results of all scenarios across three datasets.

\subsection{More Implementation Details}
\label{implementation}
\paragraph{Static-Dynamic Decomposition}
To obtain static-dynamic pixel labels, we first compute the temporal variance of the pixels of all frames for each camera. If the variance larger than a given threshold~(default is $0.03$), the pixel is dynamic, otherwise, the pixel is static. Thereby, we can obtain the dynamic map $M$ of all cameras. Then, we adopt method proposed by Swift4D~\cite{wu2025swift4d} to decompose 3D Gaussians into static and dynamic 3D Gaussians. Specifically, a learnable dynamic indictor $d$ is introduced for each 3D Gaussians to replace color $c$. The predicted dynamic map $\hat{M}$ can be obtained as follows:
\begin{equation}
\hat{M}(x) = Sigmoid\left( \sum_{m \in N} d_m \alpha_m(x) \prod_{j=1}^{m-1} (1 -\alpha_j(x)) \right)
\end{equation}
Then, we can use pseudo ground truth $M$ to supervise predicted dynamic map $\hat{M}$. Note that during training, all other Gaussian attributes are fixed. A binary cross-entropy loss is introduced to encourage the pixel value in $\hat{M}$ to be close to zero or one, thereby optimizing the dynamic indictor $d$ to span $(-\infty, +\infty)$. If $d$ exceeds a threshold $\phi$, these Gaussians are classified as dynamic Gaussian $\mathcal{G}^d$, otherwise, they are considered static Gaussians $\mathcal{G}^s$. The binary cross-entropy loss can be defined as:
\begin{equation}
\mathcal{L} = \mathbb{E}\left[ -M(x)\log(\hat{M}(x)) - (1-M(x))\log(1- \hat{M}(x)) \right]
\end{equation}
This static-dynamic decomposition training stage needs 3000 iterations and costs about 70 seconds. After training, if a Gaussian with dynamic indictor $d$ larger than 6, this Gaussian is dynamic Gaussian, otherwise is static Gaussian.

\paragraph{Interpolation-based Deformation field} The channel number of time features is set to $16$, and their dimensions are reshaped from $W \times H \times 16$ to $4W \times 4H$ for compression. Gaussian attribute decoders, i.e., $D_p$, $D_{cov}$, $D_o$, and $D_c$ for Ours(3DGS) and $D_p$, $D_{feat}$, $D_{offset}$, and $D_{scale}$ for Ours(Scaffold-GS), consists of two linear layers, each containing 64 neurons. Bias are disabled for all MLPs. The learning rate for the static and dynamic Gaussians is same to that used in the original 3DGS. The learning rate of the time features is $0.001$ with continuous learning rate decay to $0.00001$. The learning rates of Ours(3DGS) for the deformation field networks are as follows: $D_p$ from $0.005$ to $0.00005$, $D_{cov}$ at $0.04$, $D_o$ from $0.002$ to $0.00002$, and $D_c$ from $0.008$ to $0.00005$. The learning rates of Ours(Scaffold-GS) for the deformation field networks are as follows: $D_p$ from $0.005$ to $0.00005$, $D_{feat}$ from $0.004$ to $0.0004$, $D_{offset}$ from $0.005$ to $0.00005$, and $D_{scale}$ from $0.005$ to $0.00005$.

\subsection{Long duration video reconstruction}
\label{sec:long}
\begin{figure}[!t]
   \centering

   \subfloat[1st frame]{
      \includegraphics[width=0.24\textwidth]{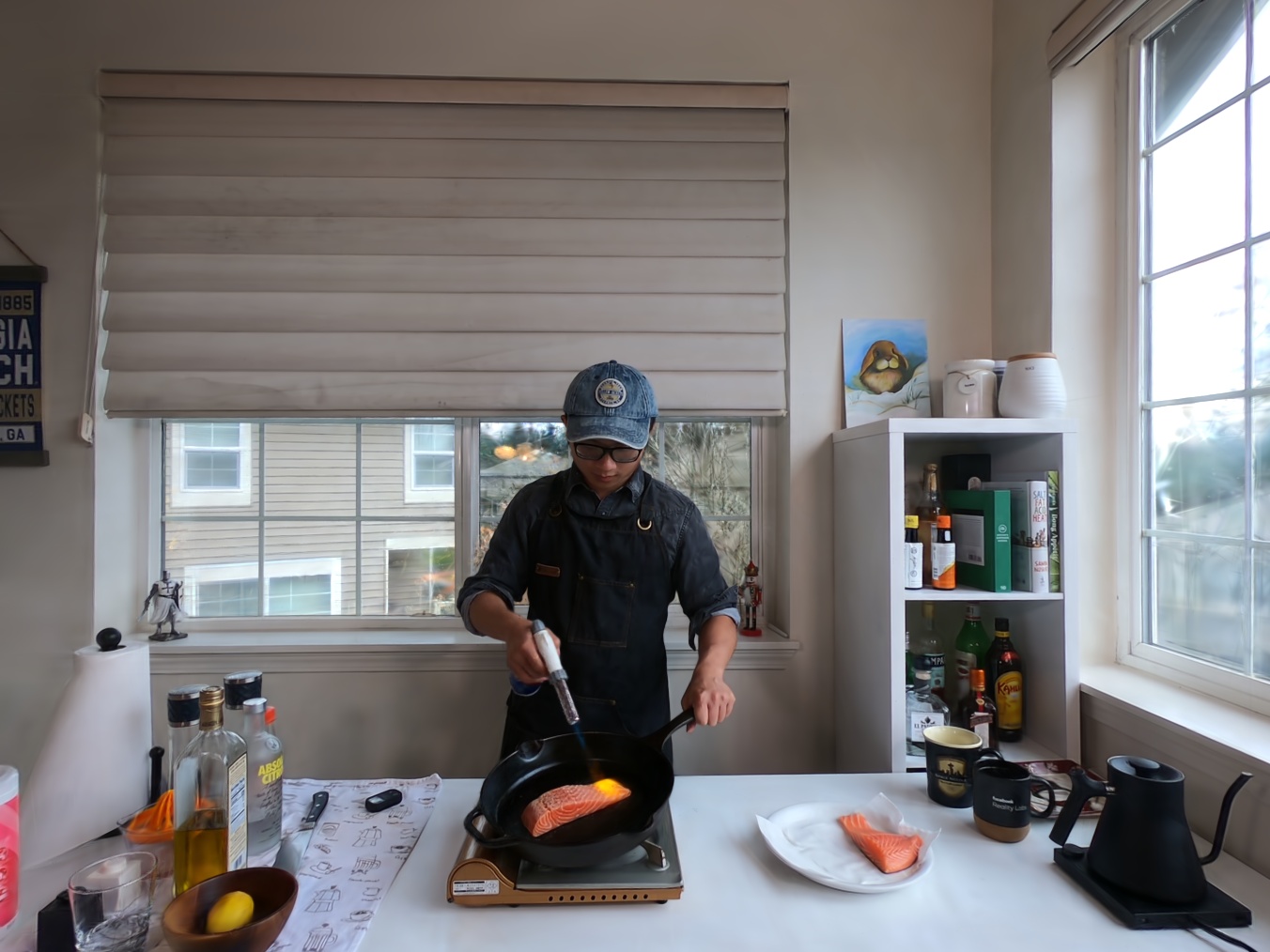}
   }
   \subfloat[155th frame]{
      \includegraphics[width=0.24\textwidth]{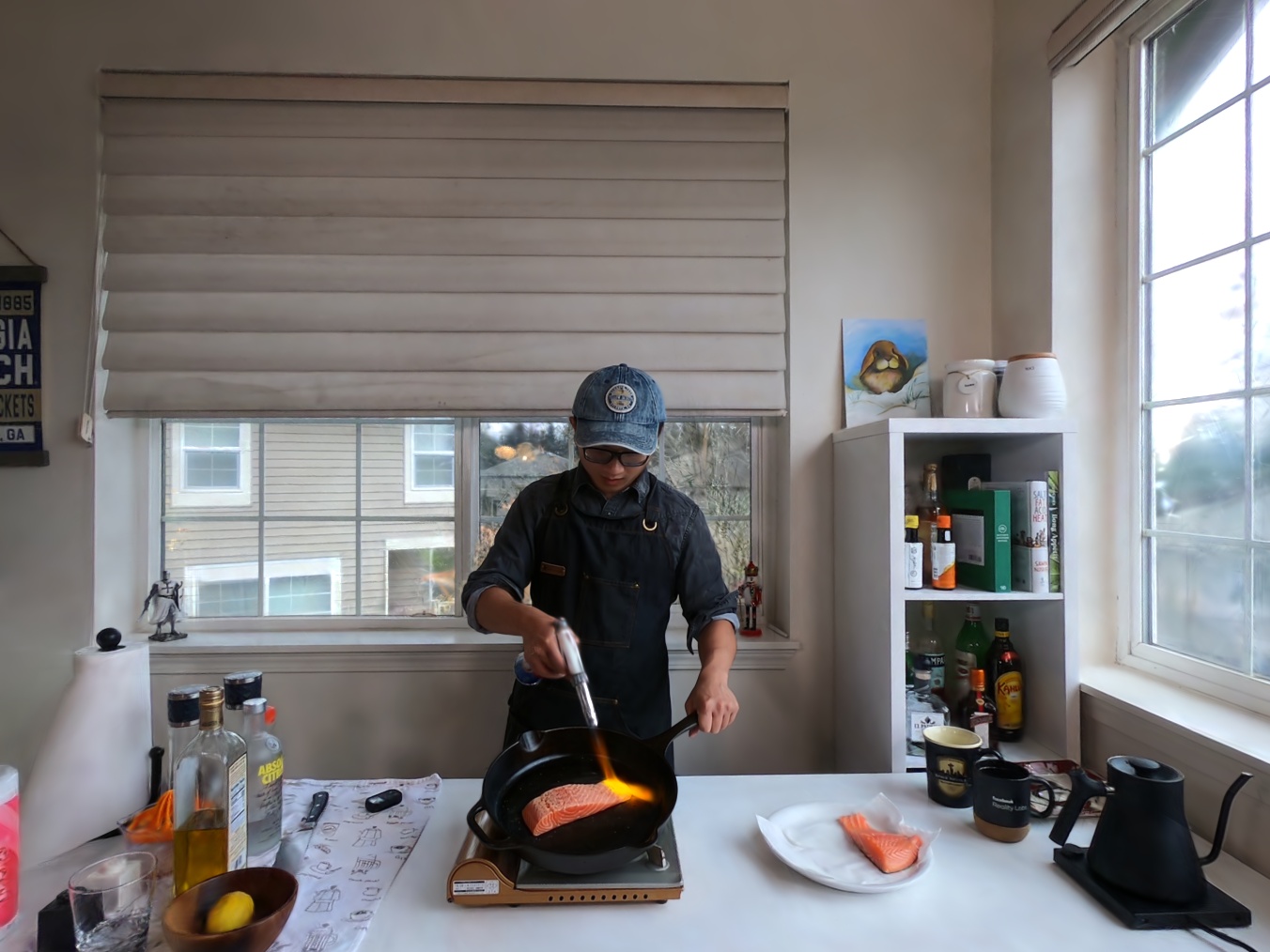}
   }
   \subfloat[299th frame]{
      \includegraphics[width=0.24\textwidth]{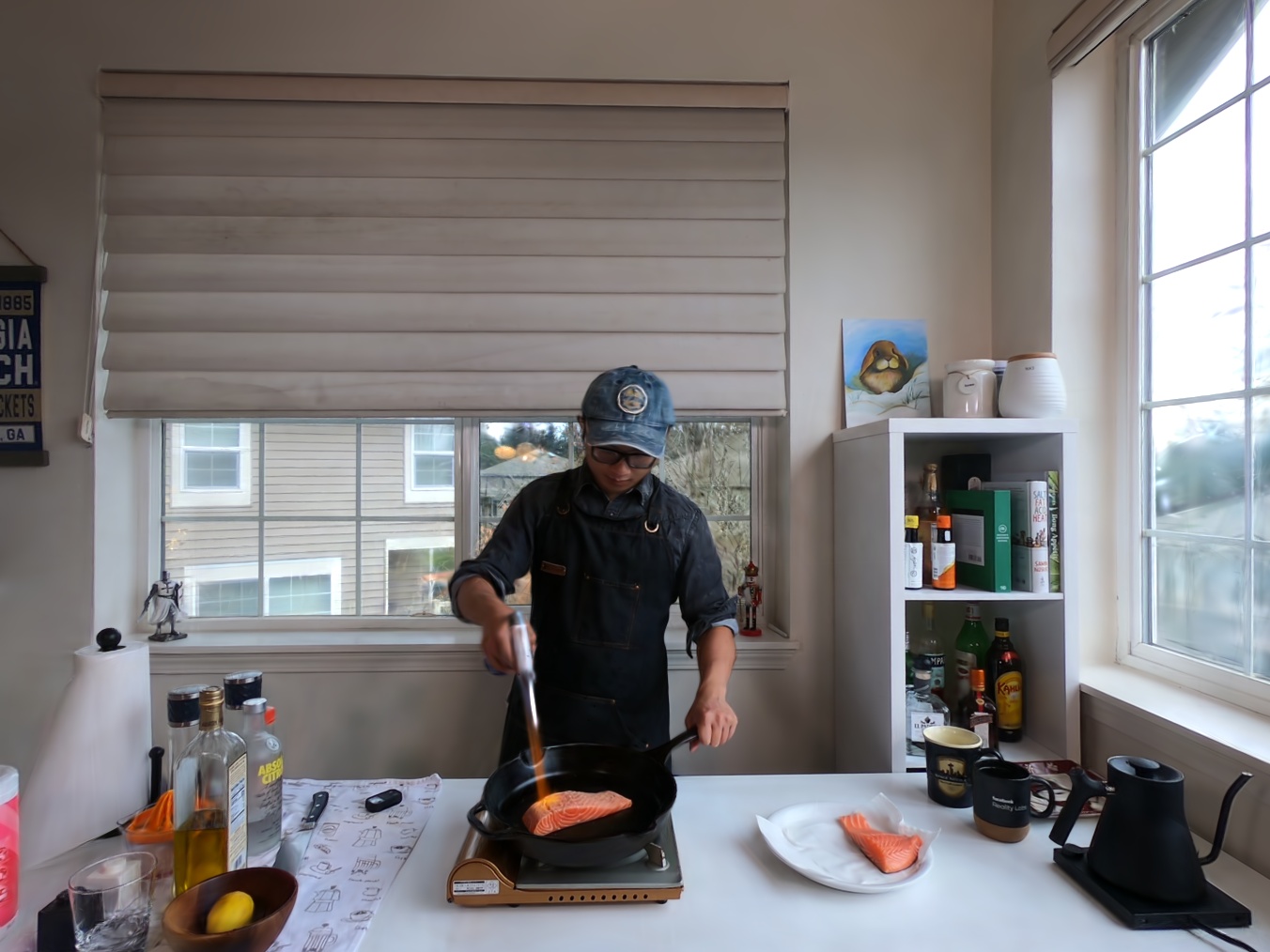}
   } 
    \subfloat[520th frame]{
      \includegraphics[width=0.24\textwidth]{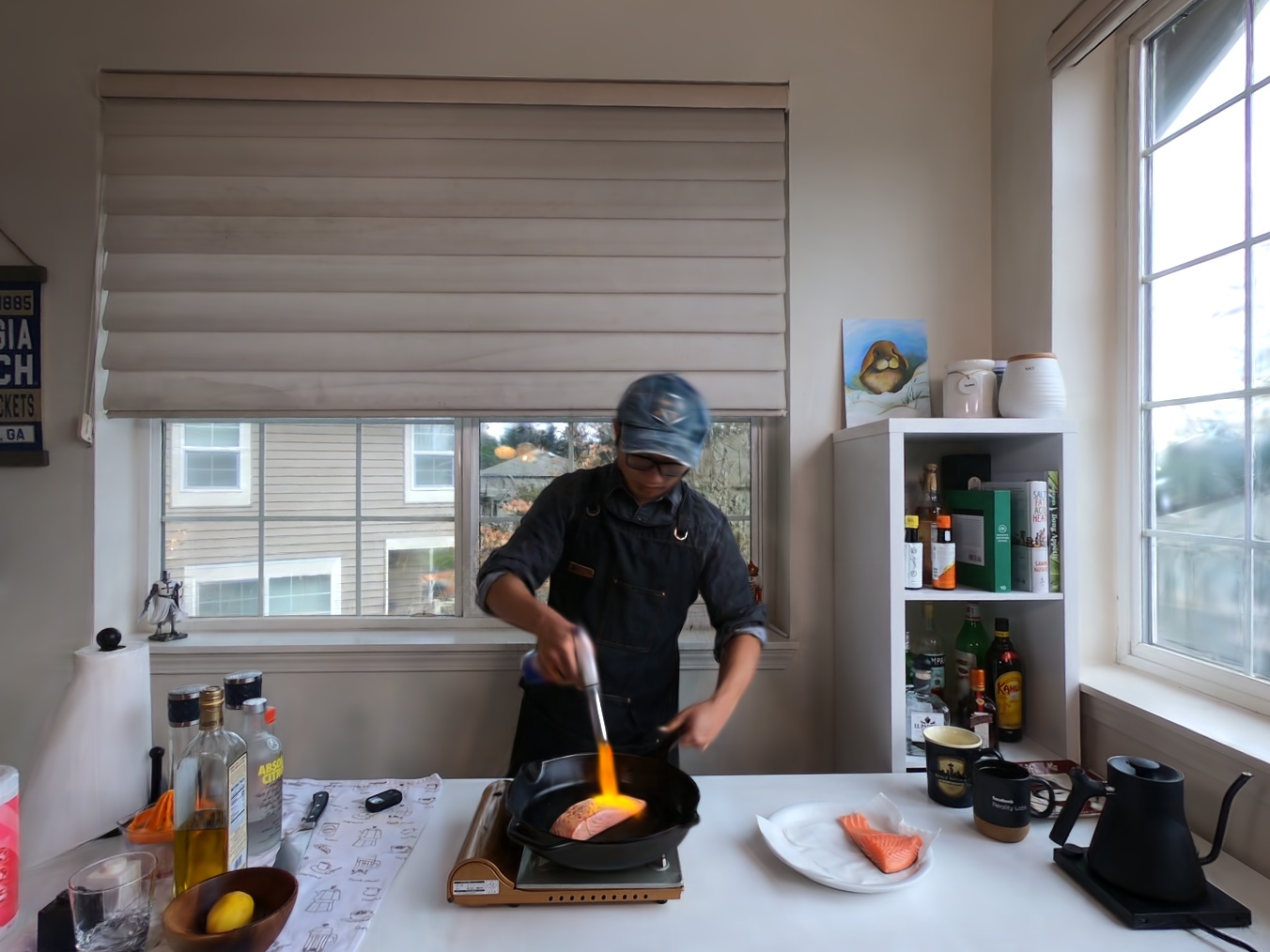}
   } 
   \qquad
      \subfloat[666th frame]{
      \includegraphics[width=0.24\textwidth]{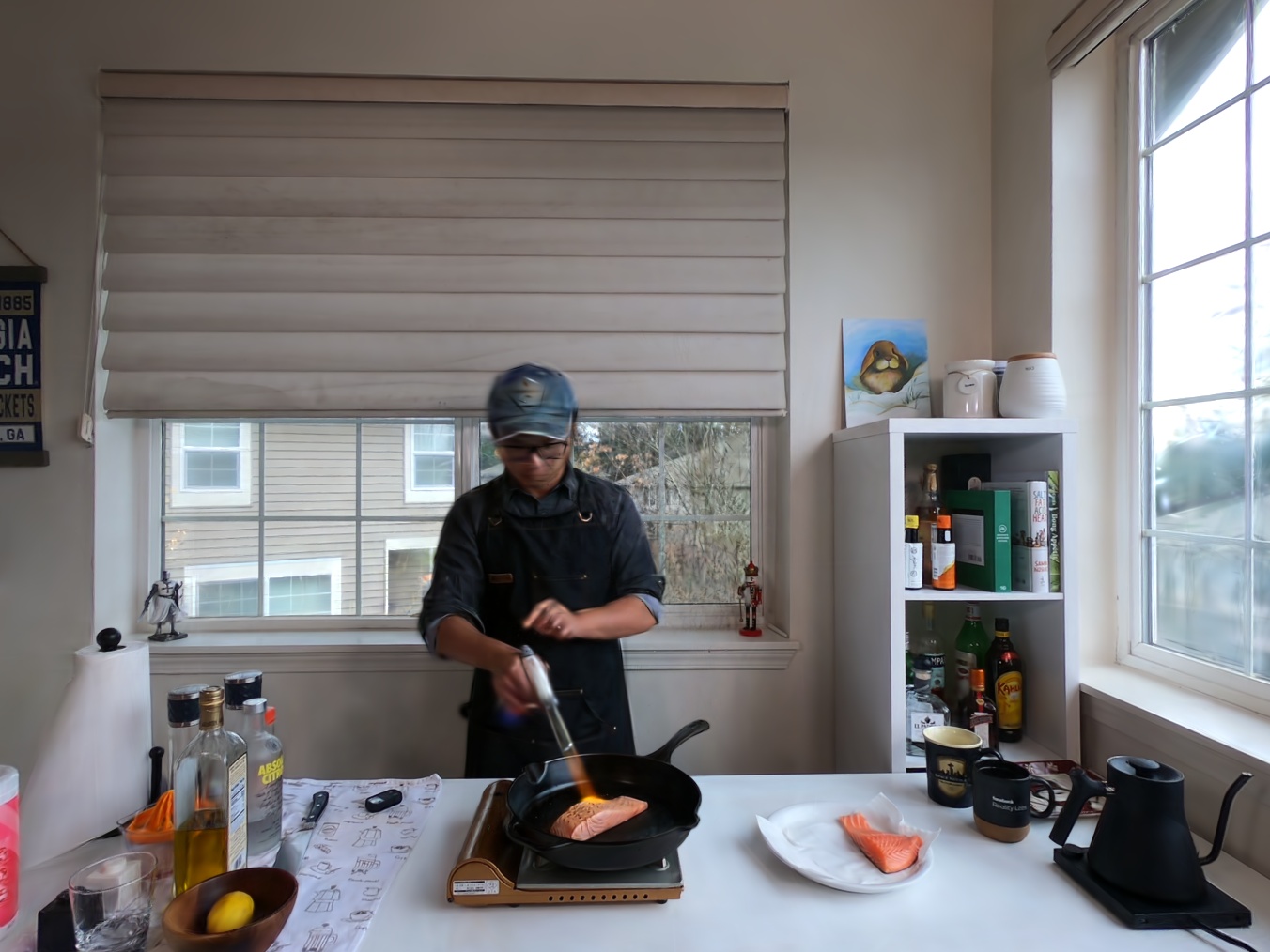}
   }
   \subfloat[782th frame]{
      \includegraphics[width=0.24\textwidth]{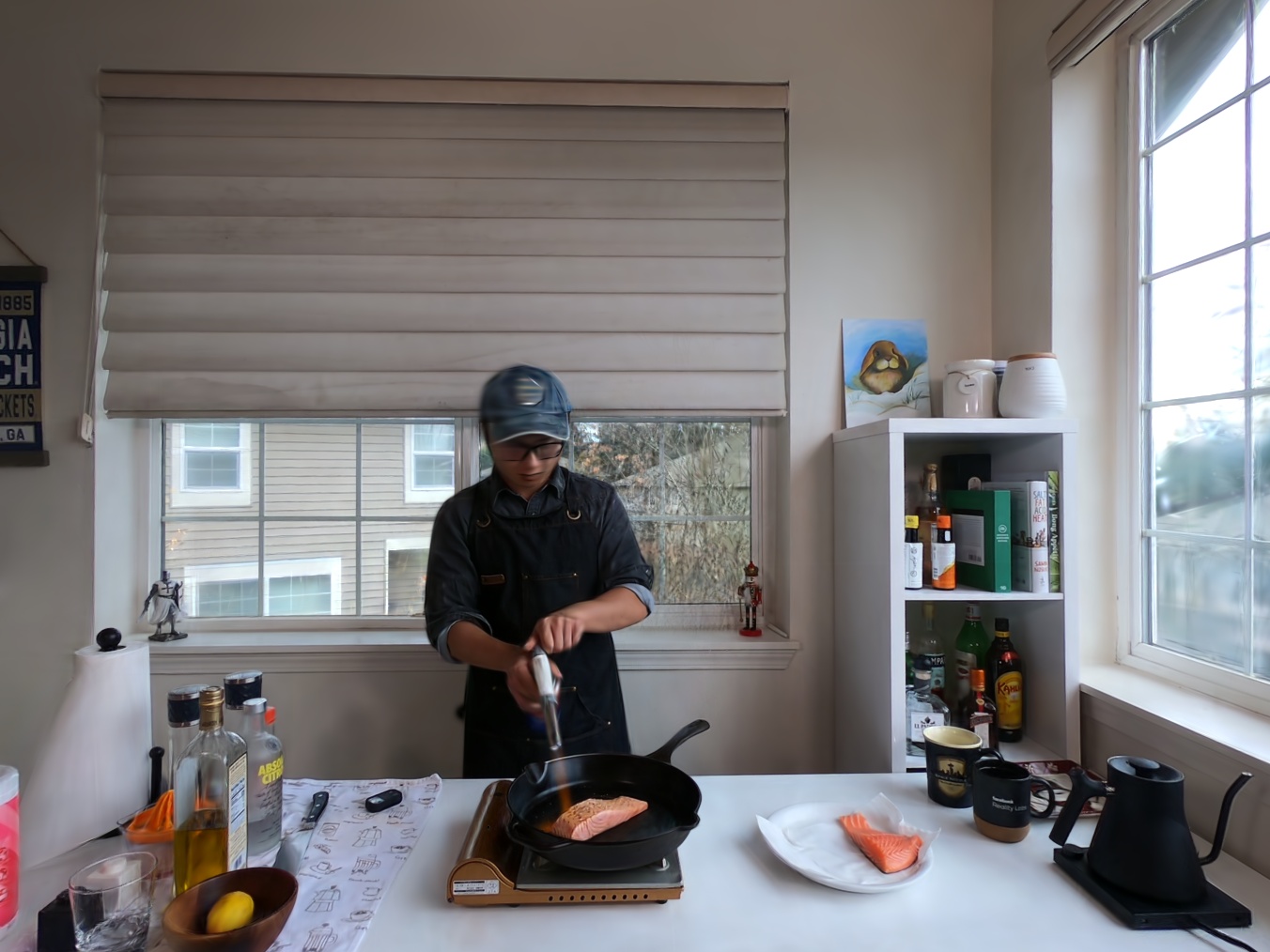}
   }
   \subfloat[908th frame]{
      \includegraphics[width=0.24\textwidth]{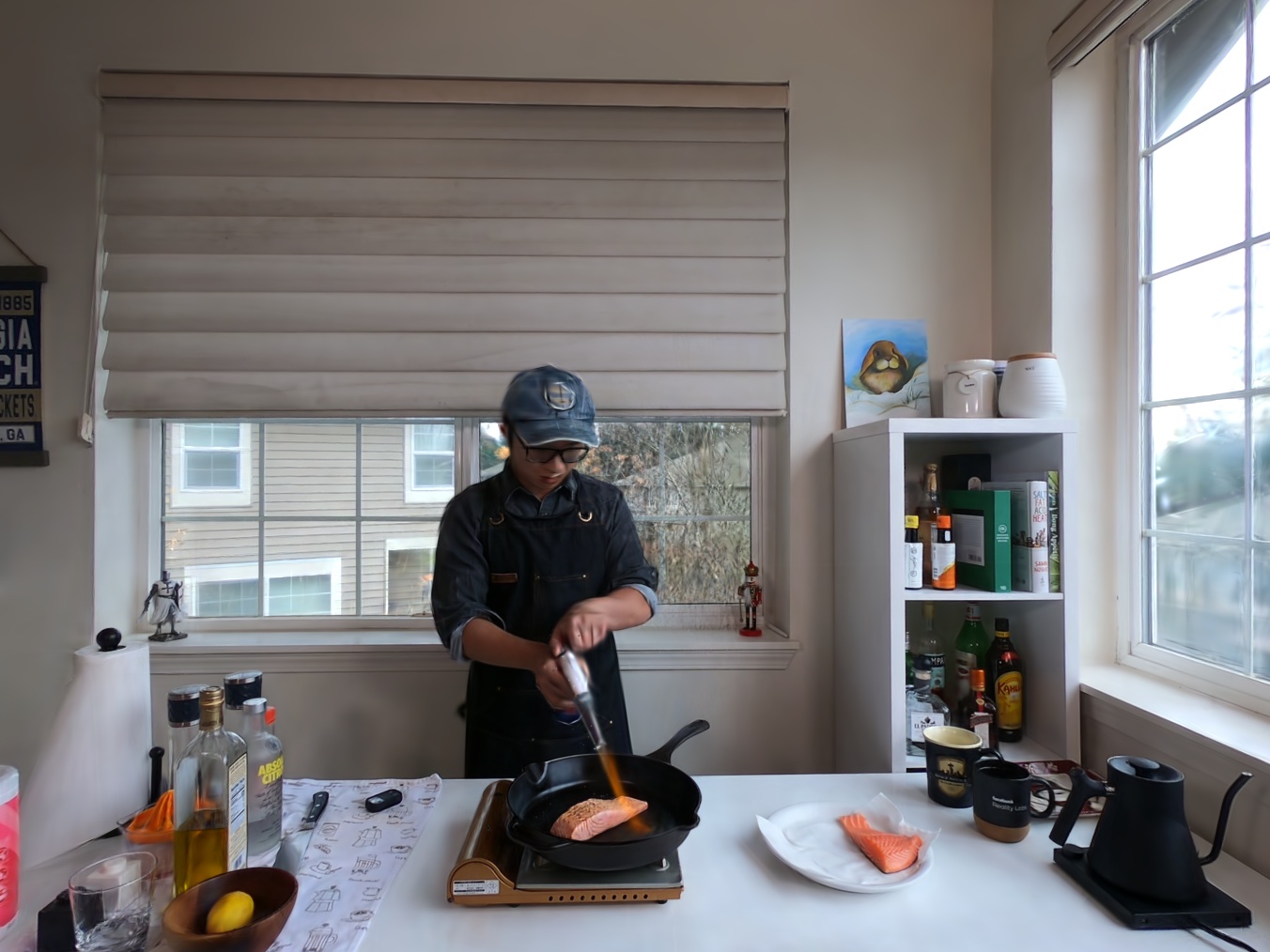}
   } 
    \subfloat[1049th frame]{
      \includegraphics[width=0.24\textwidth]{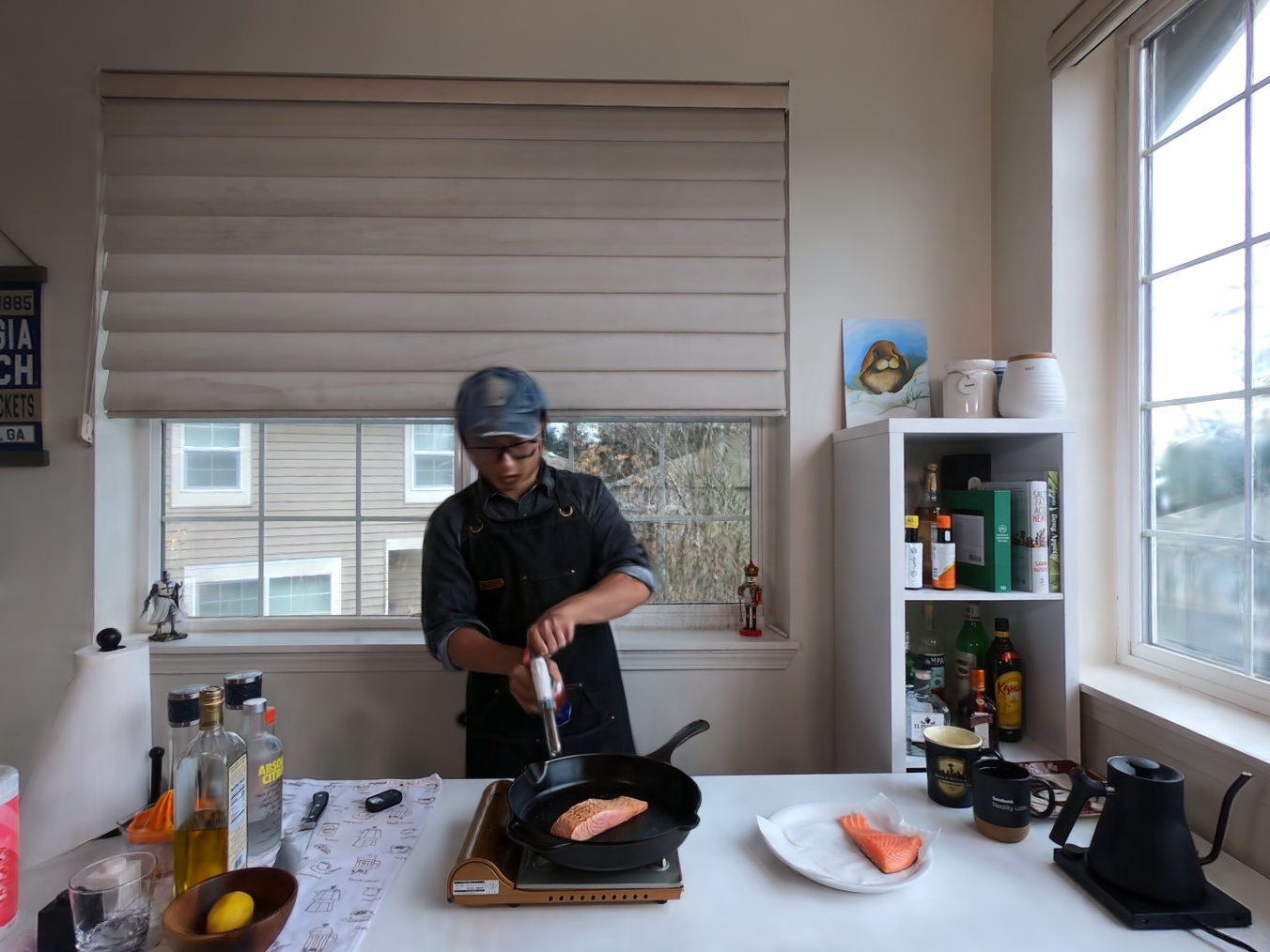}
   } 
   
   \caption{Qualitative results of the long video on \textit{Flame Salmon} scene of N3DV dataset.}
   \label{fig:long}
   \vspace{-0.2cm}
\end{figure}

We conduct an experiment using a longer video sequences~(1200 frames) from the \textit{flame salmon} scene in N3DV dataset, as shown in Fig.~\ref{fig:long}. All of subsequent GOPs are refined based on previous GOP through the GOP-by-GOP training strategy. The quantitative results are represented in Table~\ref{tab:long}. Note that 4DGS, 4DGaussians, and Ex4DGS are derived from Ex4DGS~\cite{lee2024fully}. 3DGStream and HiCoM, marked with $*$, represent our re-implementation. Table~\ref{tab:long} demonstrates that our GOP-by-GOP training strategy effectively adapts to long video sequences and achieves superior performance compared with existing SOTA methods. Moreover, the average frame size remains at 90KB and has not increased due to the appearance of long sequences. 3DGStream and HiCoM exhibited poor performance in this long video scenario. The accumulation of errors over time resulted in their performance being significantly lower than both our method. 

\begin{table}[t]
\centering
\caption{Quantitative results of the long video on \textit{Flame Salmon}.}
\label{tab:long}
\begin{tabular}{c|cccc}
\toprule
Method      & PSNR  & SSIM  & \begin{tabular}[c]{@{}c@{}}Storage\\ (KB)\end{tabular} & \begin{tabular}[c]{@{}c@{}}Streamable \end{tabular} \\ \midrule
4DGS~\cite{yang2023gs4d}                  & 26.26 & 0.897 & 5402           & $\times$         \\
4DGaussians~\cite{wu20244d}               & 28.37 & 0.903 & 64             & $\times$         \\
Ex4DGS~\cite{lee2024fully}                & 28.77 & 0.919 & 334            & $\times$          \\
3DGStream$^{*}$~\cite{sun20243dgstream}   & 27.11 & 0.898 & 8151            & $\checkmark$          \\ 
HiCoM$^{*}$~\cite{gao2024hicom}           & 26.75 & 0.872 & 14724            & $\checkmark$           \\ 
\midrule
Ours(3DGS)                                & 29.08 & 0.913 & 89.96          & $\checkmark$           \\ \bottomrule
\end{tabular}
\end{table}

\subsection{Performance of parallel training}
\label{sec:parallel}

\begin{table}[t]
\caption{Quantitative comparison of parallel GOP training on N3DV dataset.}
\label{tab:pt-lang}
\centering
\begin{tabular}{c|cccccc}
\toprule
Method           & mIoU $\uparrow$ & mAcc $\uparrow$ & PSNR $\uparrow$ & SSIM $\uparrow$  & \begin{tabular}[c]{@{}c@{}}Train \\ (s) $\downarrow$\end{tabular} & \begin{tabular}[c]{@{}c@{}}Storage \\ (KB) $\downarrow$\end{tabular} \\ \midrule

4DLangSplat~\cite{li20254d}    & 75.53 & 75.53   & 30.99 & 0.935 & 72 & 431                     \\
Ours(3DGS)         & 82.93 & 96.22   & 32.15 & 0.944 & 54  & 97.11                   \\ 
Ours-PT(3DGS)      & 85.24 & 94.00   & 32.17 & 0.944 & 26  & 95.31                   \\ 
Ours(Scaffold-GS)  & 84.69 & 96.00   & 32.04 & 0.945 & 42  & 46.33                   \\ 
Ours-PT(Scaffold-GS)  & 85.12 & 94.44   & 32.02 & 0.945 & 19  & 46.54                   \\ \bottomrule
\end{tabular}%
\vspace{-4mm}
\end{table}

Table~\ref{tab:pt-lang} presents the parallel GOP training performance of language-embedded FVV. It can be seen that due to without cumulative error per GOP, mIoU is higher and image quality shows almost no difference. In particular, we reduced the training time per frame from 72s to 26s and 19s, representing reductions of 2.7x and 3.8x, respectively. We also demonstrate the FVV reconstruction performance without language embeddings in Section~\ref{additional-fvv-quant}.

\subsection{Deforming decoded Gaussian attributes of Scaffold-GS}
\label{sec:deforming}

\begin{table}[ht]
\caption{Quantitative comparison of deforming anchor attributes or decoded Gaussian attributes on N3DV dataset.}
\label{tab:deform}
\centering
\begin{tabular}{c|cccc}
\toprule
Method           & PSNR $\uparrow$ & SSIM $\uparrow$  & LPIPS $\downarrow$ & \begin{tabular}[c]{@{}c@{}}Storage \\ (KB) $\downarrow$\end{tabular} \\ \midrule

Deform-gaussian       & 29.37 & 0.9314 & 0.1618 & 24.73                     \\
Ours(ScaffoldGS)      & 31.93 & 0.9448 & 0.1408 & 42.42                     \\ \bottomrule
\end{tabular}%
\end{table}

As shown in Table~\ref{tab:deform}, we demonstrate the results of deforming the decoded Gaussian attributes, which exhibits poor performance and fails to learn Gaussian motion. Unlike 3DGS, where attributes are explicitly learned, Scaffold-GS attributes are obtained through MLP decoding. In other words, time-varying Gaussian attributes are obtained by adding the results of two groups of MLPs. However, the interpolation-based deformation field proposed in this paper is independent of Anchor attributes, making the two MLPs unaware of each other and causing optimization difficulties. Instead, this paper first learns the changes in anchor attributes, and then uses the deformed anchor attributes to decode the final Gaussian attributes. This approach makes the backpropagation of gradients linear, allowing the two groups of MLPs to be optimized more effectively.

\subsection{Additional Quantitative Results}
\label{appendix-experiment}

\begin{table}[tp]
\caption{Performance of Ours(3DGS) under different interval and GOP length on N3DV dataset.}
\label{tab:gvn1}
\resizebox{\columnwidth}{!}{%
\begin{tabular}{c|ccc|ccc|ccc|ccc}
\toprule
Interval & \multicolumn{3}{c|}{2} & \multicolumn{3}{c|}{5} & \multicolumn{3}{c|}{10} & \multicolumn{3}{c}{20} \\
GOP      & PSNR   & SSIM  & Size  & PSNR   & SSIM  & Size  & PSNR   & SSIM   & Size  & PSNR   & SSIM   & Size \\ \midrule
30       & 32.10 & 0.946  & 225.4 & 32.03 & 0.944  & 133.0 & 32.20 & 0.945  & 132.2 & -      & -    \\
60       & 31.88 & 0.944  & 169.5 & 32.11 & 0.945  & 119.5 & 32.26 & 0.945  & 92.3  & 31.76 & 0.943  & 75.6 \\
100      & 31.79 & 0.943  & 152.6 & 32.17 & 0.945  & 106.0 & 31.97 & 0.944  & 72.6  & 32.08 & 0.944  & 59.0 \\ \bottomrule
\end{tabular}%
}
\end{table}

\begin{table}[tp]
\caption{Performance of Ours(Scaffold-GS) under different interval and GOP length on N3DV dataset.}
\label{tab:gvn2}
\resizebox{\textwidth}{!}{%
\begin{tabular}{c|ccc|ccc|ccc|ccc}
\toprule
Interval & \multicolumn{3}{c|}{2} & \multicolumn{3}{c|}{5} & \multicolumn{3}{c|}{10} & \multicolumn{3}{c}{20} \\
GOP      & PSNR   & SSIM  & Size  & PSNR   & SSIM  & Size  & PSNR   & SSIM   & Size  & PSNR   & SSIM   & Size \\ \midrule
30       & 32.17  & 0.947 & 162.1 & 32.27  & 0.947 & 87.42 & 32.00  & 0.945  & 42.15 & -      & -      & -    \\
60       & 32.14  & 0.947 & 169.2 & 32.18  & 0.947 & 98.33 & 31.93  & 0.945  & 42.42 & 32.03  & 0.946  & 52.65 \\
100      & 32.10  & 0.946 & 178.5 & 32.17  & 0.947 & 107.8 & 32.04  & 0.946  & 68.54 & 31.83  & 0.945  & 52.11  \\ \bottomrule
\end{tabular}%
}
\end{table}

\subsubsection{Ablation studies under different interval and GOP length}
Table~\ref{tab:gvn1} and Table~\ref{tab:gvn2} present the effects of time feature intervals and GOP length on performance for the N3DV dataset. The results reveal that the average frame size decreases as both the interval and GOP length increase. However, optimal reconstruction quality is achieved with a GOP length of 60 and an interval of 10, which aligns with the default configuration used in our experiments. Notably, a GOP length of 30 yields reconstruction quality comparable to the default configuration. Moreover, performance does not degrade significantly even when the GOP length is increased to 100. This demonstrates that our proposed interpolated-based deformation field effectively handles a wide range of motion dynamics. 
Increasing the keyframe interval (e.g., from 10 to 20) does not further reduce frame size. Therefore, a keyframe interval of 10 and a GOP length of 60 represent the optimal trade-off between performance and frame size.

\subsubsection{Additional FVV Quantitative Comparisons}
\label{additional-fvv-quant}

\begin{table}[t!]
  \caption{More quantitative comparison on N3DV dataset.}
  \label{tab:appendix-n3dv}
  \resizebox{\textwidth}{!}{%
  \begin{tabular}{ccccccccc}
    \toprule
    Method & PSNR $\uparrow$ & SSIM $\uparrow$ & LPIPS $\downarrow$
           & \begin{tabular}[c]{@{}c@{}}Storage\\(KB) $\downarrow$\end{tabular}
           & \begin{tabular}[c]{@{}c@{}}Train\\(s) $\downarrow$\end{tabular}
           & \begin{tabular}[c]{@{}c@{}}Decode\\(ms) $\downarrow$\end{tabular}
           & \begin{tabular}[c]{@{}c@{}}Render\\(ms) $\downarrow$\end{tabular}
           & FPS $\uparrow$ \\
    \midrule

    & \multicolumn{8}{c}{\textbf{3DGS-based representations}} \\

    TeTriRF~\cite{wu2024tetrirf}
      & 30.07 & 0.9003 & 0.2986 &
        \cellcolor[HTML]{FF9E98}65.89 &
        31.79 & 149.3 & 653.59 & 1.53 \\
    3DGStream~\cite{sun20243dgstream}
      & 30.73 & 0.9348 & 0.1470 & 8205 & 
      17.07 &
      6.51 & 13.89 & 72 \\
    HiCoM~\cite{gao2024hicom}
      & 31.31 & 0.9390 & 0.1475 & 10704 &
        \cellcolor[HTML]{FFFC9E}10.35 &
        \cellcolor[HTML]{FF9E98}0 &
        \cellcolor[HTML]{FFCE93}6.13 &
        \cellcolor[HTML]{FFCE93}163 \\
    QUEEN~\cite{girish2024queen}
      & 31.79 &
        \cellcolor[HTML]{FFCE93}0.9449 &
        0.1415 &
        20882 &
        \cellcolor[HTML]{FFCE93}8.70 &
        \cellcolor[HTML]{FFCE93}0.58 &
        \cellcolor[HTML]{FF9E98}3.98 &
        \cellcolor[HTML]{FF9E98}251 \\
    4DGC~\cite{hu20254dgc}
      & 31.53 & 0.9412 & 
      \cellcolor[HTML]{FFFC9E}0.1431 & 784 & 67.37 & 2.57 & 12.66 & 79 \\
    OR2~\cite{yun2025compensating}
      & 30.73 & 0.9336 & 0.1577 & 8005 & 42.53 &
        \cellcolor[HTML]{FFFC9E}1.20 &
        \cellcolor[HTML]{FFFC9E}7.46 &
        \cellcolor[HTML]{FFFC9E}134 \\
    StreamSTGS~\cite{ke2025streamstgs}
      & \cellcolor[HTML]{FF9E98}32.30 & 
      0.9436 & 
      0.1474 & 
      174 & 66.75 & 8.63 & 10.00 & 100 \\
    Ours(3DGS)
      & \cellcolor[HTML]{FFCE93}32.26 &
        \cellcolor[HTML]{FF9E98}0.9452 &
        \cellcolor[HTML]{FF9E98}0.1388 &
        \cellcolor[HTML]{FFCE93}92.04 & 
        18.67 & 2.27 & 9.35 & 107 \\
    Ours-PT(3DGS)
      & \cellcolor[HTML]{FFFC9E}32.17 &
        \cellcolor[HTML]{FFFC9E}0.9447 &
        \cellcolor[HTML]{FFCE93}0.1405 &
        \cellcolor[HTML]{FFFC9E}94 & 
        \cellcolor[HTML]{FF9E98}6.82 & 2.34 & 9.52 & 105 \\

    \midrule
    & \multicolumn{8}{c}{\textbf{Scaffold-GS-based representations}} \\

    GIFStream~\cite{li2025gifstream}
      & 30.76 & 0.9368 & 0.1470 &
        \cellcolor[HTML]{FF9E98}38.04 &
        22.25 & 
        1.97 &
        \cellcolor[HTML]{FFCE93}10.31 & 
        \cellcolor[HTML]{FF9E98}97 \\
    iFVC~\cite{tang2025compressing}
      & 31.38 & 
        0.9419 &
        \cellcolor[HTML]{FF9E98}0.1357 &
        114 &
        15.75 &
        68.24 &
        \cellcolor[HTML]{FF9E98}8.93 &
        14.65\\
    Ours(ScaffoldGS)
      & \cellcolor[HTML]{FF9E98}31.93 &
        \cellcolor[HTML]{FF9E98}0.9448 &
        \cellcolor[HTML]{FFCE93}0.1408 &
        \cellcolor[HTML]{FFCE93}42.42 &
        \cellcolor[HTML]{FFCE93}20.81 &
        \cellcolor[HTML]{FF9E98}1.94 &
        11.32 & 
        \cellcolor[HTML]{FFCE93}88.27\\
    Ours-PT(ScaffoldGS)
      & \cellcolor[HTML]{FFCE93}31.93 &
        \cellcolor[HTML]{FFCE93}0.9446 &
        0.1412 &
        42.65 &
        \cellcolor[HTML]{FF9E98}8.22 &
        \cellcolor[HTML]{FFCE93}1.96 &
        11.42 & 
        87.6\\

    \bottomrule
  \end{tabular}}
  \vspace{-0.2cm}
\end{table}

\begin{table}[t!]
  \caption{More quantitative comparison on MeetRoom dataset.}
  \label{tab:meet}
  \resizebox{\textwidth}{!}{%
  \begin{tabular}{ccccccccc}
    \toprule
    Method & PSNR $\uparrow$ & SSIM $\uparrow$ & LPIPS $\downarrow$
           & \begin{tabular}[c]{@{}c@{}}Storage\\(KB) $\downarrow$\end{tabular}
           & \begin{tabular}[c]{@{}c@{}}Train\\(s) $\downarrow$\end{tabular}
           & \begin{tabular}[c]{@{}c@{}}Decode\\(ms) $\downarrow$\end{tabular}
           & \begin{tabular}[c]{@{}c@{}}Render\\(ms) $\downarrow$\end{tabular}
           & FPS $\uparrow$ \\
    \midrule

    3DGStream~\cite{sun20243dgstream}
      & 26.41 & 0.8990 & 0.2370 & 4108 & 
      10.91 &
      3.03 & 8.23 & 121.44 \\
    HiCoM~\cite{gao2024hicom}
      & 26.69 & 0.9035 & 0.2315 & 5535 &
        \cellcolor[HTML]{FFCE93}6.18 &
        \cellcolor[HTML]{FF9E98}0 &
        \cellcolor[HTML]{FFCE93}3.64 &
        \cellcolor[HTML]{FFCE93}275 \\
    QUEEN~\cite{girish2024queen}
      & 26.34 &
        \cellcolor[HTML]{FF9E98}0.9185 &
        0.2164 &
        11201 &
        \cellcolor[HTML]{FF9E98}4.32 &
        \cellcolor[HTML]{FFCE93}0.35 &
        \cellcolor[HTML]{FF9E98}2.01 &
        \cellcolor[HTML]{FF9E98}497 \\
    4DGC~\cite{hu20254dgc}
      & 27.11 & 0.9093 & 
      0.2306 & 1195 & 48.54 & 1.97 & 9.06 & 110 \\
    StreamSTGS~\cite{ke2025streamstgs}
      & \cellcolor[HTML]{FFCE93}27.41 & 
      \cellcolor[HTML]{FFCE93}0.9181 & 
      0.2157 & 
      143 & 28.96 & 6.57 & 7.93 & 100 \\

    GIFStream~\cite{li2025gifstream}
      & 24.37 & 0.8807 & 0.2387 &
        \cellcolor[HTML]{FF9E98}29.11 &
        16.80 & 
        \cellcolor[HTML]{FFFC9E}1.84 &
        6.64 & 
        126 \\
    iFVC~\cite{tang2025compressing}
      & \cellcolor[HTML]{FFFC9E}27.29 & 
        0.9120 &
        \cellcolor[HTML]{FF9E98}0.2036 &
        170 &
        9.06 &
        55.87 &
        \cellcolor[HTML]{FFFC9E}5.67 &
        17.90\\
    Ours(3DGS)
      & \cellcolor[HTML]{FF9E98}27.48 &
        \cellcolor[HTML]{FFFC9E}0.9179 &
        \cellcolor[HTML]{FFFC9E}0.2080 &
        \cellcolor[HTML]{FFFC9E}75.92 & 
        \cellcolor[HTML]{FFFC9E}6.93 & 2.40 & 6.02 & 
        \cellcolor[HTML]{FFFC9E}166 \\
    Ours(ScaffoldGS)
      & 27.28 &
        0.9172 &
        \cellcolor[HTML]{FFCE93}0.2059 &
        \cellcolor[HTML]{FFCE93}49.31 &
        9.18 &
        2.47 &
        8.13 & 
        123\\

    \bottomrule
  \end{tabular}}
  \vspace{-0.2cm}
\end{table}

Table~\ref{tab:appendix-n3dv} and Table~\ref{tab:meet} supplement Tables 3 and 4 in the manuscript, showing the three metrics: decoding latency, rendering latency, and training time. While HiCoM and QUEEN outperform our method in training time, decoding speed, and rendering speed, our reconstruction quality~(i.e., PSNR, SSIM, and LPIPS) is significantly superior. For StreamSTGS, we achieve the same reconstruction quality at only one-third the training time and decoding speed. Experimental results based on Scaffold-GS representation show that our method comprehensively outperforms existing benchmarks. Moreover, based on the RD Caves results in the manuscript, our method outperforms these baseline methods in reconstruction quality at the same frame size. 

We also demonstrate the performance of GOP parallel training on the N3DV dataset without language embedding, namely Ours-PT(3DGS) and Ours-PT(Scaffold-GS). As shown in Table~\ref{tab:appendix-n3dv}, our proposed GOP parallel training can effectively reduce training time while having a negligible impact on reconstruction quality and FPS.

\subsubsection{Additional Baseline Comparisons}
Table~\ref{tab:all} presents a comprehensive comparisons against additional dynamic scene reconstruction baselines that not support streaming. As we can see, our method is quite competitive in all metrics compared to current methods for SoTA dynamic scenes reconstruction methods that do not support streaming.

\begin{table}[!t]
\caption{Quantitative comparisons with non-streamable baselines on the N3DV Datasets.}
\label{tab:all}
\resizebox{\textwidth}{!}{%
    \begin{tabular}{cccccccc}
    \toprule
    Scene            & PSNR $\uparrow$ & SSIM $\uparrow$ & LPIPS $\downarrow$ & \begin{tabular}[c]{@{}c@{}}Storage\\ (MB) $\downarrow$\end{tabular} & FPS $\uparrow$ & \begin{tabular}[c]{@{}c@{}}Training\\ (s) $\downarrow$\end{tabular} & Streamable \\
    \midrule
    HyperReel~\cite{attal2023hyperreel}   & 31.10 & 0.928 & -     & 1.2   & 2.0  & 104           & $\times$          \\
    HexPlanes~\cite{cao2023hexplane}      & 31.70 & -     & -     & 0.8   & 0.21 & 144           & $\times$          \\
    KPlanes~\cite{fridovich2023k}         & 31.63 & -     & -     & 1.0   & 0.15 & 48            & $\times$          \\
    MixVoxels~\cite{wang2022mixed}        & 30.81 & -     & -     & 1.7   & 38   & 16            & $\times$          \\
    NeRFPlayer~\cite{song2023nerfplayer}  & 30.69 & 0.932 & 0.209 & 17.1  & 0.05 & 72            & $\checkmark$      \\
    StreamRF~\cite{li2022streaming}       & 30.68 & -     & -     & 31.4  & 8.3  & 15            & $\checkmark$      \\ \midrule
    4DGS~\cite{yang2023gs4d}              & 32.01 & -     & -     & 29    & 30   & 76            & $\times$          \\
    4D-GS~\cite{wu20244d}     & 31.11 & 0.938 & 0.141 & 0.11  & 30   & 9             & $\times$          \\
    STGS~\cite{li2024spacetime}           & 31.62 & 0.946 & -     & 0.6  & 140  & 30            & $\times$          \\
    4D-Rotor~\cite{duan20244d}            & 31.62 & 0.94  & -     & -     & 277  & 12            & $\times$          \\
    DN-4DGS~\cite{lu2024dn}               & 32.02 & 0.944 & -     & 0.37  & 15   & 10            & $\times$          \\
    Ex4DGS~\cite{lee2024fully}            & 32.11 & 0.94  & -     & 0.38  & -    & -             & $\times$          \\
    Swift4D~\cite{wu2025swift4d}           & 32.23 & -     & -     & 0.4   & 125  & 5             & $\times$         \\ 
    TimeFormer(4D-GS)~\cite{jiang2025timeformer}  & 31.82 & 0.941     & -     & 0.3   & 38  & 19             & $\times$         \\ 
    TimeFormer(STGS)~\cite{jiang2025timeformer}  & 32.23 & 0.947     & -     & 0.52   & 89  & 60             & $\times$         \\ 
    \midrule

    Ours(3DGS)                           & 32.27 & 0.945 & 0.139 & 0.09  & 111  & 18            & $\checkmark$      \\
    Ours(Scaffold-GS)                    & 31.93 & 0.945 & 0.141 & 0.04  & 88  & 20.81            & $\checkmark$      \\
    \bottomrule
    \end{tabular}%
}
\vspace{-0.2cm}
\end{table}

\subsubsection{Per-scene performance}
We show results for each scene in both the N3DV and MeetRoom datasets of various metrics, including PSNR, SSIM, LPIPS(VGG), storage(the average frame size), FPS, decoding time, training time in Table~\ref{tab:app-n3dv} and Table~\ref{tab:app-meet}.

\begin{table}[!t]
\caption{Per-scene performance for the N3DV dataset.}
\centering
\label{tab:app-n3dv}
\begin{tabular}{cccccccc}
    \toprule
    Scene            & PSNR $\uparrow$ & SSIM $\uparrow$ & LPIPS $\downarrow$ & \begin{tabular}[c]{@{}c@{}}Storage\\ (KB) $\downarrow$\end{tabular} & FPS $\uparrow$ & \begin{tabular}[c]{@{}c@{}}Decoding\\ (ms) $\downarrow$\end{tabular} & \begin{tabular}[c]{@{}c@{}}Training\\ (s) $\downarrow$\end{tabular} \\
    \midrule
    & \multicolumn{7}{c}{\textbf{Ours(3DGS)}} \\
    Coffee Martini   & 29.15 & 0.92 & 0.151 & 97.4        & 84 & 1.9        & 20                                                     \\
    Cook Spinach     & 33.35 & 0.95 & 0.141 & 88.9        & 114  & 2.2        & 18                                                     \\
    Cut Roasted Beef & 33.87 & 0.96 & 0.139 & 99.8        & 116 & 2.6        & 18                                                     \\
    Flame Salmon     & 29.22 & 0.92 & 0.142 & 120         & 78  & 2.7        & 20                                                     \\
    Flame Steak      & 33.78 & 0.96 & 0.132 & 54.1        & 131  & 1.7        & 17                                                     \\
    Sear Steak       & 34.21 & 0.96 & 0.128 & 92.2        & 118  & 2.4        & 19                                                     \\ \midrule
    & \multicolumn{7}{c}{\textbf{Ours(Scaffold-GS)}} \\
    Coffee Martini   & 28.01 & 0.92 & 0.152 & 56.3        & 76 & 1.8        & 20                                                     \\
    Cook Spinach     & 33.11 & 0.95 & 0.143 & 39.2        & 80  & 2.0        & 18                                                     \\
    Cut Roasted Beef & 33.92 & 0.96 & 0.142 & 44.7        & 93 & 2.0        & 23                                                     \\
    Flame Salmon     & 28.86 & 0.92 & 0.145 & 48.8         & 81  & 2.1        & 23                                                     \\
    Flame Steak      & 33.58 & 0.96 & 0.131 & 36.3        & 94  & 2.1        & 22                                                     \\
    Sear Steak       & 34.13 & 0.96 & 0.131 & 29.0        & 104  & 1.5        & 19                                                     \\
    
    \bottomrule
\end{tabular}%
\vspace{-0.2cm}
\end{table}

\begin{table}[!t]
\caption{Per-scene performance on the MeetRoom dataset.}
\label{tab:app-meet}
\centering
\begin{tabular}{cccccccc}
    \toprule
    Scene            & PSNR $\uparrow$ & SSIM $\uparrow$ & LPIPS $\downarrow$ & \begin{tabular}[c]{@{}c@{}}Storage\\ (KB) $\downarrow$\end{tabular} & FPS $\uparrow$ & \begin{tabular}[c]{@{}c@{}}Decoding\\ (ms) $\downarrow$\end{tabular} & \begin{tabular}[c]{@{}c@{}}Training\\ (s) $\downarrow$\end{tabular} \\
    \midrule
    & \multicolumn{7}{c}{\textbf{Ours(3DGS)}} \\
    Discussion & 28.06 & 0.92 & 0.207 & 84.7     & 160 & 2.3       & 6.8         \\
    Trimming   & 27.84 & 0.92 & 0.203 & 58.9     & 174 & 1.9       & 7.2         \\
    Vrheadset  & 26.53 & 0.91 & 0.214 & 84.1     & 164 & 2.3       & 6.8         \\
    \midrule
    & \multicolumn{7}{c}{\textbf{Ours(Scaffold-GS)}} \\
    Discussion & 27.81 & 0.92 & 0.207 & 75.4     & 121 & 3.8       & 10.5         \\
    Trimming   & 27.66 & 0.92 & 0.197 & 42.8     & 120 & 2.0       & 8.2         \\
    Vrheadset  & 26.37 & 0.91 & 0.214 & 29.7     & 128 & 1.5       & 8.9         \\
    \bottomrule
\end{tabular}%
\vspace{-0.2cm}
\end{table}

\subsubsection{Per-GOP performance}
\begin{table}[!t]
\caption{Per-GOP performance of Ours(3DGS) on the N3DV dataset.}
\label{tab:per-gop}
\centering
    \begin{tabular}{ccccccc}
    \toprule
    Scene            & GOP1  & GOP2  & GOP3  & GOP4  & GOP5  & Average \\ \midrule
    \multicolumn{7}{c}{PSNR}                                           \\
    Coffee Martini   & 28.98 & 29.34 & 29.29 & 29.15 & 29.00 & 29.15   \\
    Cook Spinach     & 33.35 & 33.49 & 33.33 & 33.11 & 33.45 & 33.34   \\
    Cut Roasted Beef & 33.63 & 33.84 & 34.08 & 33.89 & 33.92 & 33.87   \\
    Flame Salmon     & 29.16 & 29.16 & 29.22 & 29.30 & 29.25 & 29.22   \\
    Flame Steak      & 34.03 & 33.99 & 33.56 & 33.64 & 33.65 & 33.77   \\
    Sear Steak       & 34.18 & 34.18 & 34.58 & 34.11 & 34.06 & 34.22   \\ \midrule
    \multicolumn{7}{c}{SSIM}                                           \\
    Coffee Martini   & 0.92  & 0.92  & 0.92  & 0.92  & 0.92  & 0.92    \\
    Cook Spinach     & 0.95  & 0.95  & 0.95  & 0.95  & 0.95  & 0.95    \\
    Cut Roasted Beef & 0.96  & 0.96  & 0.96  & 0.96  & 0.95  & 0.96    \\
    Flame Salmon     & 0.92  & 0.92  & 0.92  & 0.92  & 0.92  & 0.92    \\
    Flame Steak      & 0.96  & 0.96  & 0.96  & 0.96  & 0.96  & 0.96    \\
    Sear Steak       & 0.96  & 0.96  & 0.96  & 0.96  & 0.96  & 0.96    \\ \midrule
    \multicolumn{7}{c}{LPIPS}                                          \\
    Coffee Martini   & 0.152 & 0.150 & 0.150 & 0.150 & 0.151 & 0.151   \\
    Cook Spinach     & 0.140 & 0.141 & 0.142 & 0.142 & 0.142 & 0.141   \\
    Cut Roasted Beef & 0.135 & 0.135 & 0.141 & 0.139 & 0.142 & 0.138   \\
    Flame Salmon     & 0.143 & 0.142 & 0.142 & 0.141 & 0.141 & 0.142   \\
    Flame Steak      & 0.134 & 0.134 & 0.134 & 0.131 & 0.129 & 0.133   \\
    Sear Steak       & 0.128 & 0.128 & 0.126 & 0.128 & 0.128 & 0.128   \\ \bottomrule
    \end{tabular}%
\vspace{-0.2cm}
\end{table}

\begin{table}[!t]
\caption{Per-GOP performance of Ours(Scaffold-GS) on the N3DV dataset.}
\label{tab:per-gop-2}
\centering
    \begin{tabular}{ccccccc}
    \toprule
    Scene            & GOP1  & GOP2  & GOP3  & GOP4  & GOP5  & Average \\ \midrule
    \multicolumn{7}{c}{PSNR}                                           \\
    Coffee Martini   & 28.23 & 28.10 & 27.94 & 27.91 & 27.85 & 28.01   \\
    Cook Spinach     & 33.22 & 33.12 & 33.14 & 32.94 & 33.13 & 33.11   \\
    Cut Roasted Beef & 34.04 & 33.98 & 34.01 & 34.01 & 33.56 & 33.92   \\
    Flame Salmon     & 28.70 & 28.89 & 28.93 & 28.91 & 28.85 & 28.86   \\
    Flame Steak      & 33.99 & 33.78 & 33.41 & 33.43 & 33.27 & 33.58   \\
    Sear Steak       & 34.36 & 34.33 & 34.04 & 33.95 & 33.96 & 34.13   \\ \midrule
    \multicolumn{7}{c}{SSIM}                                           \\
    Coffee Martini   & 0.92  & 0.92  & 0.92  & 0.92  & 0.92  & 0.91    \\
    Cook Spinach     & 0.95  & 0.95  & 0.95  & 0.95  & 0.95  & 0.95    \\
    Cut Roasted Beef & 0.96  & 0.96  & 0.96  & 0.96  & 0.95  & 0.96    \\
    Flame Salmon     & 0.92  & 0.92  & 0.92  & 0.92  & 0.92  & 0.92    \\
    Flame Steak      & 0.96  & 0.96  & 0.96  & 0.96  & 0.96  & 0.96    \\
    Sear Steak       & 0.96  & 0.96  & 0.96  & 0.96  & 0.96  & 0.96    \\ \midrule
    \multicolumn{7}{c}{LPIPS}                                          \\
    Coffee Martini   & 0.151 & 0.150 & 0.152 & 0.152 & 0.153 & 0.152   \\
    Cook Spinach     & 0.143 & 0.142 & 0.144 & 0.143 & 0.144 & 0.143   \\
    Cut Roasted Beef & 0.138 & 0.139 & 0.144 & 0.145 & 0.145 & 0.142   \\
    Flame Salmon     & 0.146 & 0.145 & 0.145 & 0.145 & 0.146 & 0.145   \\
    Flame Steak      & 0.131 & 0.131 & 0.133 & 0.131 & 0.129 & 0.131   \\
    Sear Steak       & 0.130 & 0.130 & 0.131 & 0.131 & 0.131 & 0.131   \\ \bottomrule
    \end{tabular}%
\vspace{-0.2cm}
\end{table}

Table~\ref{tab:per-gop} and Table~\ref{tab:per-gop-2} show results for each GOP on N3DV dataset. As the GOP increases, the reconstruction quality of our method does not decrease, indicating that GOP-by-GOP training strategy can effectively solve the performance degradation caused by accumulated errors in frame-by-frame training.

\subsection{Additional Qualitative Results}
\label{appendix-qualitative}
\subsubsection{Qualitative Results of 4D Frame Interpolation}
\begin{figure}[tp]
   \centering

   \subfloat[106th - reconstructed]{%
      \begin{tikzpicture}
         \node[draw=green, line width=1pt, inner sep=0pt] (img1) {\includegraphics[width=0.24\textwidth]{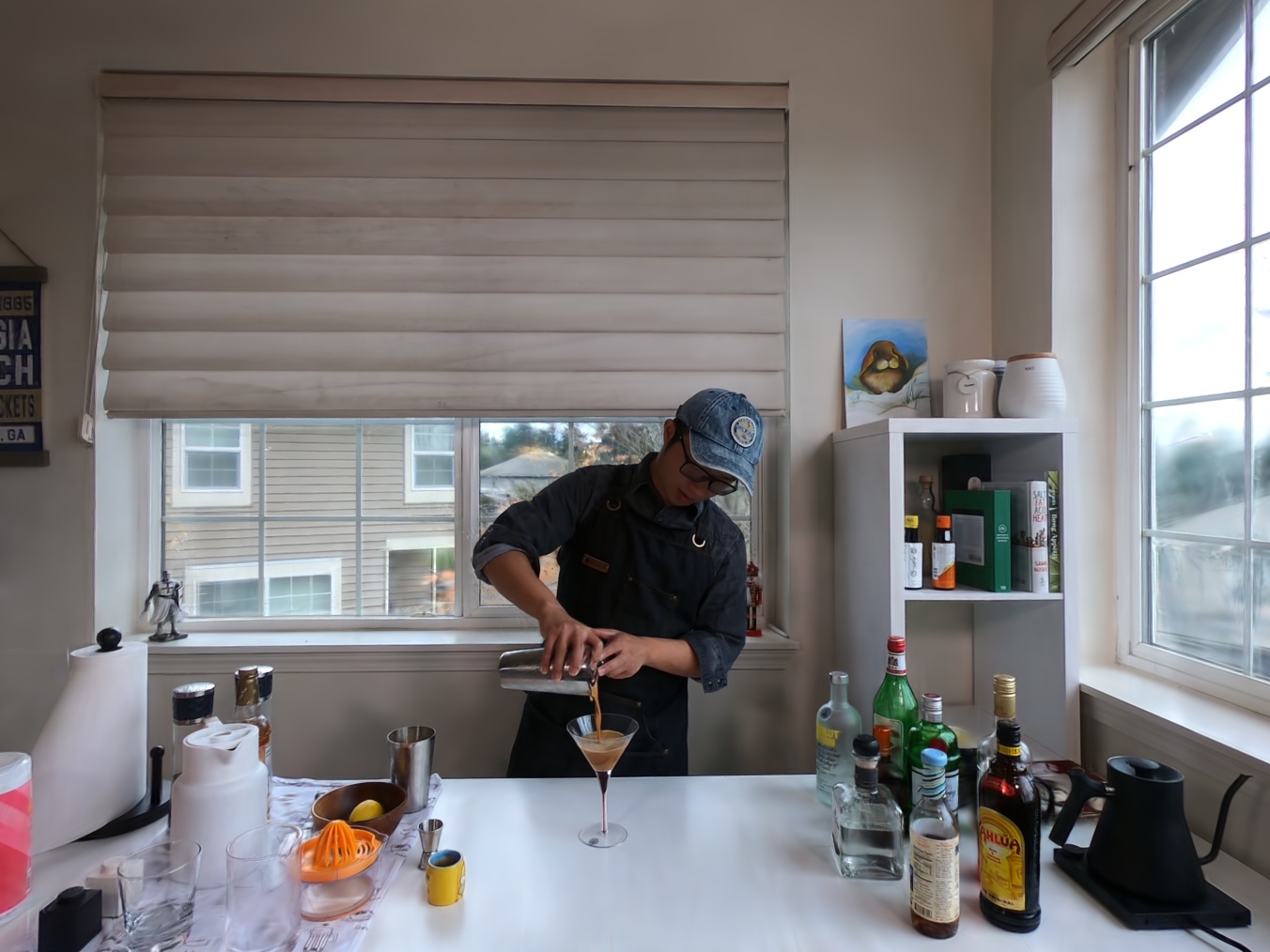}};
      \end{tikzpicture}
   }
   \subfloat[107th - interpolated]{%
      \begin{tikzpicture}
         \node[draw=red, line width=1pt, inner sep=0pt] (img2) {\includegraphics[width=0.24\textwidth]{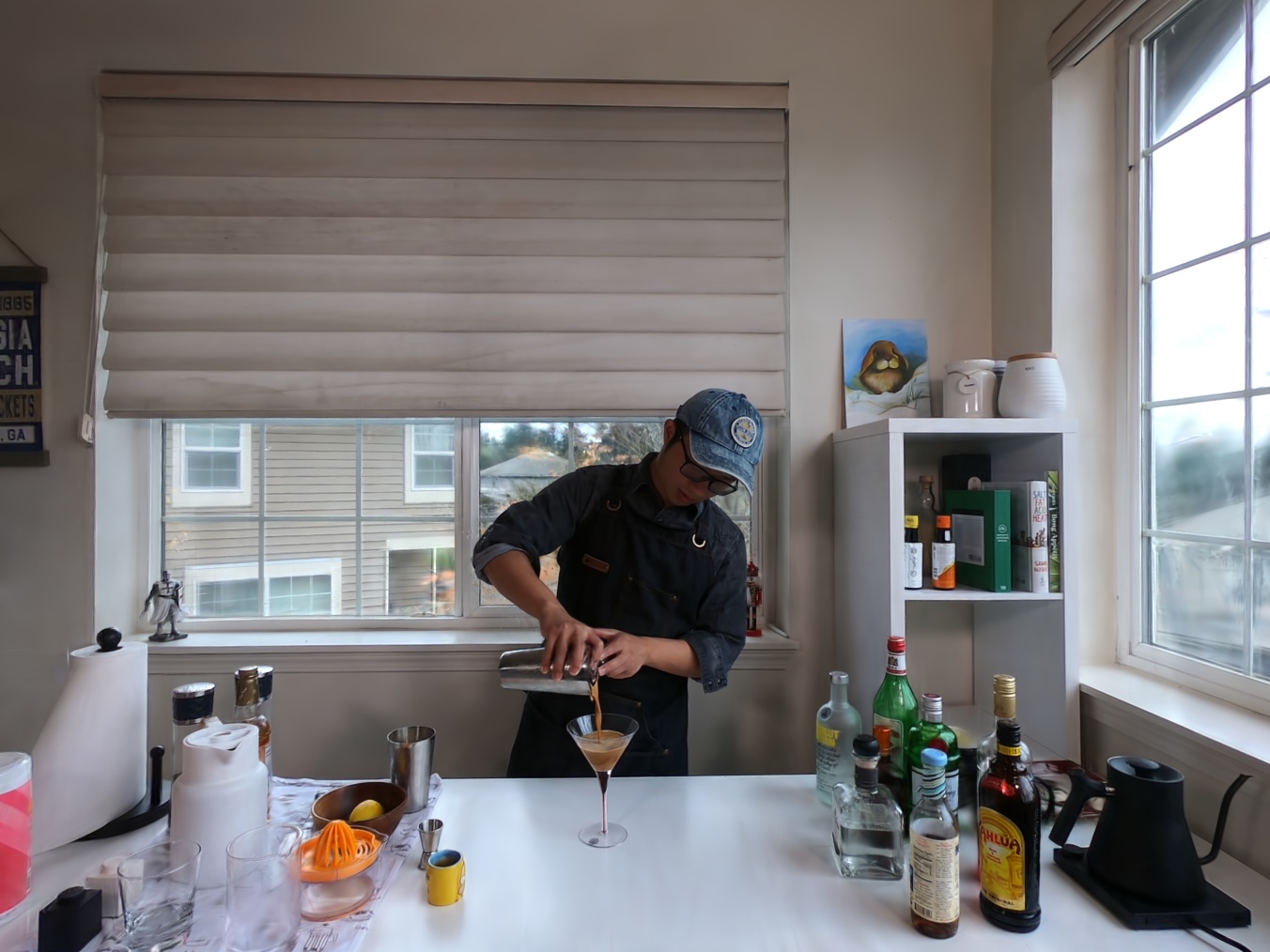}};
      \end{tikzpicture}
   }
   \subfloat[108th - reconstructed]{%
      \begin{tikzpicture}
         \node[draw=green, line width=1pt, inner sep=0pt] (img3) {\includegraphics[width=0.24\textwidth]{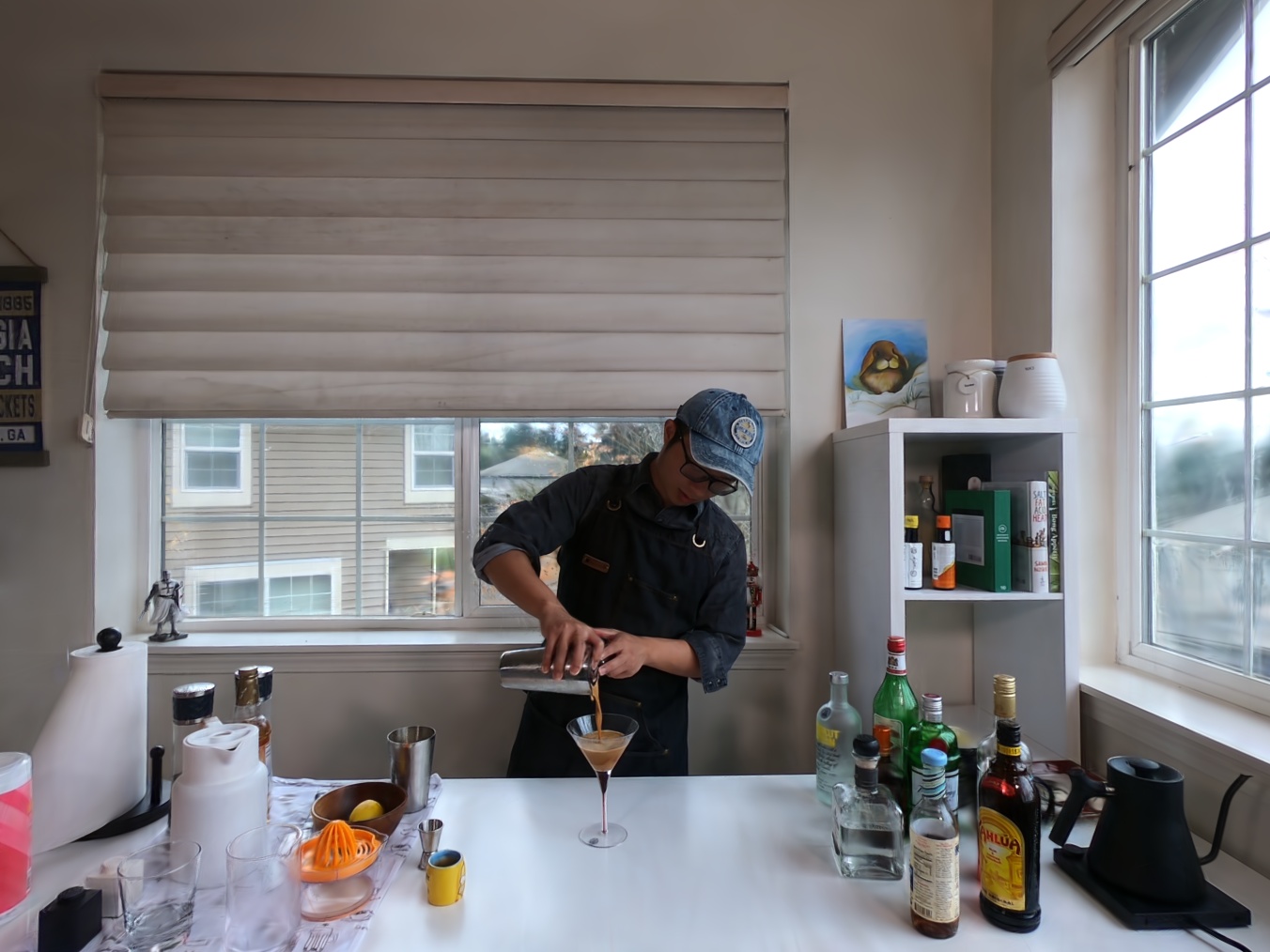}};
      \end{tikzpicture}
   } 
   \subfloat[109th - interpolated]{%
      \begin{tikzpicture}
         \node[draw=red, line width=1pt, inner sep=0pt] (img4) {\includegraphics[width=0.24\textwidth]{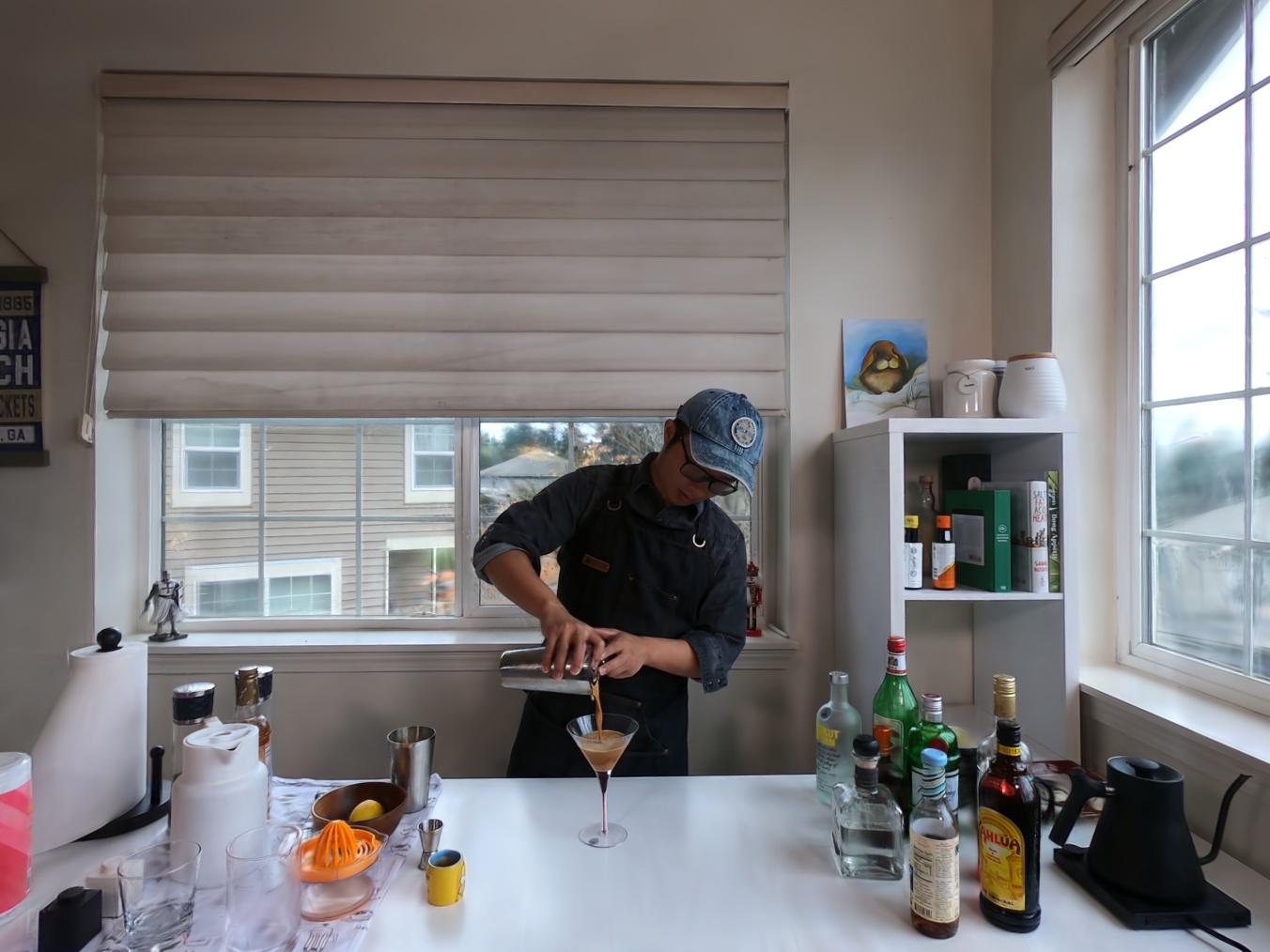}};
      \end{tikzpicture}
   } 
   \qquad
    \subfloat[106th - interpolated]{%
      \begin{tikzpicture}
        \node[draw=red, line width=1pt, inner sep=0pt] (img5)
        {\includegraphics[width=0.24\textwidth]{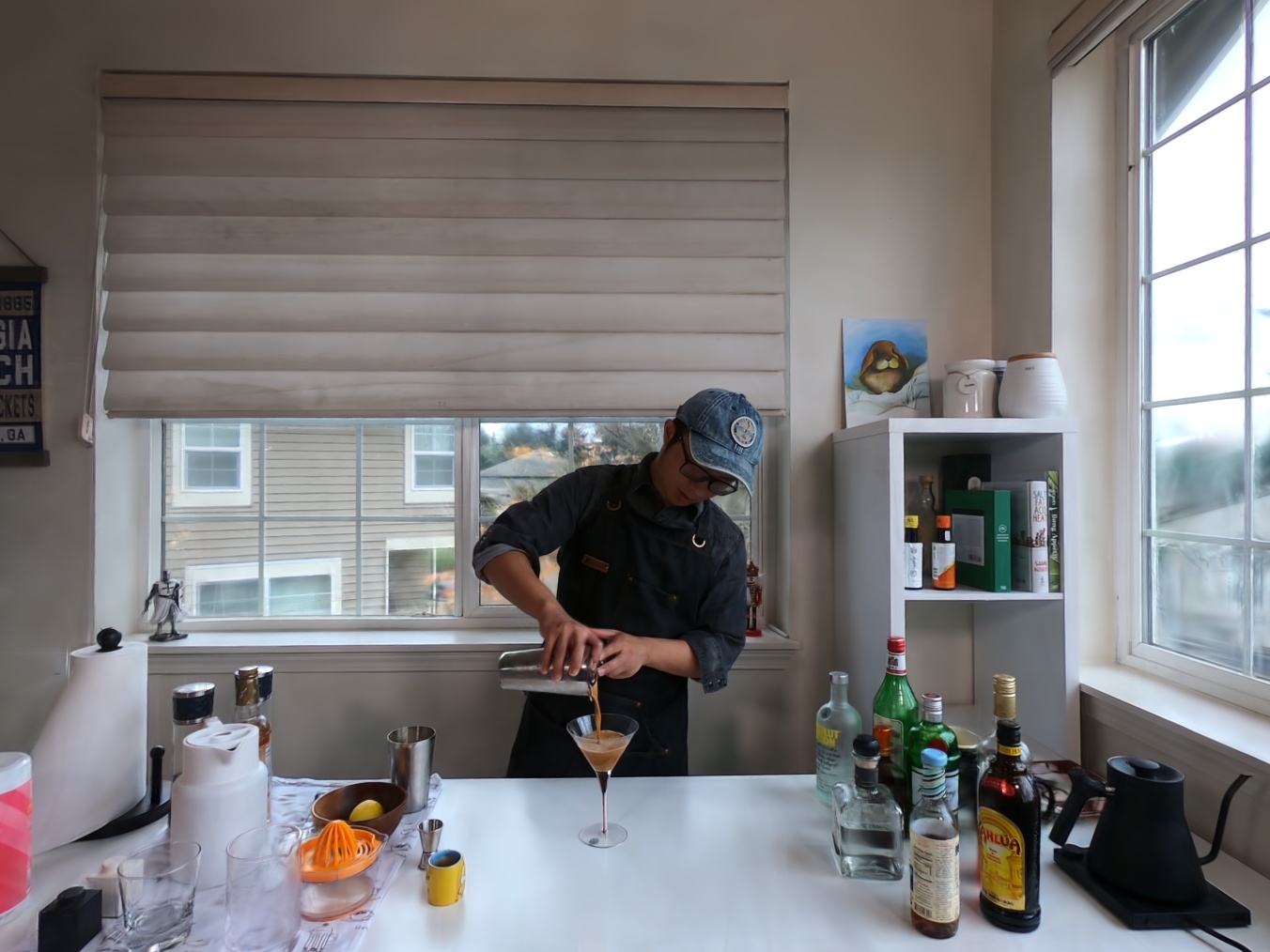}};
      \end{tikzpicture}
   }
   \subfloat[107th - interpolated]{%
      \begin{tikzpicture}
        \node[draw=red, line width=1pt, inner sep=0pt] (img6)
        {\includegraphics[width=0.24\textwidth]{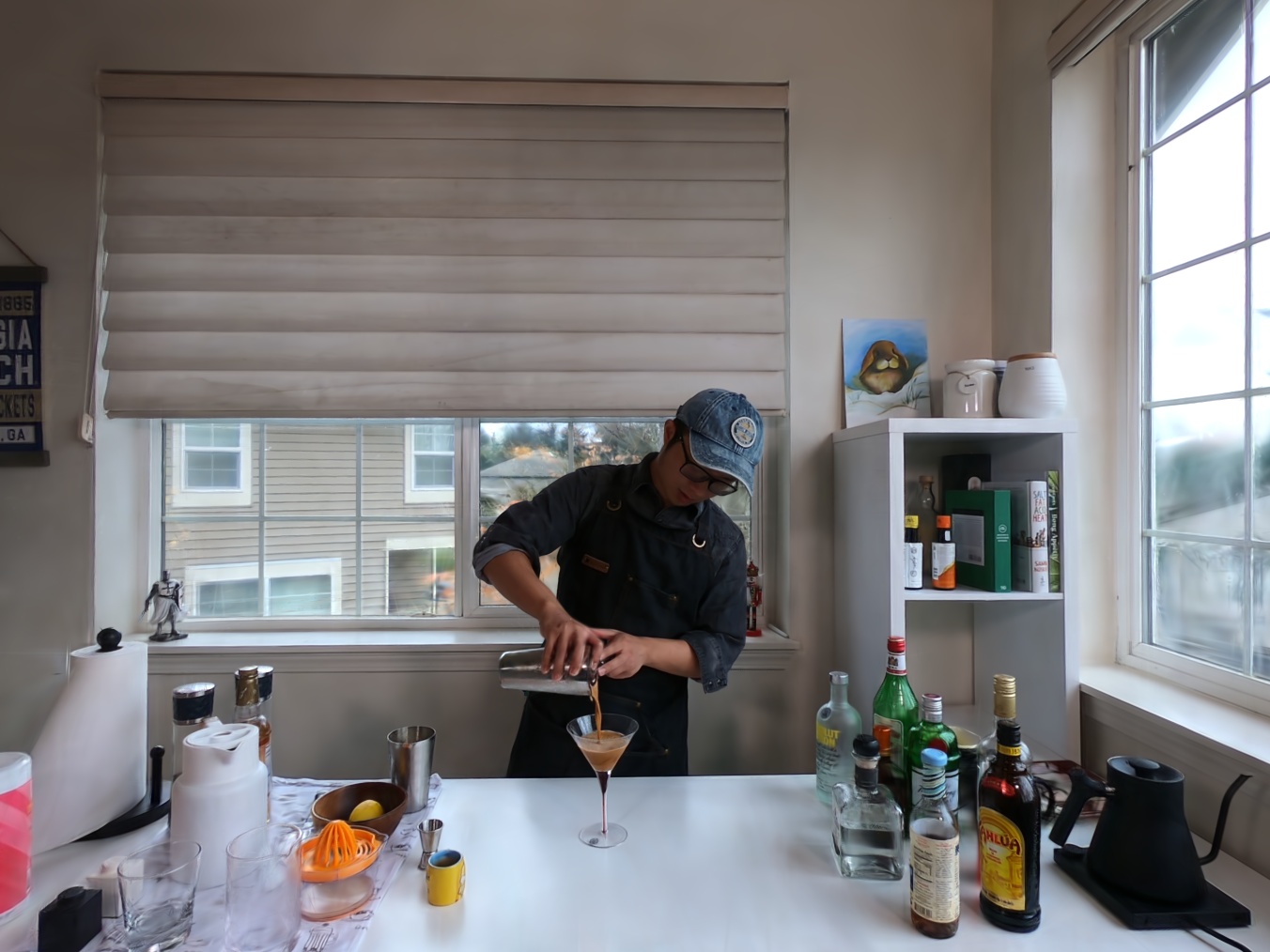}};
      \end{tikzpicture}
   }
   \subfloat[108th - reconstructed]{%
      \begin{tikzpicture}
        \node[draw=green, line width=1pt, inner sep=0pt] (img7)
        {\includegraphics[width=0.24\textwidth]{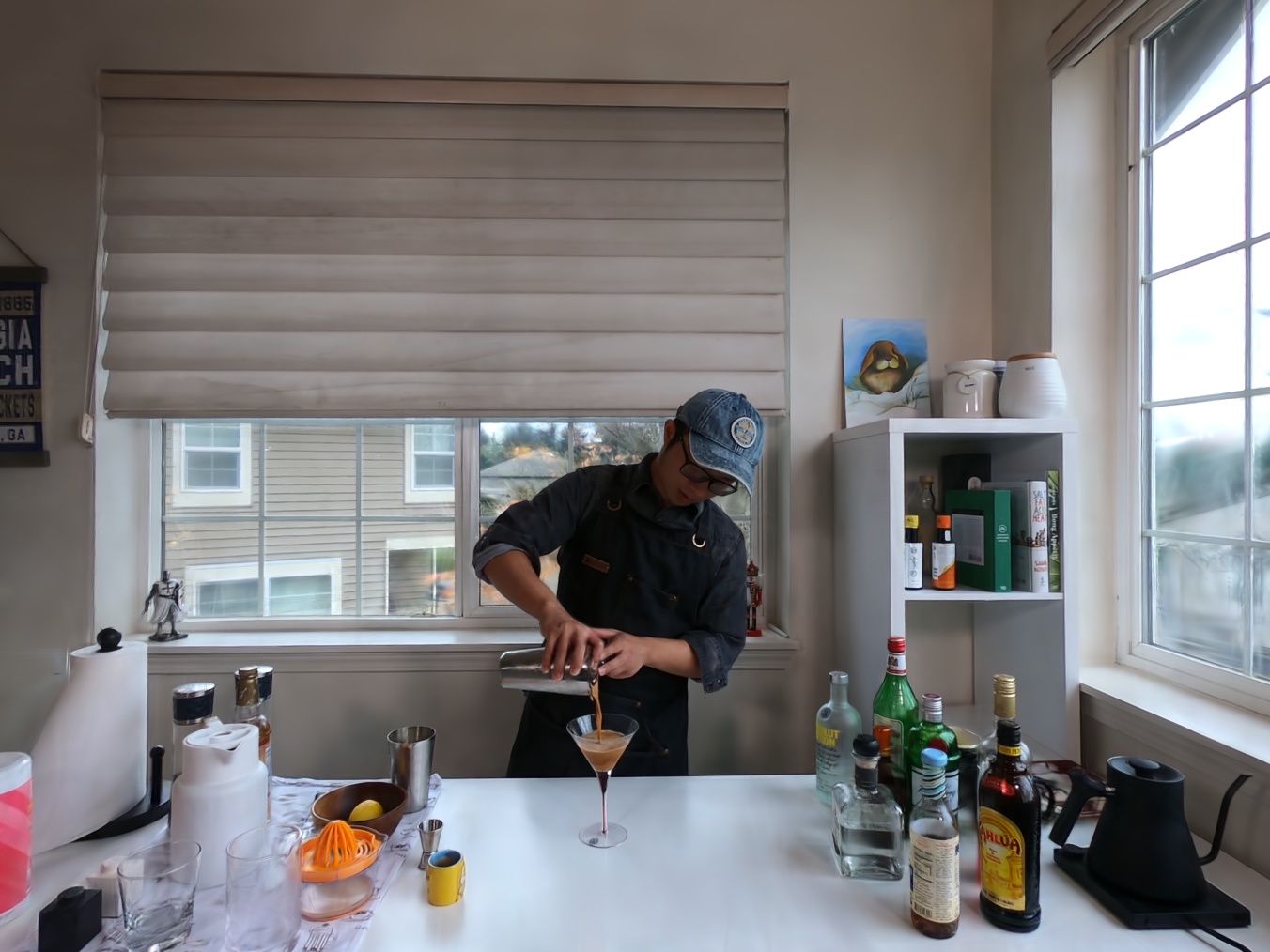}};
      \end{tikzpicture}
   } 
    \subfloat[109th - interpolated]{%
      \begin{tikzpicture}
        \node[draw=red, line width=1pt, inner sep=0pt] (img8)
        {\includegraphics[width=0.24\textwidth]{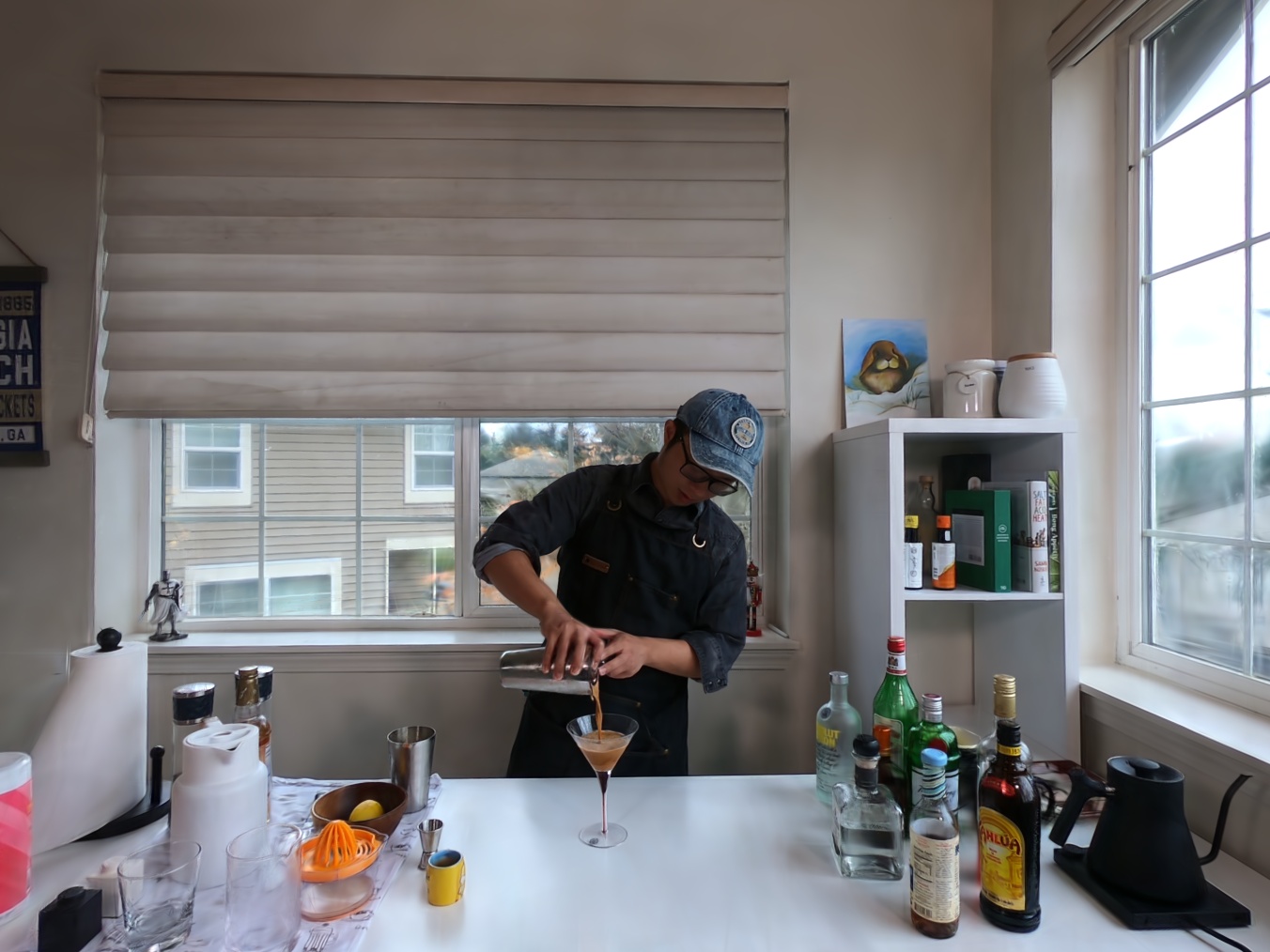}};
      \end{tikzpicture}
   } 
   \caption{Qualitative result of Ours(3DGS)'s frame interpolation on \textit{Coffee Martini} scene of N3DV dataset.}
   \label{fig:flerp-coffee}
   \vspace{-0.2cm}
\end{figure}

\begin{figure}[tp]
   \centering
   \subfloat[261th - reconstructed]{%
      \begin{tikzpicture}
         \node[draw=red, line width=1pt, inner sep=0pt] (img1) {\includegraphics[width=0.24\textwidth]{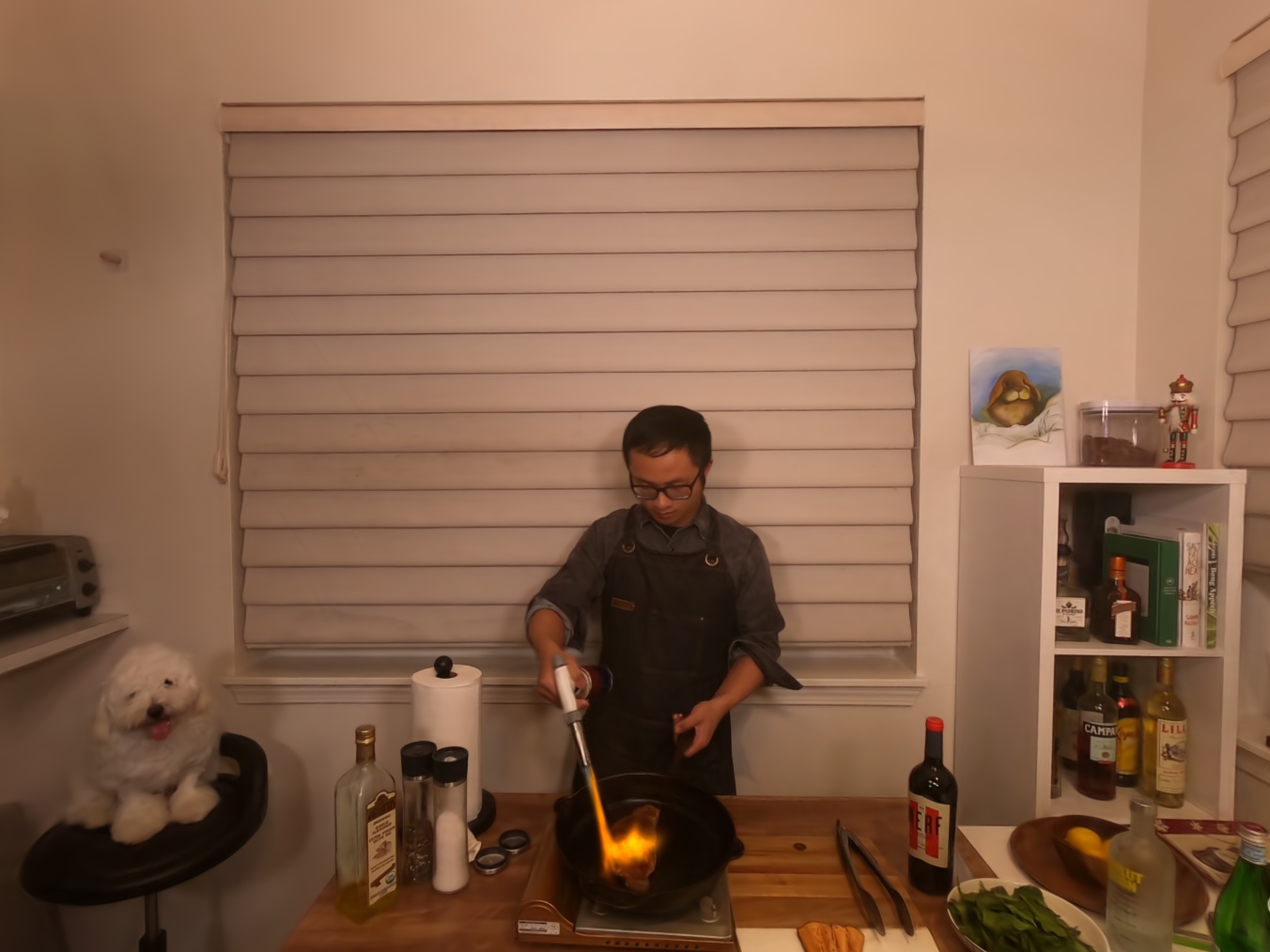}};
      \end{tikzpicture}
   }
   \subfloat[262th - interpolated]{%
      \begin{tikzpicture}
         \node[draw=green, line width=1pt, inner sep=0pt] (img2) {\includegraphics[width=0.24\textwidth]{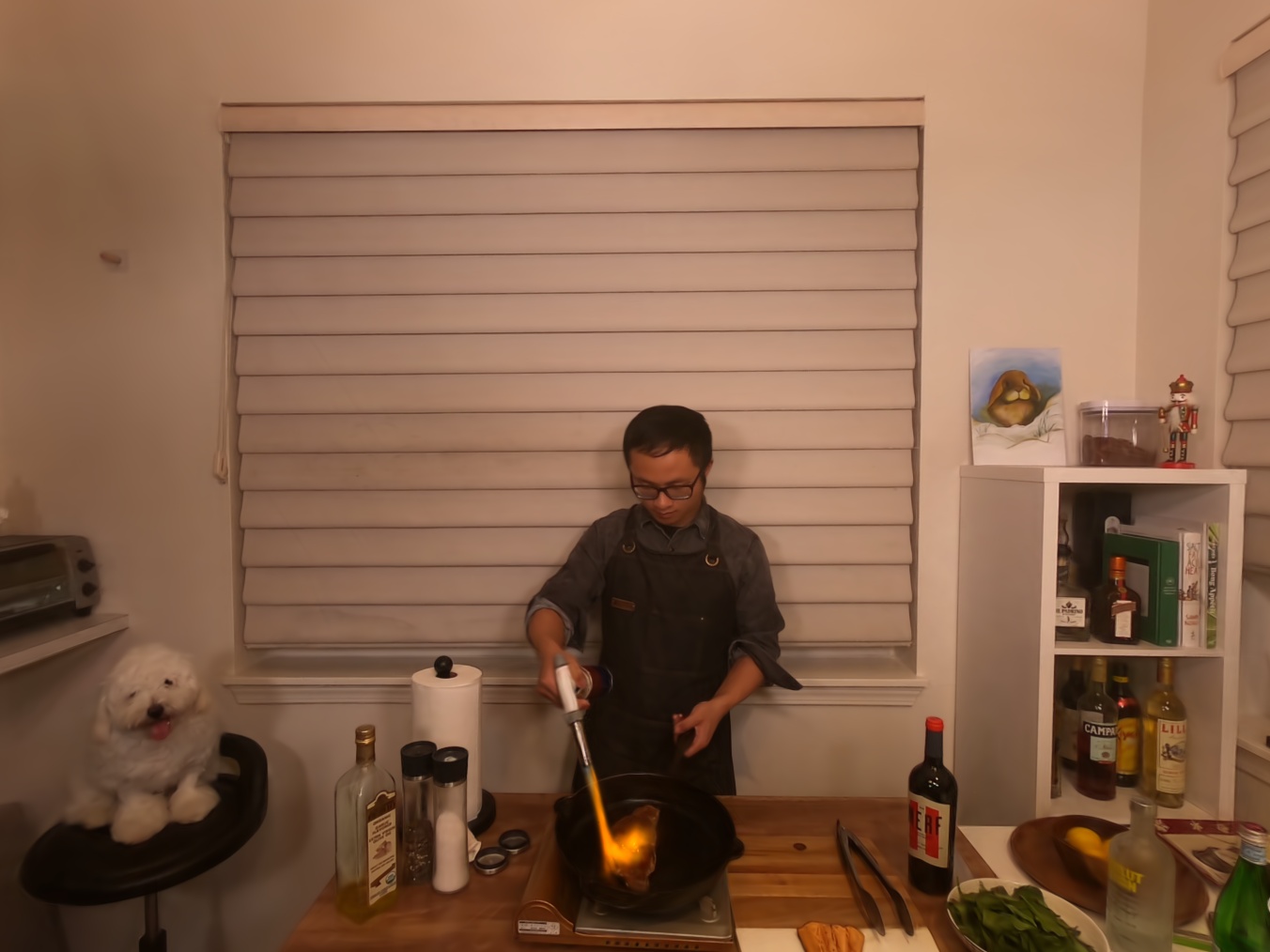}};
      \end{tikzpicture}
   }
   \subfloat[263th - reconstructed]{%
      \begin{tikzpicture}
         \node[draw=red, line width=1pt, inner sep=0pt] (img3) {\includegraphics[width=0.24\textwidth]{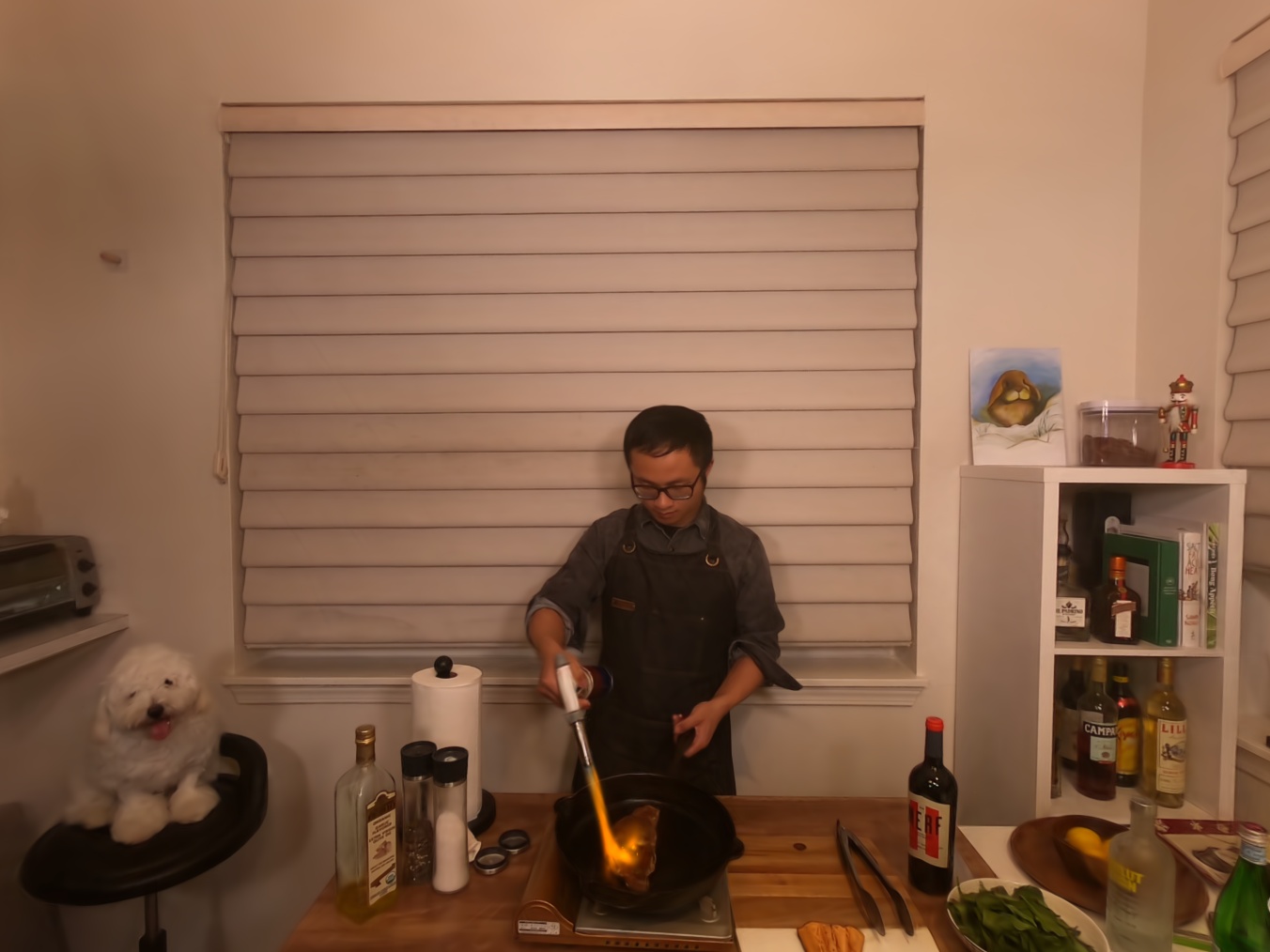}};
      \end{tikzpicture}
   } 
   \subfloat[264th - interpolated]{%
      \begin{tikzpicture}
         \node[draw=green, line width=1pt, inner sep=0pt] (img4) {\includegraphics[width=0.24\textwidth]{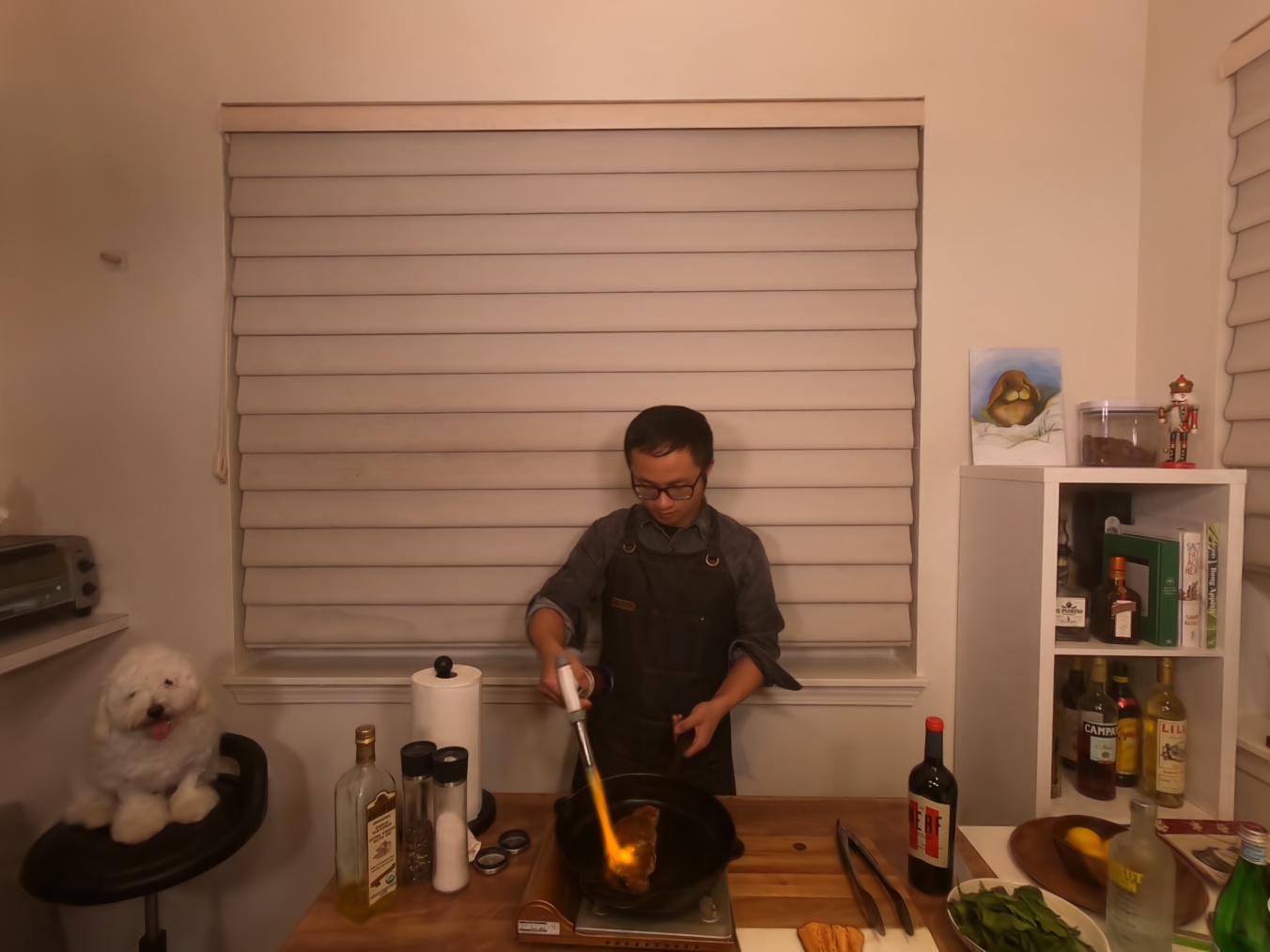}};
      \end{tikzpicture}
   } 
   \qquad
    \subfloat[261th - reconstructed]{%
      \begin{tikzpicture}
        \node[draw=green, line width=1pt, inner sep=0pt] (img5)
        {\includegraphics[width=0.24\textwidth]{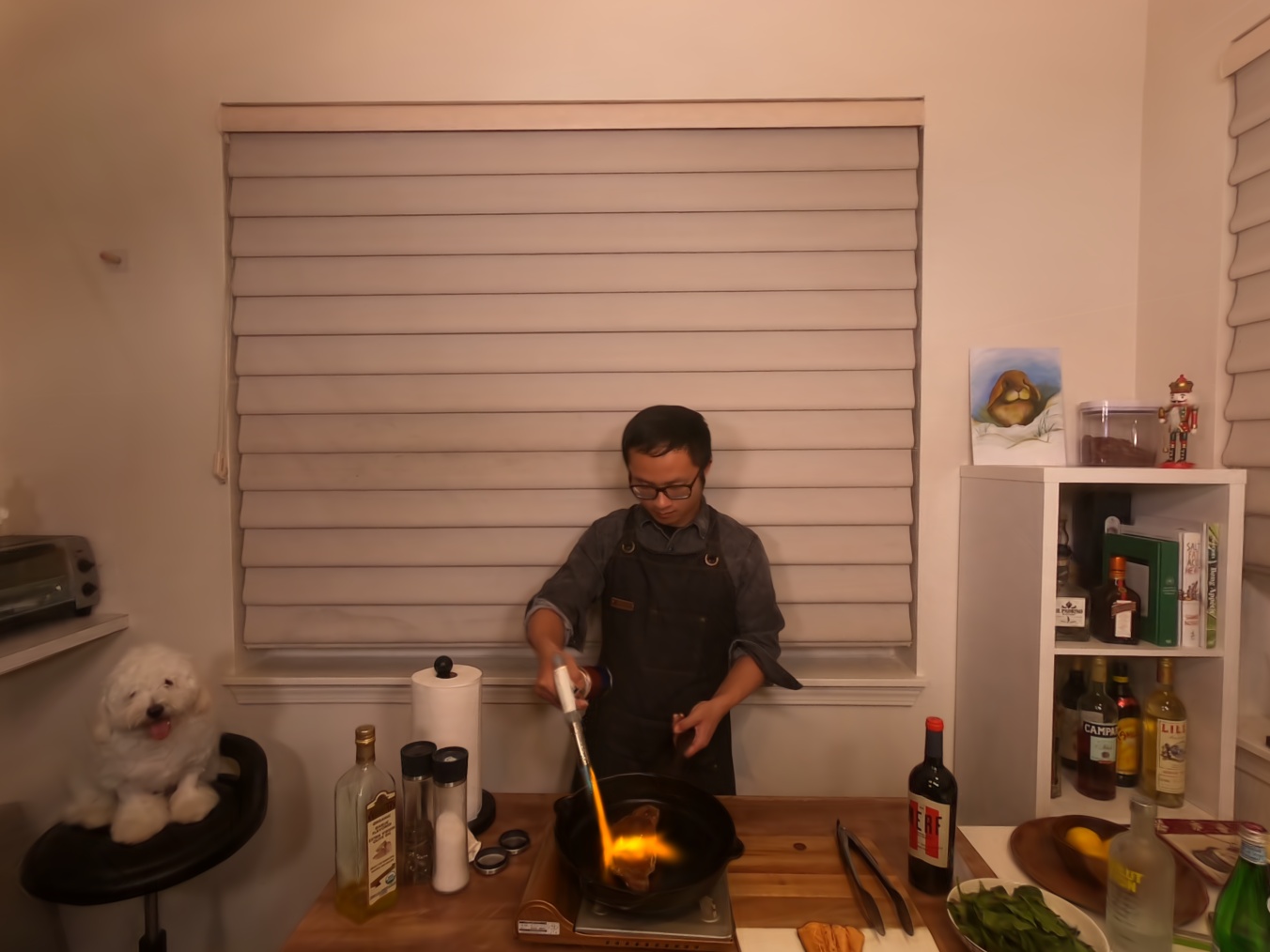}};
      \end{tikzpicture}
   }
   \subfloat[262th - interpolated]{%
      \begin{tikzpicture}
        \node[draw=red, line width=1pt, inner sep=0pt] (img6)
        {\includegraphics[width=0.24\textwidth]{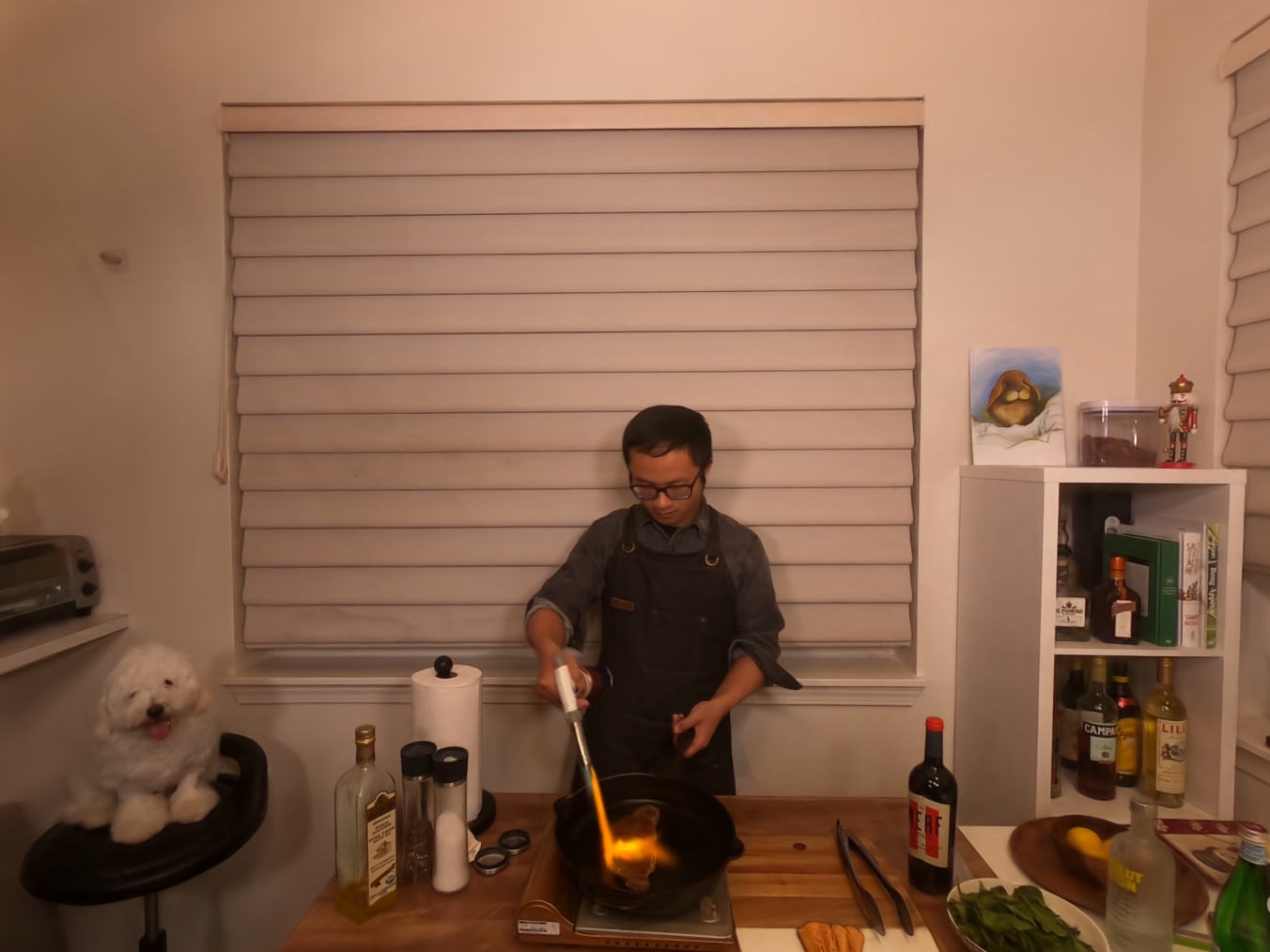}};
      \end{tikzpicture}
   }
   \subfloat[263th - interpolated]{%
      \begin{tikzpicture}
        \node[draw=red, line width=1pt, inner sep=0pt] (img7)
        {\includegraphics[width=0.24\textwidth]{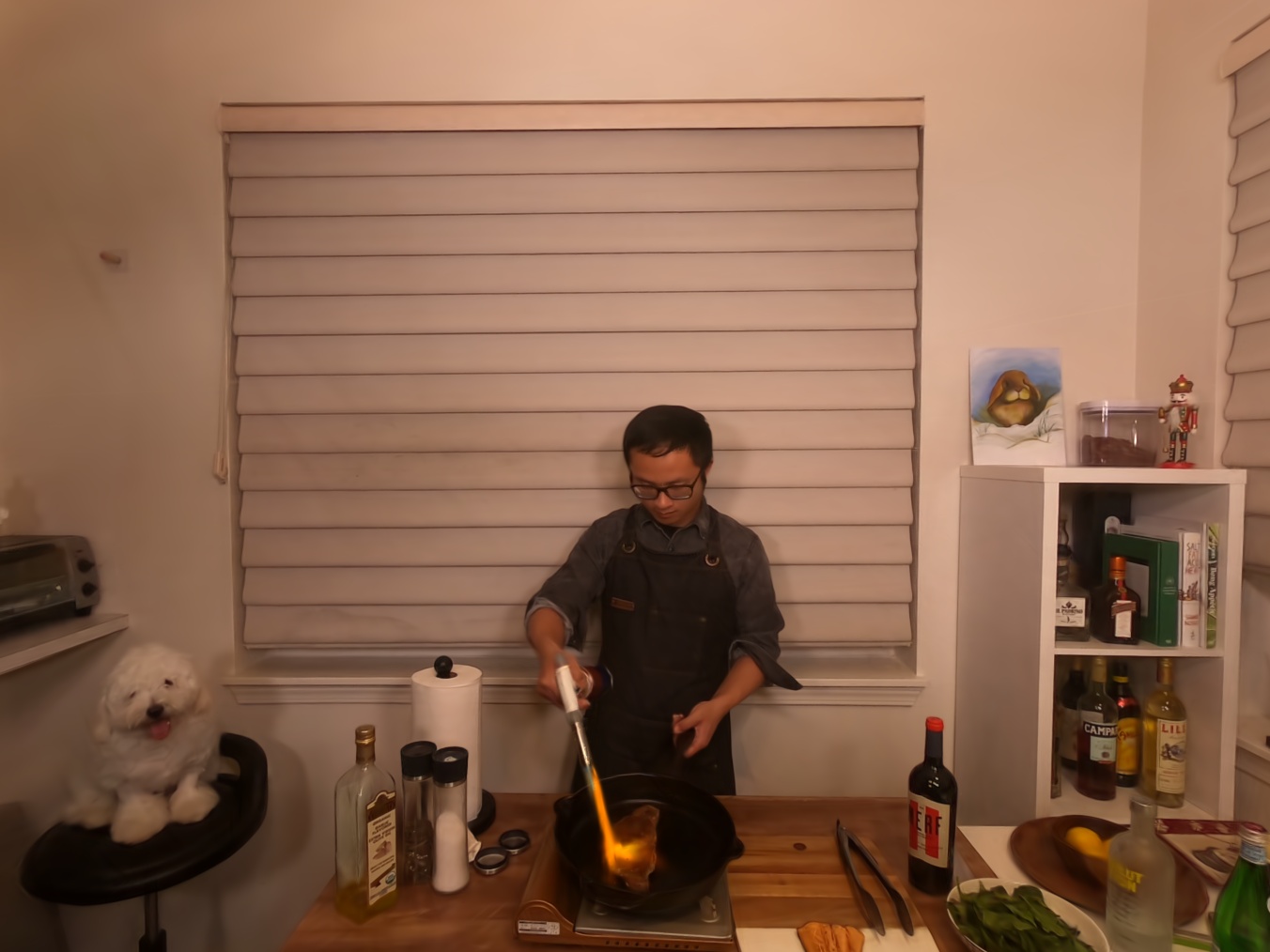}};
      \end{tikzpicture}
   } 
    \subfloat[264th - interpolated]{%
      \begin{tikzpicture}
        \node[draw=red, line width=1pt, inner sep=0pt] (img8)
        {\includegraphics[width=0.24\textwidth]{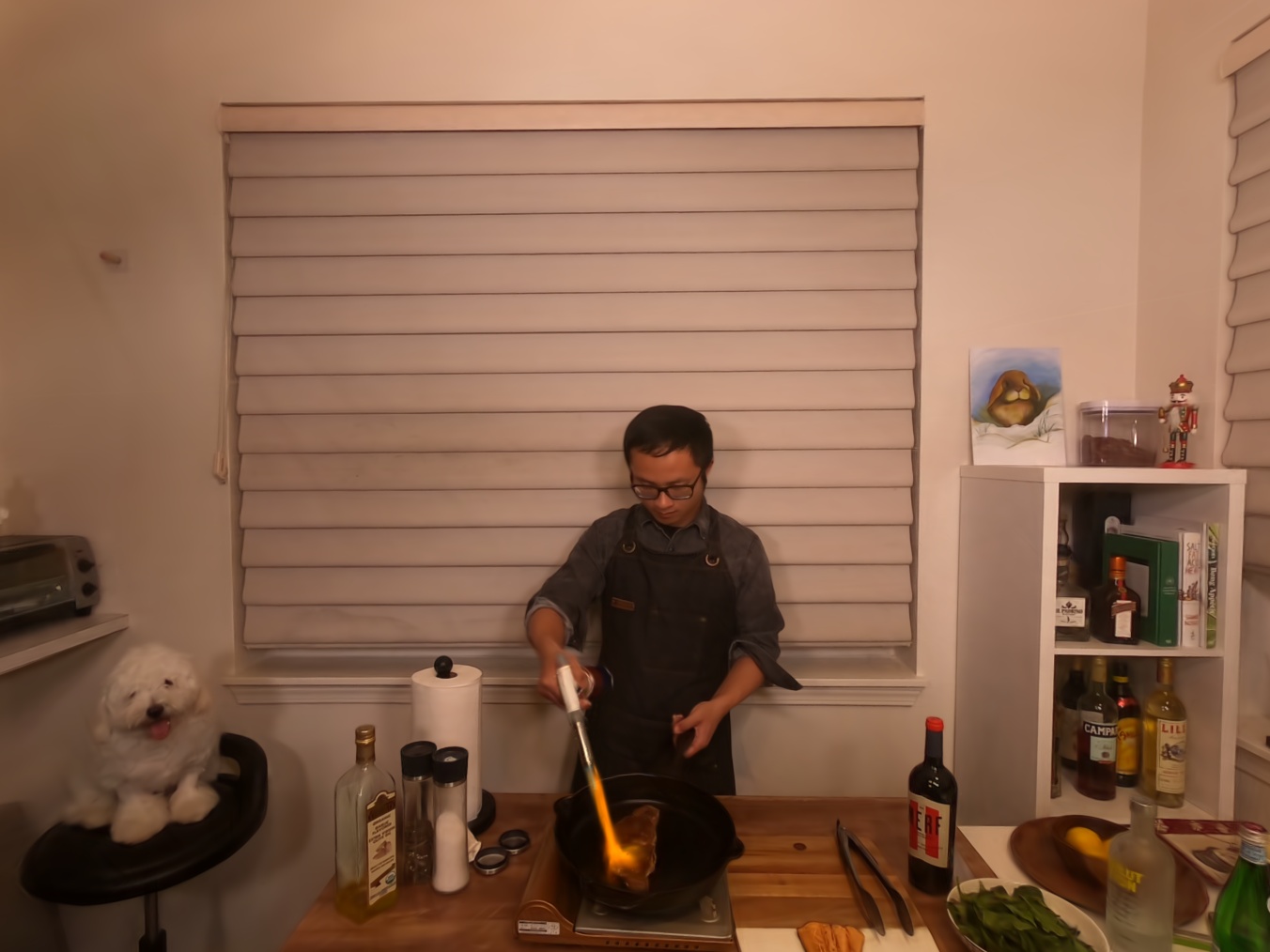}};
      \end{tikzpicture}
   } 
   \caption{Qualitative result of Ours(Scaffold-GS)'s frame interpolation on \textit{Flame Steak} scene of N3DV dataset.}
   \label{fig:flerp-steak}
   \vspace{-0.2cm}
\end{figure}

As shown in Fig.~\ref{fig:flerp-coffee} and Fig.~\ref{fig:flerp-steak}, subfigures (a) to (d) are the results of interpolating one frame, and subfigures (e) to (h) are the results of interpolating two frames. Ours(3DGS) and Ours(Scaffold-GS) successfully interpolated the motion in the dynamic scenes. It is worth noting that we rely solely on the interpolation-based deformation field to achieve frame interpolation, without using any external priors. Therefore, there is still room for further improvement in interpolation performance, which we will explore further in the future.

\begin{figure}[tp]
   \centering
   \subfloat[70th - reconstructed]{%
      \begin{tikzpicture}
         \node[draw=red, line width=1pt, inner sep=0pt] (img1) {\includegraphics[width=0.24\textwidth]{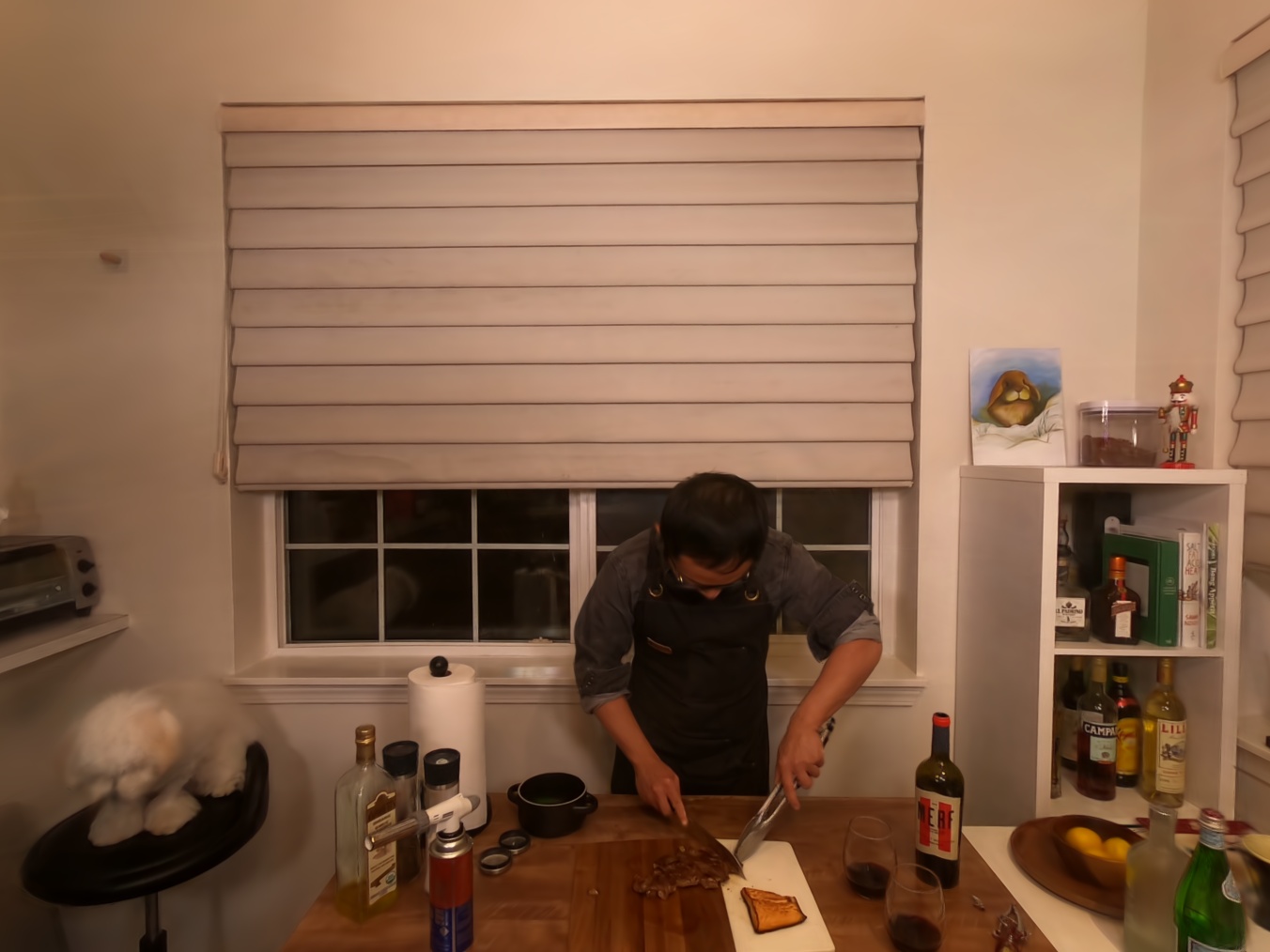}};
      \end{tikzpicture}
   }
   \subfloat[71th - interpolated]{%
      \begin{tikzpicture}
         \node[draw=green, line width=1pt, inner sep=0pt] (img2) {\includegraphics[width=0.24\textwidth]{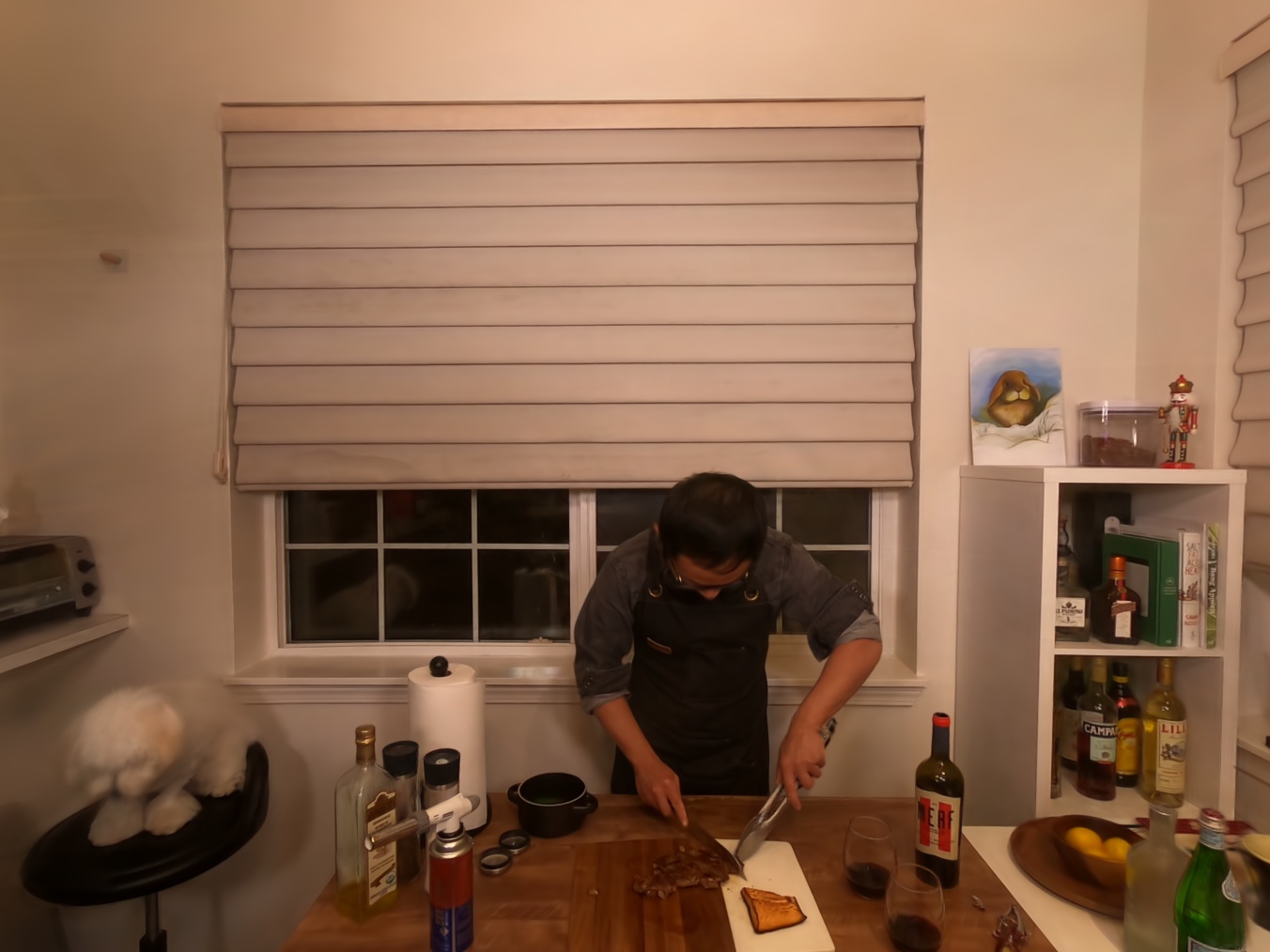}};
      \end{tikzpicture}
   }
   \subfloat[72th - reconstructed]{%
      \begin{tikzpicture}
         \node[draw=red, line width=1pt, inner sep=0pt] (img3) {\includegraphics[width=0.24\textwidth]{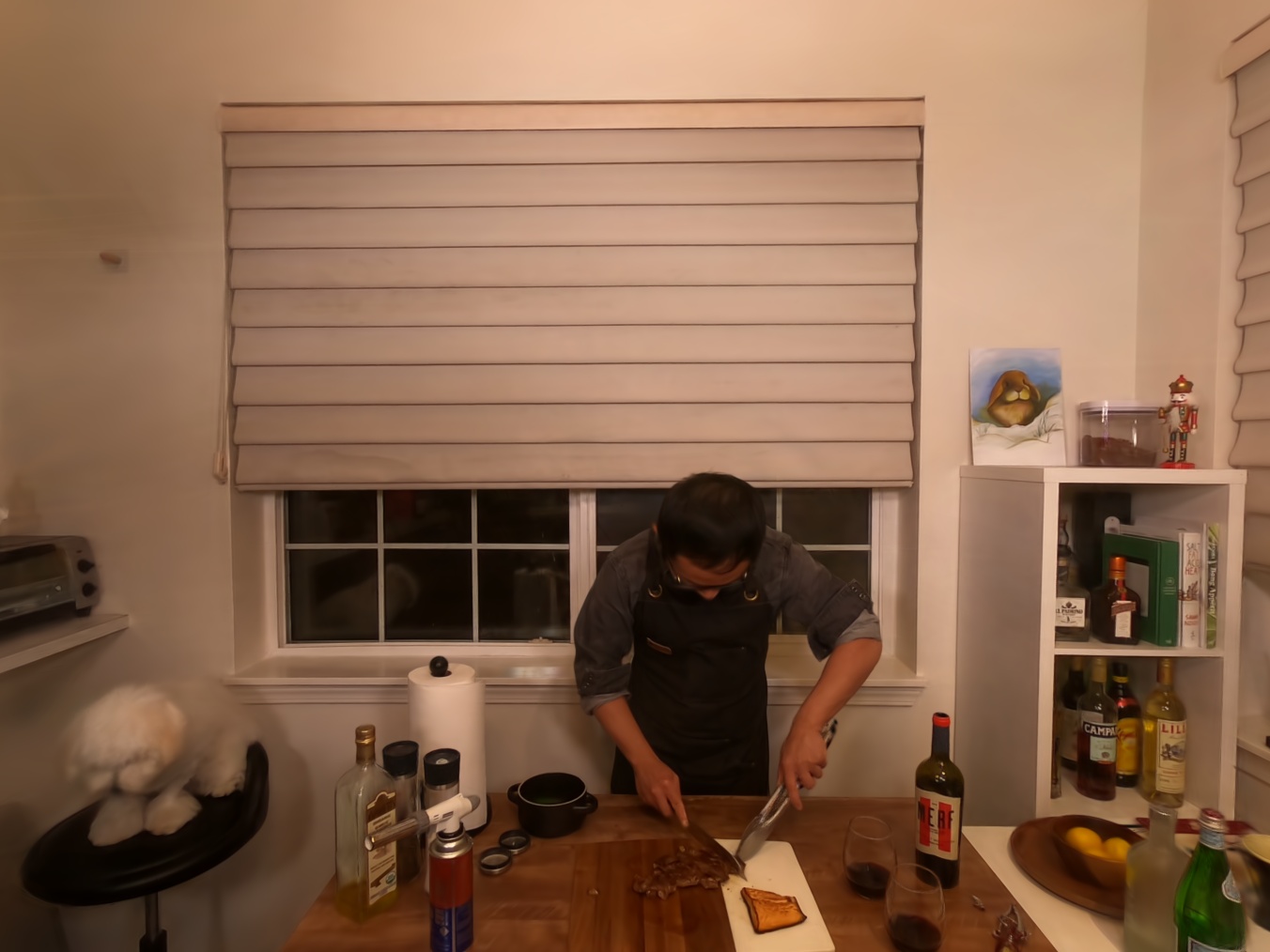}};
      \end{tikzpicture}
   } 
   \subfloat[73th - interpolated]{%
      \begin{tikzpicture}
         \node[draw=green, line width=1pt, inner sep=0pt] (img4) {\includegraphics[width=0.24\textwidth]{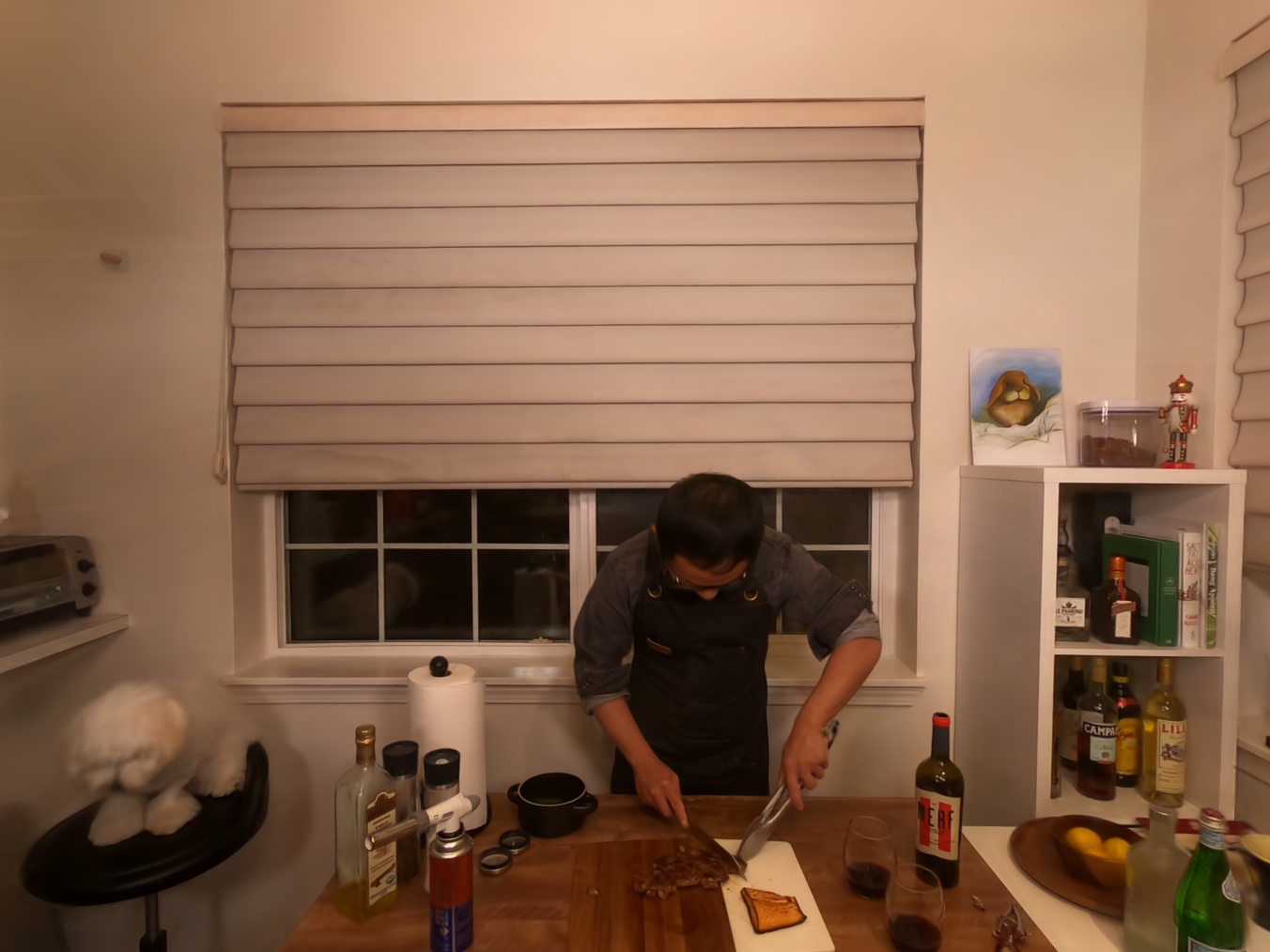}};
      \end{tikzpicture}
   } 
   \qquad
    \subfloat[70th - reconstructed]{%
      \begin{tikzpicture}
        \node[draw=green, line width=1pt, inner sep=0pt] (img5)
        {\includegraphics[width=0.24\textwidth]{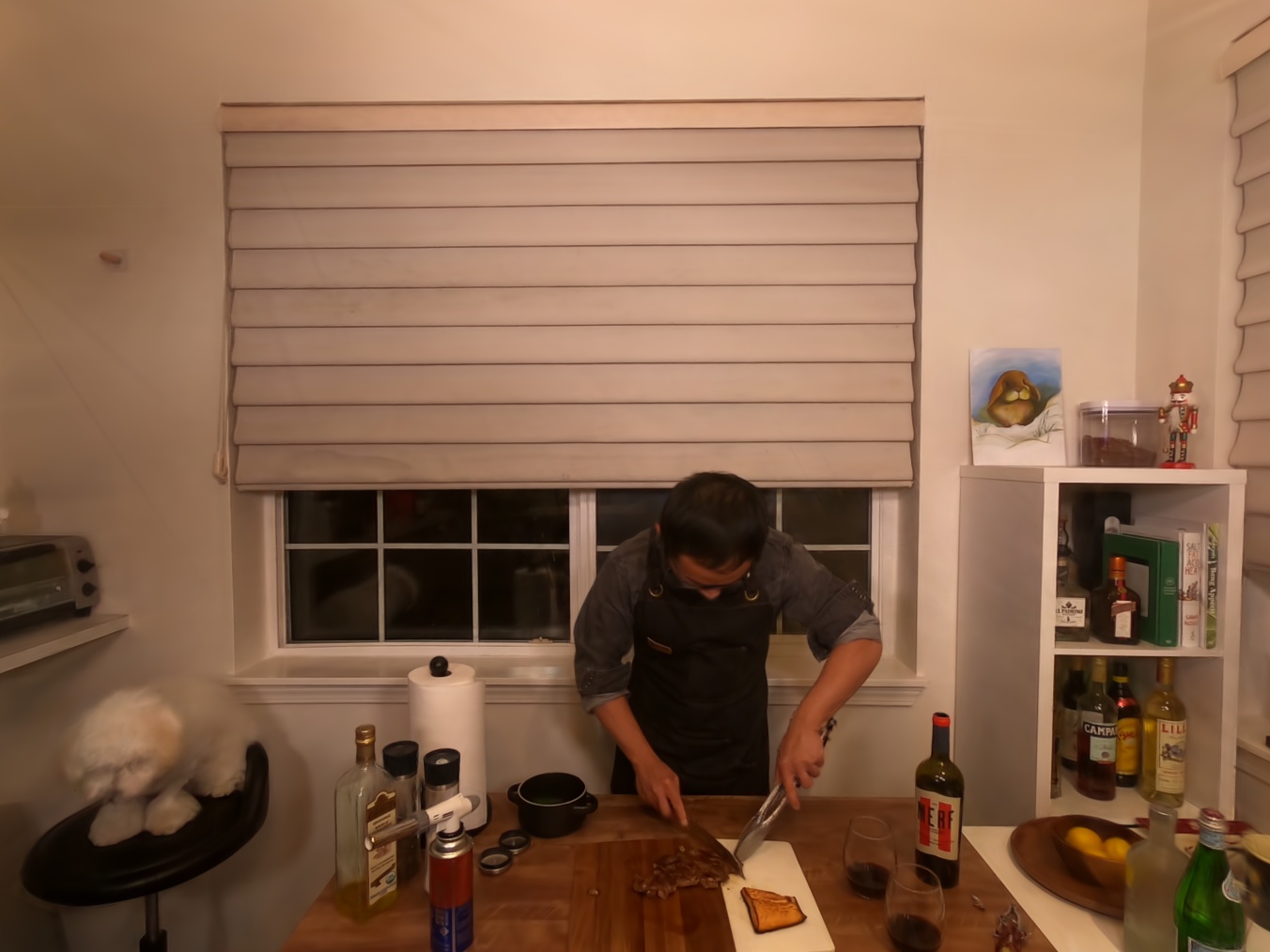}};
      \end{tikzpicture}
   }
   \subfloat[71th - interpolated]{%
      \begin{tikzpicture}
        \node[draw=red, line width=1pt, inner sep=0pt] (img6)
        {\includegraphics[width=0.24\textwidth]{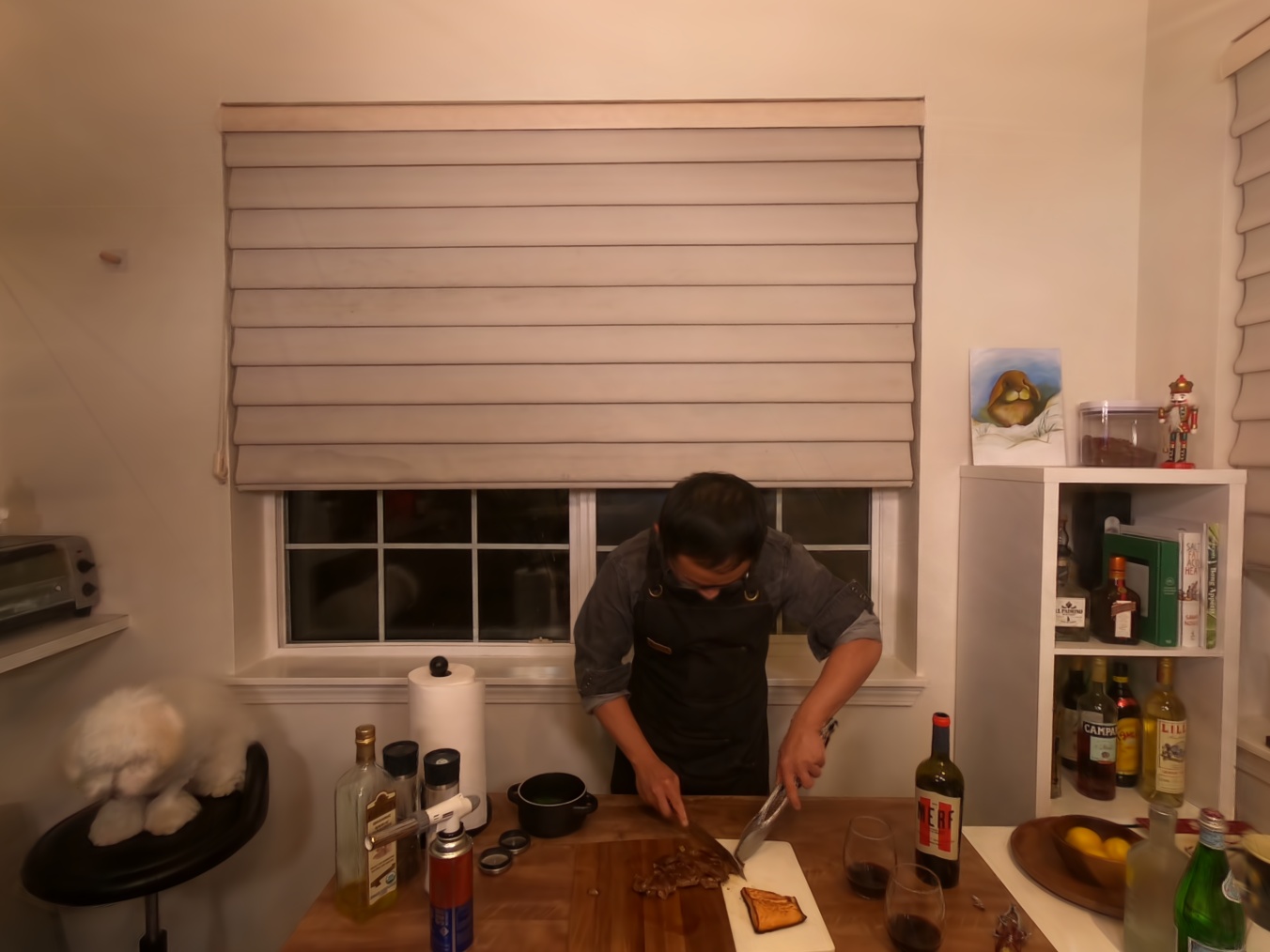}};
      \end{tikzpicture}
   }
   \subfloat[72th - interpolated]{%
      \begin{tikzpicture}
        \node[draw=red, line width=1pt, inner sep=0pt] (img7)
        {\includegraphics[width=0.24\textwidth]{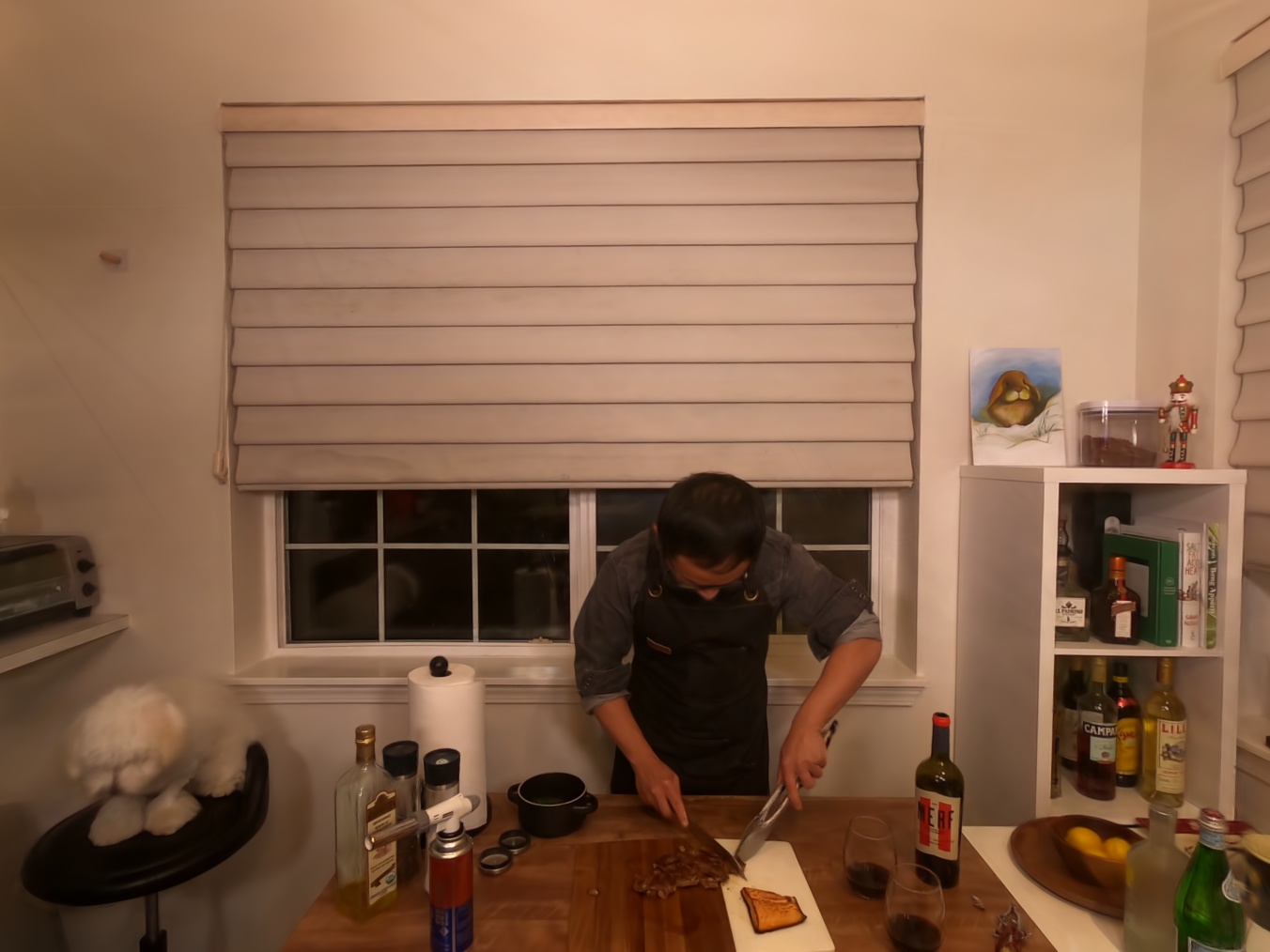}};
      \end{tikzpicture}
   } 
    \subfloat[73th - interpolated]{%
      \begin{tikzpicture}
        \node[draw=red, line width=1pt, inner sep=0pt] (img8)
        {\includegraphics[width=0.24\textwidth]{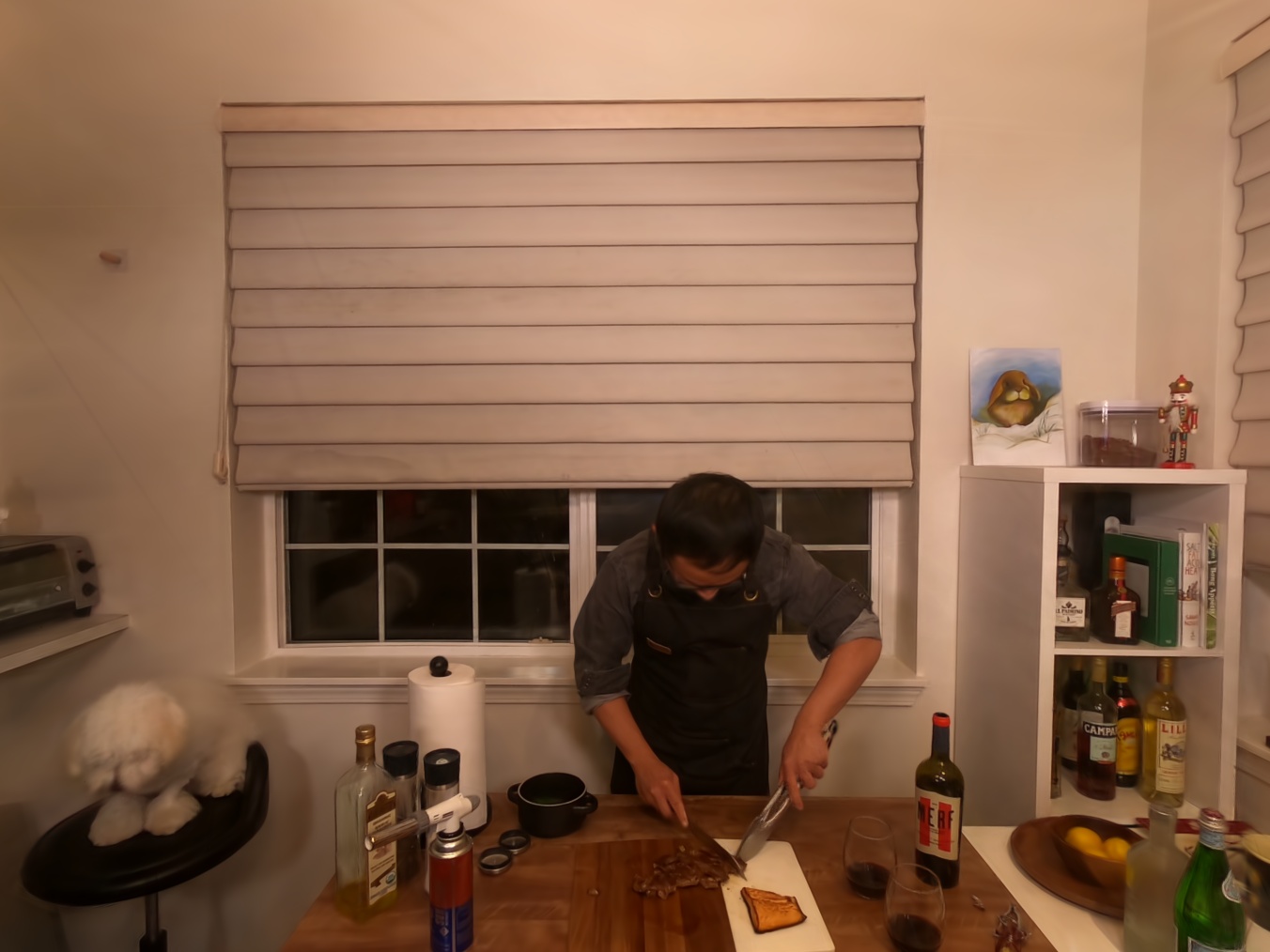}};
      \end{tikzpicture}
   } 
   \caption{Qualitative result of Ours(3DGS)'s frame interpolation on \textit{Cut Roasted Beef} scene of N3DV dataset.}
   \label{fig:flerp-cut}
   \vspace{-0.2cm}
\end{figure}

\subsubsection{FVV Qualitative Comparisons}
\begin{figure}[!t]
   \centering

   \subfloat[StreamSTGS~\cite{ke2025streamstgs}]{
      \includegraphics[width=0.24\textwidth]{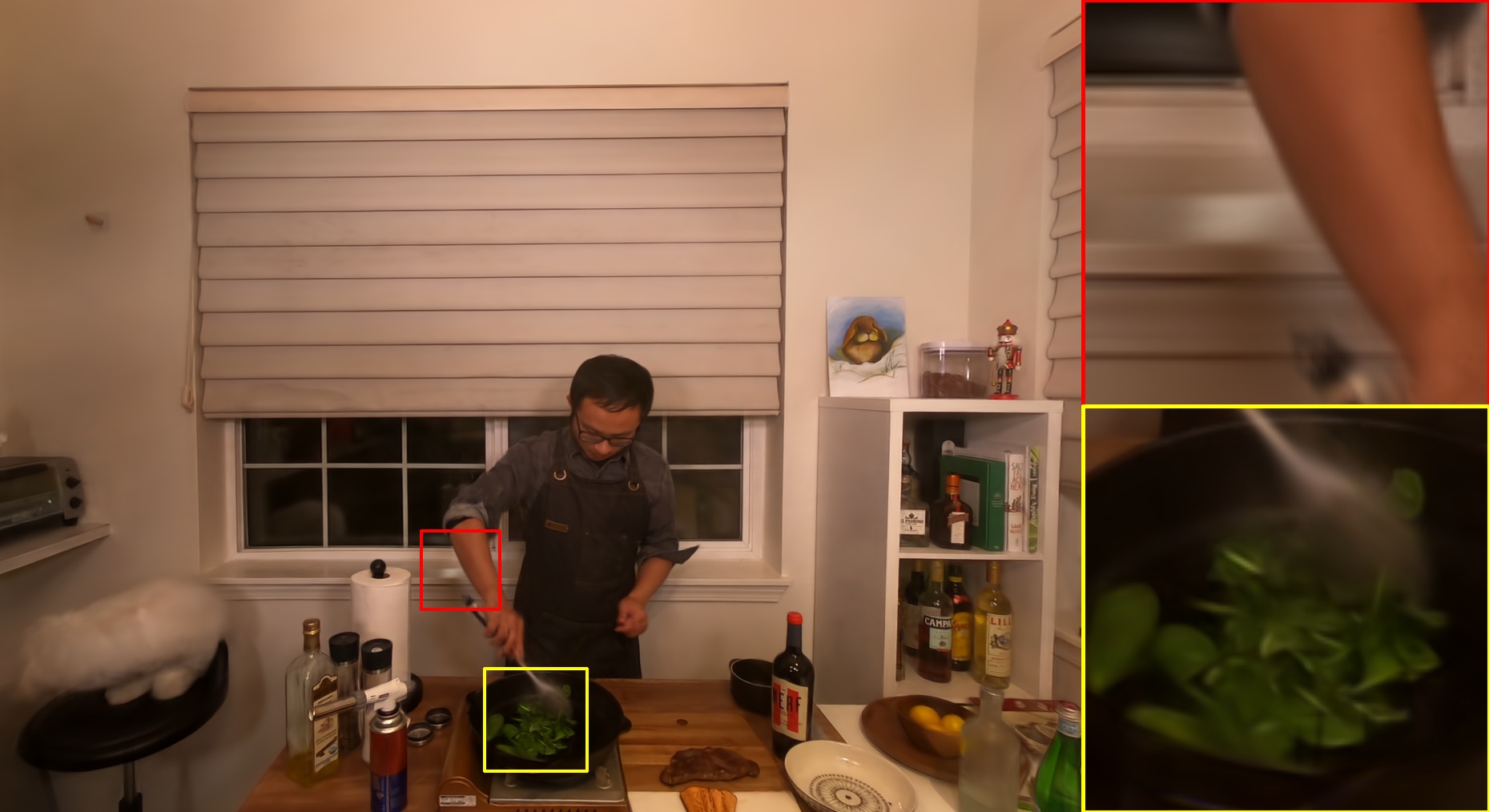}
   }
   \subfloat[QUEEN~\cite{girish2024queen}]{
      \includegraphics[width=0.24\textwidth]{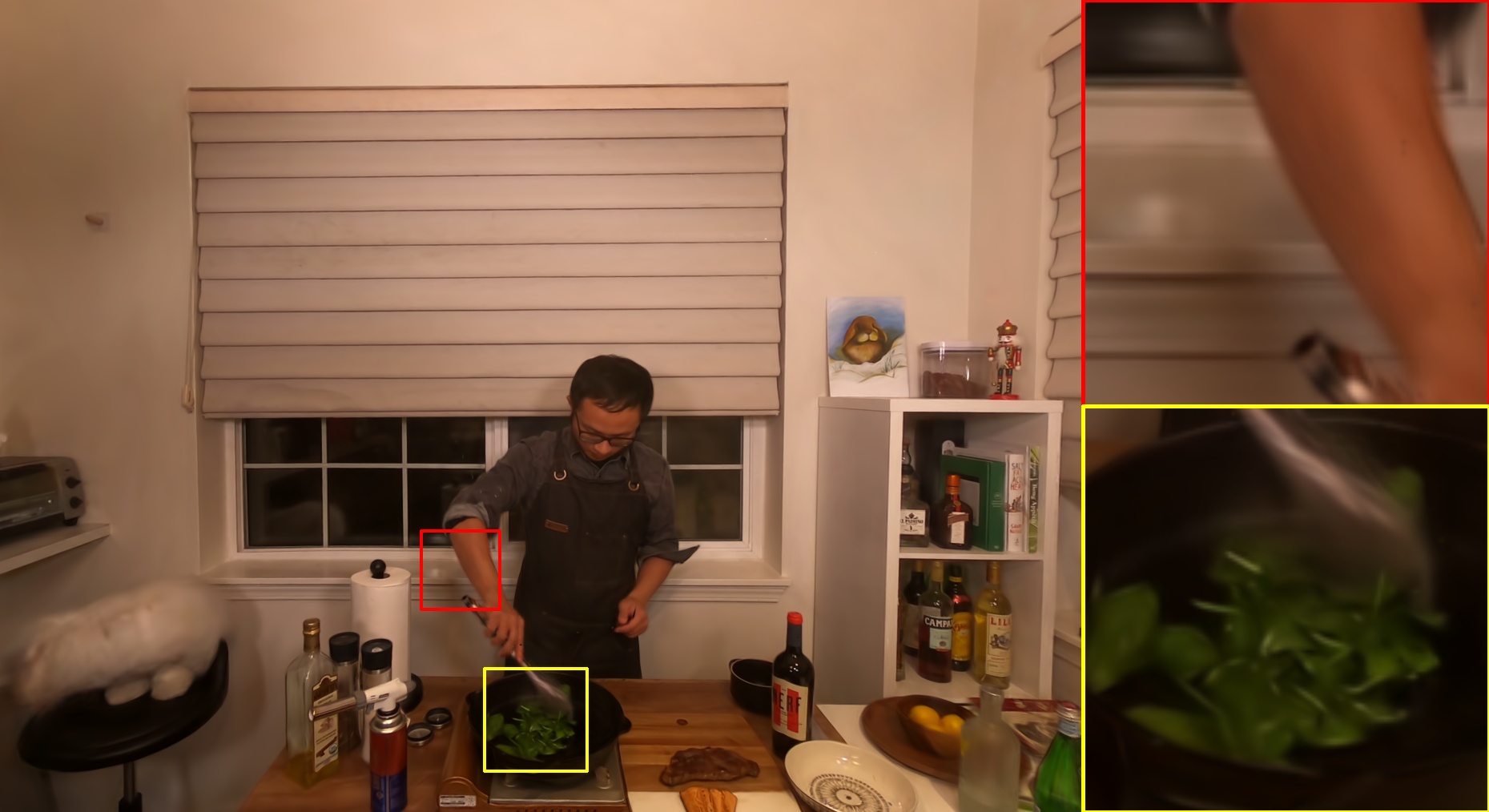}
   }
   \subfloat[HiCoM~\cite{gao2024hicom}]{
      \includegraphics[width=0.24\textwidth]{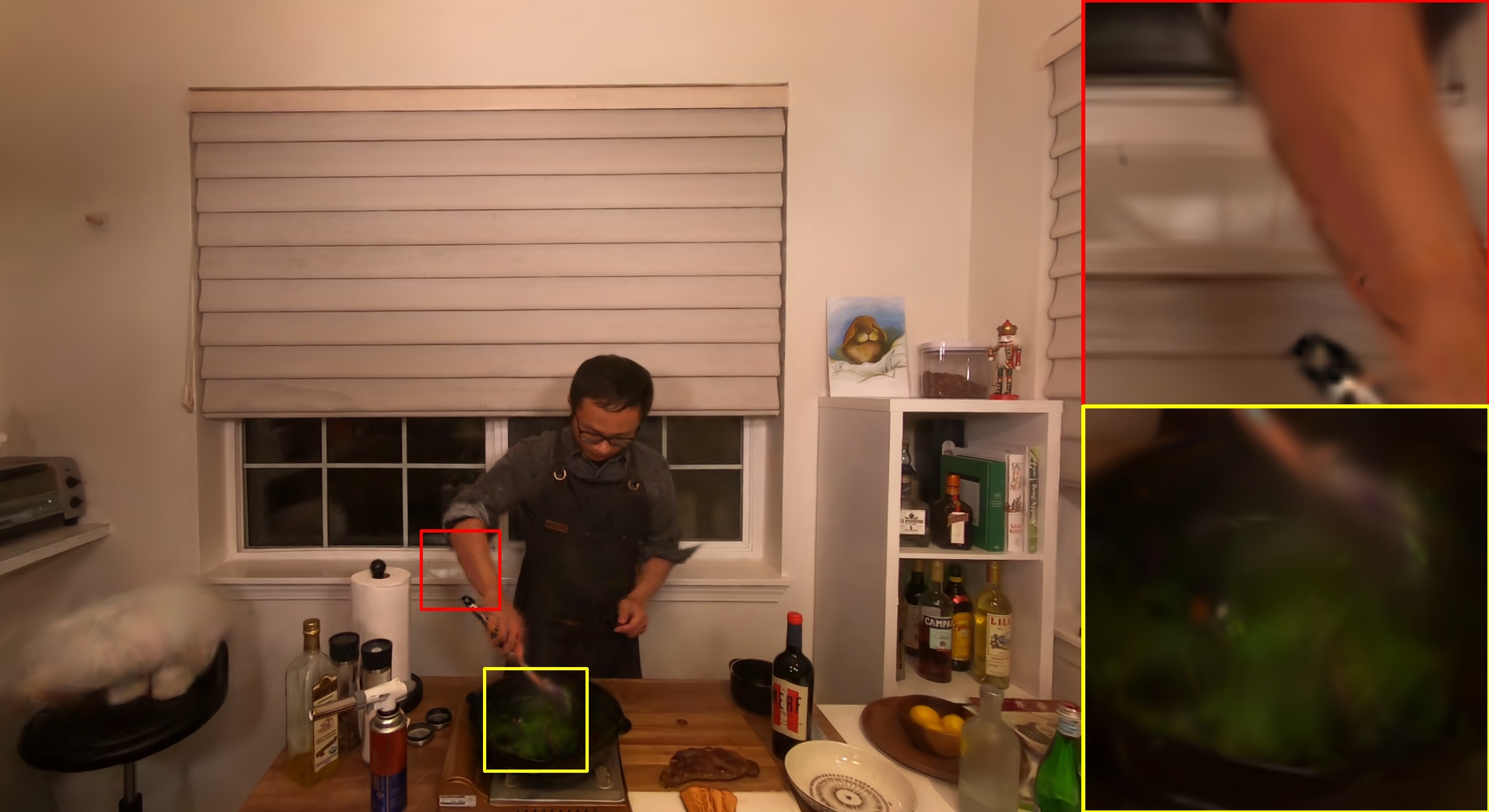}
   }
   \subfloat[Ours (3DGS)]{
      \includegraphics[width=0.24\textwidth]{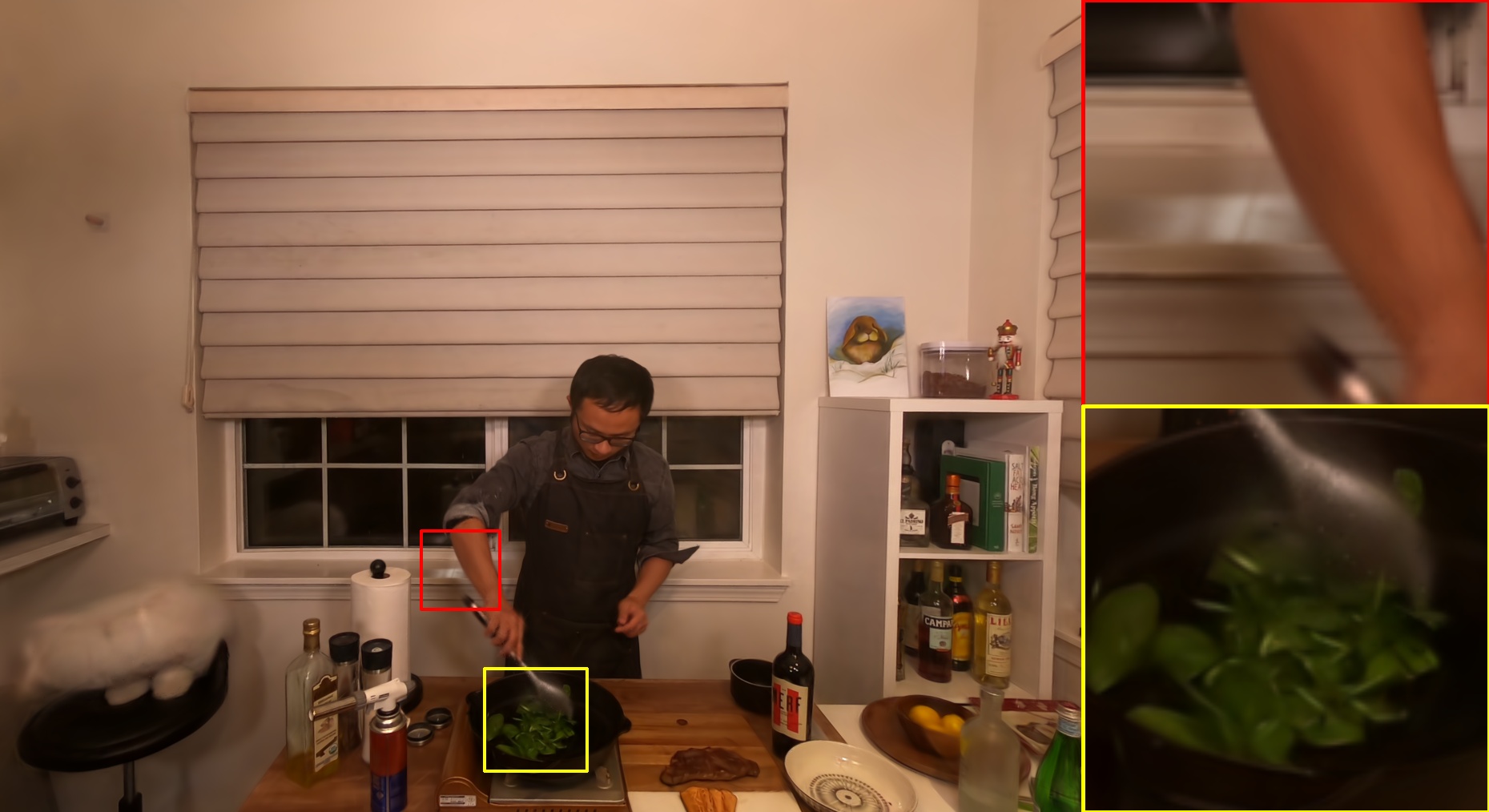}
   }
   \\[-0.2em] 


   \subfloat[GIFStream~\cite{li2025gifstream}]{
      \includegraphics[width=0.24\textwidth]{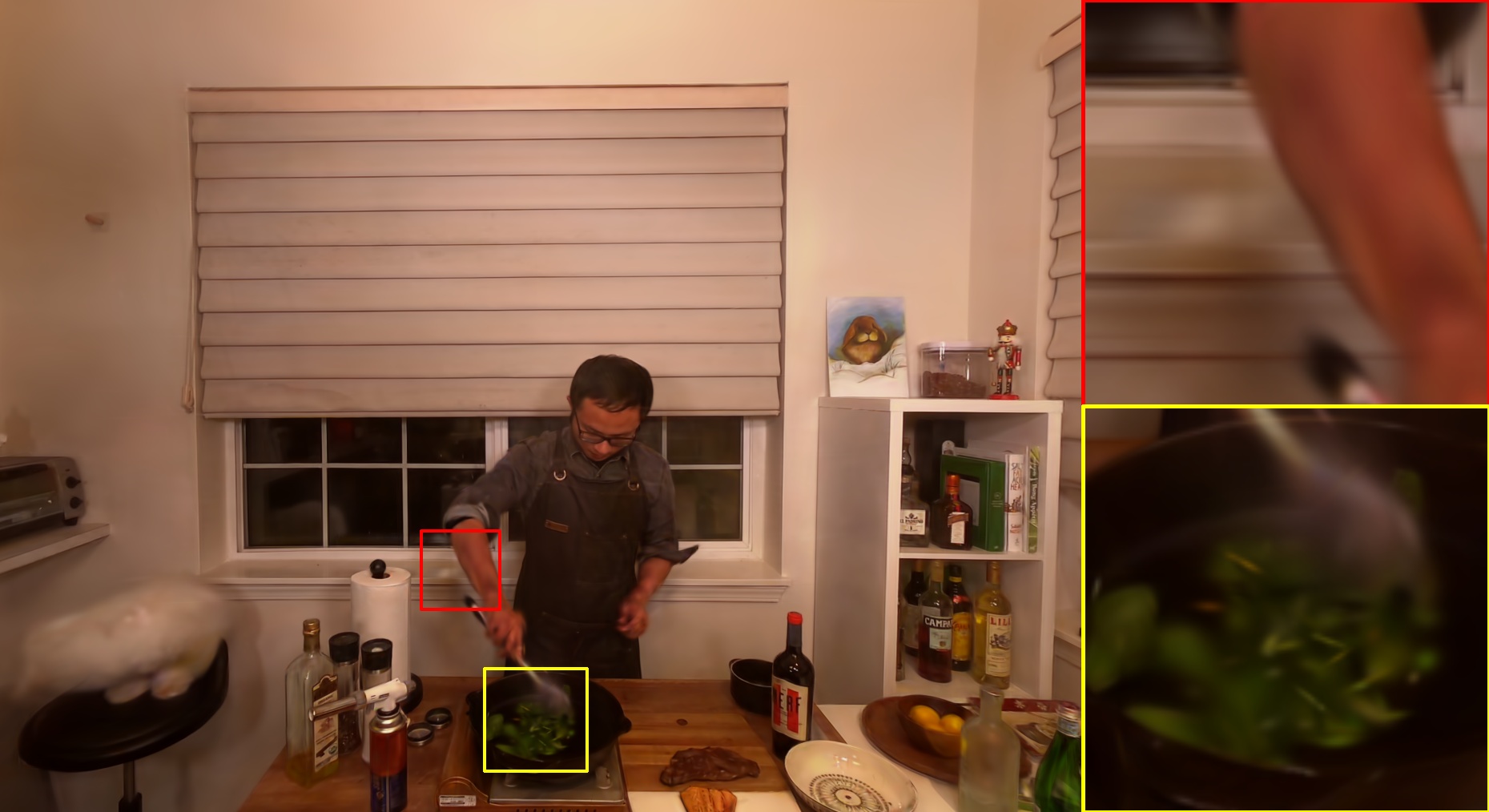}
   }
   \subfloat[iFVC~\cite{tang2025compressing}]{
      \includegraphics[width=0.24\textwidth]{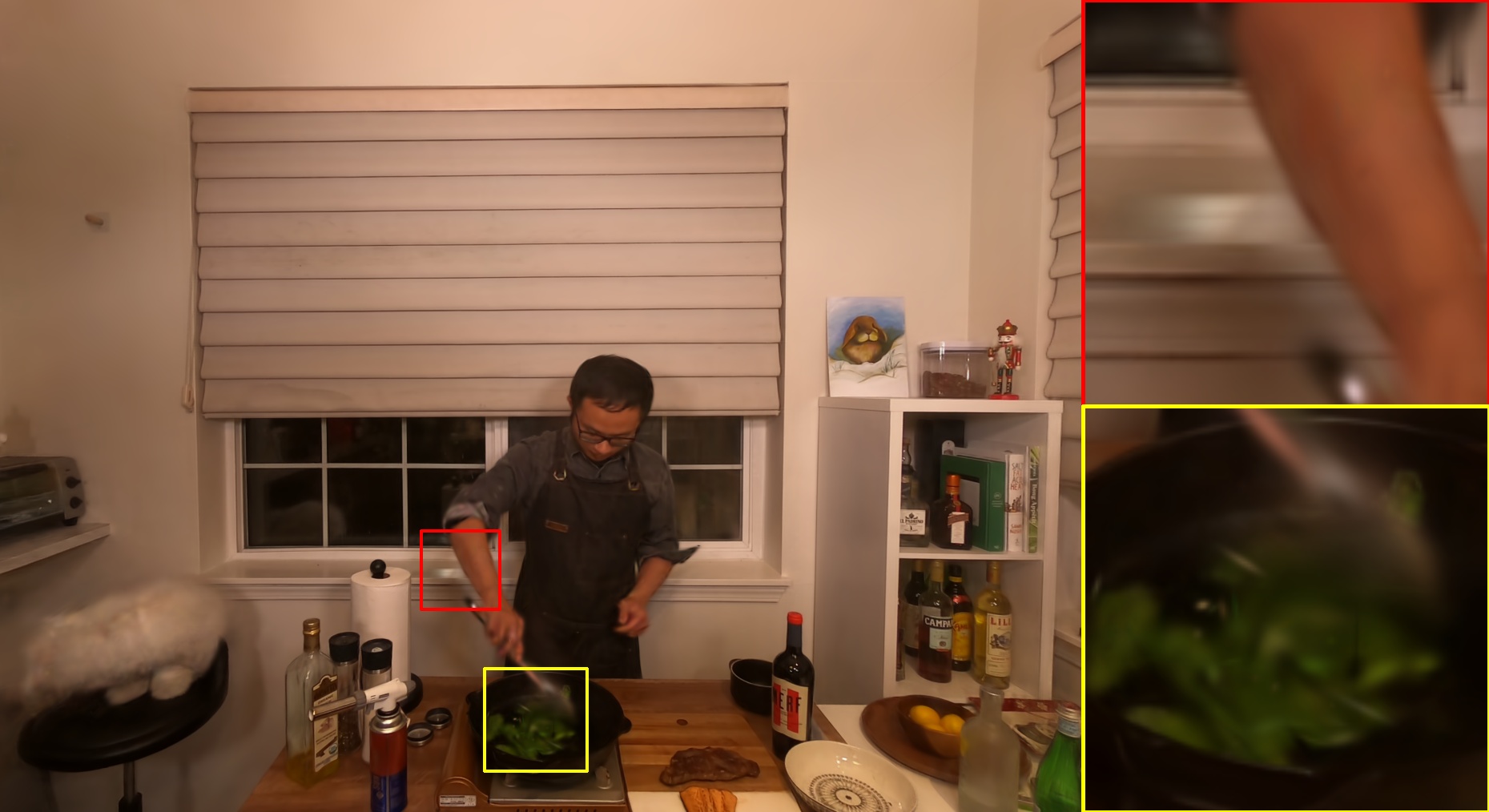}
   }
   \subfloat[Ours(ScaffoldGS)]{
      \includegraphics[width=0.24\textwidth]{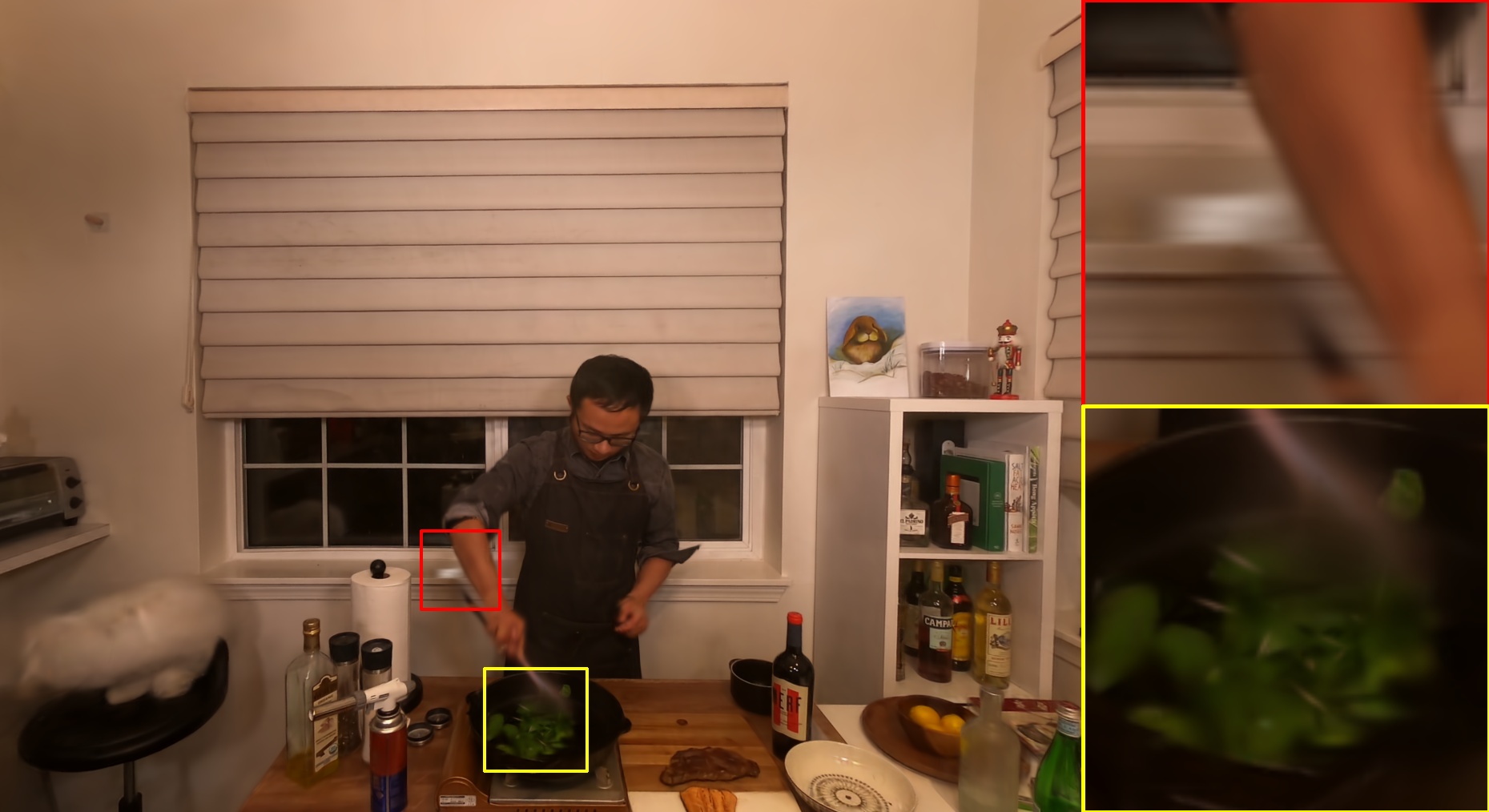}
   }
   \subfloat[Ground Truth]{
      \includegraphics[width=0.24\textwidth]{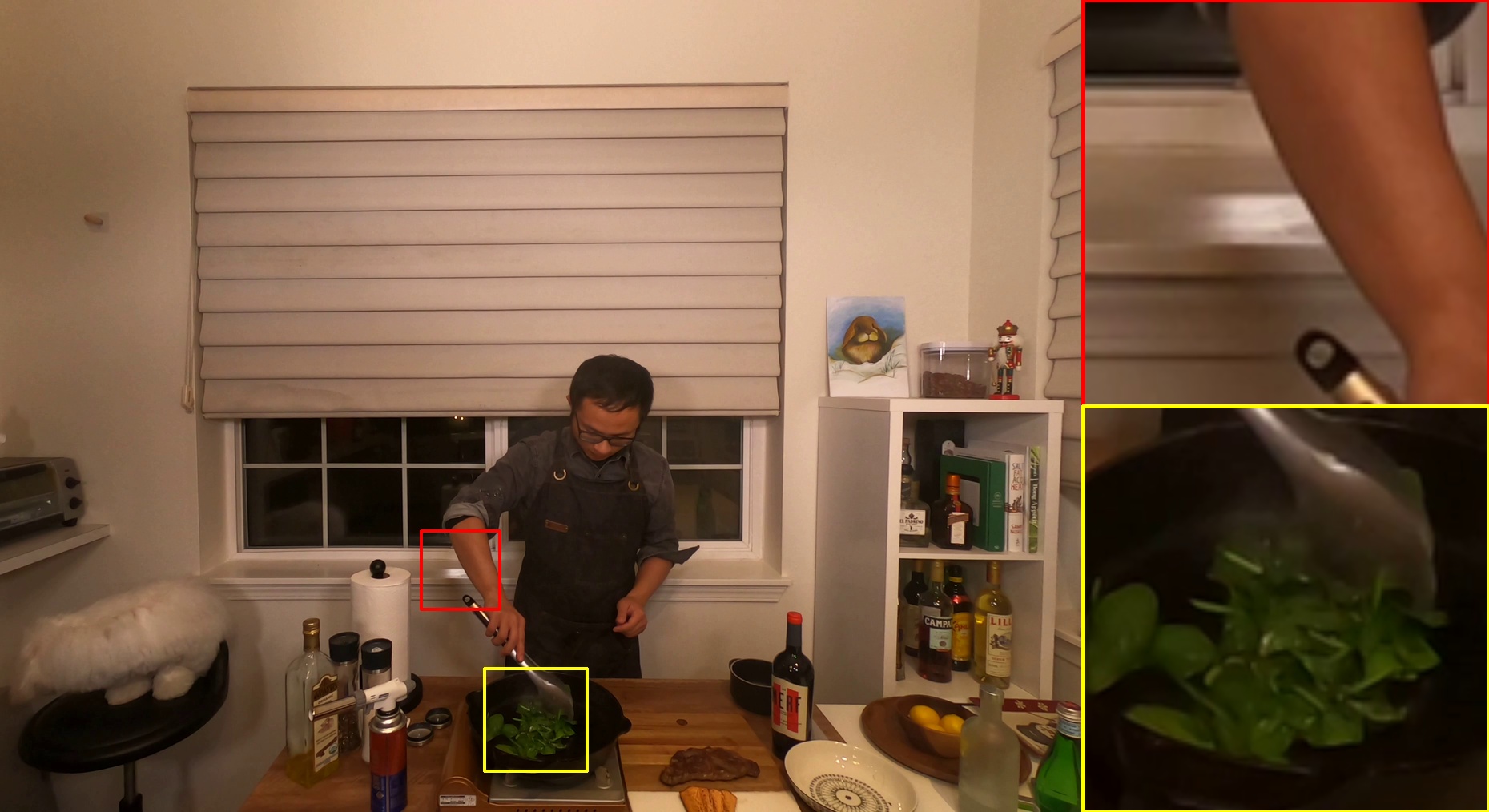}
   }
   \caption{Qualitative comparison on \textit{Cook Spinach} scene of N3DV dataset.}
   \label{fig:cook}
   \vspace{-0.2cm}
\end{figure}

\begin{figure}[!t]
   \centering

   \subfloat[StreamSTGS~\cite{ke2025streamstgs}]{
      \includegraphics[width=0.24\textwidth]{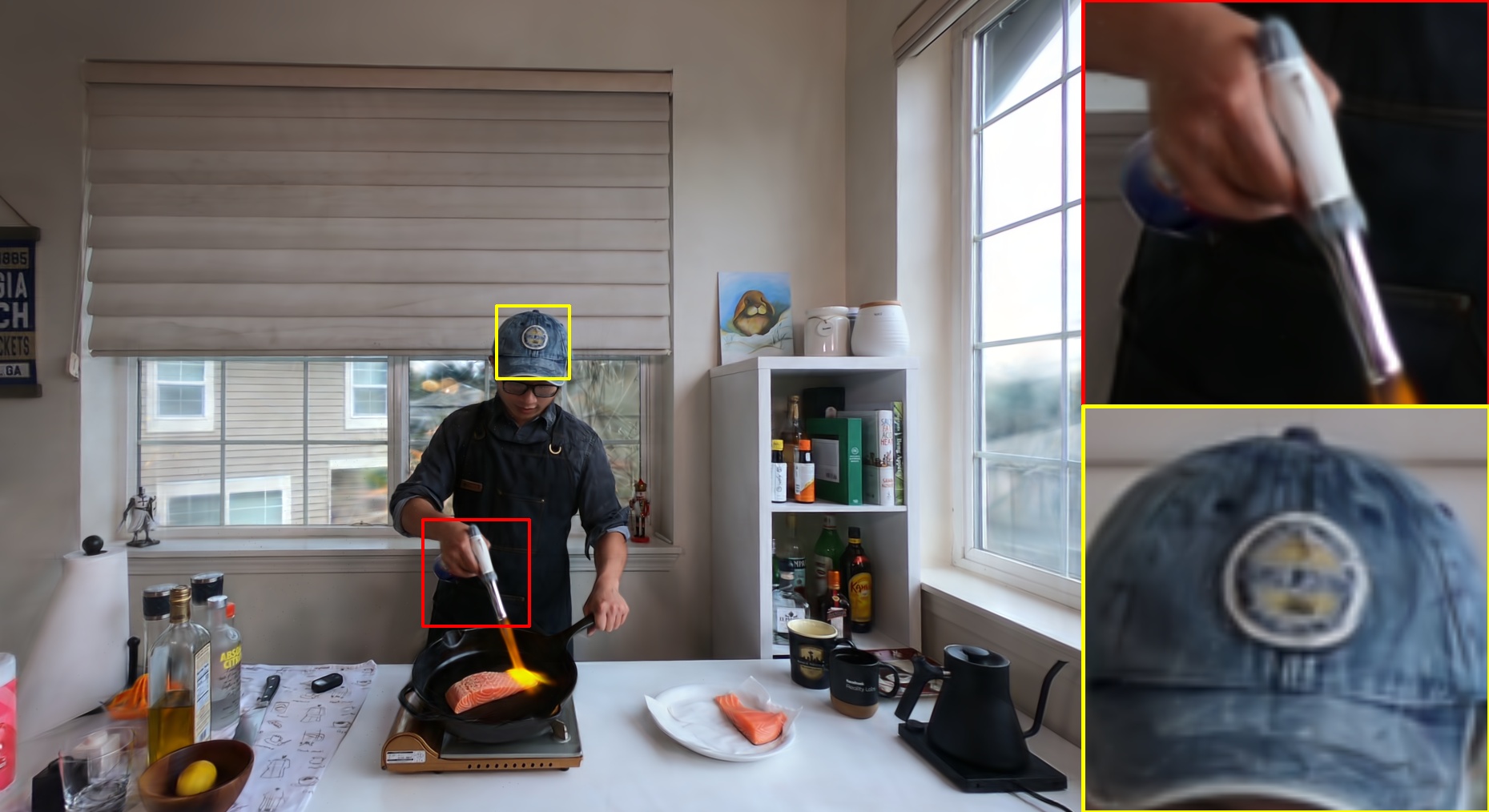}
   }
   \subfloat[QUEEN~\cite{girish2024queen}]{
      \includegraphics[width=0.24\textwidth]{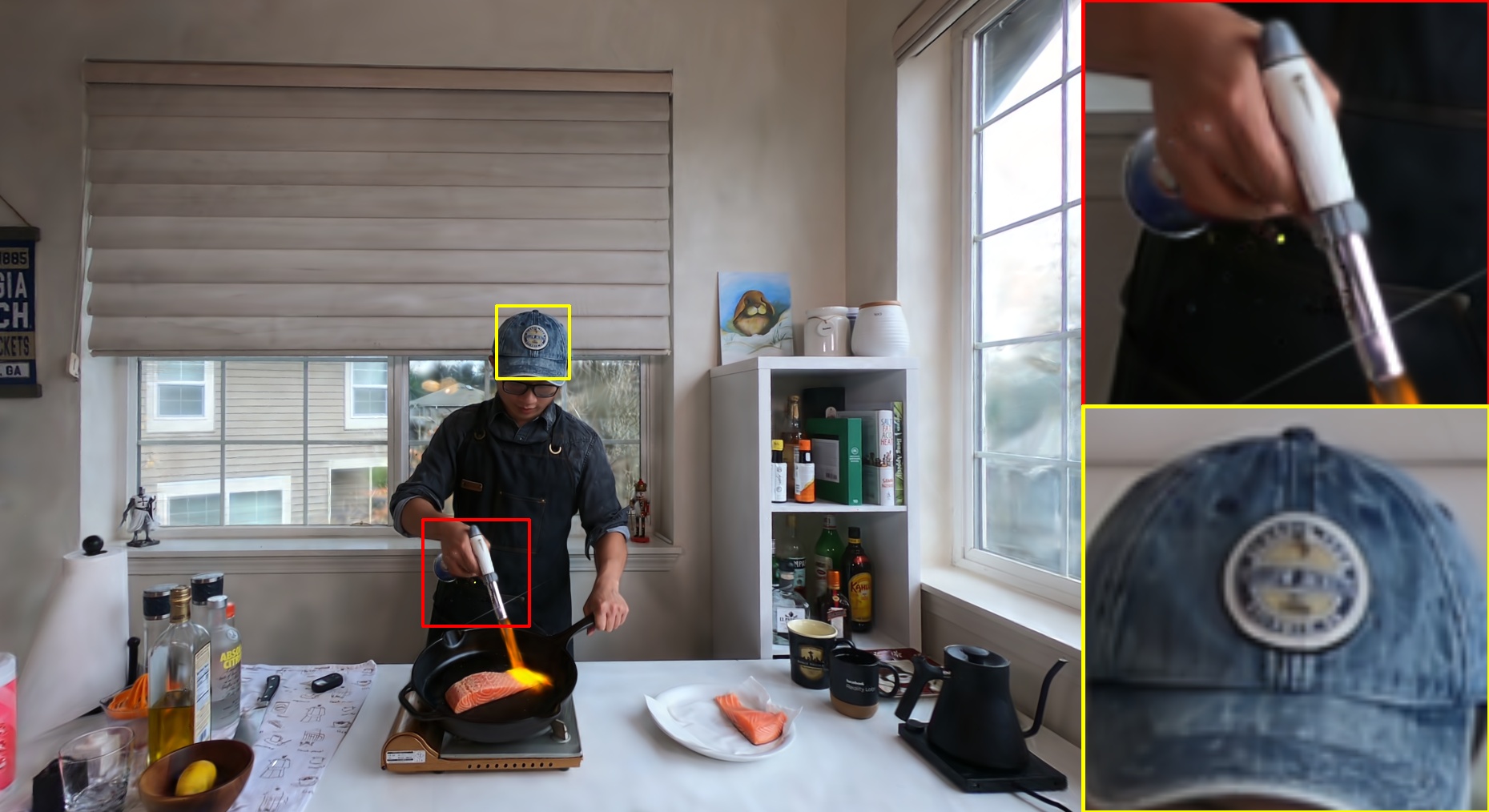}
   }
   \subfloat[HiCoM~\cite{gao2024hicom}]{
      \includegraphics[width=0.24\textwidth]{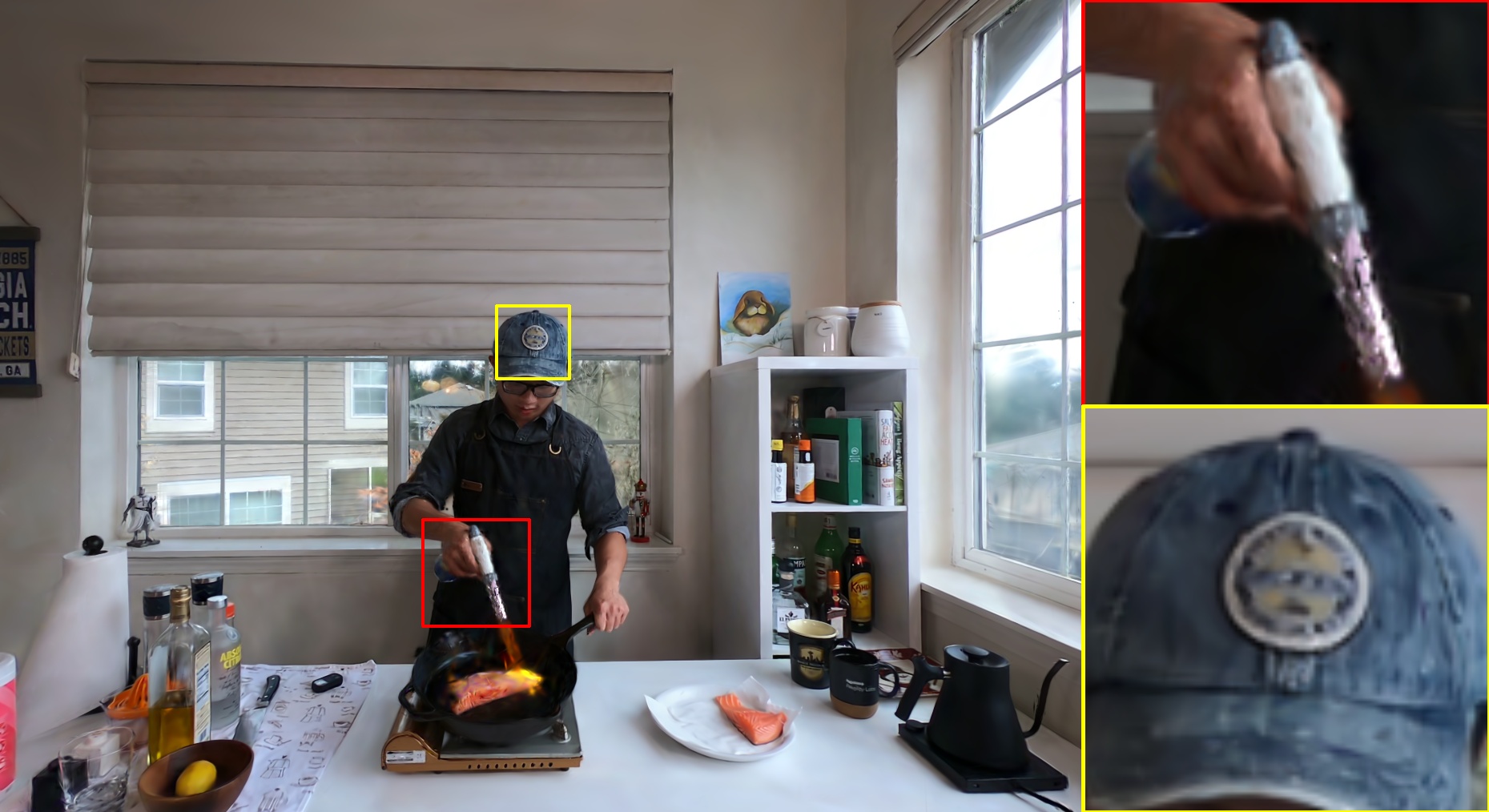}
   }
   \subfloat[Ours (3DGS)]{
      \includegraphics[width=0.24\textwidth]{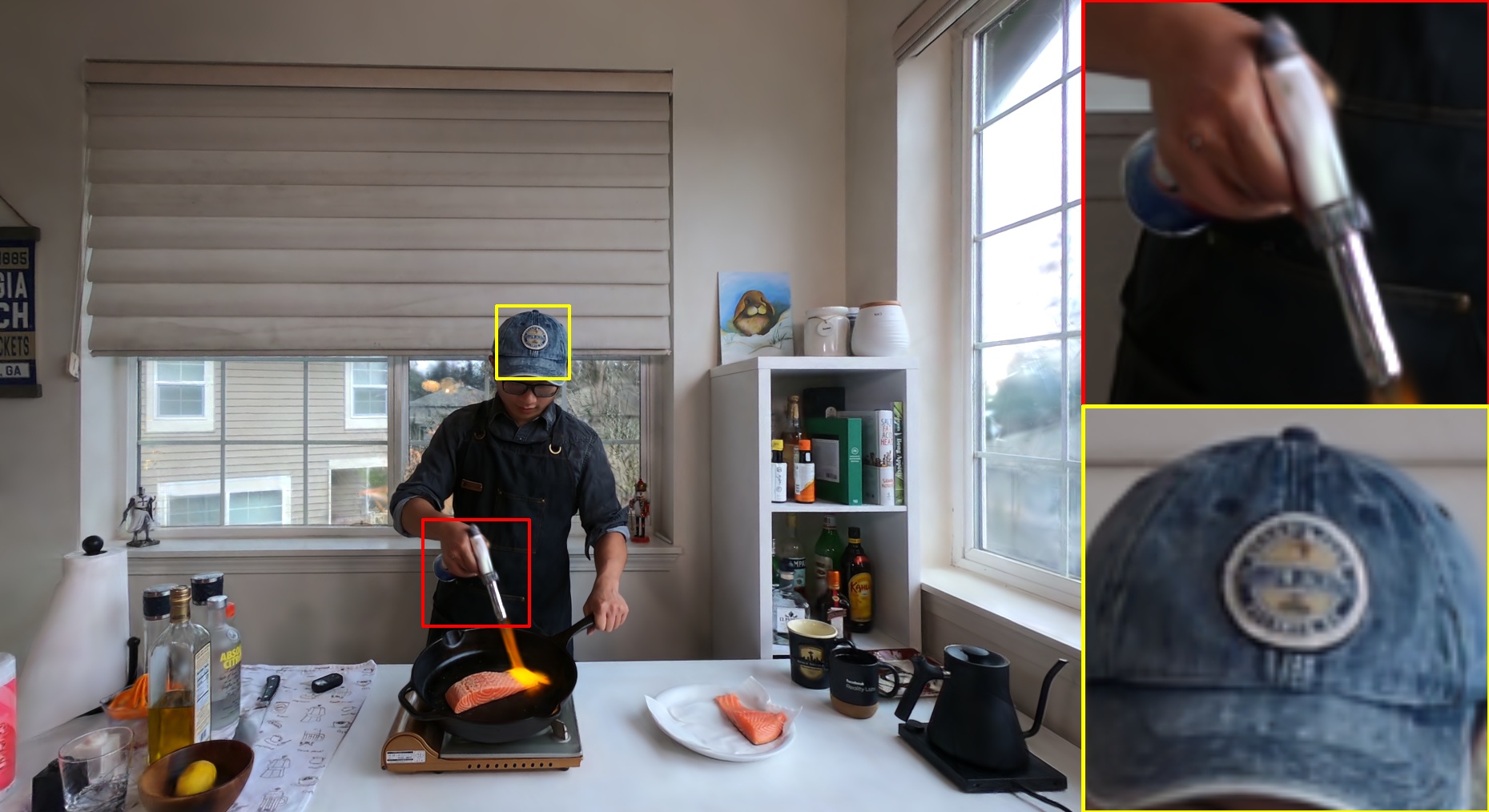}
   }
   \\[-0.2em] 

   \subfloat[GIFStream~\cite{li2025gifstream}]{
      \includegraphics[width=0.24\textwidth]{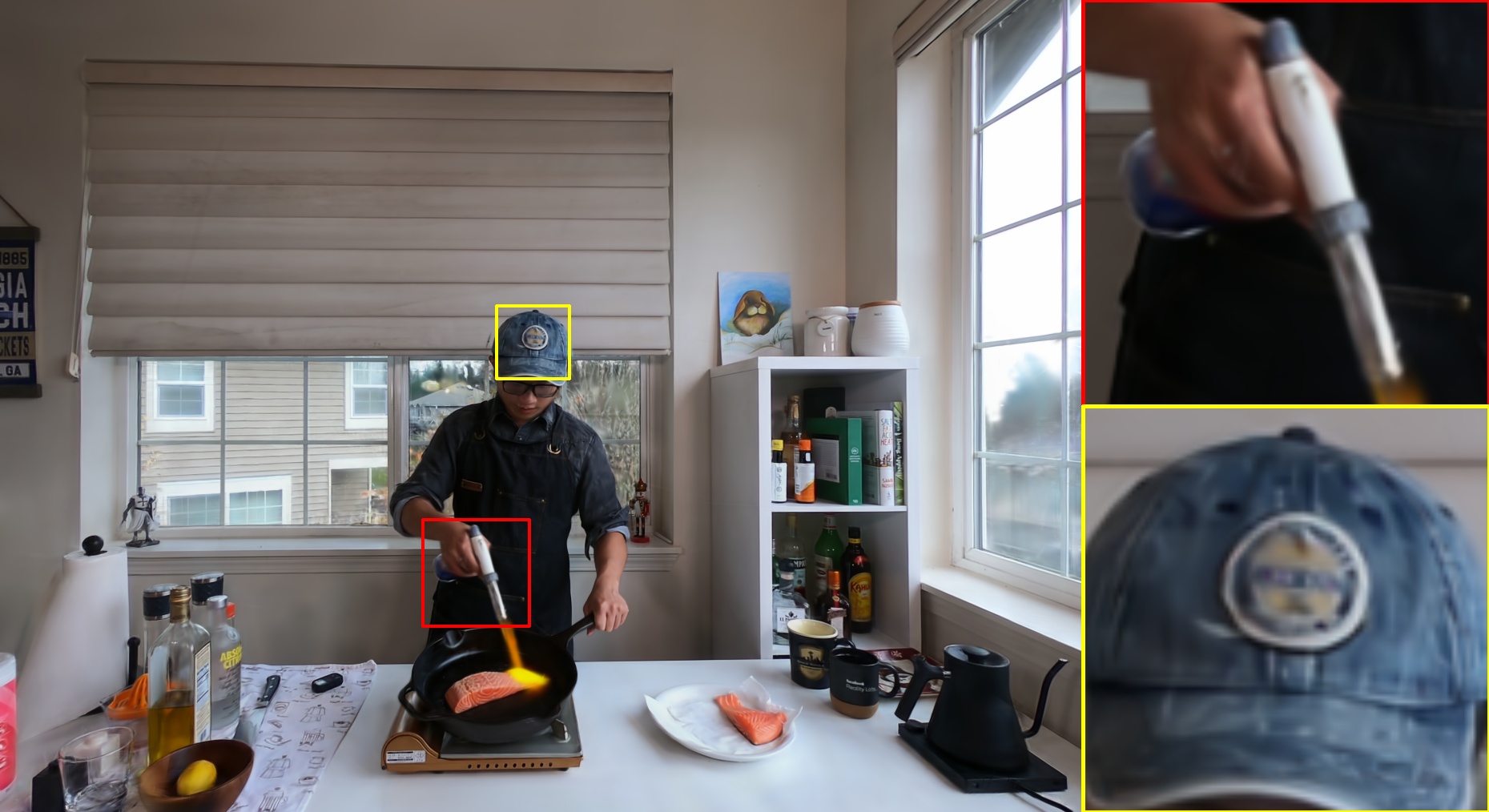}
   }
   \subfloat[iFVC~\cite{tang2025compressing}]{
      \includegraphics[width=0.24\textwidth]{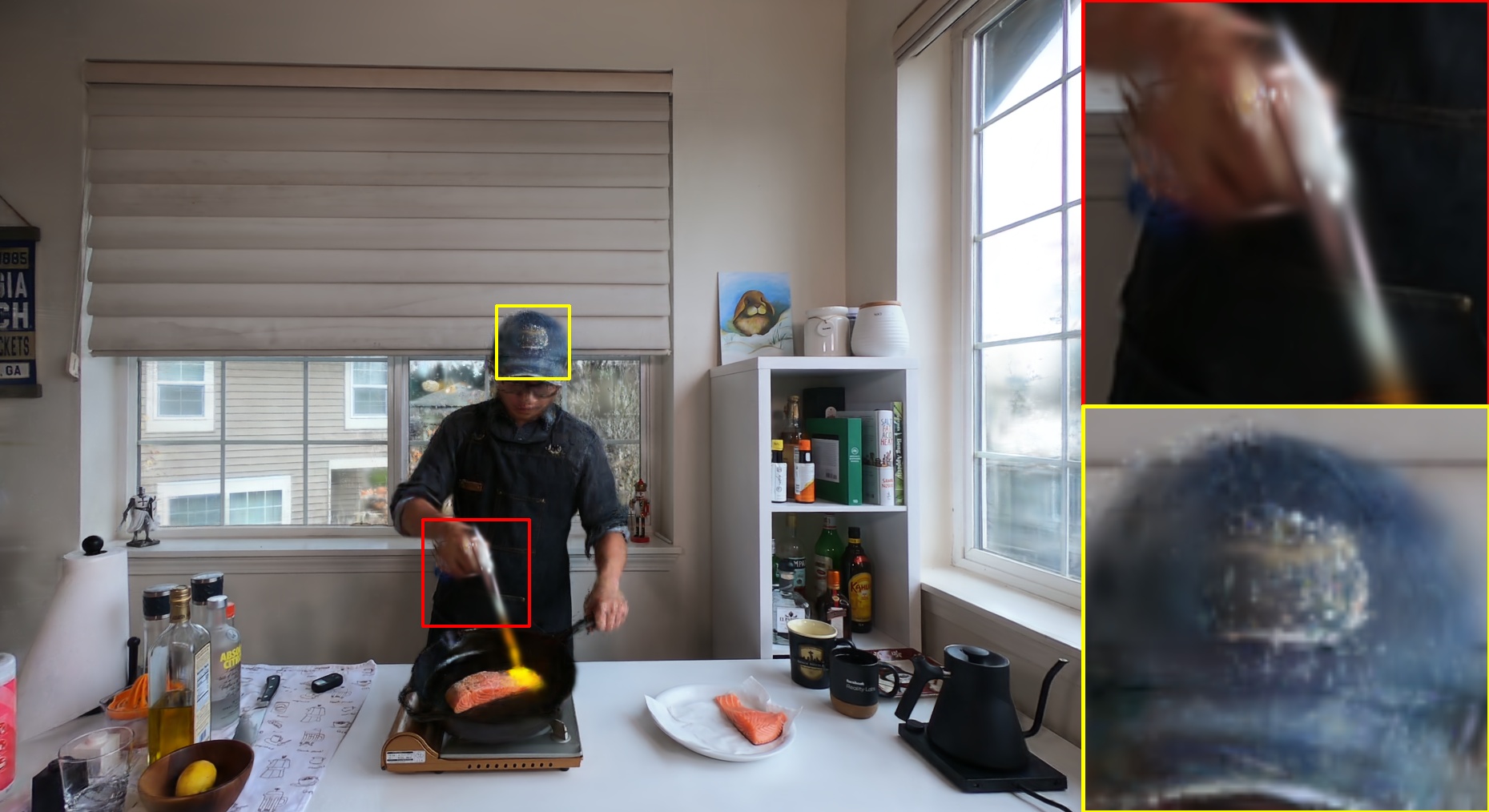}
   }
   \subfloat[Ours(ScaffoldGS)]{
      \includegraphics[width=0.24\textwidth]{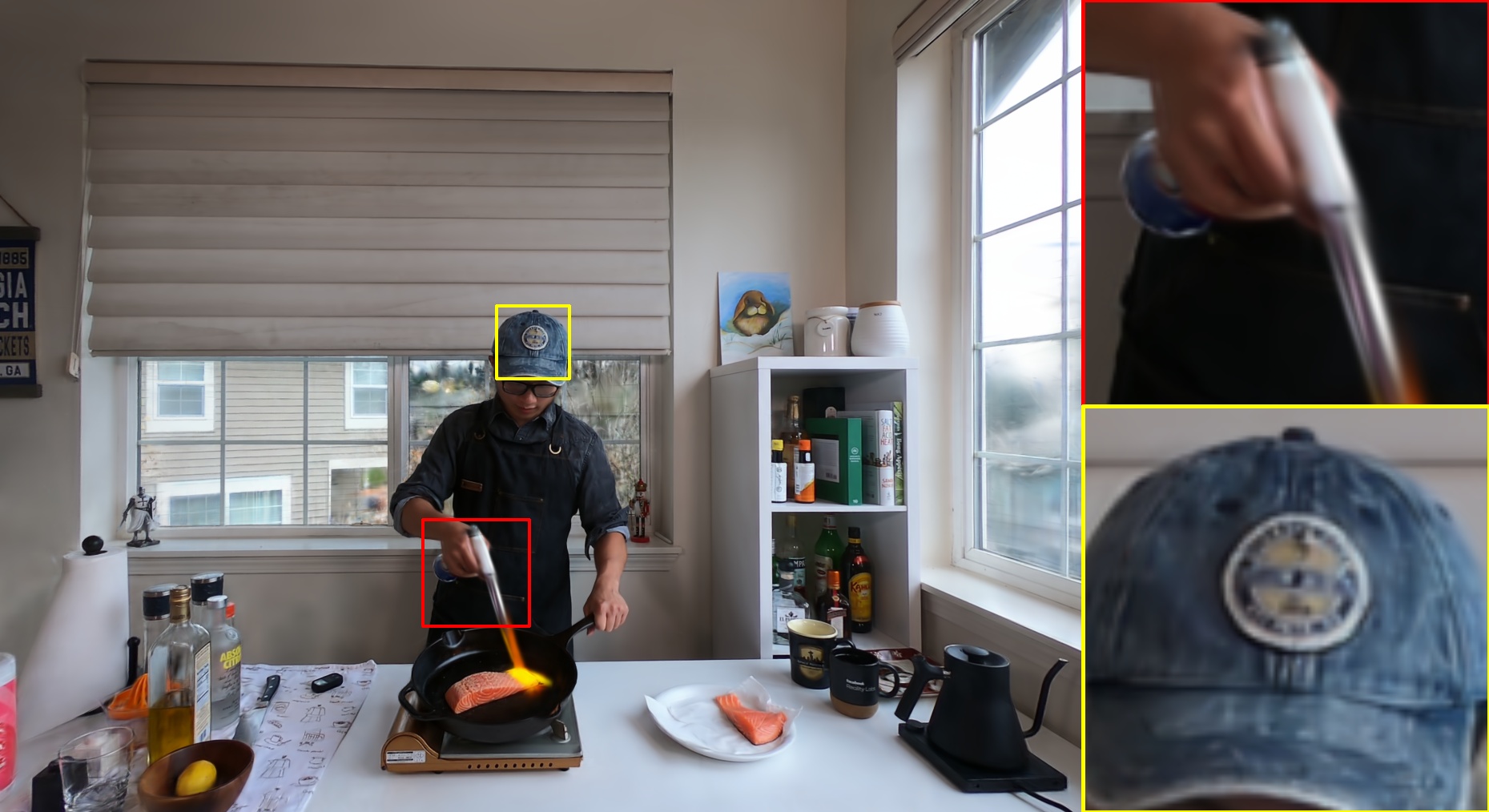}
   }
   \subfloat[Ground Truth]{
      \includegraphics[width=0.24\textwidth]{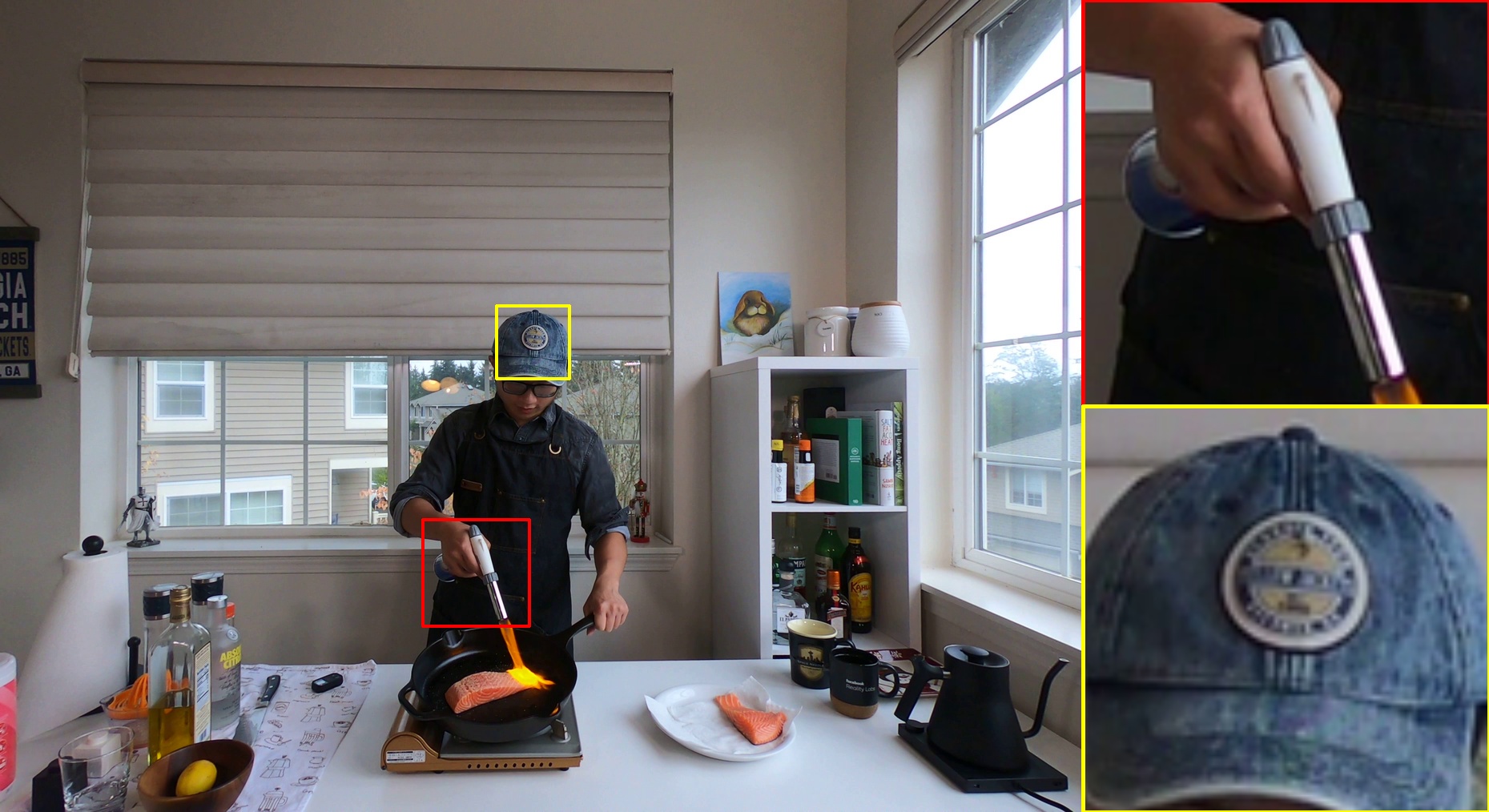}
   }
   \caption{Qualitative comparison on \textit{Flame Salmon} scene of N3DV dataset.}
   \label{fig:salmon}
   \vspace{-0.4cm}
\end{figure}

\begin{figure}[!t]
   \centering

   \subfloat[StreamSTGS~\cite{ke2025streamstgs}]{
      \includegraphics[width=0.24\textwidth]{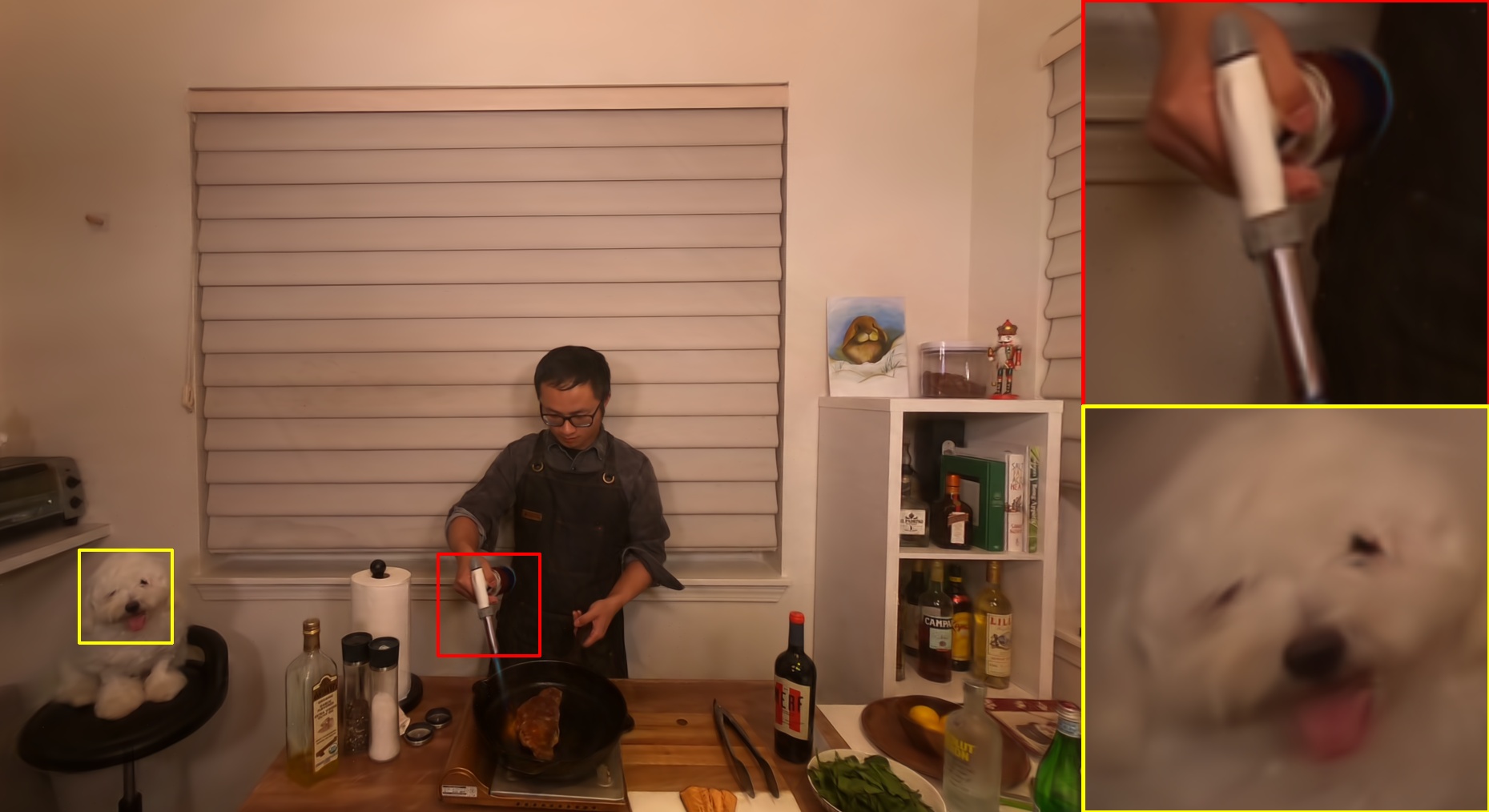}
   }
   \subfloat[QUEEN~\cite{girish2024queen}]{
      \includegraphics[width=0.24\textwidth]{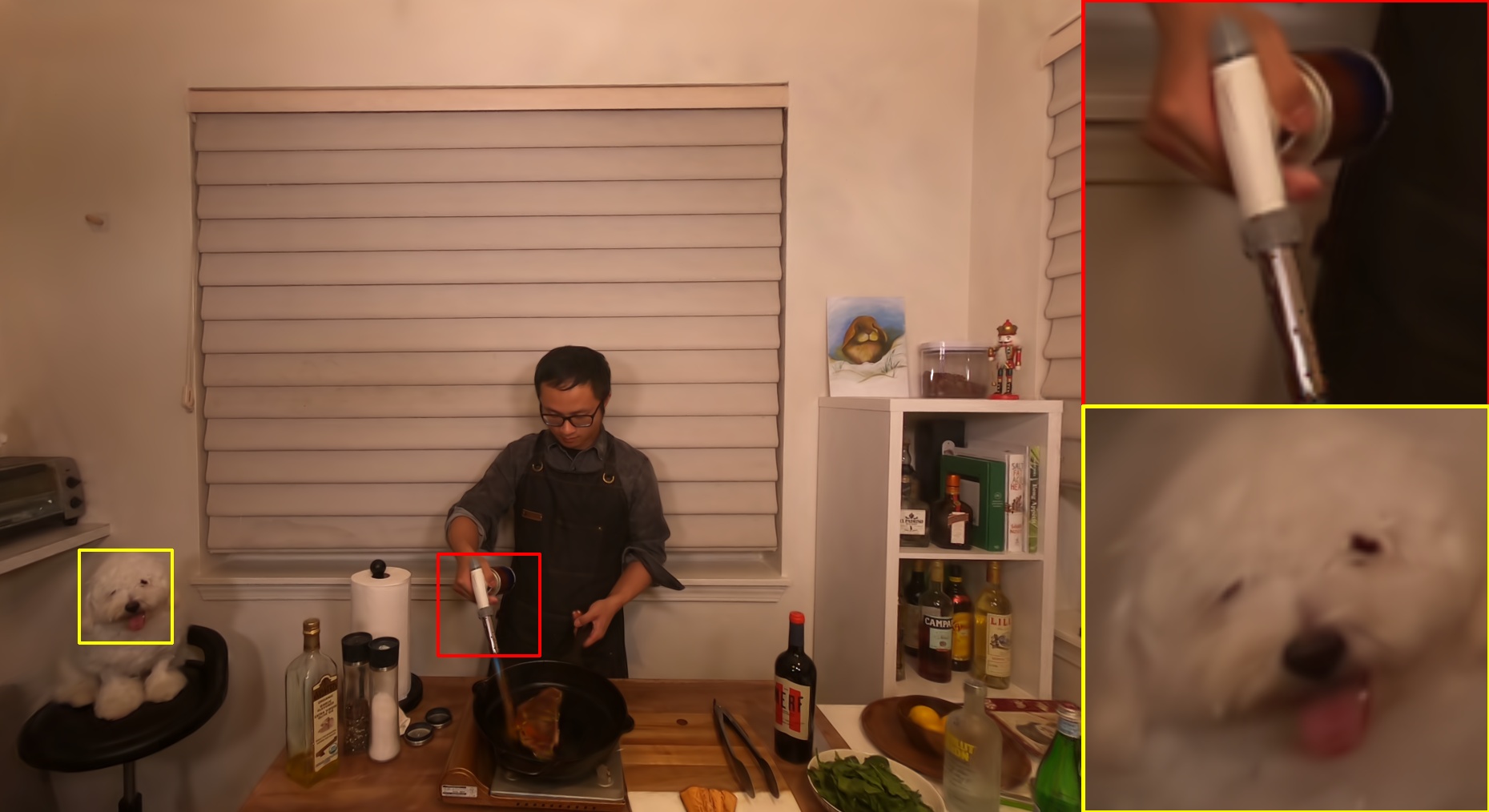}
   }
   \subfloat[HiCoM~\cite{gao2024hicom}]{
      \includegraphics[width=0.24\textwidth]{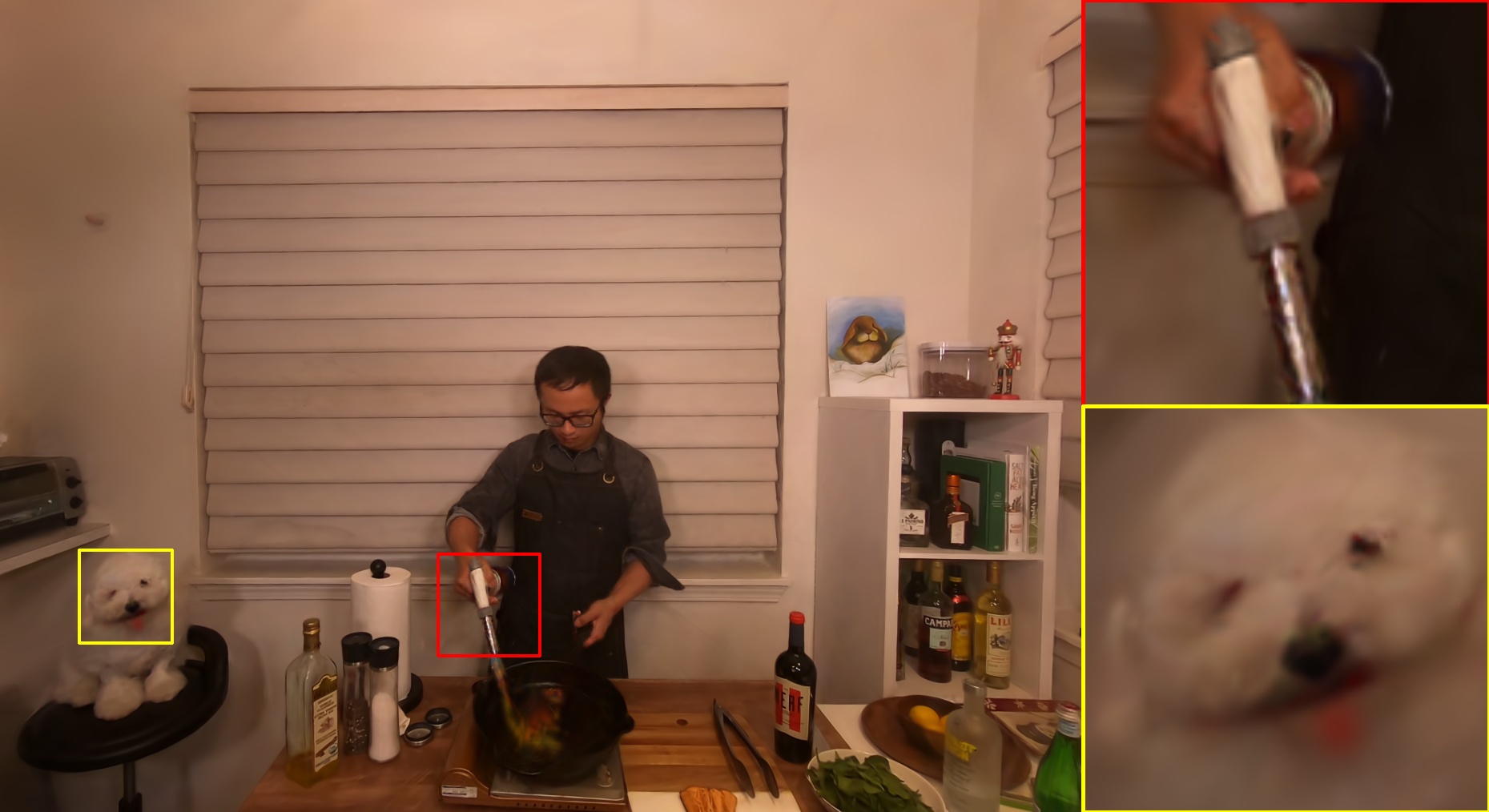}
   }
   \subfloat[Ours (3DGS)]{
      \includegraphics[width=0.24\textwidth]{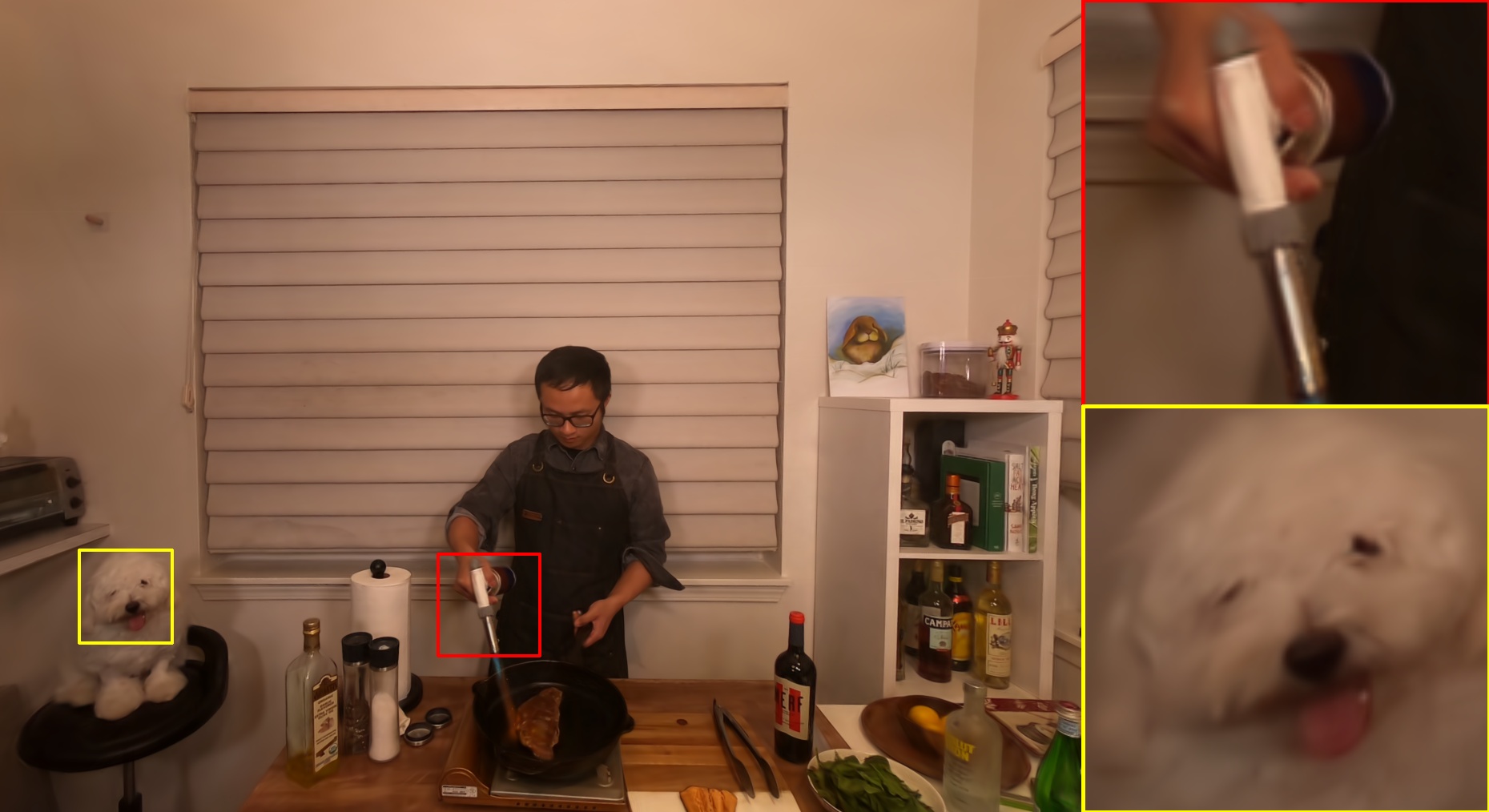}
   }
   \\[-0.2em] 


   \subfloat[GIFStream~\cite{li2025gifstream}]{
      \includegraphics[width=0.24\textwidth]{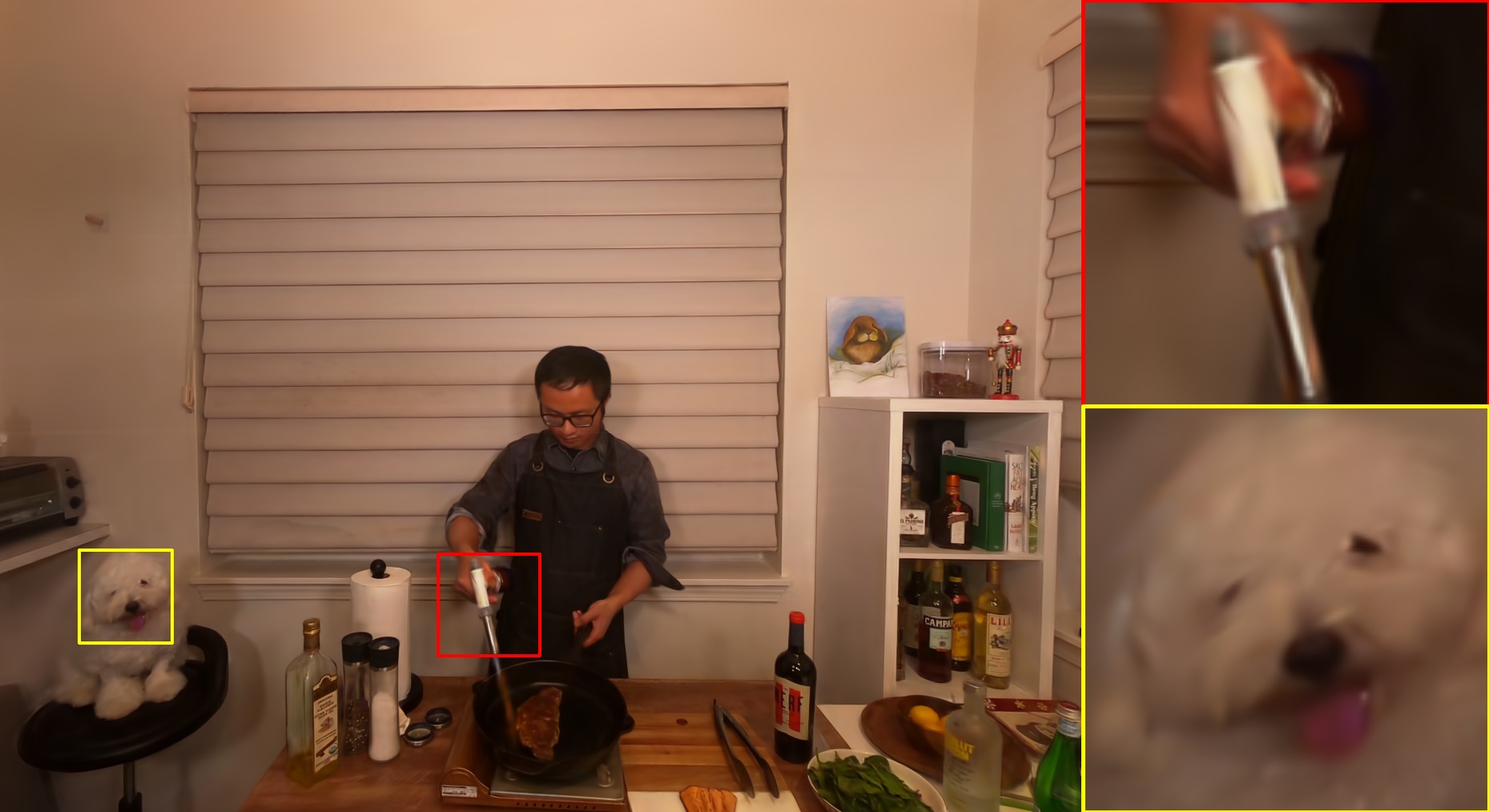}
   }
   \subfloat[iFVC~\cite{tang2025compressing}]{
      \includegraphics[width=0.24\textwidth]{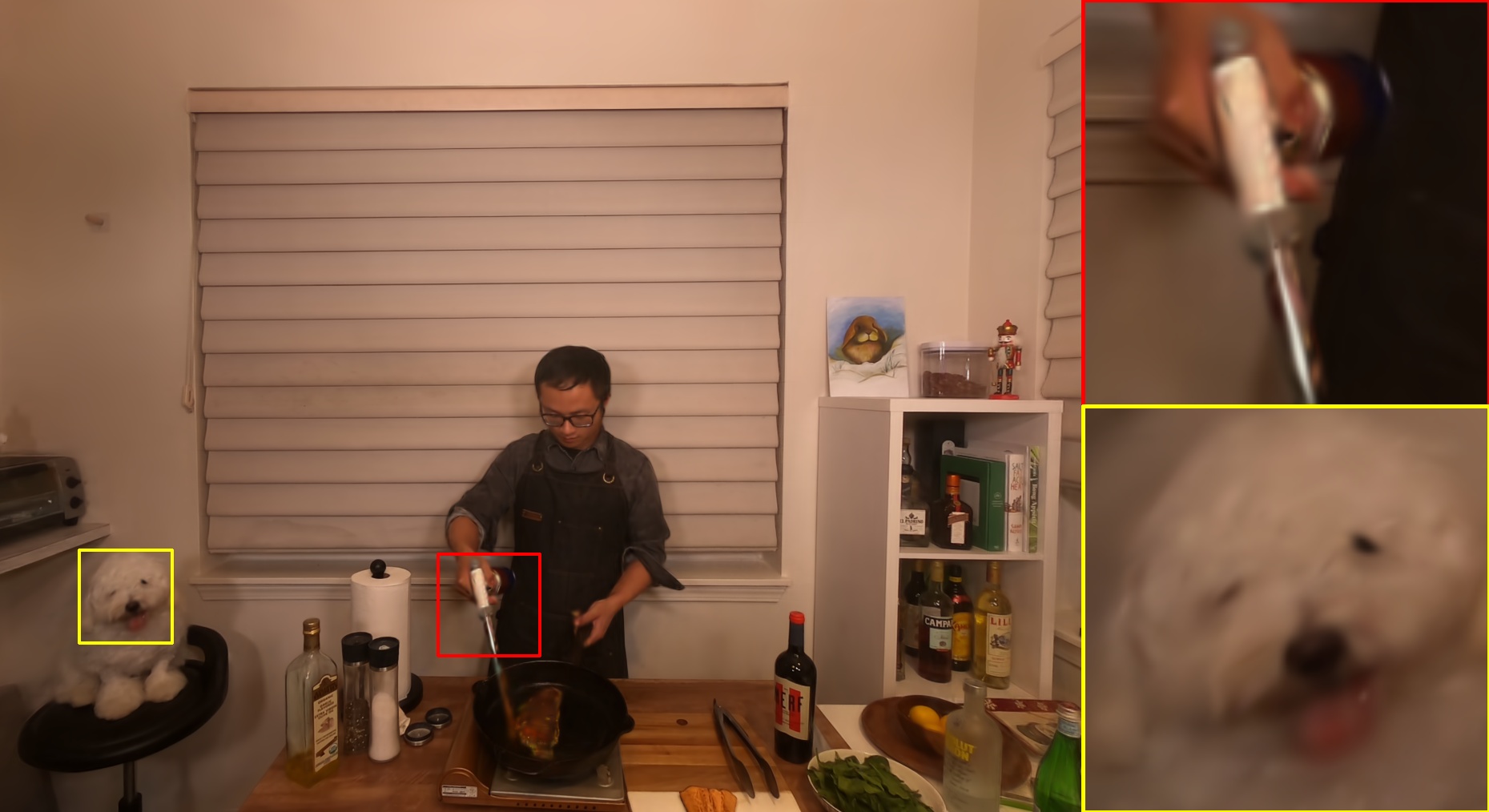}
   }
   \subfloat[Ours(ScaffoldGS)]{
      \includegraphics[width=0.24\textwidth]{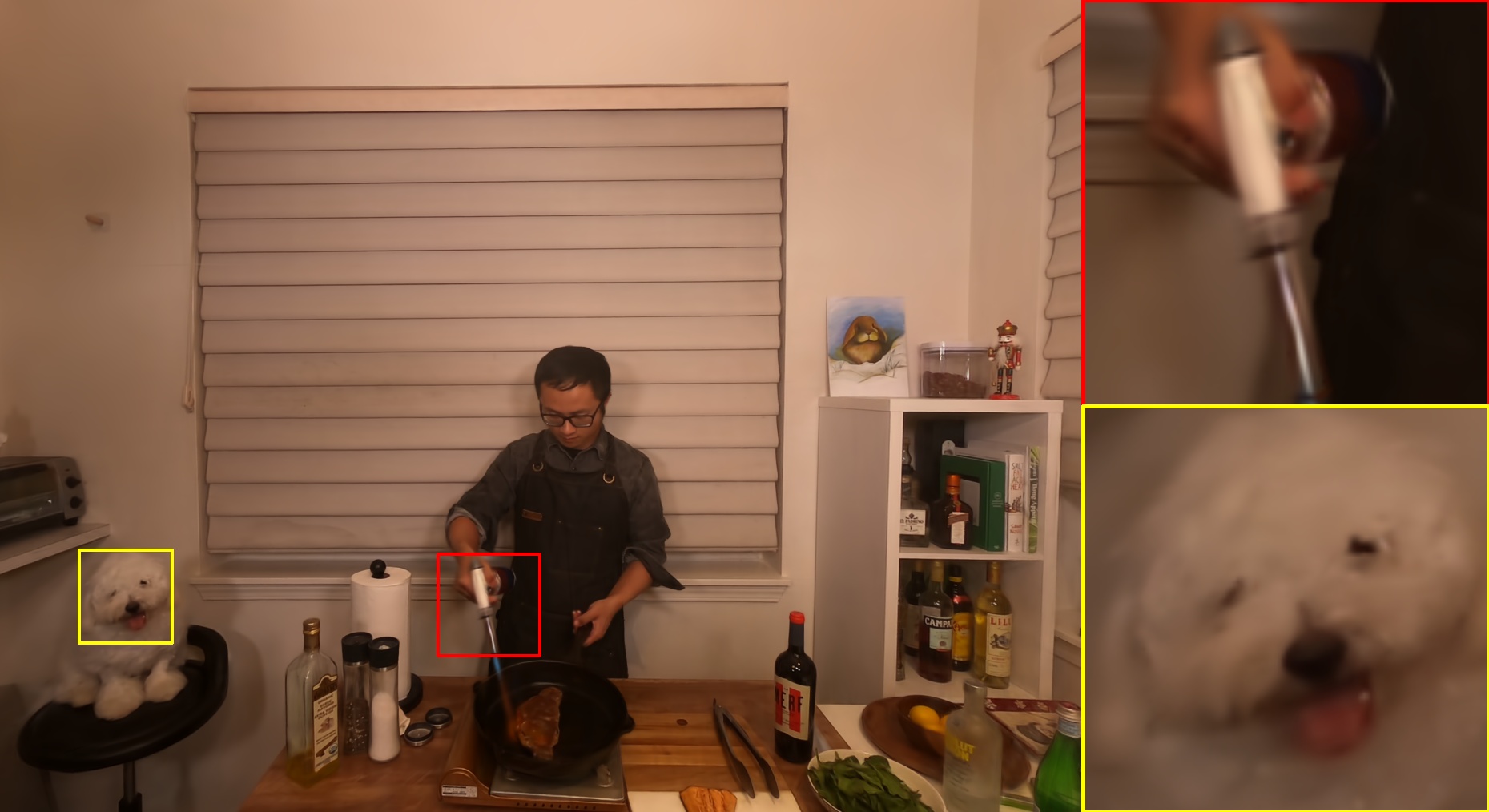}
   }
   \subfloat[Ground Truth]{
      \includegraphics[width=0.24\textwidth]{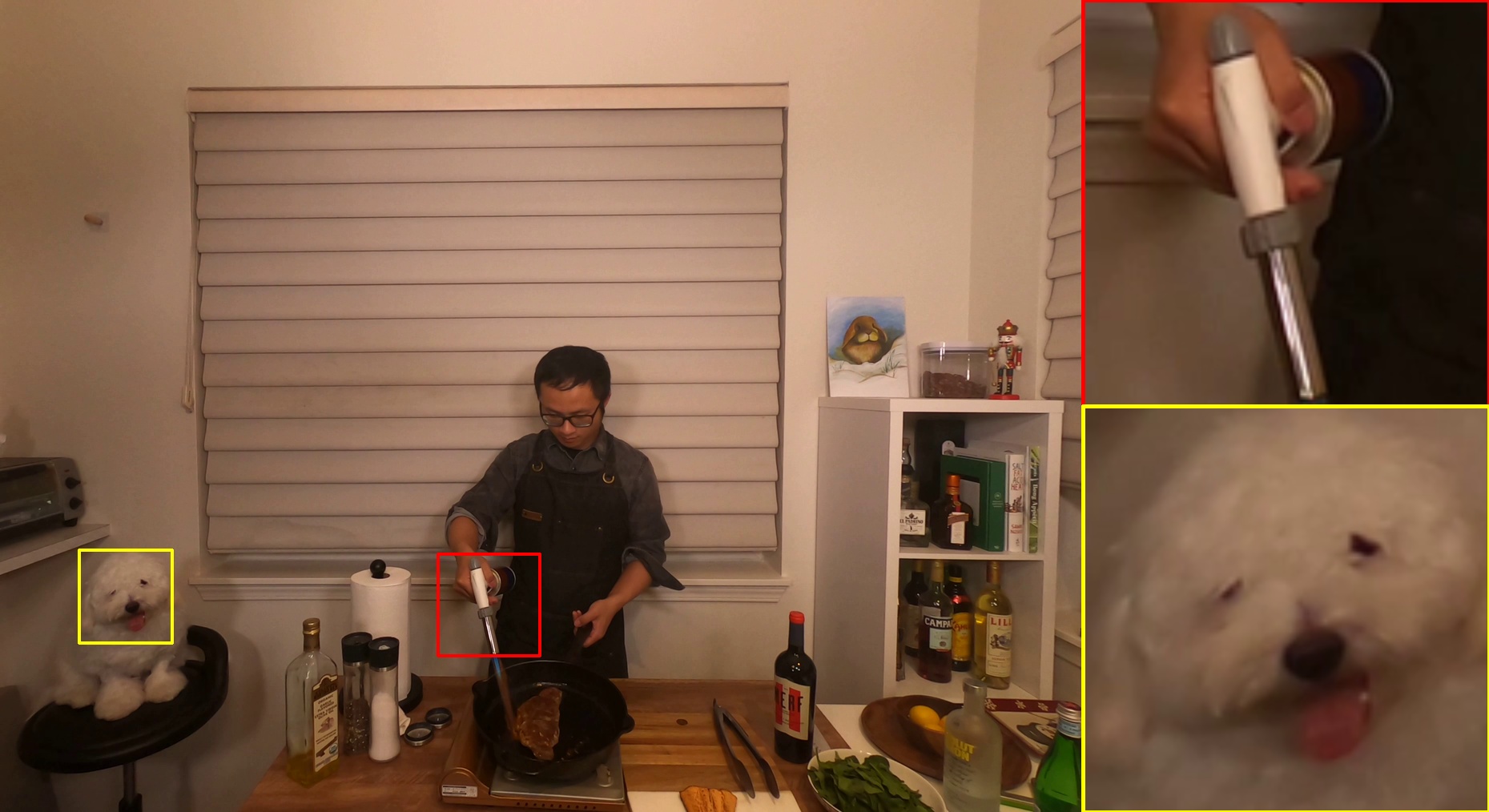}
   }
   \caption{Qualitative comparison on \textit{Flame Steak} scene of N3DV dataset.}
   \label{fig:steak}
   \vspace{-0.2cm}
\end{figure}

\begin{figure}[!t]
   \centering

   \subfloat[StreamSTGS~\cite{ke2025streamstgs}]{
      \includegraphics[width=0.24\textwidth]{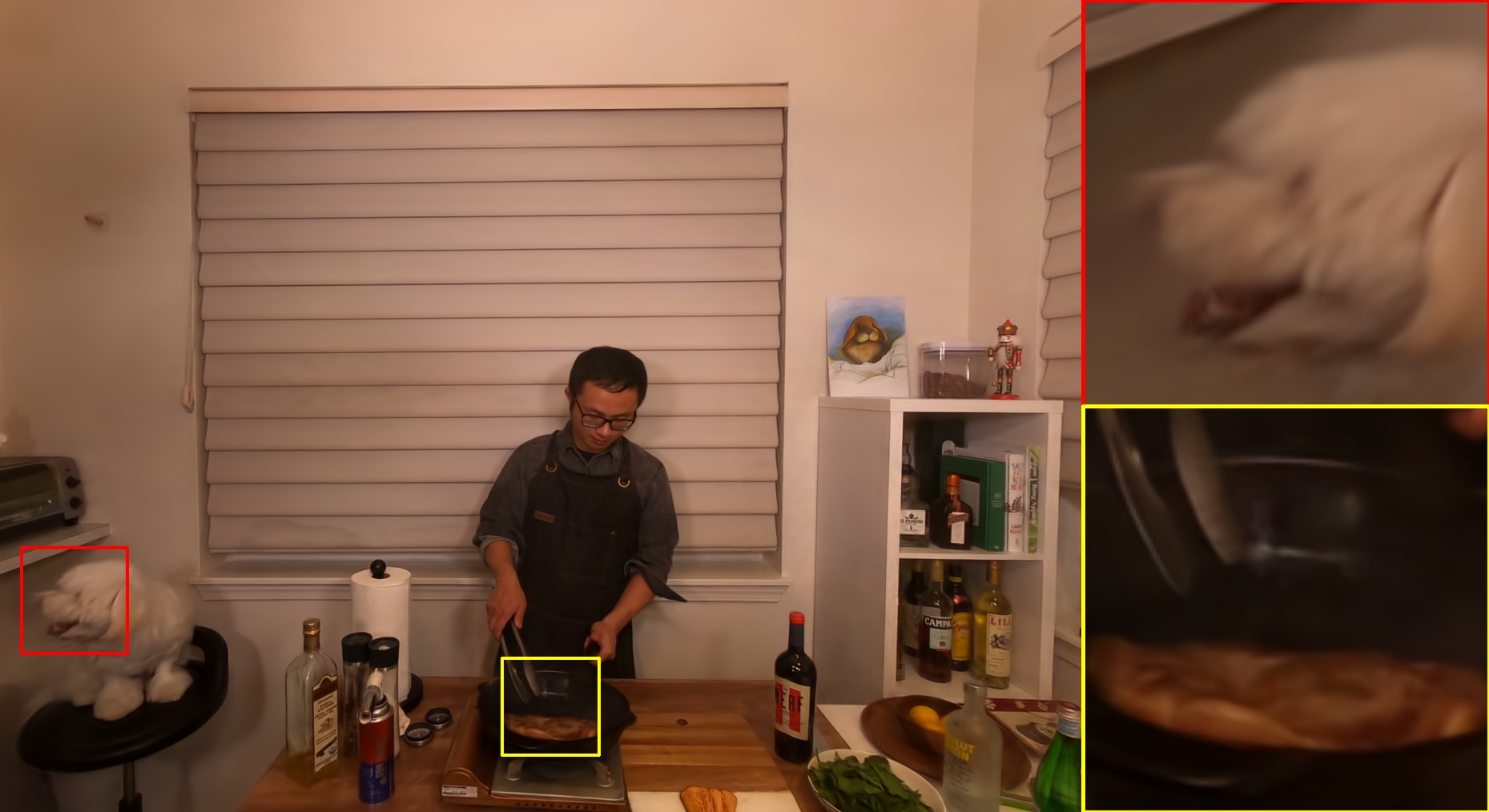}
   }
   \subfloat[QUEEN~\cite{girish2024queen}]{
      \includegraphics[width=0.24\textwidth]{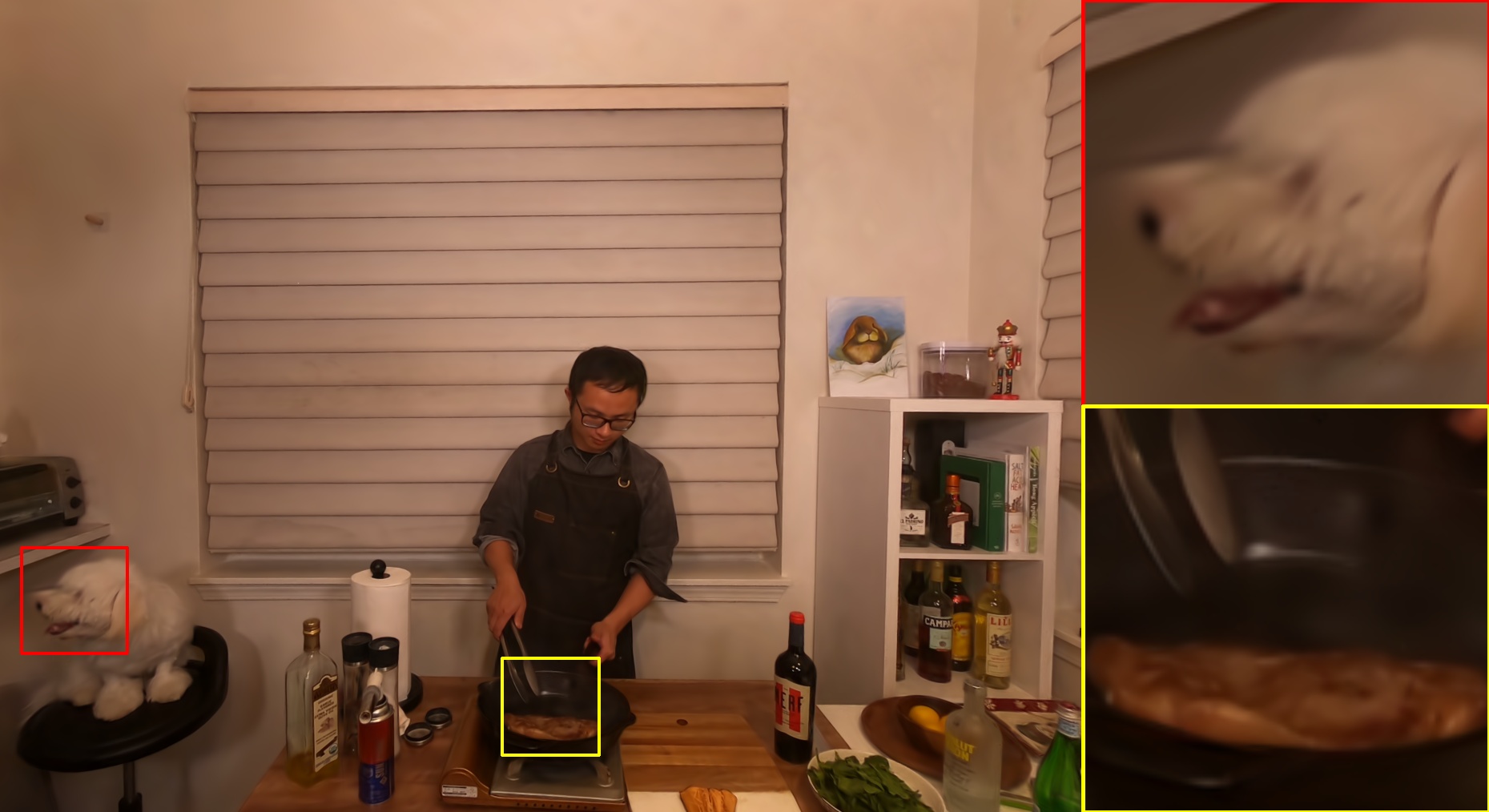}
   }
   \subfloat[HiCoM~\cite{gao2024hicom}]{
      \includegraphics[width=0.24\textwidth]{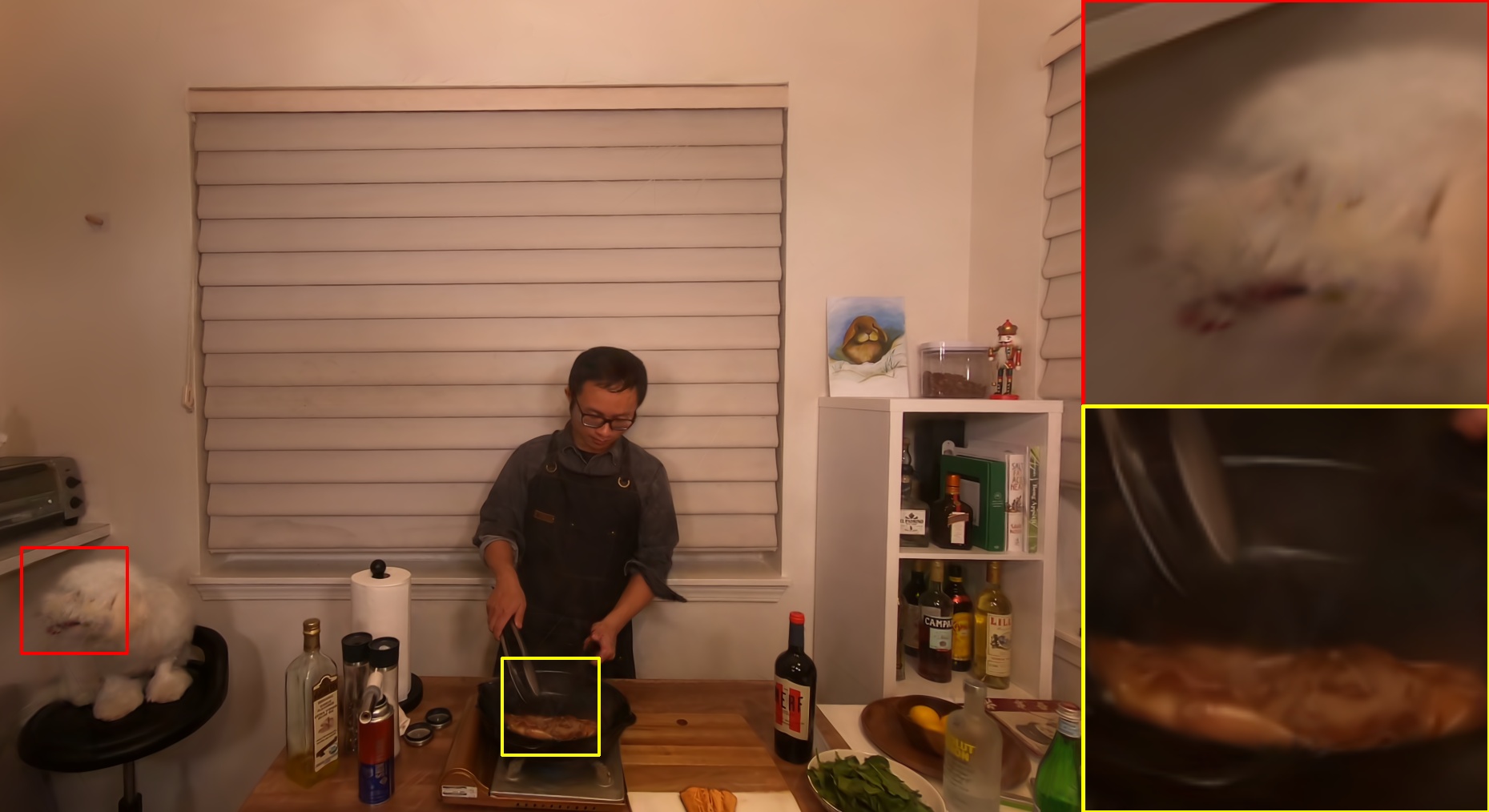}
   }
   \subfloat[Ours (3DGS)]{
      \includegraphics[width=0.24\textwidth]{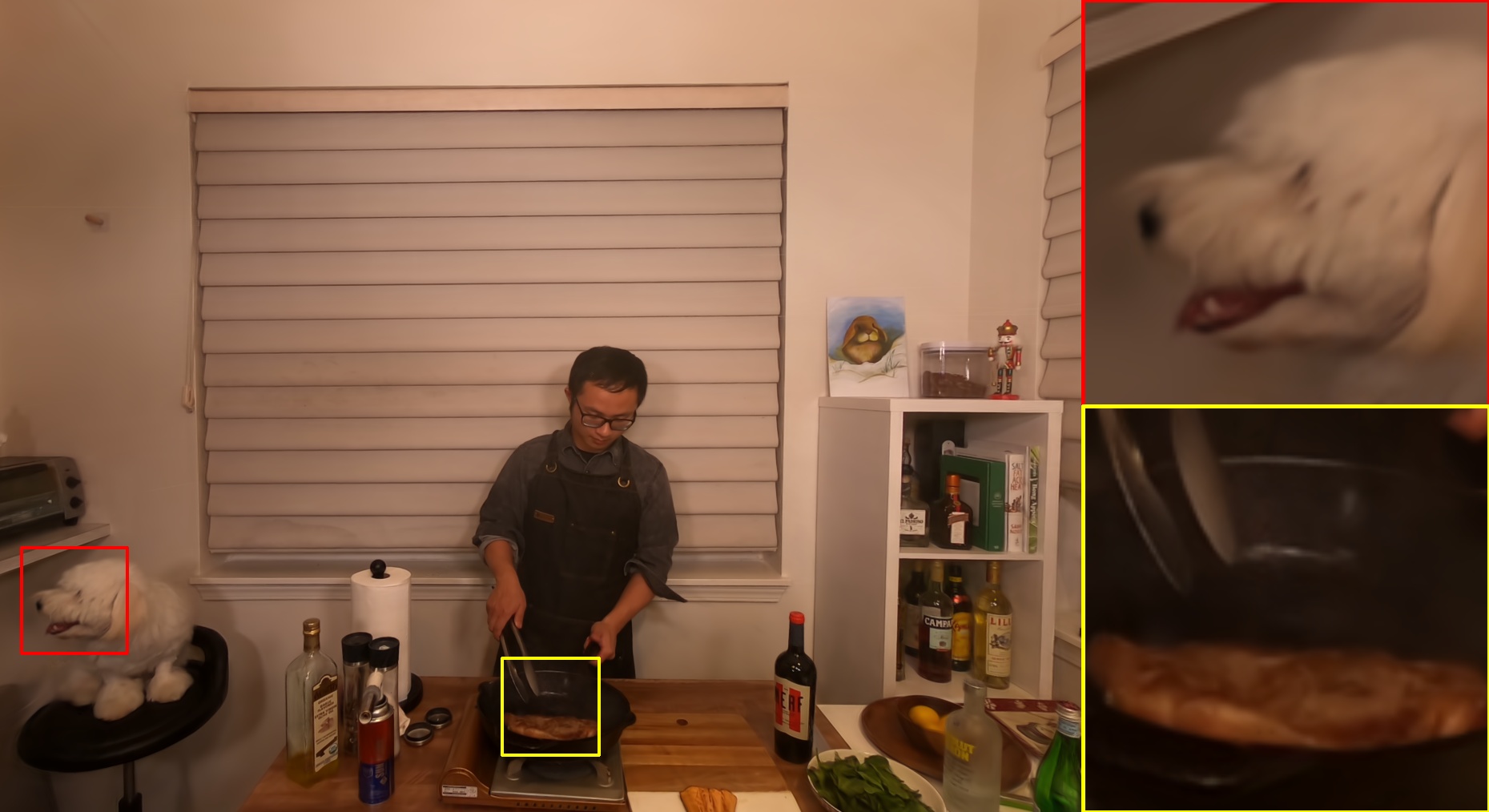}
   }
   \\[-0.2em] 


   \subfloat[GIFStream~\cite{li2025gifstream}]{
      \includegraphics[width=0.24\textwidth]{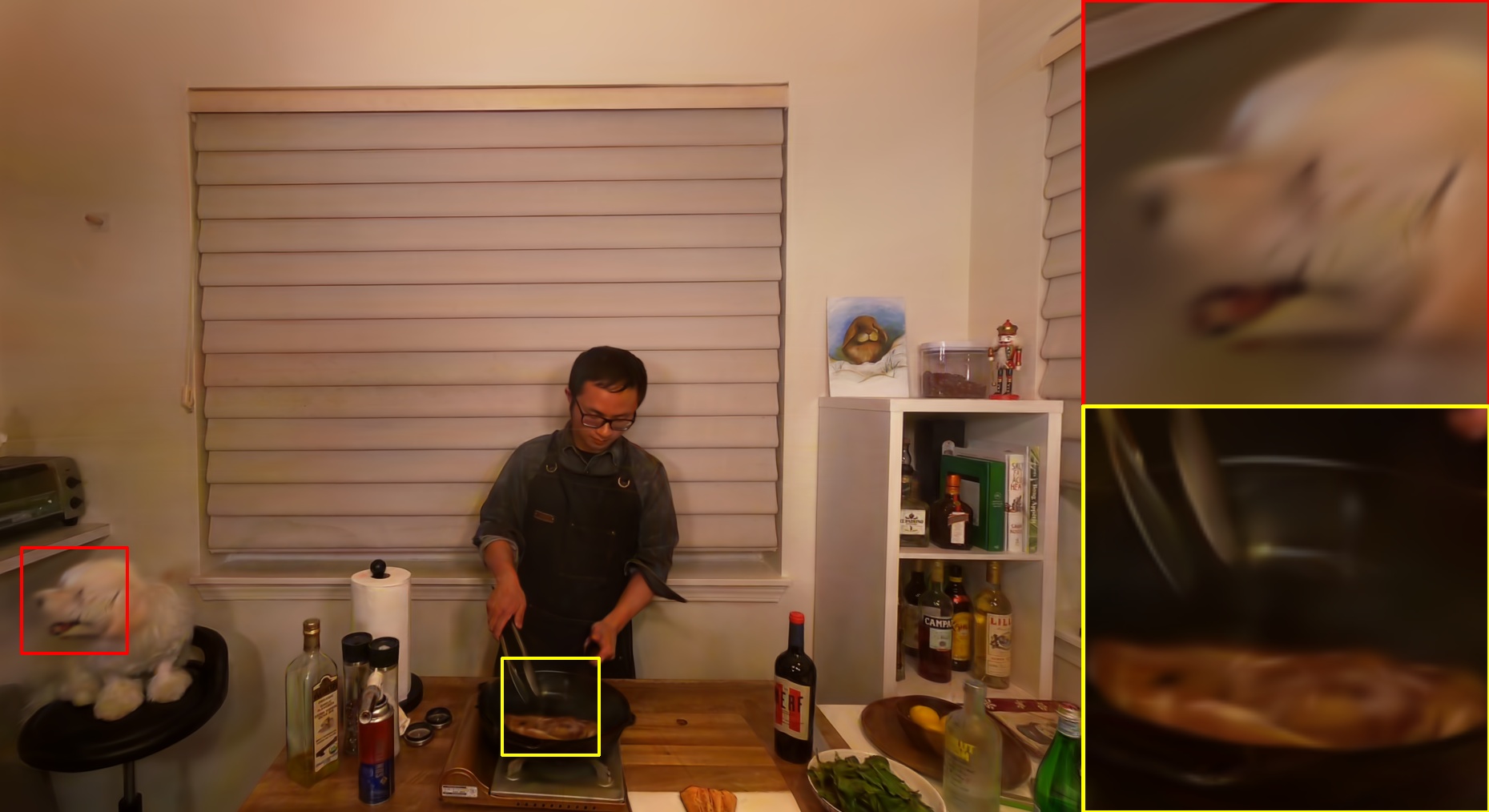}
   }
   \subfloat[iFVC~\cite{tang2025compressing}]{
      \includegraphics[width=0.24\textwidth]{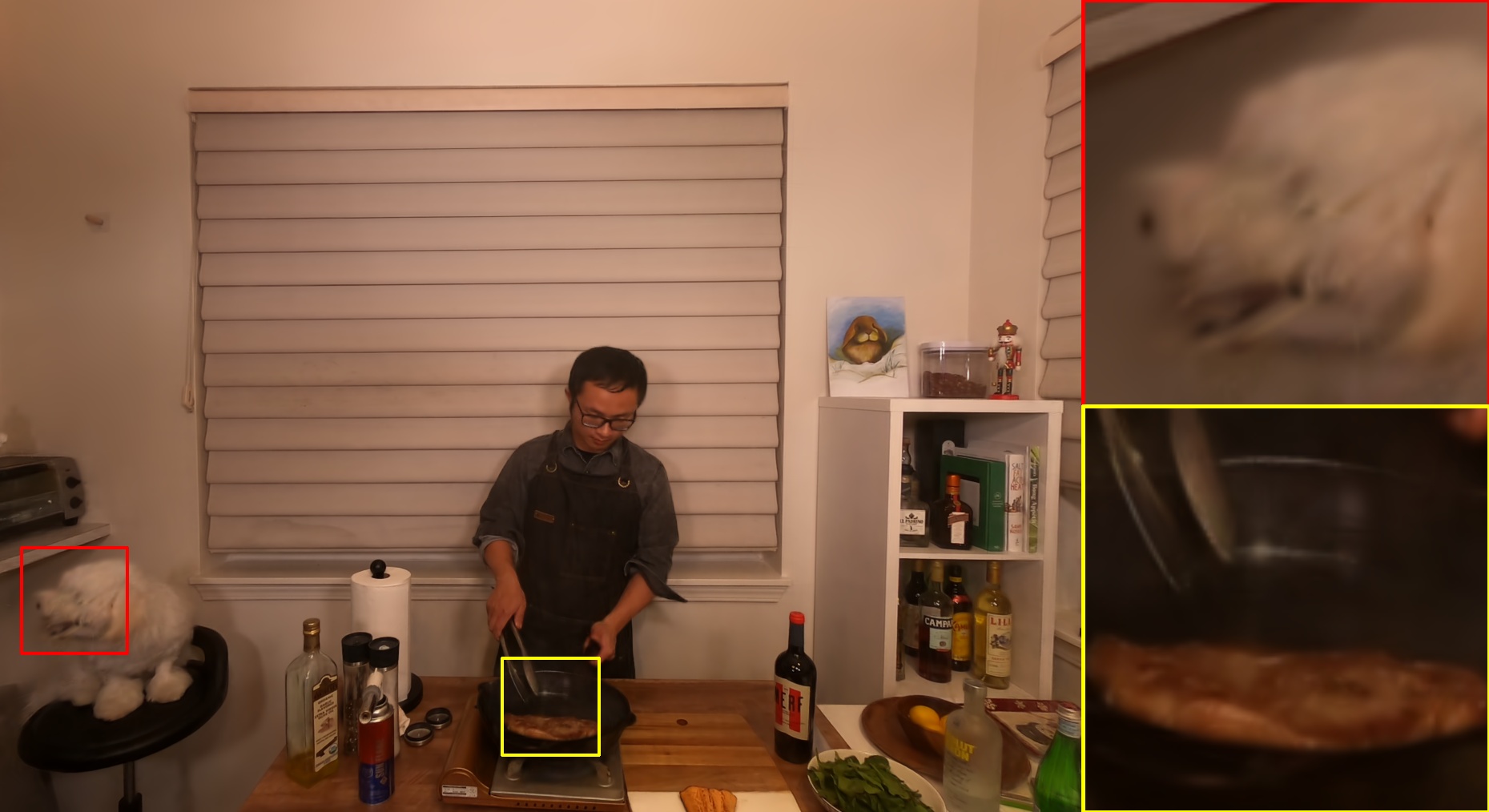}
   }
   \subfloat[Ours(ScaffoldGS)]{
      \includegraphics[width=0.24\textwidth]{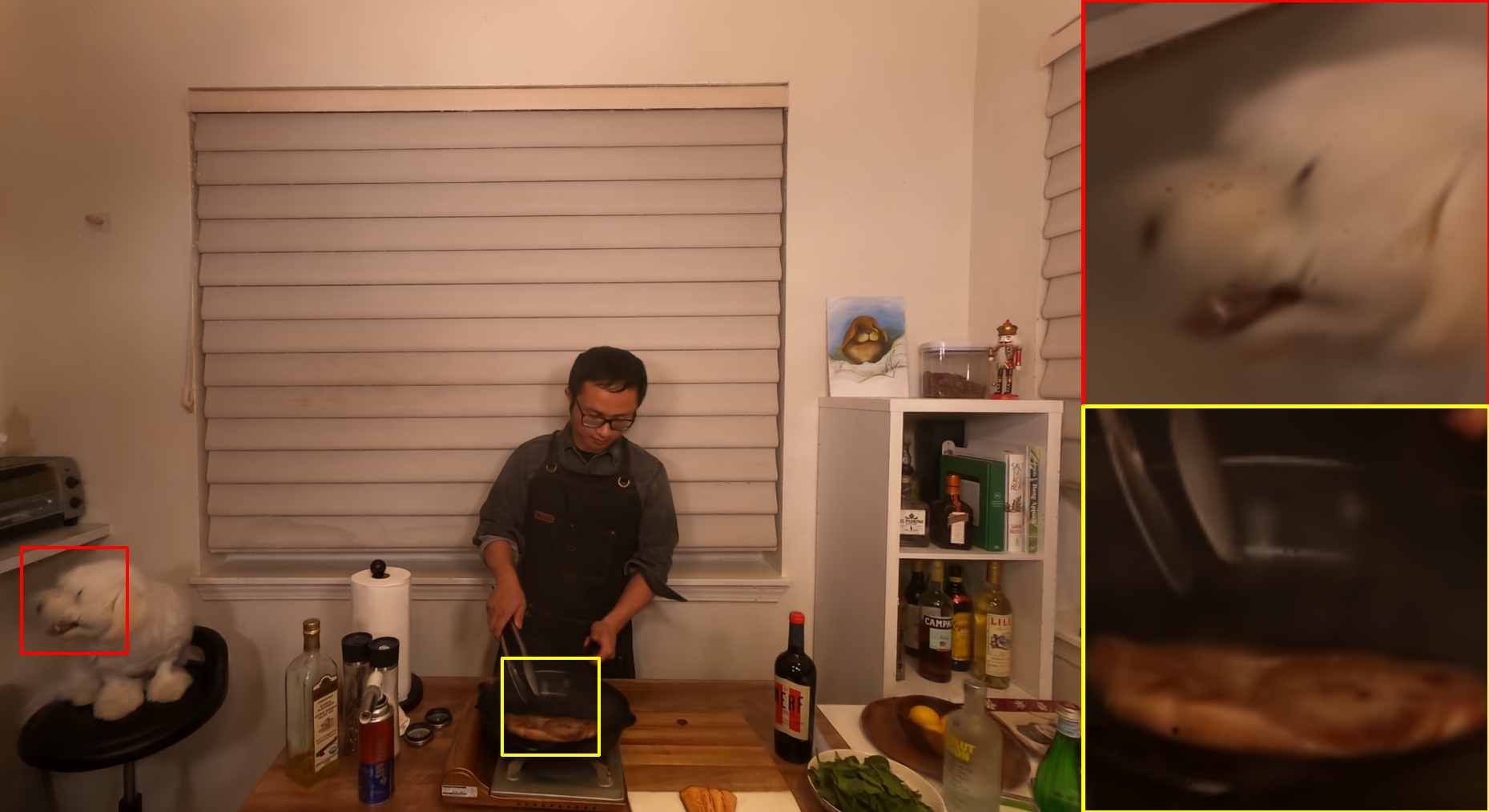}
   }
   \subfloat[Ground Truth]{
      \includegraphics[width=0.24\textwidth]{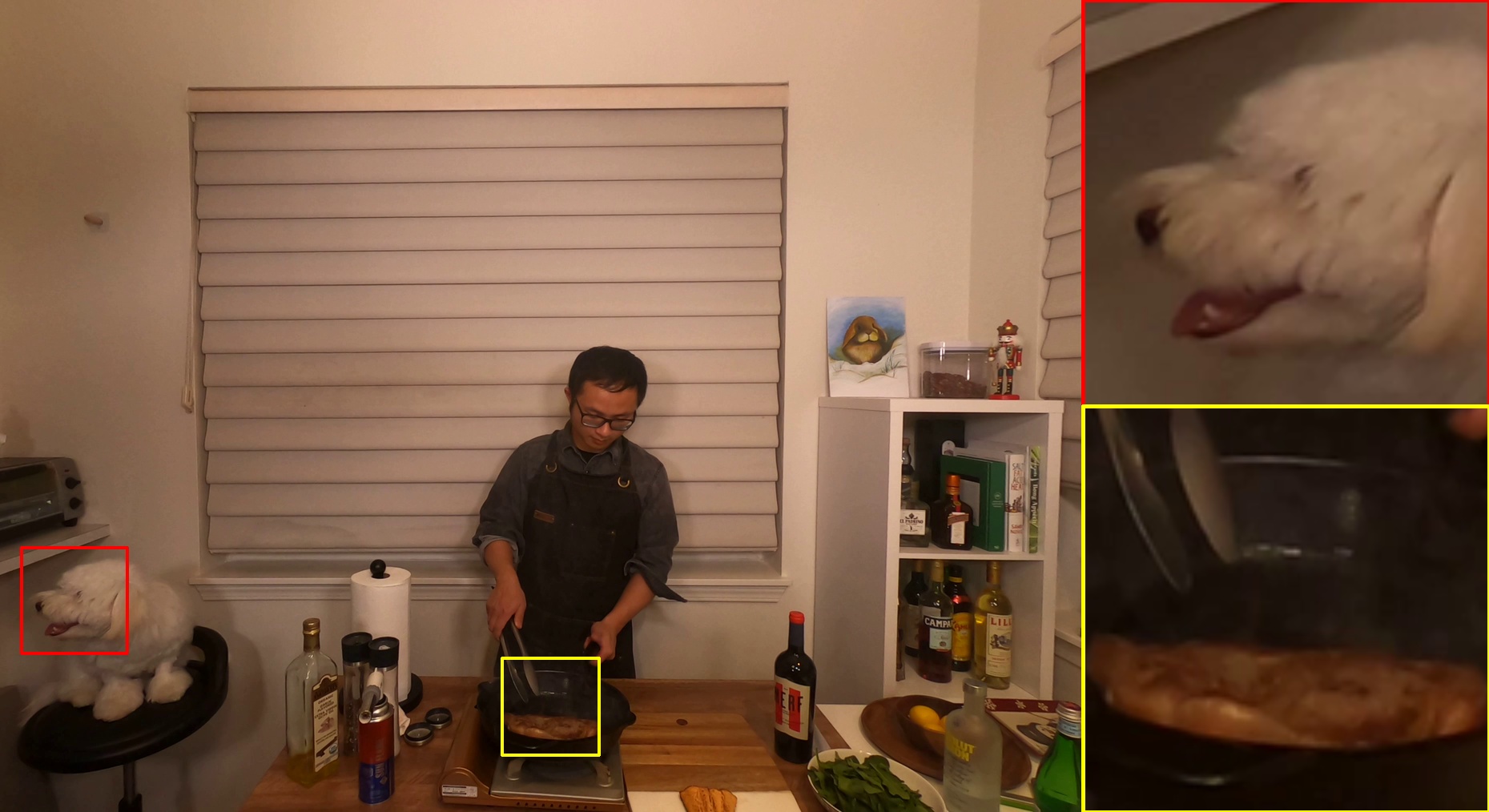}
   }
   \caption{Qualitative comparison on \textit{Sear Steak} scene of N3DV dataset.}
   \label{fig:sear}
   \vspace{-0.2cm}
\end{figure}

\begin{figure}[!t]
   \centering

   \subfloat[StreamSTGS~\cite{ke2025streamstgs}]{
      \includegraphics[width=0.24\textwidth]{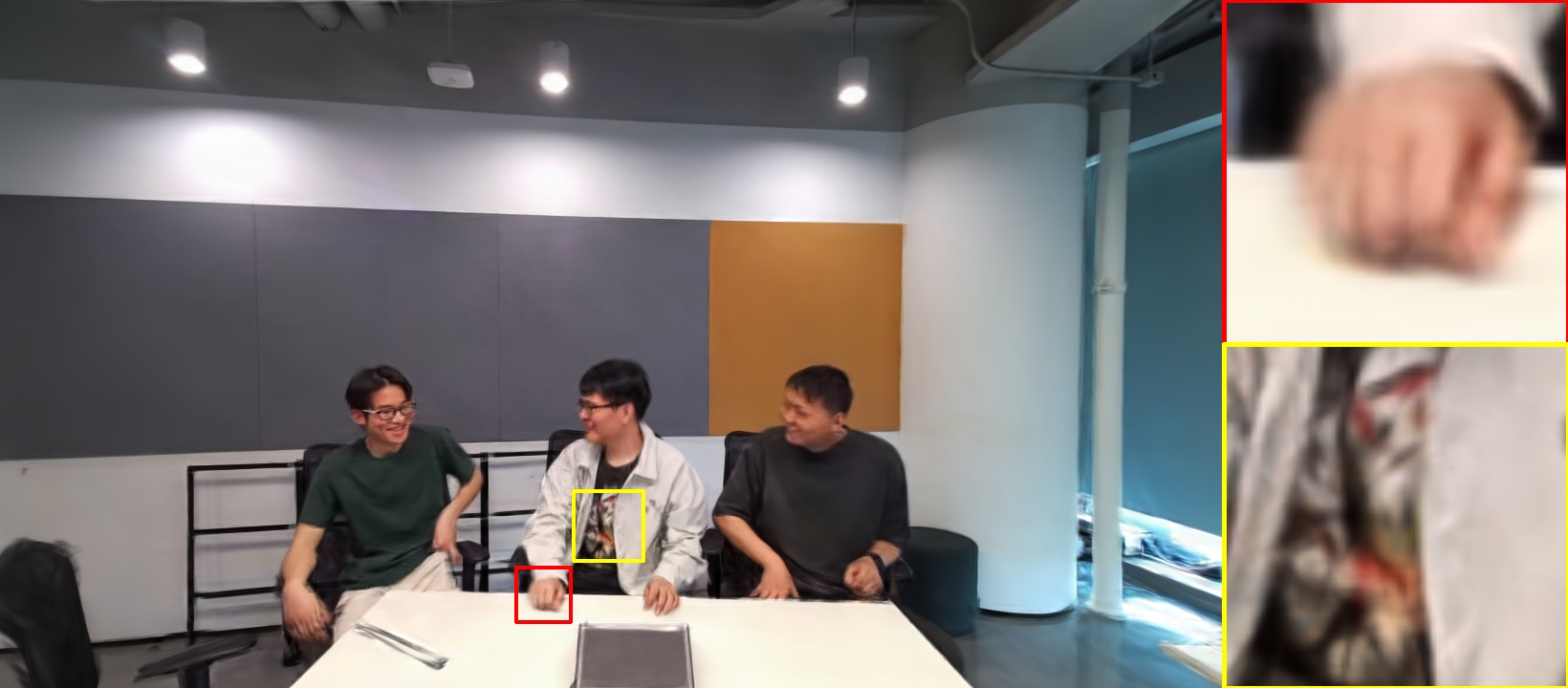}
   }
   \subfloat[QUEEN~\cite{girish2024queen}]{
      \includegraphics[width=0.24\textwidth]{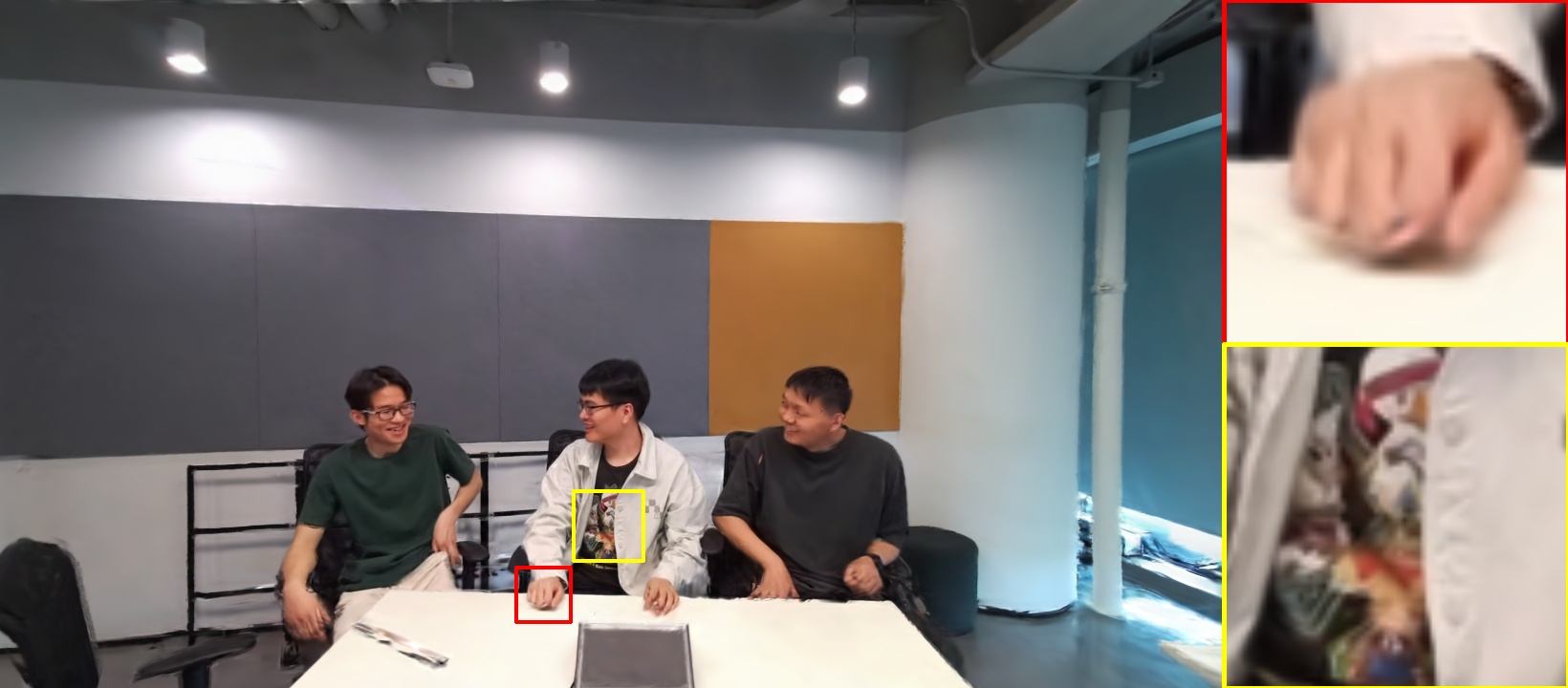}
   }
   \subfloat[HiCoM~\cite{gao2024hicom}]{
      \includegraphics[width=0.24\textwidth]{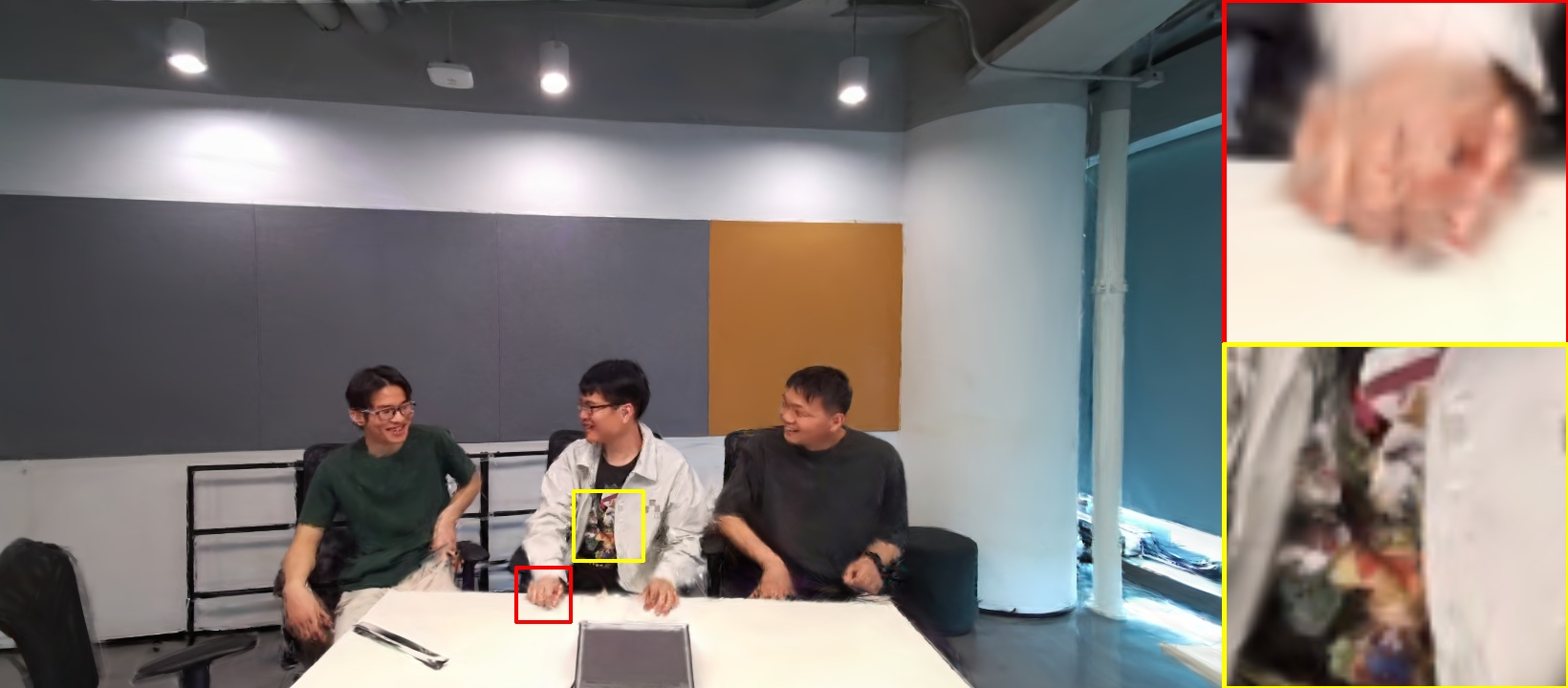}
   }
   \subfloat[Ours (3DGS)]{
      \includegraphics[width=0.24\textwidth]{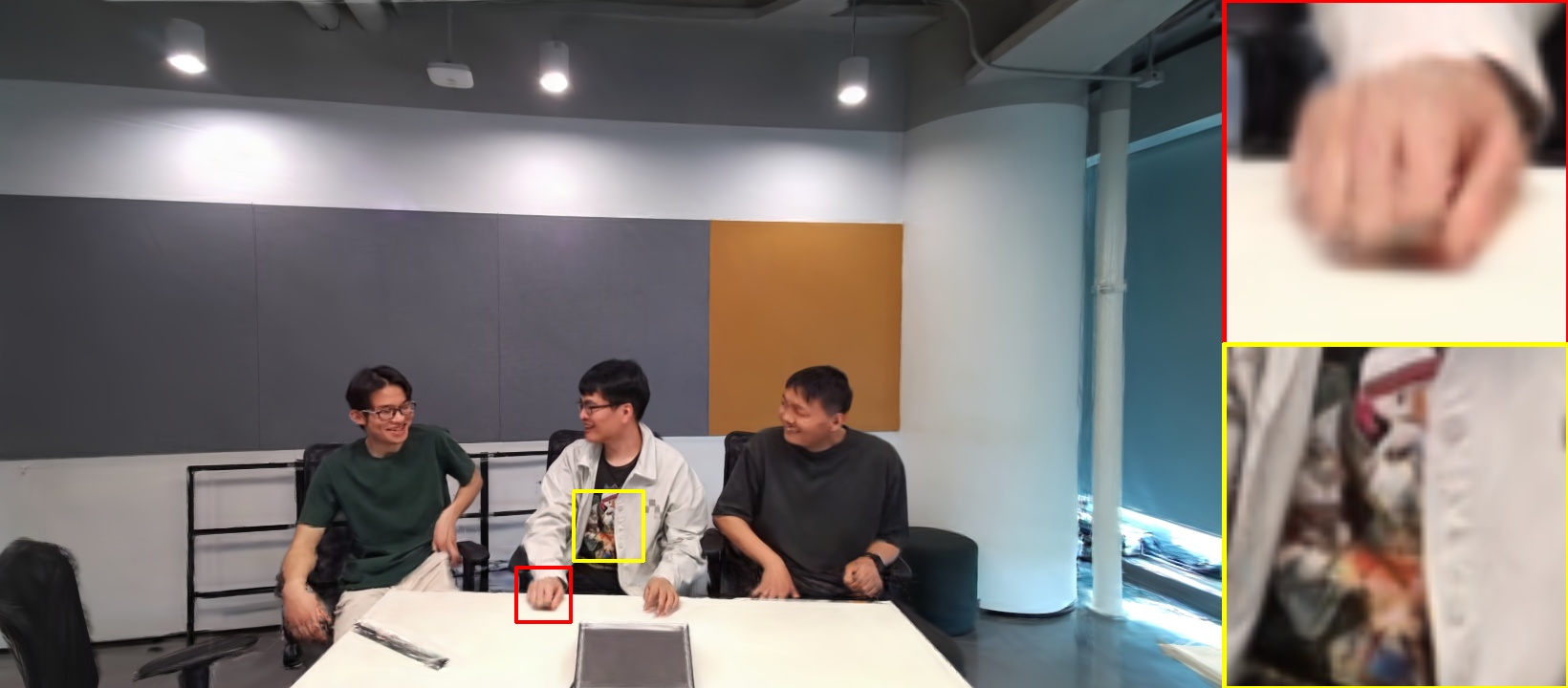}
   }
   \\[-0.2em] 


   \subfloat[GIFStream~\cite{li2025gifstream}]{
      \includegraphics[width=0.24\textwidth]{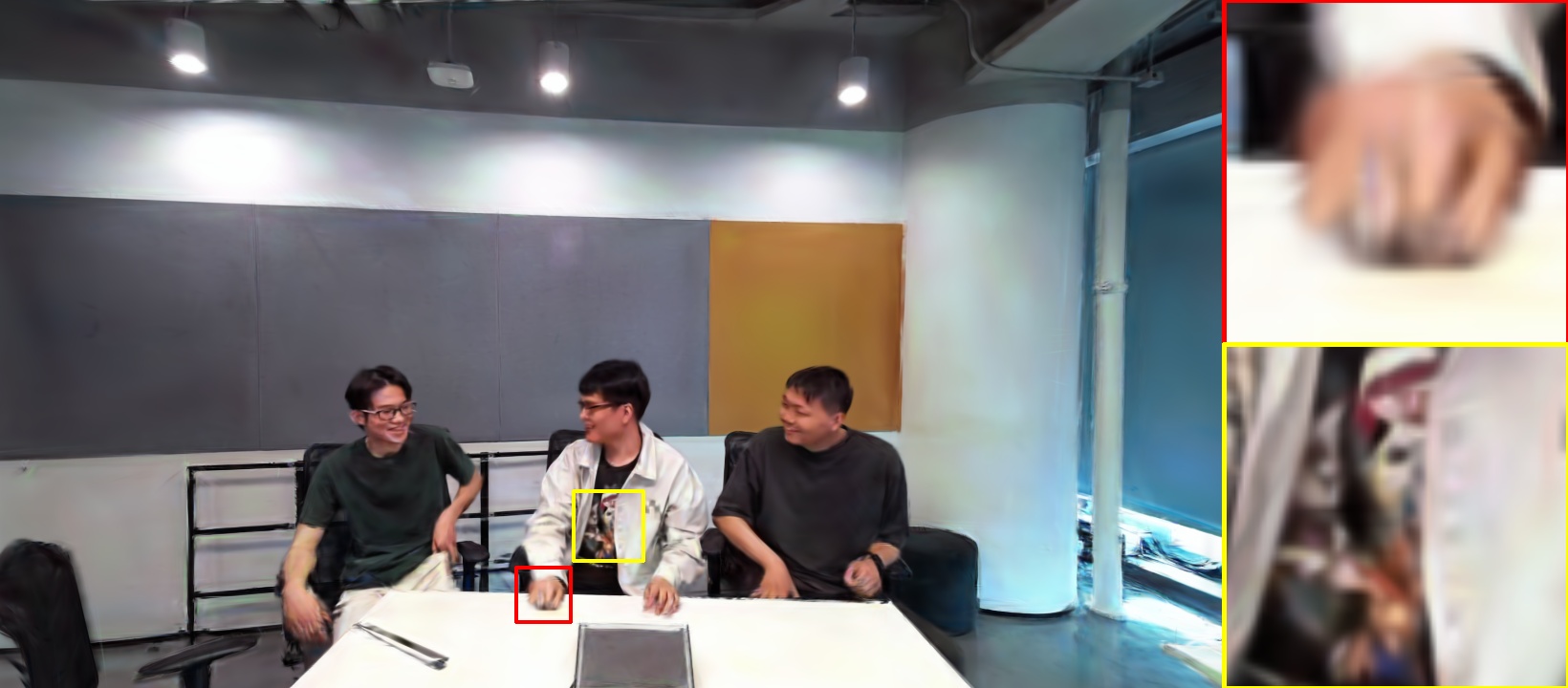}
   }
   \subfloat[iFVC~\cite{tang2025compressing}]{
      \includegraphics[width=0.24\textwidth]{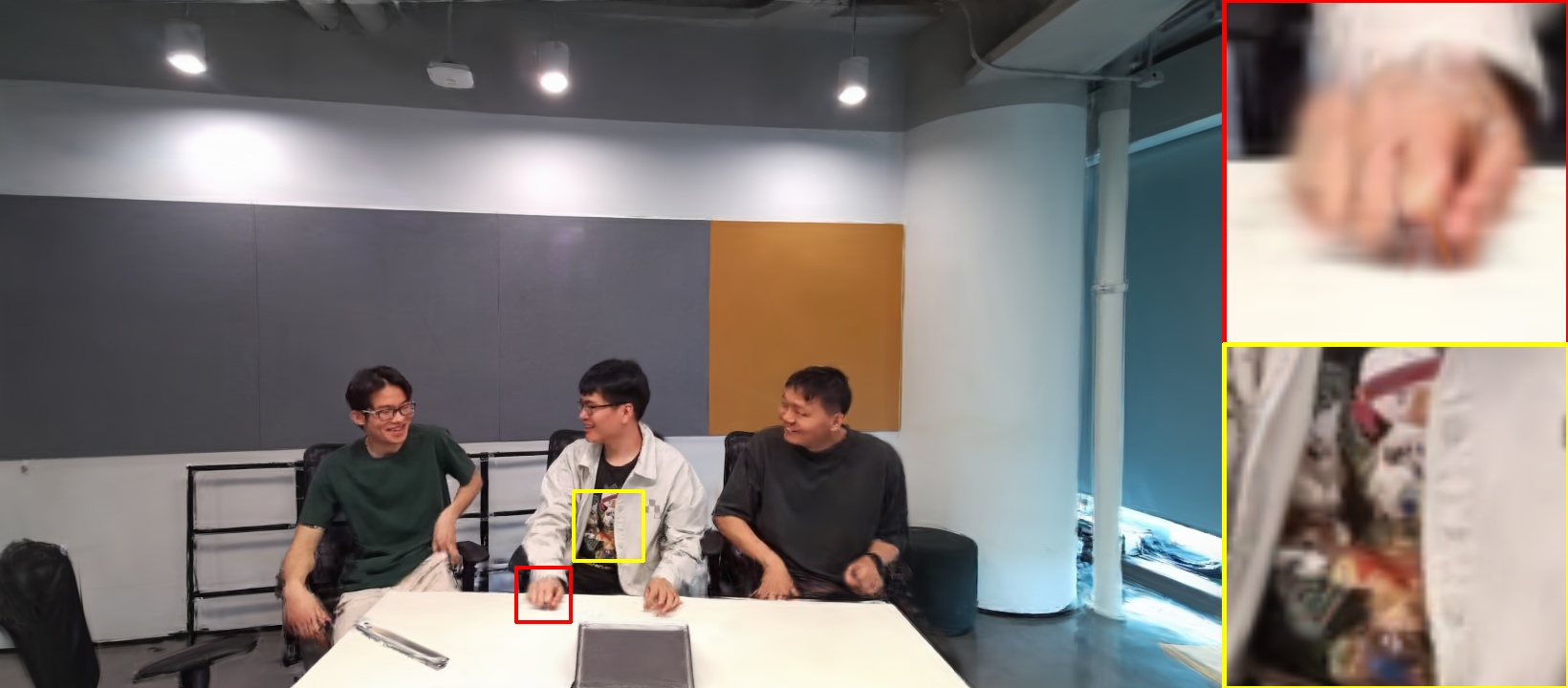}
   }
   \subfloat[Ours(ScaffoldGS)]{
      \includegraphics[width=0.24\textwidth]{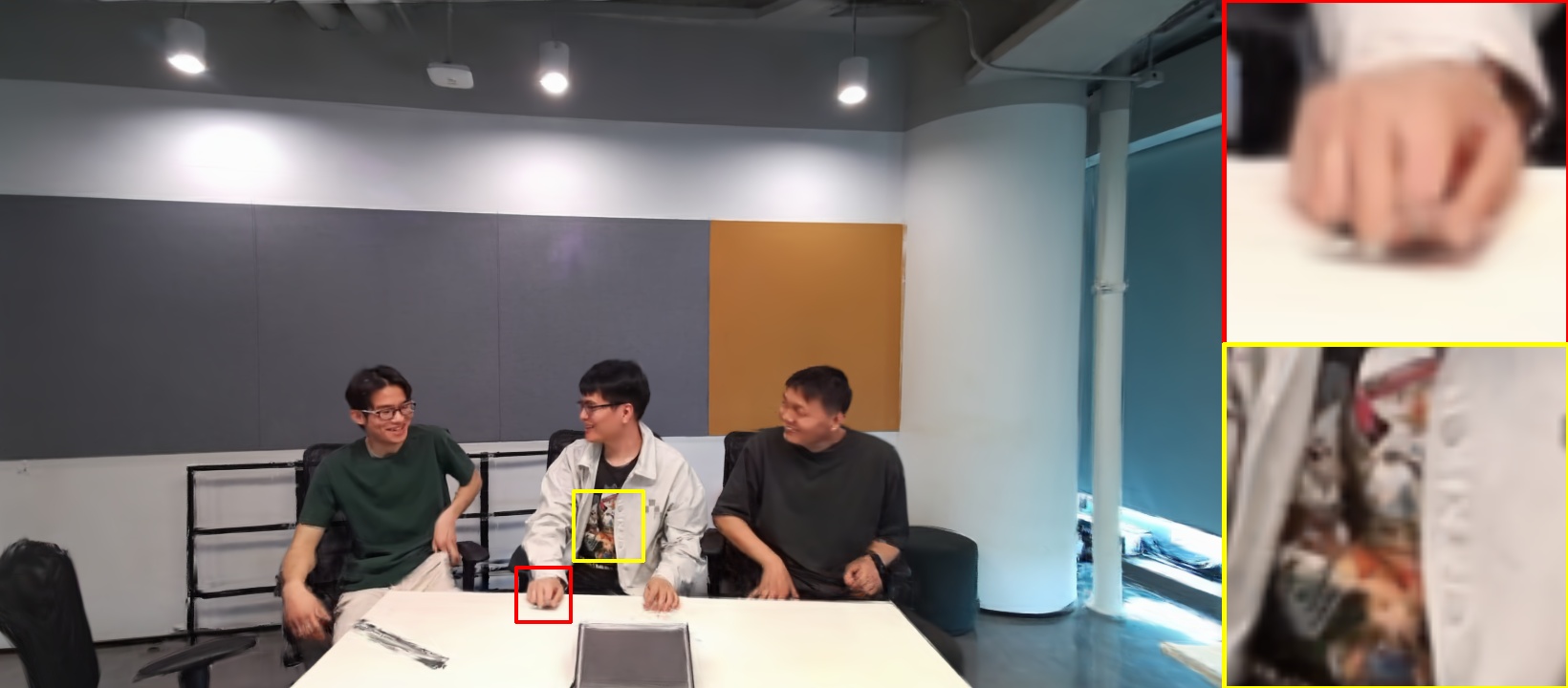}
   }
   \subfloat[Ground Truth]{
      \includegraphics[width=0.24\textwidth]{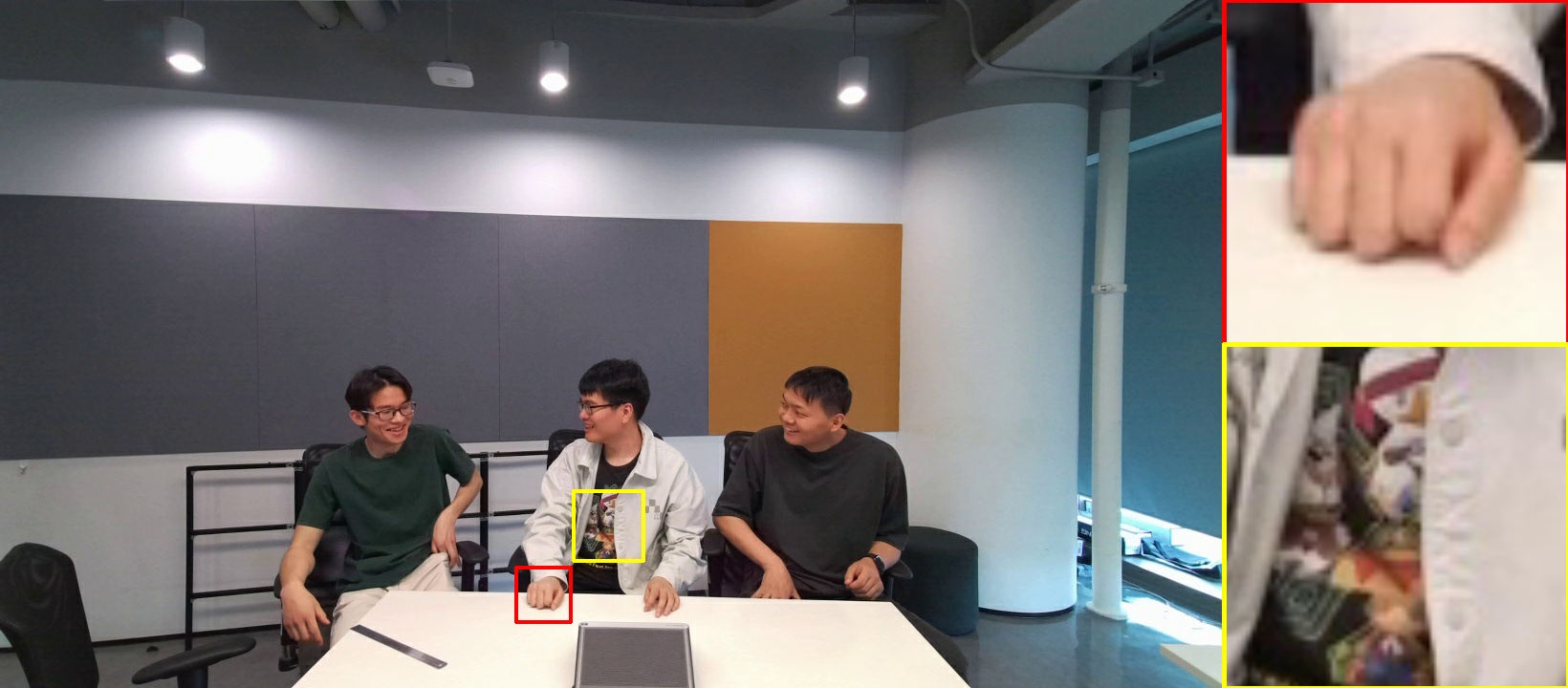}
   }
   \caption{Qualitative comparison on \textit{Discussion} scene of MeetRoom dataset.}
   \label{fig:discussion}
   \vspace{-0.2cm}
\end{figure}

\begin{figure}[!t]
   \centering

   \subfloat[StreamSTGS~\cite{ke2025streamstgs}]{
      \includegraphics[width=0.24\textwidth]{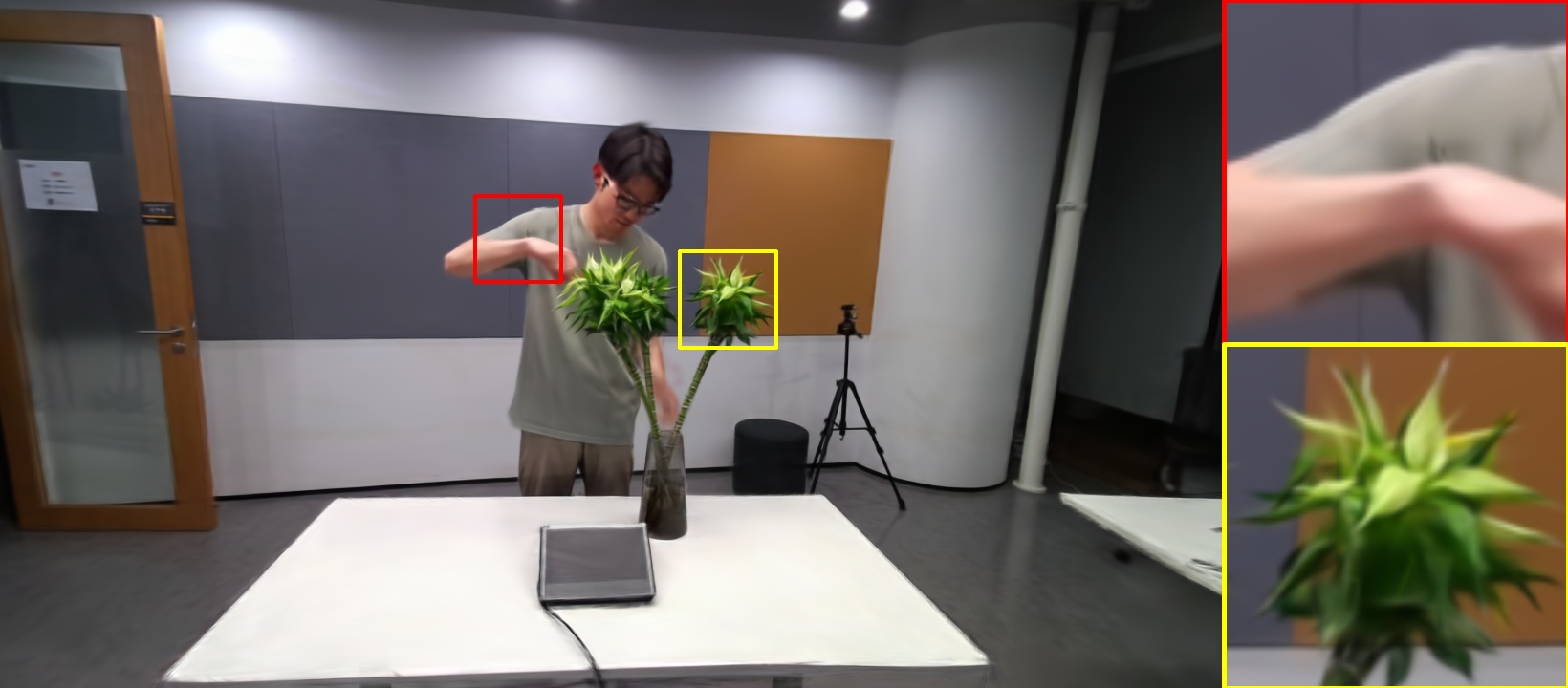}
   }
   \subfloat[QUEEN~\cite{girish2024queen}]{
      \includegraphics[width=0.24\textwidth]{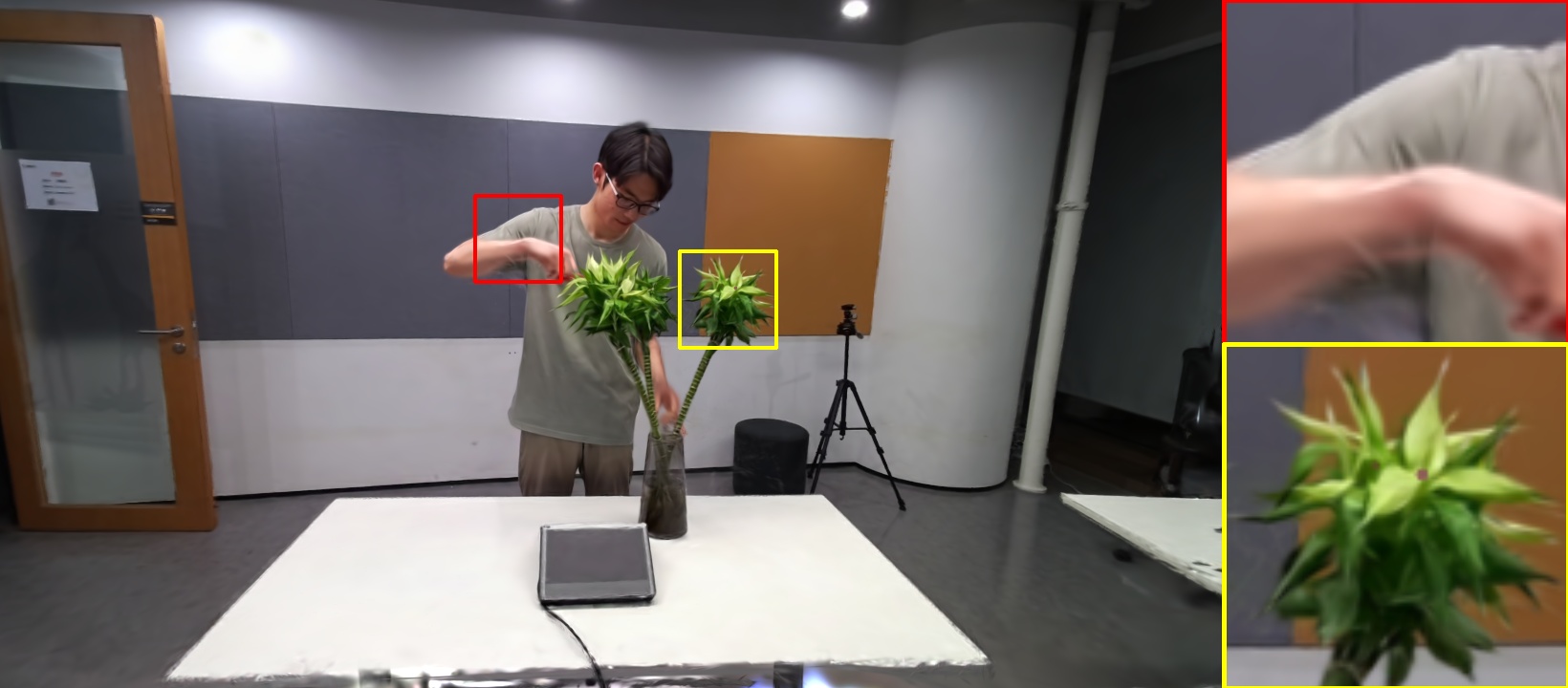}
   }
   \subfloat[HiCoM~\cite{gao2024hicom}]{
      \includegraphics[width=0.24\textwidth]{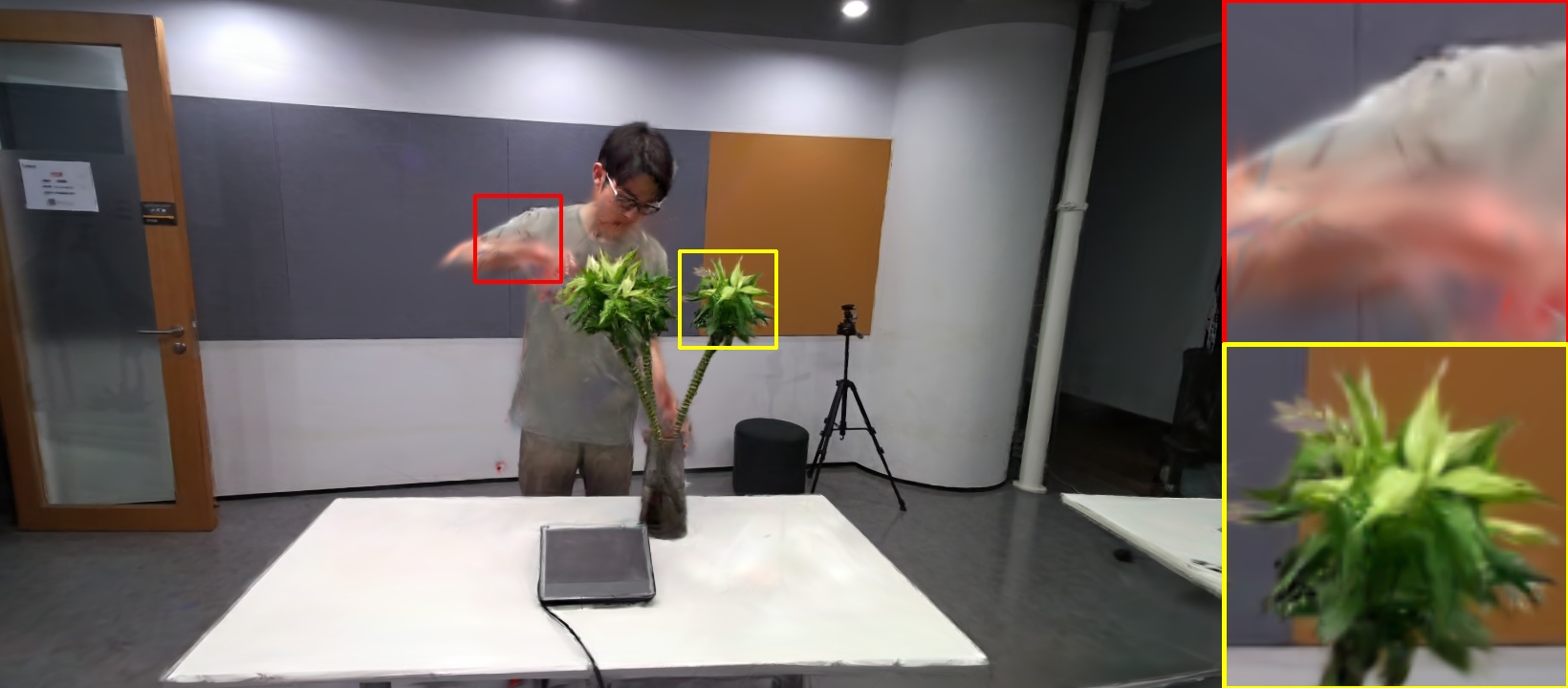}
   }
   \subfloat[Ours (3DGS)]{
      \includegraphics[width=0.24\textwidth]{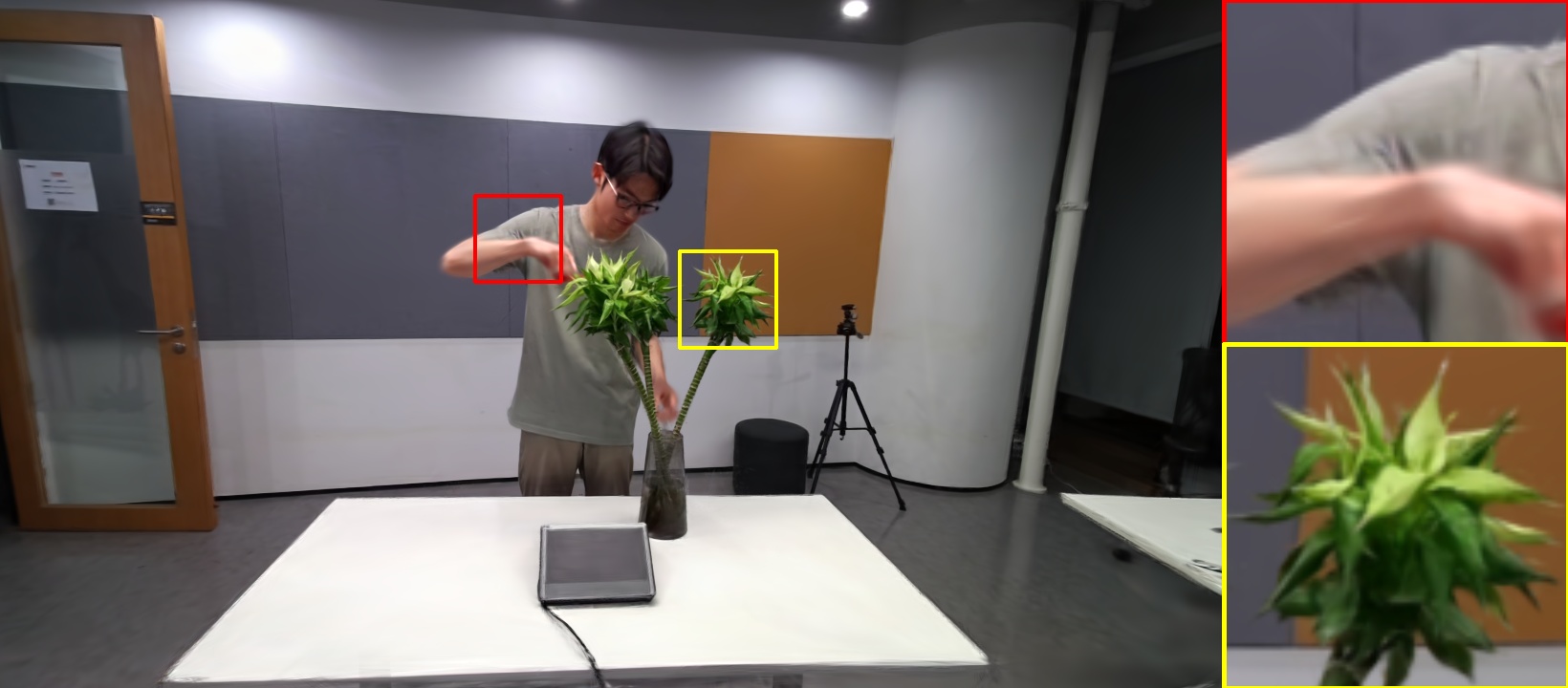}
   }
   \\[-0.2em] 


   \subfloat[GIFStream~\cite{li2025gifstream}]{
      \includegraphics[width=0.24\textwidth]{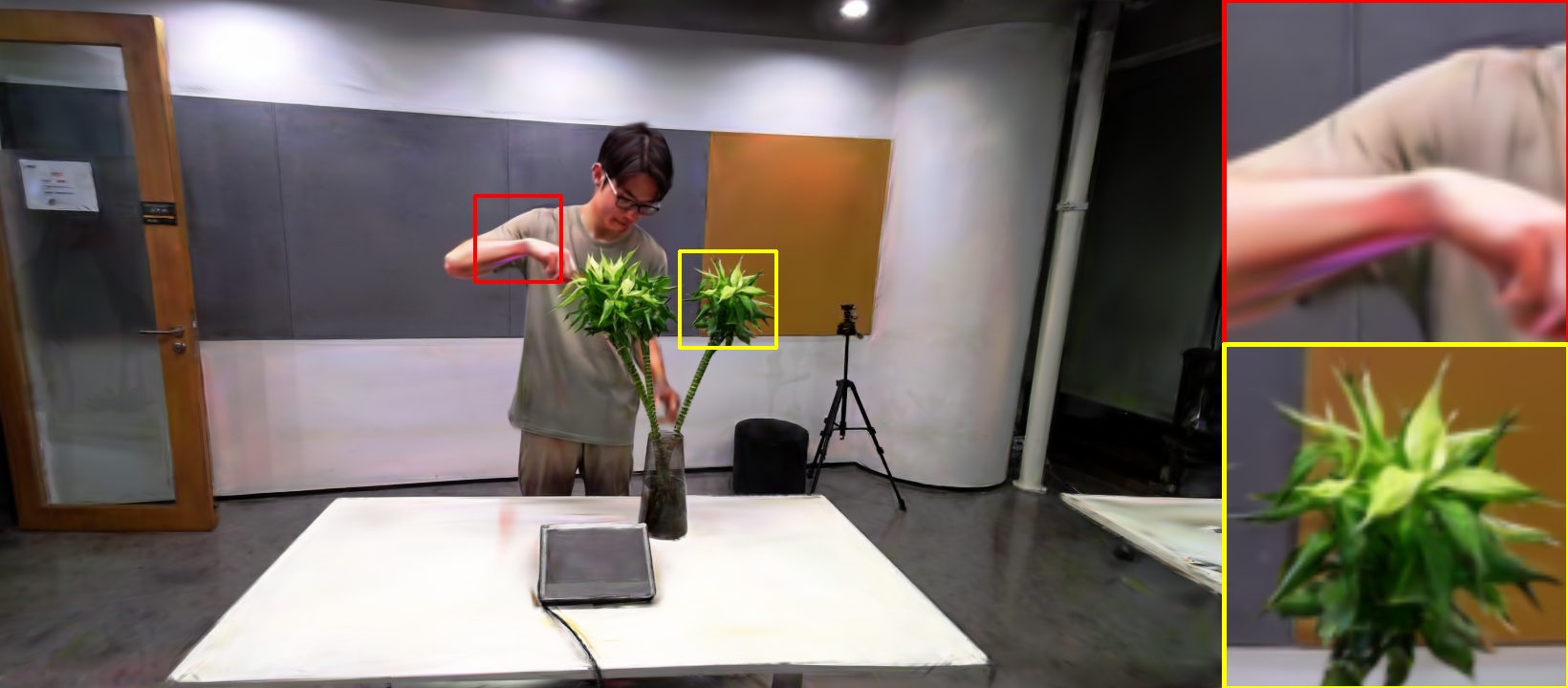}
   }
   \subfloat[iFVC~\cite{tang2025compressing}]{
      \includegraphics[width=0.24\textwidth]{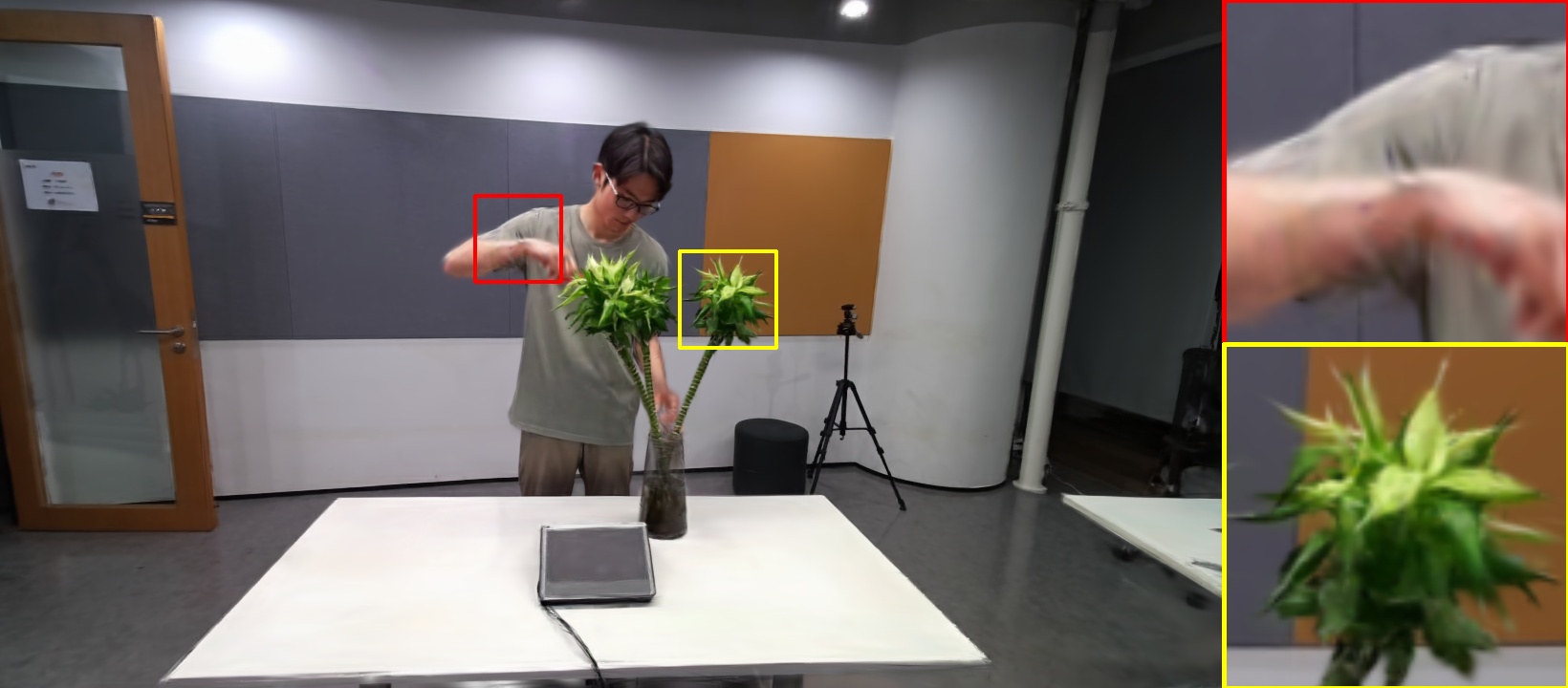}
   }
   \subfloat[Ours(ScaffoldGS)]{
      \includegraphics[width=0.24\textwidth]{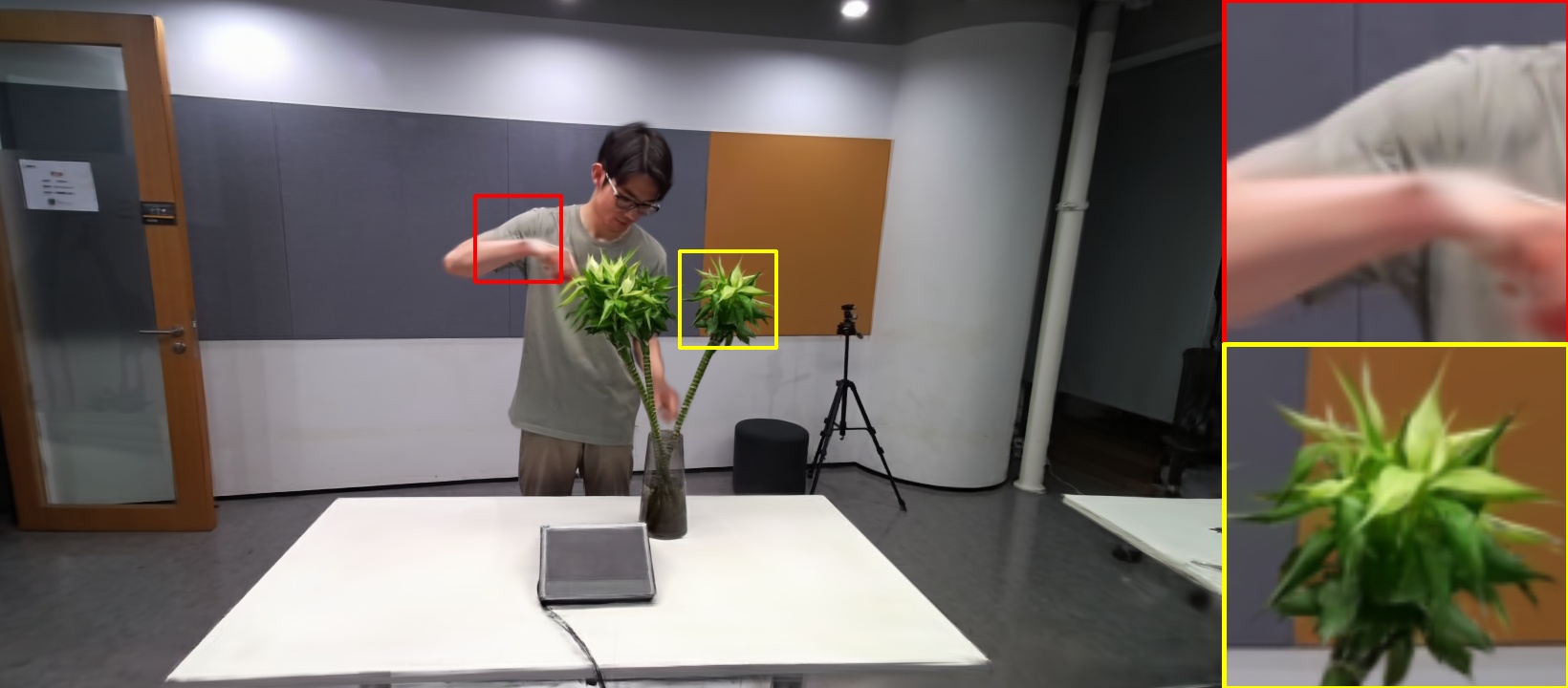}
   }
   \subfloat[Ground Truth]{
      \includegraphics[width=0.24\textwidth]{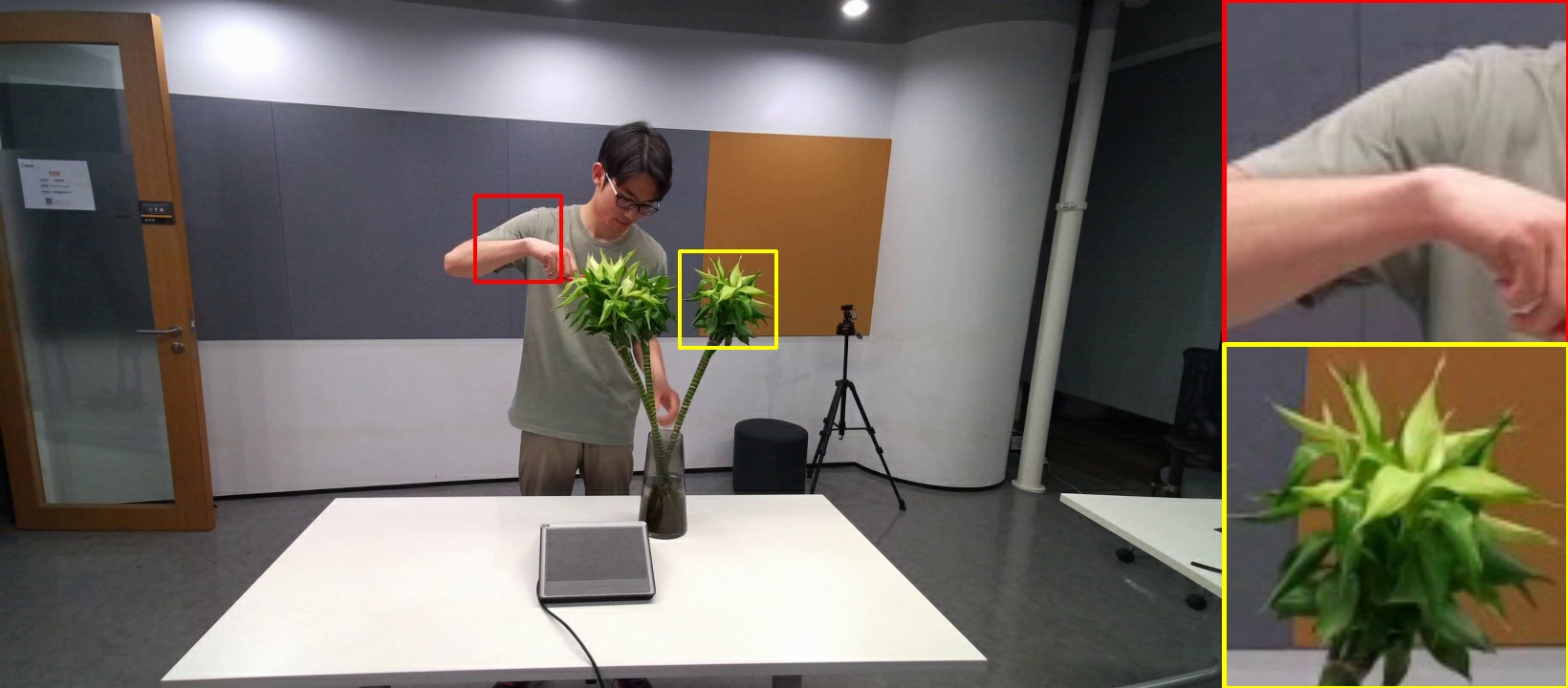}
   }
   \caption{Qualitative comparison on \textit{Trimming} scene of MeetRoom dataset.}
   \label{fig:trimming}
   \vspace{-0.2cm}
\end{figure}

\begin{figure}[!t]
   \centering

   \subfloat[StreamSTGS~\cite{ke2025streamstgs}]{
      \includegraphics[width=0.24\textwidth]{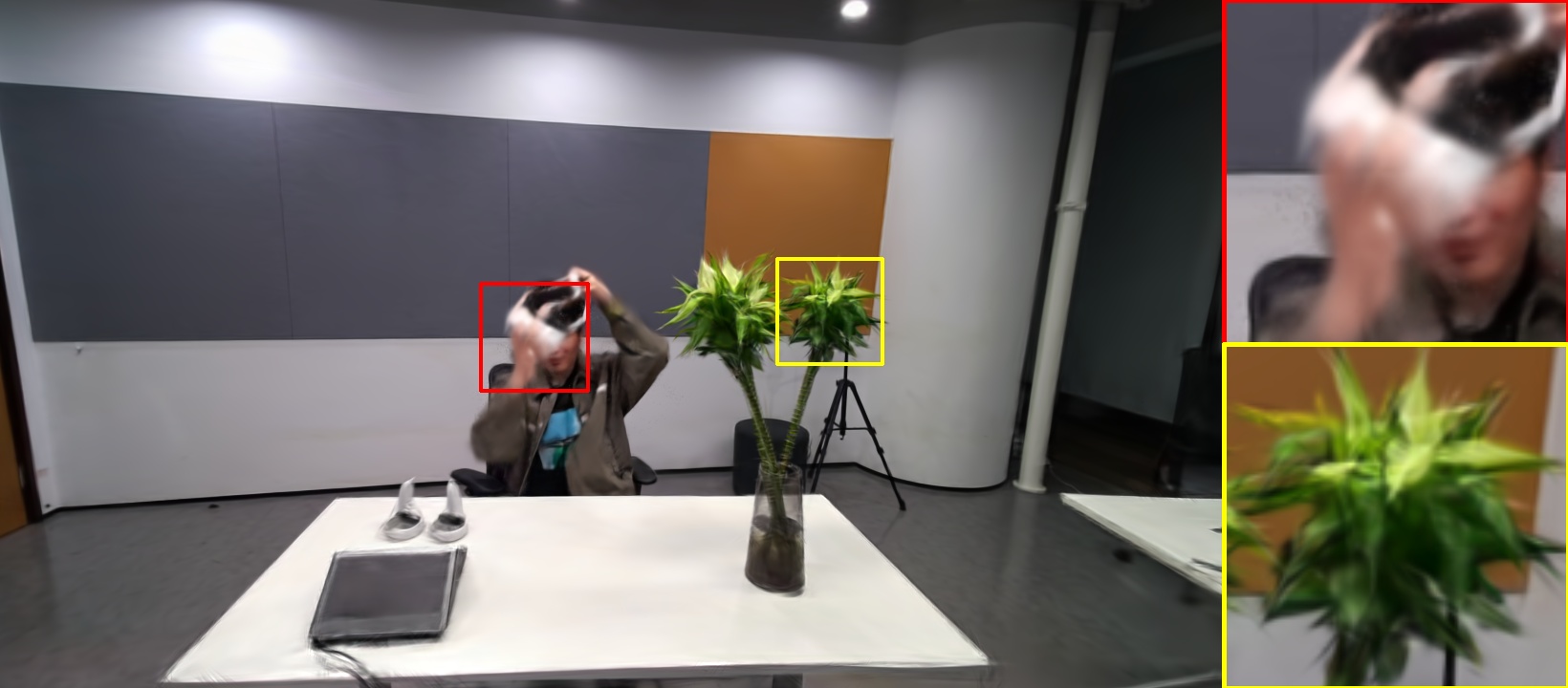}
   }
   \subfloat[QUEEN~\cite{girish2024queen}]{
      \includegraphics[width=0.24\textwidth]{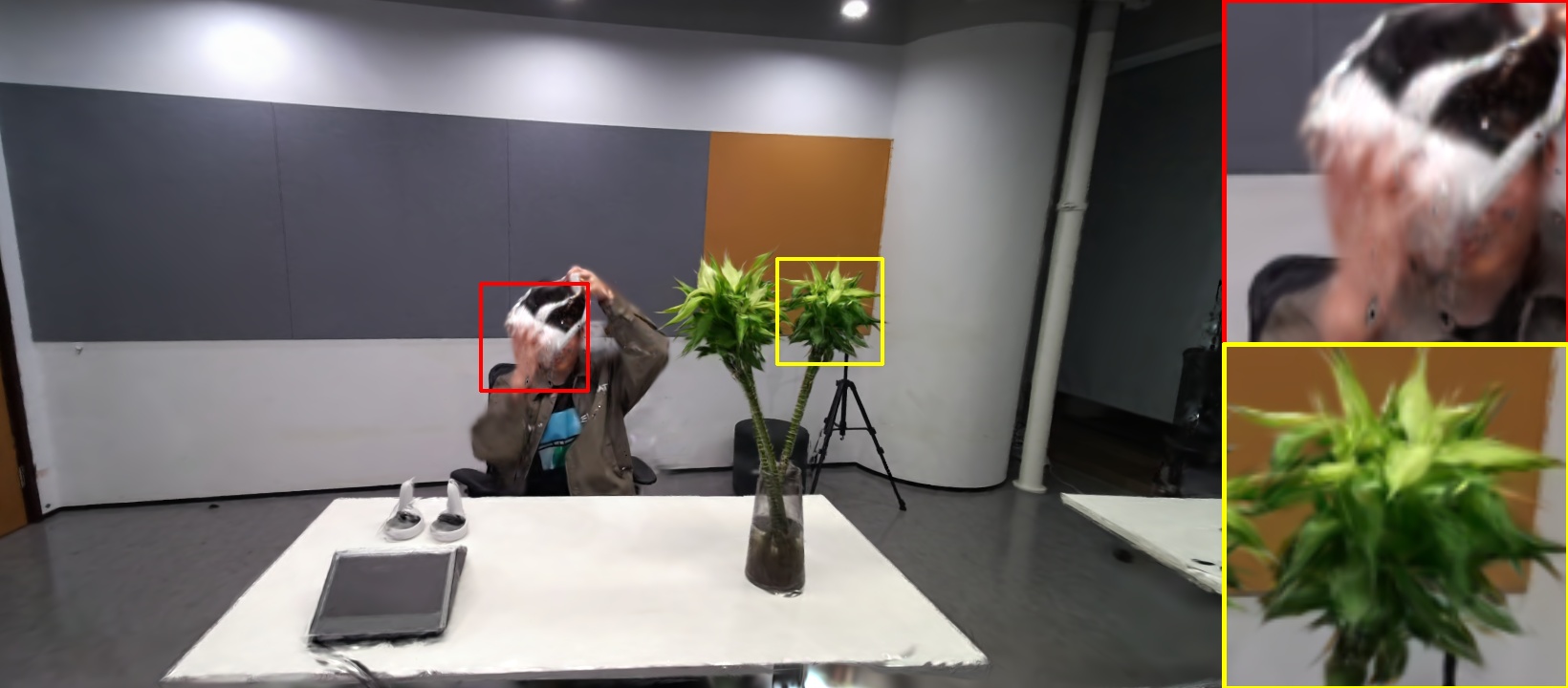}
   }
   \subfloat[HiCoM~\cite{gao2024hicom}]{
      \includegraphics[width=0.24\textwidth]{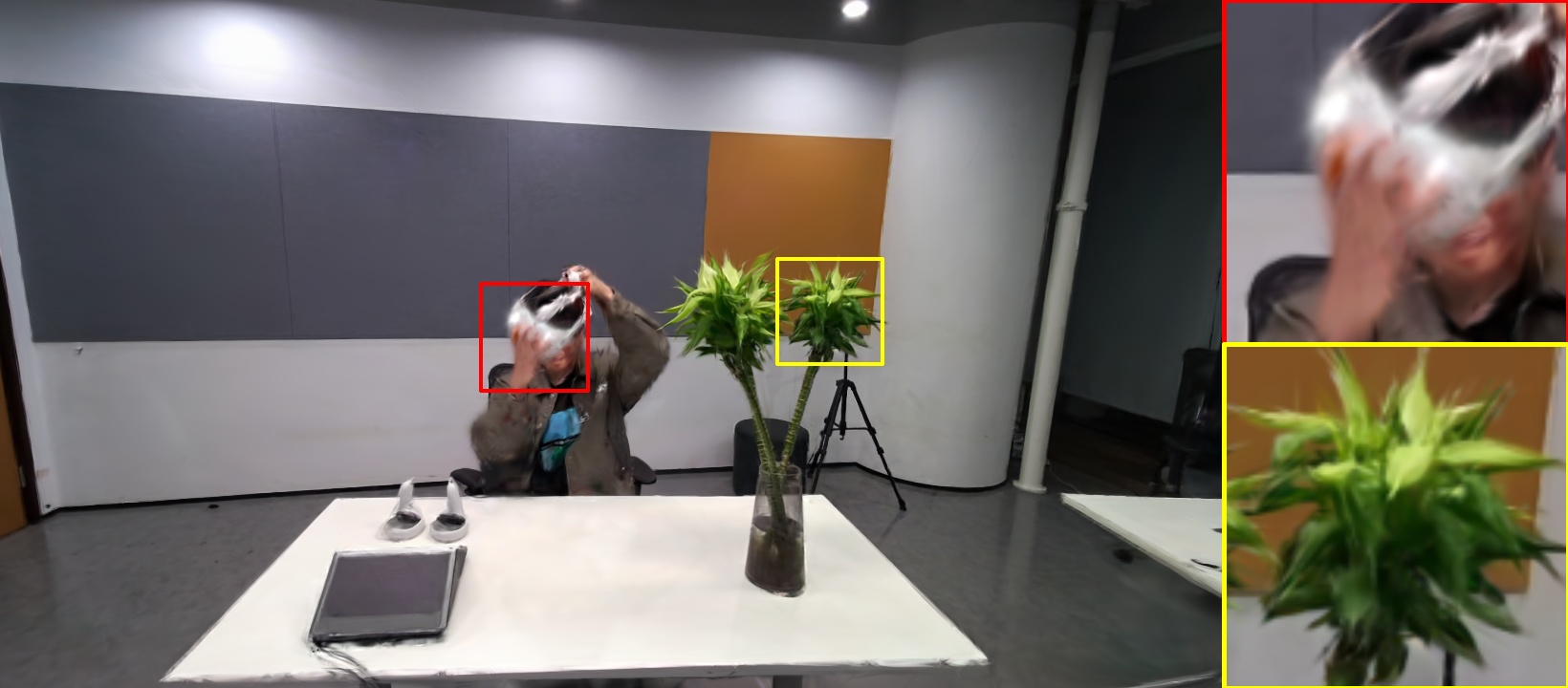}
   }
   \subfloat[Ours (3DGS)]{
      \includegraphics[width=0.24\textwidth]{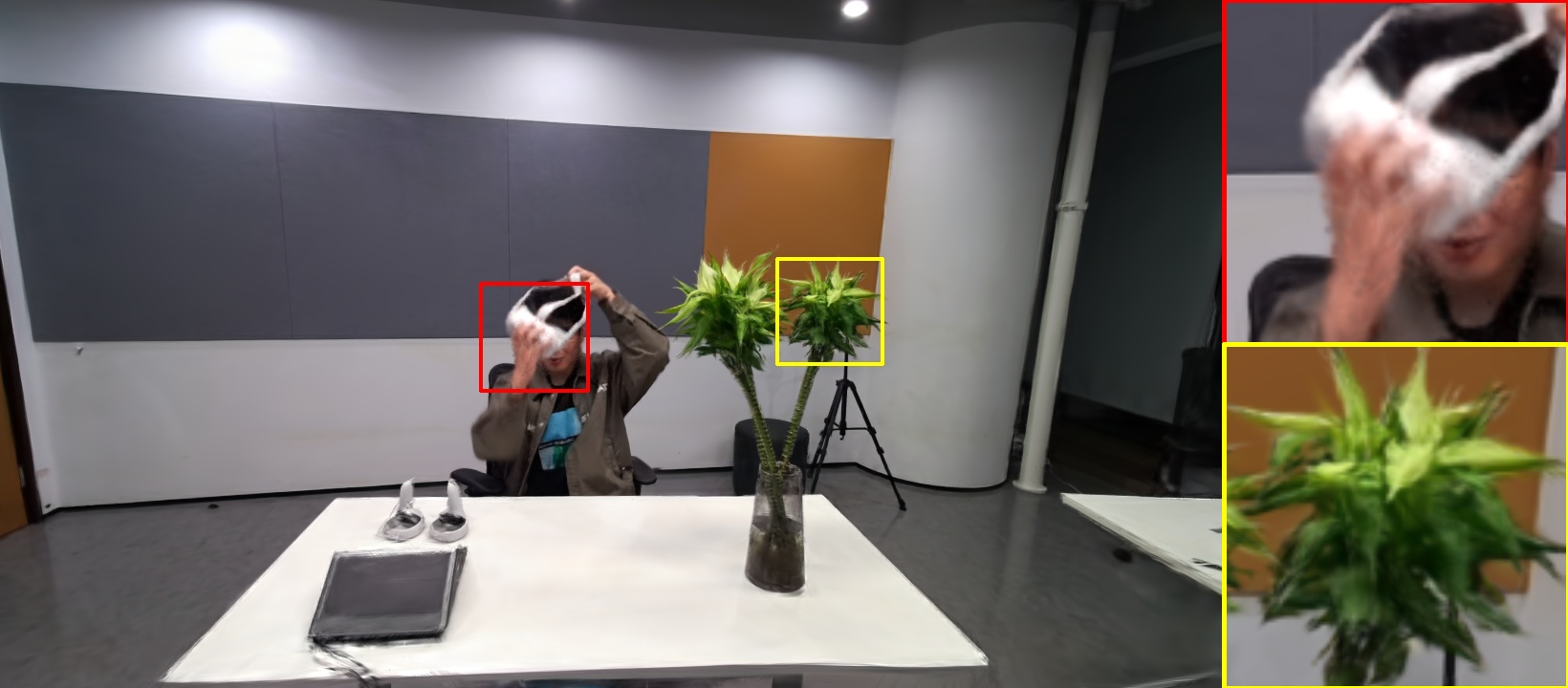}
   }
   \\[-0.2em] 


   \subfloat[GIFStream~\cite{li2025gifstream}]{
      \includegraphics[width=0.24\textwidth]{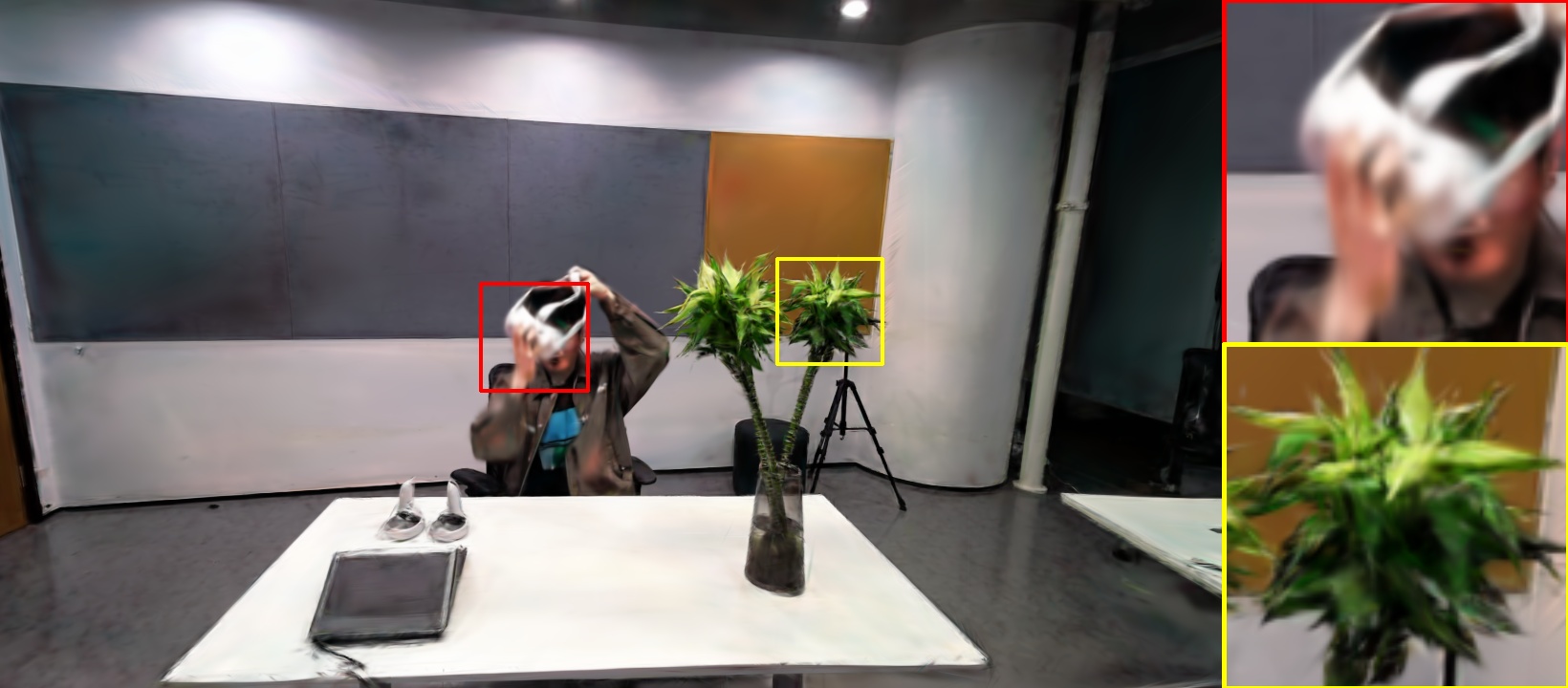}
   }
   \subfloat[iFVC~\cite{tang2025compressing}]{
      \includegraphics[width=0.24\textwidth]{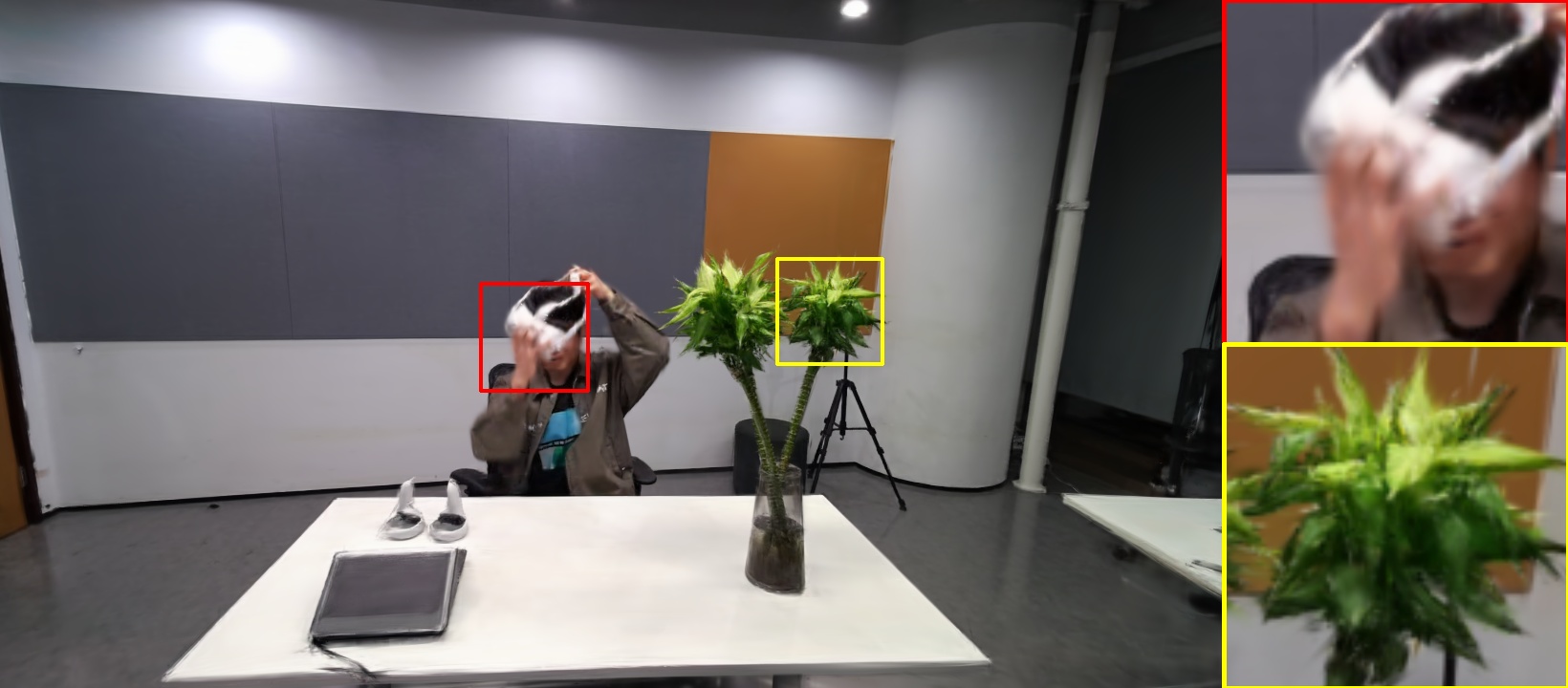}
   }
   \subfloat[Ours(ScaffoldGS)]{
      \includegraphics[width=0.24\textwidth]{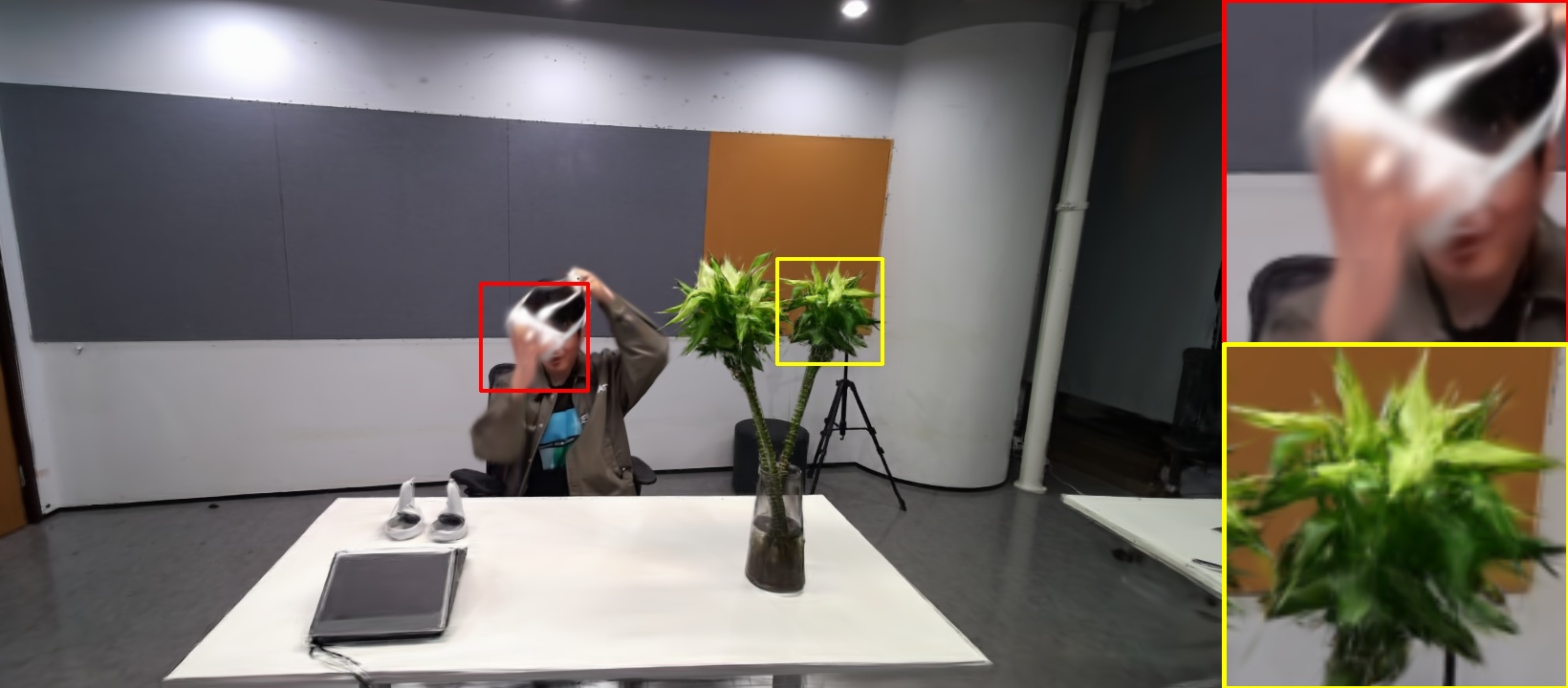}
   }
   \subfloat[Ground Truth]{
      \includegraphics[width=0.24\textwidth]{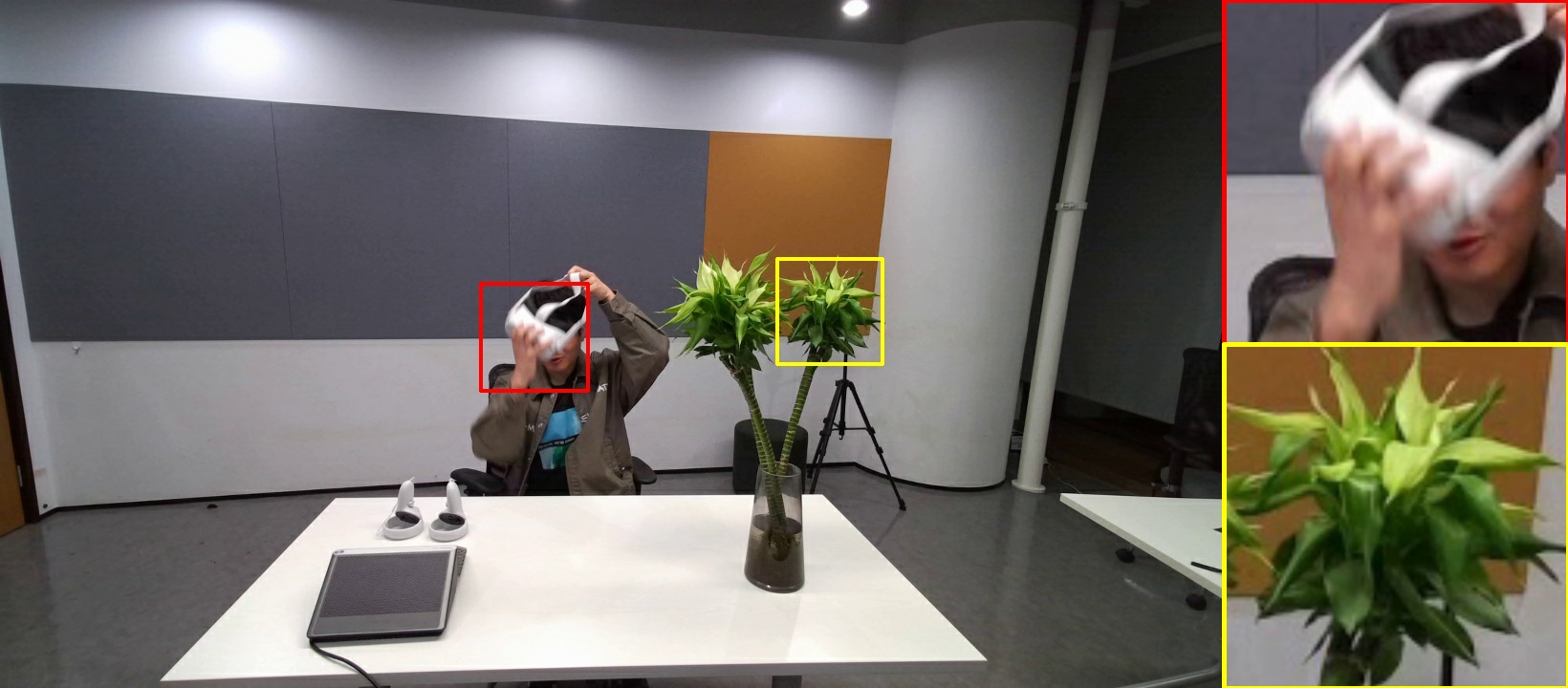}
   }
   \caption{Qualitative comparison on \textit{Vrheadset} scene of MeetRoom dataset.}
   \label{fig:vrheadset}
   \vspace{-0.2cm}
\end{figure}

\begin{figure}[!t]
   \centering

   \subfloat[iFVC~\cite{tang2025compressing}]{
      \includegraphics[width=0.32\textwidth]{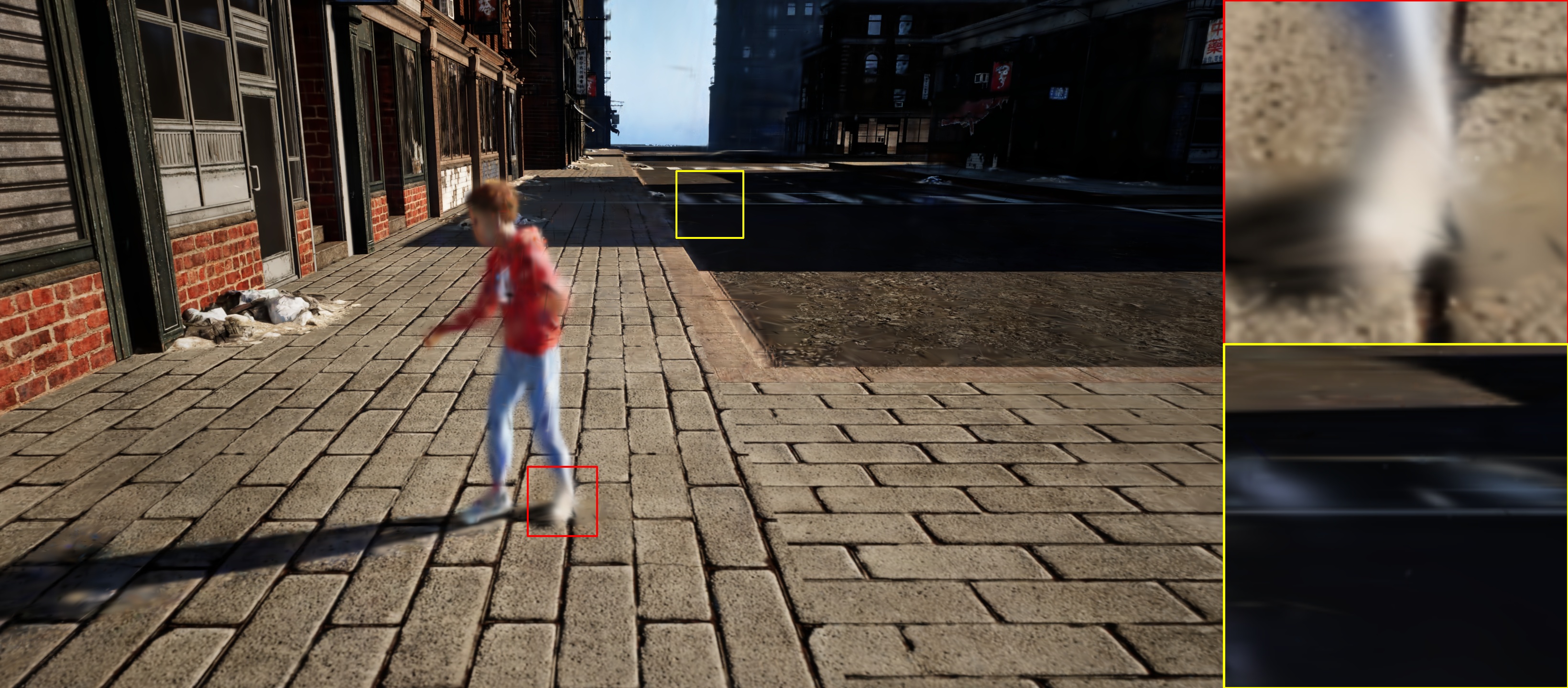}
   }
   \subfloat[Ours (Scaffold-GS)]{
      \includegraphics[width=0.32\textwidth]{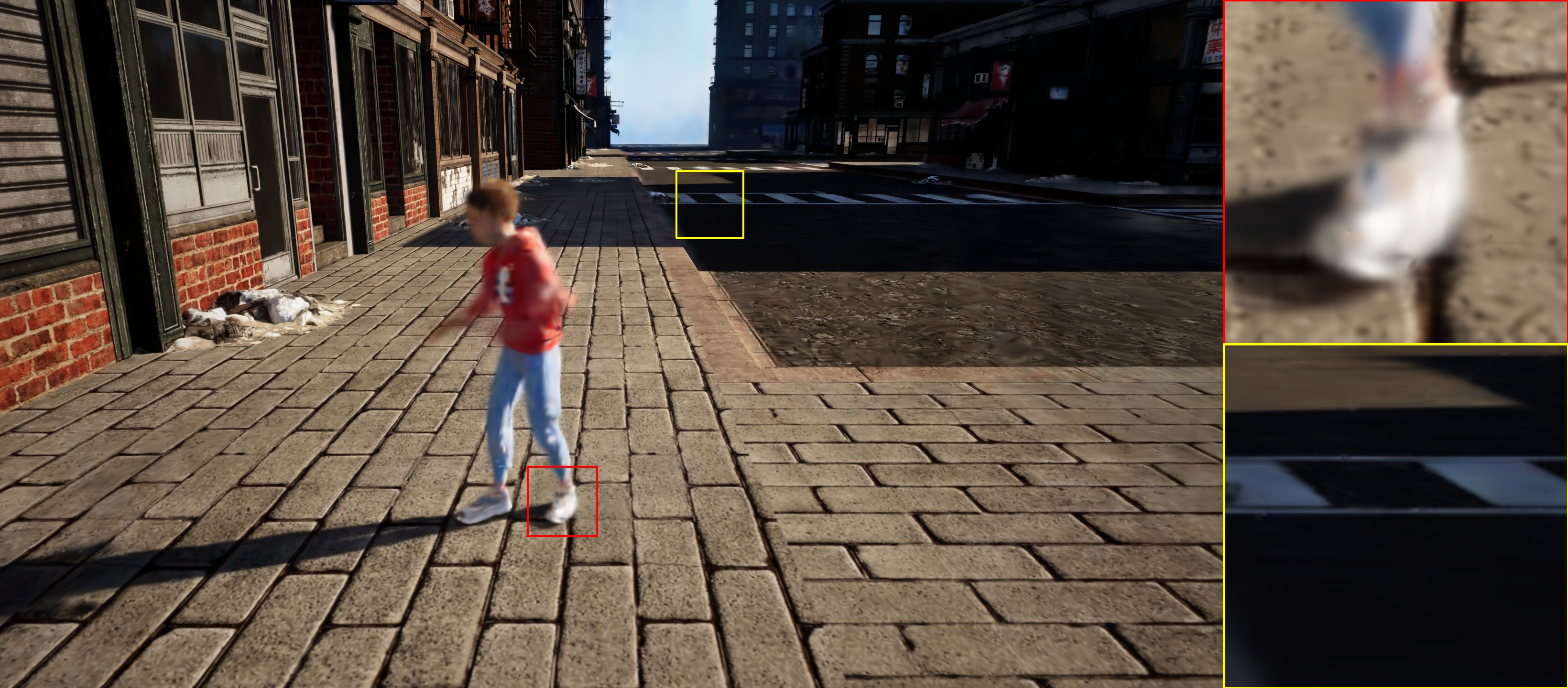}
   }
   \subfloat[Ground Truth]{
      \includegraphics[width=0.32\textwidth]{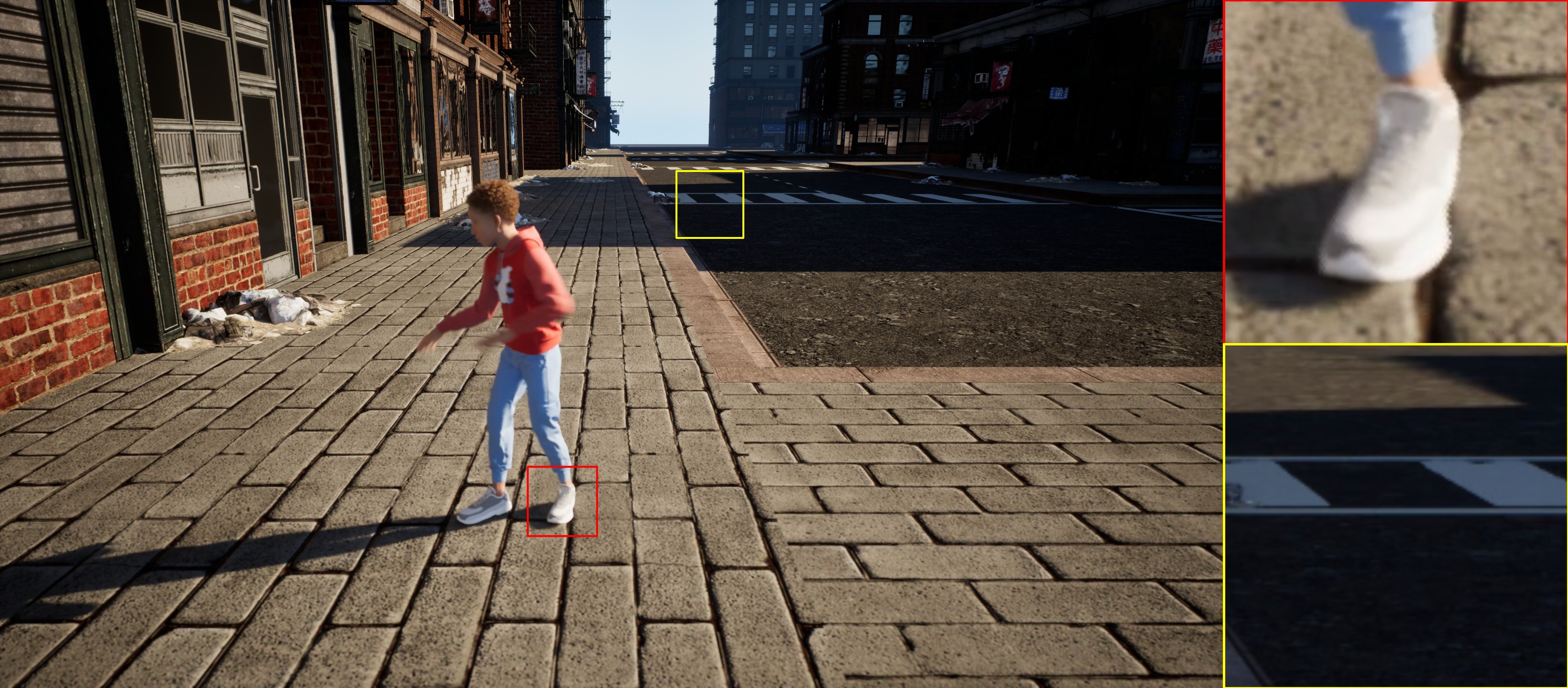}
   }
   \caption{Qualitative comparison on \textit{Dance boy city} scene of WideRange4D dataset.}
   \label{fig:danceboy}
   \vspace{-0.2cm}
\end{figure}

\begin{figure}[!t]
   \centering

   \subfloat[iFVC~\cite{tang2025compressing}]{
      \includegraphics[width=0.32\textwidth]{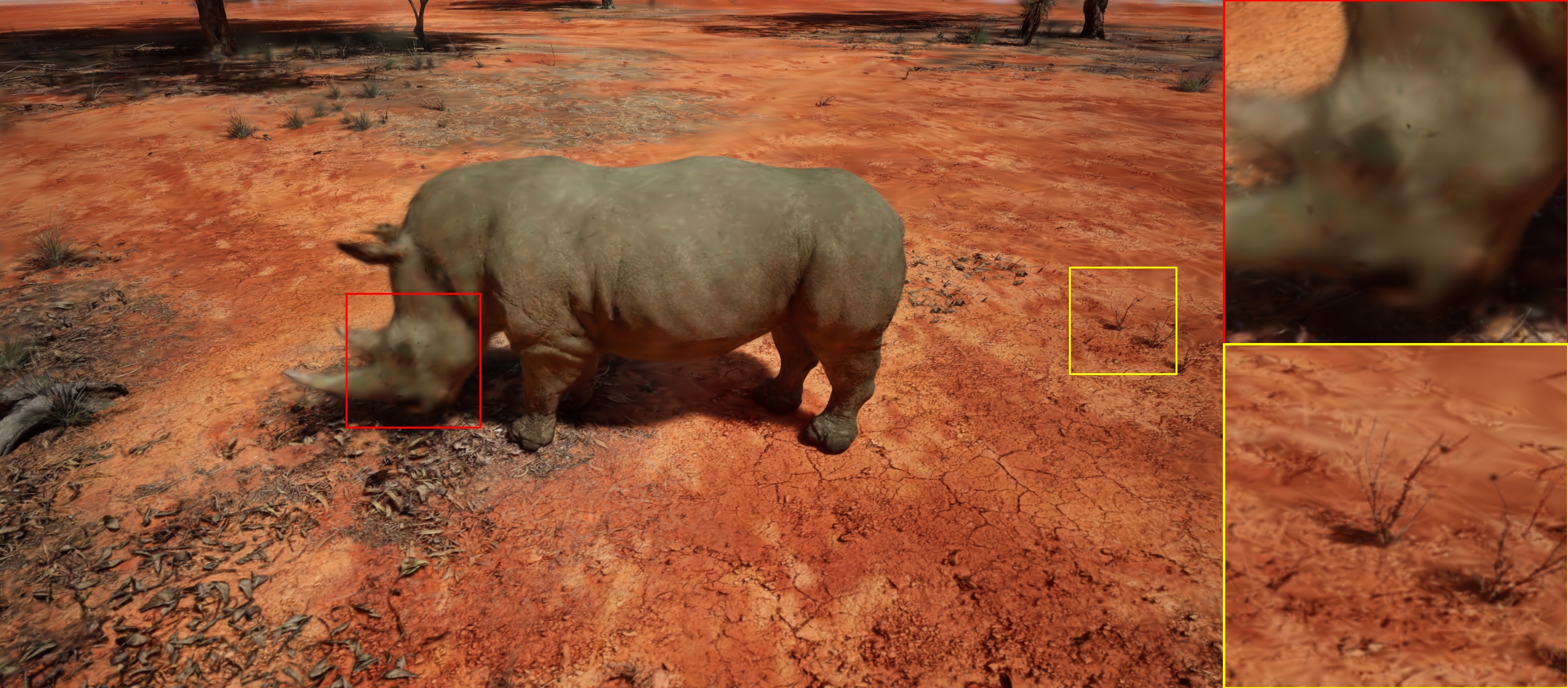}
   }
   \subfloat[Ours (Scaffold-GS)]{
      \includegraphics[width=0.32\textwidth]{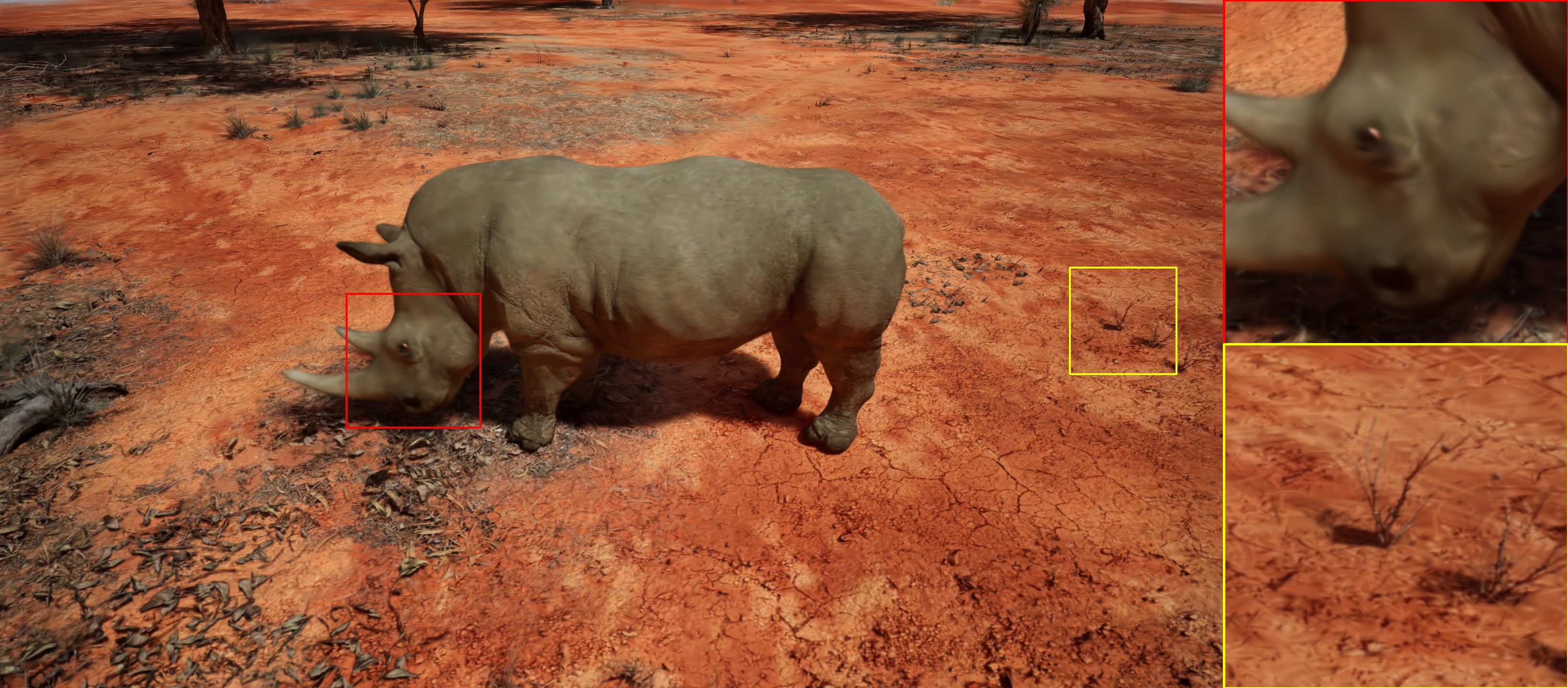}
   }
   \subfloat[Ground Truth]{
      \includegraphics[width=0.32\textwidth]{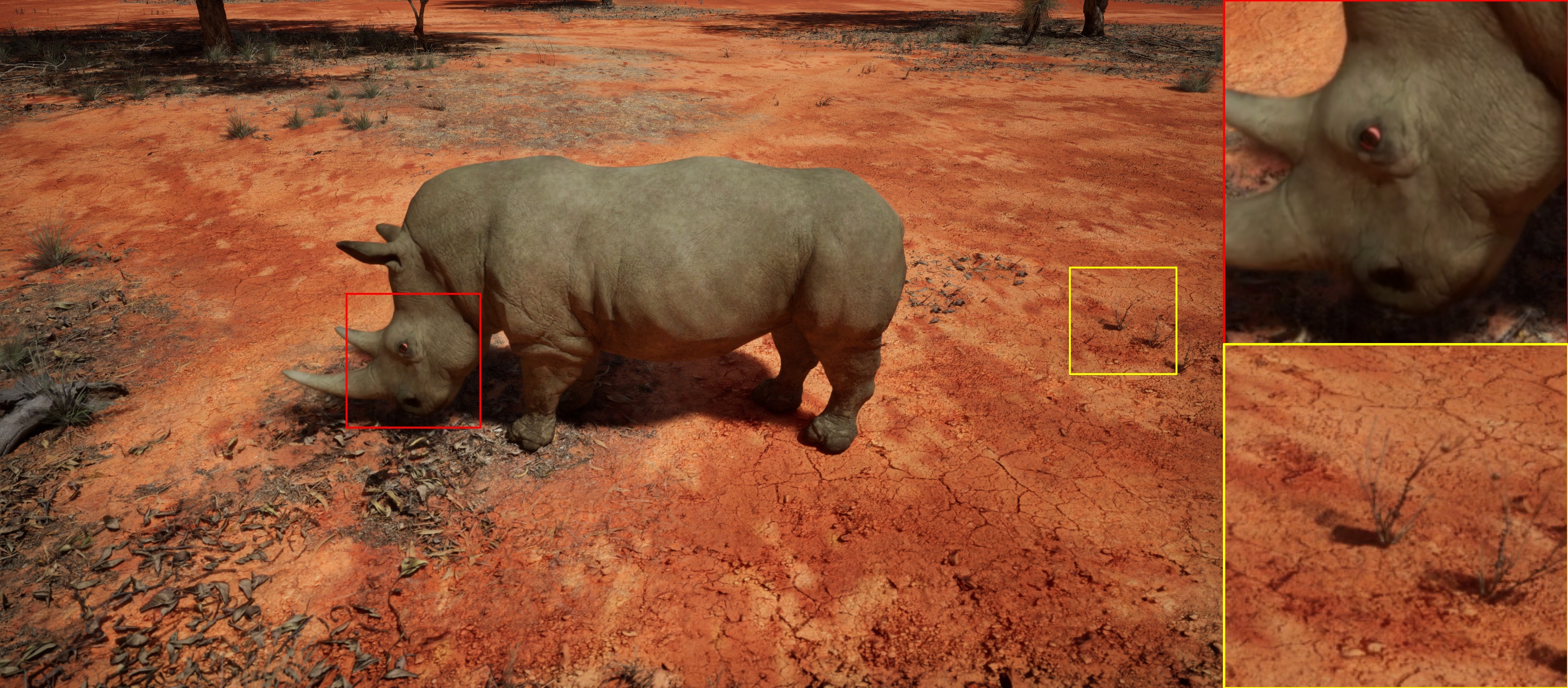}
   }
   \caption{Qualitative comparison on \textit{Findfood rhinoceros village} scene of WideRange4D dataset.}
   \label{fig:findfood}
   \vspace{-0.2cm}
\end{figure}

From Fig.~\ref{fig:cook} to Fig.~\ref{fig:findfood}, we present more qualitative comparisons. Our method effectively capture details in dynamic areas, leading to superior reconstruction quality across all real-world dynamic scenarios. We provide some videos in the supplementary materials, which demonstrate that our method successfully resolved flickering between GOPs, compared to GIFSteam~\cite{li2025gifstream} and StreamSTGS~\cite{ke2025streamstgs}.


\end{document}